\documentclass[12pt,twoside]{mitthesis}
\usepackage{lgrind}
\usepackage{cmap}
\usepackage{cjhebrew}
\usepackage[hyphens]{url}
\usepackage{hyperref}
\usepackage{times}
\usepackage{latexsym}
\usepackage{booktabs}
\usepackage{subcaption}
\usepackage{graphicx}
\usepackage{color,soul}
\usepackage{amsmath,amssymb}
\usepackage{multicol}
\usepackage{multirow}
\usepackage{bigdelim}
\usepackage[colorinlistoftodos]{todonotes}

\usepackage{comment}
\usepackage{numprint}
\usepackage[inline]{enumitem}
\usepackage{diagbox}
\usepackage{pifont}
\usepackage[acronym]{glossaries}
\usepackage{glossary-longbooktabs}
\makeglossaries

\newacronym{rnn}{RNN}{recurrent neural network}
\newacronym{seq2seq}{seq2seq}{sequence-to-sequence}
\newacronym{lstm}{LSTM}{long short-term memory}
\newacronym{gru}{GRU}{gated recurrent unit}
\newacronym{cnn}{CNN}{convolutional neural network}
\newacronym{relu}{ReLU}{rectified linear unit}
\newacronym{ctc}{CTC}{connectionist temporal classification}
\newacronym{recnn}{RecNN}{recursive neural network}

\newacronym{sgd}{SGD}{stochastic gradient descent}
\newacronym{em}{EM}{expectation maximization}
\newacronym{lrp}{LRP}{layer-wise relevance propagation}
\newacronym{svd}{SVD}{singular value decomposition}
\newacronym{lda}{LDA}{latent Dirichlet allocation}

\newglossaryentry{softmax}{name=Softmax, description={Normalization layer used in classification tasks}}
\newglossaryentry{attention}{name=attention, description={Mechanism for aligning source and target inputs in \acrfull{seq2seq} models}}
\newglossaryentry{dropout}{name=dropout, description={Method for regularizing neural networks}}
\newglossaryentry{highway}{name=highway, description={Type of connection in neural networks}}
\newglossaryentry{feedforward}{name=feed-forward, description={}}
\newglossaryentry{adam}{name=Adam, description={An adaptive optimization method}}

\newacronym[longplural={parts-of-speech}]{pos}{POS}{part-of-speech}
\newacronym{oov}{OOV}{out-of-vocabulary}
\newacronym{sem}{SEM}{semantic}

\newglossaryentry{lemma}{name=lemma, description={A dictionary item}}
\newglossaryentry{morpheme}{name=morpheme, description={A meaningful morphological unit}}
\newglossaryentry{phoneme}{name=phoneme, description={An abstract speech class that carries meaning in a specific language}}
\newglossaryentry{phone}{name=phone, description={A speech sound}}

\newacronym{pmb}{PMB}{Groningen Parallel Meaning Bank}
\newacronym{nlp}{NLP}{natural language processing}
\newacronym{pbmt}{PBMT}{phrase-based machine translation}
\newacronym{nmt}{NMT}{neural machine translation}
\newacronym{lsa}{LSA}{latent semantic analysis}

\newacronym{asr}{ASR}{automatic speech recognition}
\newacronym{mfcc}{MFCC}{Mel-frequency cepstral coefficient}
\newacronym{wer}{WER}{word error rate}
\newacronym{fft}{FFT}{fast Fourier transform}
\newacronym{lpc}{LPC}{linear predictive coding}
\newacronym{dtw}{DTW}{dynamic time warping}
\newacronym{hmm}{HMM}{hidden Markov model}
\newacronym{gmm}{GMM}{Gaussian mixture model}
\newacronym{plp}{PLP}{perceptual linear prediction}

\newglossaryentry{deepspeech2}{name=DeepSpeech2, description={End-to-end \gls{asr} model based on \gls{ctc}}}
\newglossaryentry{hamming}{name=Hamming, description={A window function often used in speech processing}} 
\newglossaryentry{formant}{name=formant, description={A natural frequency or resonance of a speech signal}}
\newglossaryentry{cepstrum}{name=cepstral analysis, description={A common procedure for de-convolution of a speech signal}}

\newglossaryentry{f1}{name=F$_1$, description={An evaluation metric defined as the harmonic mean of precision and recall}}

\newacronym{ai}{AI}{Artificial Intelligence}
\newignoredglossary{hidden}
\glsmoveentry{asr}{hidden}

\usepackage[marginparwidth=3cm]{geometry}
\usepackage{marginnote}

\usepackage{setspace}

\usepackage{epigraph}

\usepackage[sort,compress,numbers]{natbib}

\usepackage{bm}
\usepackage{amsfonts}
\usepackage{amsthm}

\usepackage{transparent}

\usepackage{lingmacros}

\usepackage{tabularx}

\usepackage{amsmath,amsfonts,bm}

\def\figref#1{figure~\ref{#1}}
\def\Figref#1{Figure~\ref{#1}}

\def\Twofigref#1#2{Figures \ref{#1} and \ref{#2}}

\def\eqref#1{equation~\ref{#1}}

\def\Chapref#1{Chapter~\ref{#1}}

\def\1{\bm{1}}

\def\vtheta{{\bm{\theta}}}

\def\vx{{\bm{x}}}

\DeclareMathAlphabet{\mathsfit}{\encodingdefault}{\sfdefault}{m}{sl}
\SetMathAlphabet{\mathsfit}{bold}{\encodingdefault}{\sfdefault}{bx}{n}

\def\sD{{\mathbb{D}}}

\DeclareMathOperator*{\argmin}{arg\,min}
\DeclareMathOperator*{\argminB}{argmin}

\usepackage{algorithm}
\usepackage[noend]{algpseudocode}

\begin{document}

\title{Scaling Laws for Deep Learning \\ {\fontsize{16.7}{22} \selectfont }}
\def\titleabstract{Scaling Laws for Deep Learning}

\author{Jonathan S. Rosenfeld}
       \prevdegrees{B.A., Physics, Israel Institute of Technology (2003) \\ 
                    B.Sc., Electrical Engineering, Israel Institute of Technology (2003) \\ 
                    MBA., Electrical Engineering, Israel Institute of Technology (2011) \\ 
                    M.Sc. Electrical Engineering and Computer Science, \\Massachusetts Institute of Technology (2019)}
\department{Department of Electrical Engineering and Computer Science}

\degree{Doctor of Philosophy in Electrical Engineering and Computer Science}

\degreemonth{September}
\degreeyear{2021}
\thesisdate{July 16, 2021}


\supervisor{Nir Shavit}{Professor of Electrical Engineering and Computer Science}

\chairman{Leslie A. Kolodziejski}{Professor of Electrical Engineering and Computer Science \\ Chair, Department Committee on Graduate Students}

\maketitle



\cleardoublepage
\setcounter{savepage}{\thepage}
\begin{abstractpage}
%
%
%


Running faster will only get you so far --- it is generally advisable to first understand where the roads lead, then get a car ...

The renaissance of machine learning (ML) and deep learning (DL) over the last decade is accompanied by an unscalable computational cost, limiting its advancement and weighing on the field in practice. 
In this thesis we take a systematic approach to address the algorithmic and methodological limitations at the root of these costs.
We first demonstrate that DL training and pruning are predictable and governed by scaling laws --- for state of the art models and tasks, spanning image classification and language modeling, as well as for state of the art model compression via iterative pruning. Predictability, via the establishment of these scaling laws, provides the path for principled design and trade-off reasoning, currently largely lacking in the field.
We then continue to analyze the sources of the scaling laws, offering an approximation-theoretic view and showing through the exploration of a noiseless realizable case that DL is in fact dominated by error sources very far from the lower error limit. We conclude by building on the gained theoretical understanding of the scaling laws' origins. We present a conjectural path to eliminate one of the current dominant error sources --- through a data bandwidth limiting hypothesis and the introduction of Nyquist learners --- which can, in principle, reach the generalization error lower limit (e.g. 0 in the noiseless case), \textit{at finite dataset size}.

\end{abstractpage}



\begin{center}
\cjRL{lhwryy\textrm{,} b'hbh wbh`rkh}
\end{center}

\clearpage

\section*{Acknowledgments}








I am deeply grateful to my advisor and dear friend Prof. Nir Shavit. Through freedom and guidance, advice and support, criticality and friendship, the numerous challenges and successes --- this work was made possible in profound, multidimensional, ways. 

Very little during the period of time and circumstances of my PhD was typical, and I am truly blessed to have had Leslie Kolodziejski and Janet Fischer --- the lifeblood oxygenating heart of EECS --- at my busy corner and to glean inspiration through their tireless care and dedication for students everywhere. One needs stronger words than `thank you'... perhaps `you succeeded in every way, and your impact immense'   

I would like also to thank members of MIT EECS community and beyond for helpful discussions and collaborations --- Josh Tenenbaum, Sasha Rakhlin, Constantinos Daskalakis, 
Piotr Indyk, Antonio Torralba, Aleksander Madry, Tommi Jaakola, Dan Alistrah, Lu Mi, Vikas Garg, Kai Xiao, Preetum Nakkiran, Yamini Bansal and Roy Shilkrot. 

Sharing destiny (and a publication) my brothers and best friends --- Amir Rosenfeld and Yonatan Belinkov --- thank you. For bolstering me professionally and emotionally,  sharing both warmth and guidance, work and path, general and specific. 

The ultimate gratitude to my parents who have shown unconditional belief and ongoing support --- without which this dream would have remained a fantasy.

Finally to my wife and partner, Meirav... life is not tried, it is merely survived, if you dance outside the fire.

\clearpage

\section*{Bibliographic Note}
Parts of this thesis are based on prior publications. 
\Chapref{sec:Dense} is mainly based on our work published in ICLR \cite{rosenfeld2020a} (first appearing in preprint form in 2019 \cite{rosenfeld2019constructive}).  
\Chapref{sec:Pruning} is based on our work published in ICML  \cite{rosenfeld2020predictability} (first appearing in 2020 \cite{rosenfeld2020pruningpre}). 



\pagestyle{plain}
\tableofcontents
 \cleardoublepage
\addcontentsline{toc}{chapter}{List of Figures}
\listoffigures
 \cleardoublepage
\addcontentsline{toc}{chapter}{List of Tables}
\listoftables
 \cleardoublepage
 
 \setglossarystyle{long3col-booktabs}
\setlength{\glsdescwidth}{0.45\linewidth}
\setlength{\glspagelistwidth}{0.25\linewidth}

\addcontentsline{toc}{chapter}{List of Abbreviations}
\printglossary[type=\acronymtype,title=List of Abbreviations]

\addcontentsline{toc}{chapter}{Glossary}
\printglossary

\newpage
\chapter{Introduction} \label{chap:introduction}
\glsresetall

\epigraph{\textit{``During the gold rush it's a good time to be in the pick and shovel business''} \\ --- Mark Twain}



\section{A Gold Rush at the Tip of the Iceberg}
Over the last decade, artificial intelligence (AI) has seen a renaissance in the form of tremendous growth and adoption across very many industries,  proving to be a valuable tool and business enabler. Of particular note is the use of neural networks (NN) and deep learning (DL). To glean the utility of these tools, one can gauge the largest corporations in the world for which they are core --- companies such as Apple, Google, Facebook and Amazon. The pervasiveness of these technologies has spanned across industries large and small which are undergoing transformations --- with data being the new gold. 

However, the adoption of AI and DL is akin to a goldrush in fundamental manners transcending the value of data --- with storied tales of riches accompanied by risky endeavours and with little guarantees with respect to time, effort and method of attaining them. The equal parts embarrassing and potential-laden state of affairs is that there is a shortage of shovels and an understanding of how to use them. 

Concretely, with the success and heightened adoption of neural networks for real world tasks, basic challenges concerning their research, development and real life costs still loom, and to a large extent, weigh on the field in its entirety. 

Chief amongst these challenges is the \textit{unscalable computational cost of deep learning}. 
%
Cost ramifications touch numerous aspects of the field. With mounting computational, and associated time and financial costs, the ability to engage in research and development becomes concentrated at the hands of organizations with correspondingly large resources \cite{Schwartz_2020}. Even within these large entities, the soaring costs impede advancements and slow advancement rates. Advancement is weighed down also indirectly by these mounting costs as the feedback phenomena of `algorithmic rut' \cite{Barham2019MachineLS} evolves: software and accelerators are optimized for old algorithms, relegating the exploration of novel algorithms and machine learning research ideas to inefficiencies making them much harder to explore, in turn much less attractive to incorporate in software tools and accelerators.
Beyond concentration and advancement concerns, the machine learning costs exact a correspondingly increasing environmental toll \cite{Strubell_2019}. Additionally, with the cost per development iteration constantly increasing, risks such as the safety and social bias of applications reliant on machine learning become increasingly costly to mitigate through testing and verification.

At the high level, the computational cost of DL stems from compounding factors in practice: 
\begin{itemize}
    \item Large scale models and data (and hence compute budgets) are paramount de-facto for the attainment of high-performance.
    \item Compoundingly, attaining good performance, de-facto involves very many iterations --- e.g. over hyperparameters. The elephant in the room is that trial and error (perhaps in the guise of fancy methods or names) through brute force are the predominant tool deployed in the development process of real life applications of machine learning as well as the advancement of the field.
\end{itemize}

One may be tempted to think that the tools-shovels required to remedy this situation are ones which would take the existing tasks and lower their computational costs given the current development practices (i.e. ``do the same, at lower cost / more of the same, for same cost") such as accelerators specifically designed for DL tasks --- whether general or specialized. Indeed, a plethora of ventures and companies have been doing just that, with large sums of money devoted to such accelerators both within organizations which rely heavily on DL and outside these organizations (the poster child of this is the role of the GPU and dedicated hardware or DL software).

However, while there is definitely utility in lowering the cost per computation, such an endeavour alone teeters glaringly, poorly scalable. The rate at which computational demands per task increases --- fueling the remarkable advancements in machine learning --- far outpaces the rate at which cost per computation and hardware performance gains can scale, and is nearing a regime where progress will be limited to accelerator (dominated by hardware) scaling rates \cite{thompson2020computational} and optimizations of current algorithms \cite{hern2020measuring}. 

Reassuringly, lowering the cost per computation is but the tip of a broader set of venues and questions we may consider worthwhile if we set to advance the field and scope of what is realistically attainable through machine learning --- by addressing the root causes giving rise to its limitations.

\begin{figure*}[t]
  \centering
    \includegraphics[width=1\linewidth,trim={1.85cm 5.5cm 1.5cm 3cm},clip]{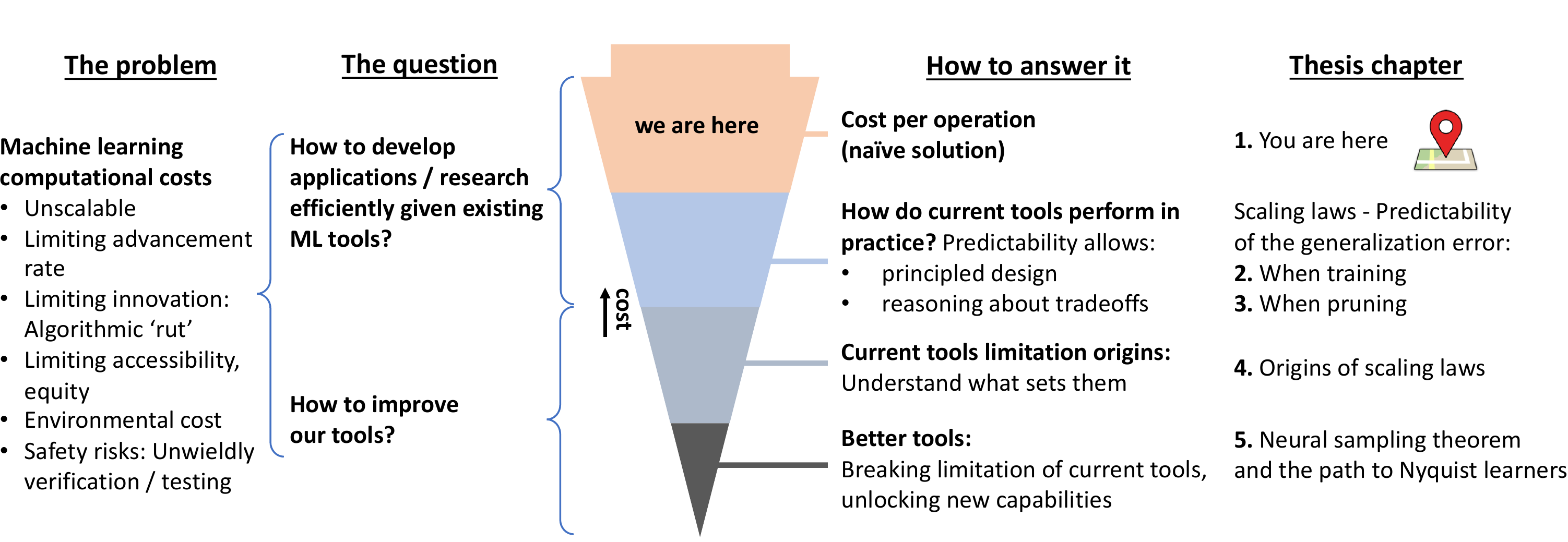}
    \caption{Thesis roadmap}
    \label{fig:roadmap_intro}

\end{figure*}

Figure \ref{fig:roadmap_intro} paints such a map of the layered efforts one may consider in order to, more broadly, tackle this target. This map has been the aspirational guide to this work and correspondingly the thesis is organized as its projection.

At a high level, we set out, first, to glean a better understanding of how machine learning behaves in practice --- and in particular, how does the generalization error (the error on unseen data, considered throughout this thesis as the metric of reference for performance) depend on different practical problem constituents. We find this dependency and show it to be predictable, following phenomenological (scaling) laws. Chapters 2 explores and establishes such laws for state of the art training scenarios across vision and language domains and Chapter 3 applies the same methodology for the pruning case. These two cases in tandem contribute to two key aspects of practical interest --- correspondingly the aspects of (1) ML-based application development through model training and (2) deployment of models for inference. As, attaining predictability of the error and problem constituents is necessary in any field to mature from trial and error to principled design and reasoning about tradeoffs, the scaling laws pave a path to such maturation in the context of wielding DL. Moreover, once armed with empirical laws governing performance, one can further query --- what gives them rise, seeking to glean insight as to how, perhaps, their limitations may be broken or broadened. In Chapter 4 we thus consider and classify the families of mechanisms which may give rise to the discovered scaling laws under a functional approximation view. We and segregate them through experimentation, by constructing a realizable teacher-student setup where we can control (eliminate) one such error source. We find the characteristic error landscape unchanged in this setting --- pointing to the other error sources dominance. Finally, we conjecturally discuss in Chapter 5 how what we have learned may pave the way to fundemental advances breaking the scaling laws currently governing DL costs.

Let us shed some more detail on the above stages:





\section{Scaling Laws --- Predictability and Principled Design}

Put simply, principled design, is design in light of an understanding of what one should expect. In the context of DL there is a gap in terms of common methodologies --- a lack of principled design --- mirrored in the following:
\begin{itemize}
    \item Training in large model and data scales becomes necessary without a principled constructive 
    path and guarantees connecting small and large scale performance. Here we use the term constructive to mean that the method specifies the model configuration completely.
    \item A relation between problem constituents such as the amount of data, model size, architectural dimensions and level of compressability (sparsity) leading to same error performance levels (or other performance metrics of interest) is not typically known. Consequently, analytical reasoning about expected costs, such as computational cost is not generally possible, and is relegated to empirical trials and heuristics.
    \item For the same reasons, reasoning about projected performance and tradeoffs between constituents is not generally possible. In other words, the performance envelope is unknown.
    
\end{itemize}

These issues share a composite common culprit --- and as such, a common remedy --- (lack of) knowledge or understanding of the laws governing machine learning in practice. Specifically, a principled (e.g. analytical) prediction means tying performance with problem constituents coupled with a constructive method specifying the model which would attain said performance if trained. 

Consequently, these gaps are symptoms of the poor understanding that, at this time, still permeates the excitingly immature nature of the machine and deep learning field. Akin to trial and error flight prior to understanding lift forces, construction prior to understanding loading capacities and heavenly bodies motion prior to Kepler's discoveries and laws of motion, machine learning is charging ahead without answers to questions such as how much  data is needed for a given performance? How much compute and how big a model? Is there a limit to the performance we can expect using these tools? Are we close? 

In Chapter \ref{sec:Dense} we explore the behavior of the generalization error of state of the art models --- across vision and language modeling domains, covering a wide range of model types and datasets. We find a common phenomological (scaling-) law governing the dependency of the generalization error (and loss) with the problem constituents --- model dimensions and dataset size. Specifically, we find a joint power law dependence of the error on data and number of parameters (induced by width / depth). Through the demonstration of accurate fits and extrapolations, we thus arrive at a predictive, constructive, tool in the practical scenario of training models. 

In Chapter \ref{sec:Pruning} we set to achieve a similar in spirit goal in the complementary setting of pruning. This setting is of particular interest both because it is a materially different setting in which to explore laws and predictability and because of applicative interest in the context of the optimization of models for deployment and inference. Employing the same methodology, we find that this case is also highly predictable, and shares power law characteristics of the scaling laws, while exhibiting pruning-specific additional insights.

\section{Origins of Scaling Laws}

Predictability in itself, as we argue, is necessarily useful in practice, but it leaves one both intellectually curious and practically hungry --- if there is a, predictable, performance envelope, where does it come from? In Chapter \ref{sec:towards_origin} we explore the potential types of sources of the generalization error and show that two out of three of these archetypical sources dominate the error, but not a third. Specifically, the error is not dominated by a hypothesis mismatch between the true and model class, but is dominated by approximation uncertainty and learning algorithm (optimization) deficiencies. 

\section{A Path to Better Learning}

The insights gained regarding the limitations associated with the sources of error giving rise to the found scaling laws, allow us to consider the conditions under which they may be breached. Chapter \ref{sec:towards_nyquist} is of a future looking and conjectural nature. Of particular interest is the notion of data sufficiency and an hypothesized class of agorithms which may reach minimal generalization error for finite detests.











\section{Related Work}
\paragraph{Model Scaling:}
A number of studies have explored the effect of model scaling on performance. 
For instance, image classification networks can be scaled by depth~\citep[number of layers;][]{he2016deep} or width~\citep[number of channels;][]{zagoruyko2016wide,howard2017mobilenets}. 
More recently, 
\citet{tan2019efficientnet} 
demonstrated how scaling width, depth, and input resolution has combined positive effects larger than  scaling each factor in isolation. However, this relationship has yet to be quantified in a predictive form -- by how much will error change with model scaling?
In this work, we focus on finding a constructive functional form for determining the model given a specified performance. 

\paragraph{Data Scaling:}
It has long been recognized that more data improves performance, and various studies report such trends in both computer vision~\citep[e.g.,][]{zhu2012we,sun2017revisiting} and language processing tasks~\citep[e.g.,][]{banko2001mitigating,talmor-berant-2019-multiqa}. 
A number of prior studies  observed power-law relations between the generalization error and training data size~\citep{cho2015much,miceli-barone-etal-2017-regularization,johnson-etal-2018-predicting}. 
Most relevant to our work, \citet{hestness2017deep}  explored the effect of data size on the generalization error in vision, language, and speech  tasks, and observed a strikingly consistent power-law behavior in a large set of experiments. However, while these studies point to the empirical \textit{existence} of a power law in terms of data, they do not offer tools for  predicting the performance given a specified model.
Nor do they offer low-cost methods to specify the model configuration which would attain the power law with data dependency.
Indeed, \citeauthor{hestness2017deep}\ had to search over models and their configurations at large scale to exhibit their findings, incurring prohibitive computational costs.

In contrast, we demonstrate a constructive recipe, where we directly predict the test performance at large scale and specify the full model configuration which attains it (with no need for large-scale search), given performance at small scale. Following our work, \cite{kaplan2020scaling} at OpenAI present very similar results in a subset of the scenarios we consider (language modeling) and corroborate them at a much larger scale. Further work from OpenAI takes these results to the largest scale in current existence in GPT3 \cite{brown2020language}.

\paragraph{Predicting Model Performance:}
Since training models at full data/model scale may be computationally prohibitive, a line of work tries to predict the performance of a given model  on a given dataset, without training the model, for example by using a bank  of previously trained models, dataset, and their associated performances~\citep{istrate2019tapas}.
Others have proposed to estimate performance on small data~\citep{klein2017fast} or model sizes~\citep{zoph2018learning,real2019aging} in the context of neural architecture search (NAS). In this case, the small-scale evaluation is used to compare models at small cost, to expedite the search process; see~\cite{elsken2019neural} for a recent survey. 
Our work complements previous approaches by demonstrating a functional form that can predict large-scale performance from small-scale measurements. Moreover, our method may be integrated in NAS, addressing some of its current limitations (as discussed in section \ref{sec:discussion}). 

\paragraph{Theoretical Error Bounds and Scaling Laws Origins:}
Much attention has been given to theoretical explanations of the generalization capabilities of deep neural networks~\citep{
neyshabur2017exploring,neyshabur2017pac,allen2018learning,allen2018convergence,arora2018stronger}, preceding this work. Bounds of the generalization error have persistently proved loose or vacvuous with dependencies going counter to practice (e.g. error increasing with model size). 
Posterior to the findings in this work (Chapters 2--4), theoretical interest in explaining the scaling laws has followed and broadened a similar approximation-centric \cite{Sharma2020ANS,Bahri2021ExplainingNS} analysis yielding a rigorous theoretical understanding of the scaling laws power dependency, but leaving unexplained the error plateauing (low error limit) mechanism --- see Chapter \ref{sec:towards_origin} where we engage with the origins of the scaling laws.

\section{Contributions}

\textbf{Chapter \ref{sec:Dense}:}
\begin{itemize} 
    \item We establish a scaling law governing the generalization error: A \textit{joint} functional form of the generalization error landscape---as dependent on both data and model size---with few, interpretable degrees of freedom.
    \item Constructive: Direct and complete specification (via a scaling policy) 
    of the model configuration attaining said generalization error across model and dataset sizes.
    \item Highly accurate approximation of error measurements across model and data scales via the functional form, evaluated on different models, datasets, and tasks.
    
    \item Highly accurate error prediction from small  to large model and data. 
\end{itemize}

\textbf{Chapter \ref{sec:Pruning}:}
\begin{itemize}
    \item We develop a scaling law that accurately estimates the error when pruning a single network with Iterative Magnitude Pruning (IMP).
    \item We observe and characterize an \emph{invariant} that allows error-preserving interchangeability among depth, width, and pruning density.
    \item Using this invariant, we extend our single-network scaling law into a joint scaling law that predicts the error of all members of a network family at all dataset sizes and all pruning densities.
    \item In doing so, we demonstrate that there is structure to the behavior of the error of iteratively magnitude-pruned networks that we can capture explicitly with a simple functional form and interpretable parameters.
    \item Our scaling law enables a framework for reasoning analytically about IMP, allowing principled design and reasoning about tradeoffs in the pruning context as well.
\end{itemize}

\textbf{Chapter \ref{sec:towards_origin}:}
\begin{itemize}
    \item We introduce an approximation-centric view of the sources of error: delineating the role of realizability, uncertainty, learning deficiency and noise.
    \item Through the construction of a realizable noiseless teacher-student setup we show that realizability is not the dominant error source; rather, it is uncertainty and learning deficiency.
\end{itemize}

\textbf{Chapter \ref{sec:towards_nyquist}:}
\begin{itemize}
    \item We discuss the information theoretic circumstances under which perfect reconstruction (eliminating the uncertainty error source) are possible and introduce a Neural Sampling Conjecture.
    \item We introduce the Bandlimited Data Hypothesis --- hypothesizing that the data (target function) over the data manifold is bandlimited. 
    \item Under the Bandlimited Data Hypothesis and Neural Sampling Conjecture, we point to the existence of Nyquist learners --- learners for which the uncertainty error source is eliminated, and thus can reach the lower error limit (which can be as low as 0 in the noiseless case) at finite dataset size.
\end{itemize}
\chapter{Scaling Laws in the Wild} \label{sec:Dense}
\epigraph{\textit{``Without proper experiments I conclude nothing.''} \\ --- Johannes Kepler}
\glsresetall
In this chapter, which is heavily based on our work as published in \cite{rosenfeld2020a} we establish a constructive prediction of the generalization error. This chapter concentrates on the case of training dense models and explores the relationship (scaling laws) between model degrees of freedom (e.g. depth, width), dataset size and generalization error. We formulate such scaling laws and show that we achieve both high accuracy fits and good extrapolation accuracy. 

\section{Introduction}
In this work we circle back to the fundamental question:  \textit{what is the (functional) relation between generalization error and model and dataset sizes}? Critically, we capitalize on the concept of model scaling in its strictest form: we consider the case where there is some given scaling policy 
that completely defines how to scale up a model from small to large scales. We include in this context all model parameters, such that traversing from one scale (in which all parameters are known) to another requires no additional resources for specifying the model (e.g., architecture search/design). 

We empirically explore the behavior of the generalization error over a wide range of datasets and models in vision and language tasks. While the error landscape seems fairly complex at first glance,
we observe the emergence of several key characteristics shared across benchmarks and domains. Chief among these characteristics is the emergence of regions where power-law behavior approximates the error well both with respect to data size, when holding model size fixed, and vice versa.

Motivated by these observations, we establish criteria which a function approximating the error landscape  should meet. 
We propose an intuitive candidate for such a function and evaluate its quality, both in explaining the observed error landscapes and in extrapolating from small scale (seen) to large scale (unseen) errors. 
Critically, our functional approximation of the error depends on both model and data sizes. 
We find that this function leads to a high quality fit and extrapolation. For instance, the mean and standard deviation of the relative errors are under 2\% when fitting across all scales investigated and under 5\%  when extrapolating from a slimmed-down model (1/16 of the parameters) on a fraction of the training data (1/8 of the examples) on the ImageNet \citep{russakovsky2015imagenet} and WikiText-103 \citep{merity2016pointer} datasets, with similar results for other datasets. 


\section{Experimental Setup}

\paragraph{Notation:} Let $\sD_n = \{  \vx_i,y_i \}_{i=1}^{n}$ denote a labeled (training) dataset with $n$ samples or datapoints.
Let $f_m$ denote a neural network whose size is the number of parameters $m$, such that $\hat{y} = f_m(\vx)$ is the predicted label.
Let $\epsilon \left(n,m  \right)$ be the generalization error as a function of $n$ and $m$, measured by a performance metric (e.g., top-1 accuracy or cross-entropy loss) on a held-out test set. We refer to this error function as the \textit{error landscape}. 

\subsection{Scaling Policies}

\paragraph{Dataset scaling:}
We wish to scale datasets while preserving the original distribution.
For image classification, we uniformly subsample all classes by a constant ratio, thus  preserving the relative sample size per class. We limit the maximal sub-sampling to avoid eradicating any class.
For language modeling, where the number of classes (vocabulary items) has a very long tail distribution, we randomly sample sentences such that the total number of sampled words will be a certain fraction of the original dataset. 
Table~\ref{tab:stats} reports the data scales we use. 
In all tasks the held-out test set remains untouched for evaluating the error.

\paragraph{Model scaling:} We are critically interested in a method where moving across scales is defined by some scaling function, such that no additional significant computation would be incurred. We thus  consider the case where the model architecture is given and the model size determines how to scale it. 
For instance, one may scale width (number of channels in convolutional networks, hidden state size in recurrent networks), depth (number of layers),  do compound scaling~\citep{tan2019efficientnet},  or more generally define a function tying the model degrees of freedom and size. 
We focus primarily on width scaling in our experiments; the model scales are reported in Table~\ref{tab:stats}. We also perform selected depth scaling to demonstrate flexibility with respect to the scaling method. 

\paragraph{Hyper-parameters:}
For similar reasons we wish to avoid hyper-paramater search at large scales, and thus 
avoid the temptation to tune hyper-parameters accordingly (learning rate, regularization, etc.). Therefore, we hold all hyper-parameters fixed. 
This 
enables us to construct a functional form that fits the error landscape and can be used to predict the error across scales while completely defining the model attaining it. 

\subsection{Tasks, Models, Optimizers and Datasets}

We experiment with both vision and language tasks.  We use 6  benchmark datasets for  image classification and 3 for language modeling. For image classification, we train ResNet~\citep{he2016deep} and WRN models~\citep{zagoruyko2016wide} with stochastic gradient decent (SGD). In section \ref{sec:arch_optim_var} we explore the effect of varying architectures and optimizers for a fixed task (CIFAR100), adding VGG16 \citep{simonyan2014very} and DenseNet \citep{huang2017densely} models trained with both Adam \citep{kingma2015adam} and SGD. 
For language modeling, we train AWD-LSTM~\citep{merity2018regularizing} and Transformer-XL models~\citep{dai-etal-2019-transformer} with SGD and Adam optimizers respectively. Summary statistics are shown in Table~\ref{tab:stats}, along with the range of explored scales.
Appendix~\ref{app:data-models} gives additional information.

\begin{table}[t]
\centering
\caption{The datasets and models used in this work, along with their original training data size and the range of explored scales. For more information, see appendix~\ref{app:data-models}.}
\begin{subtable}[b]{\textwidth}
\centering
\caption{Training data size (number of words) and model size (number of parameters excluding word embeddings) for language modeling tasks. 
}
\small
\begin{tabular}{l r c p{1.25cm} l r c p{1.25cm}}
\toprule
 Dataset & Size ($N$) &  & Scales ($n$) & Base Model  & Size ($M$) & & Scales ($m$)  \\ 
 \cmidrule(lr){1-4} \cmidrule(lr){5-8}
PTB & 0.9M & \rdelim\}{3}{*} & \multirow{3}{1.3cm}{$2^{-k} N$, $0 \leq k \leq 5$} & AWD-LSTM & 20M & \rdelim\}{3}{*} & \multirow{3}{1.3cm}{$4^{-k}M$, $0 \leq k \leq 6$} \\ 
WikiText-2 & 2M & &  & AWD-LSTM & 20M &  &\\ 
WikiText-103 & 100M & &  & Transformer-XL & 41M & & \\ 
\bottomrule 
\end{tabular}
\label{tab:stats-language}
\end{subtable} \vspace{10pt}
\begin{subtable}[b]{\textwidth}
 \vspace{10pt}
\caption{Training data size (number of images) and model size (number of parameters) for image classification tasks. }
\centering
\small
\begin{tabular}{l r c p{1.75cm} l r c p{3.2cm}}
\toprule
 Dataset & Size ($N$) &  & Scales ($n$) & Base Model  & Size ($M$) & & Scales ($m$)  \\ 
 \cmidrule(lr){1-4} \cmidrule(lr){5-8}
ImageNet        & 1.2M & & $2^{-k} N$,  $0 \leq k \leq 6$  & ResNet-50 & 25.5M & &$4^{-k} M$,  $0 \leq k \leq 6$  \\
CIFAR10        & 60K  & \rdelim\}{5}{*} & \multirow{5}{1.5cm}{$2^{-k} N$, $0 \leq k \leq 5$} &  WRN-44-16 &  0.7M  & & $4^{-k} M$,  $-3 \leq k \leq 4$  \\

CIFAR100     & 60K &  &  & WRN-44-16 & 0.7M  & \rdelim\}{4}{*} & \multirow{4}{2cm}{$4^{-k} M$, $-2 \leq k \leq 4$} \\
DTD             & 5640 & &  & WRN-44-16     & 0.7M  & & \\ 
Aircraft        & 10K  & &  & WRN-44-16     & 0.7M  & & \\ 
UCF101          & 13K  & &  & WRN-44-16     & 0.7M  & & \\ 
\bottomrule 

\end{tabular}
\label{tab:stats-vision}
\end{subtable}
\label{tab:stats}
\end{table}

\clearpage

\begin{figure}[t]
\centering
    \begin{subfigure}[b]{0.45\linewidth}   
        \centering 
        \includegraphics[width=\linewidth
        ,trim={0 0 0 1.3cm}
        ,clip]{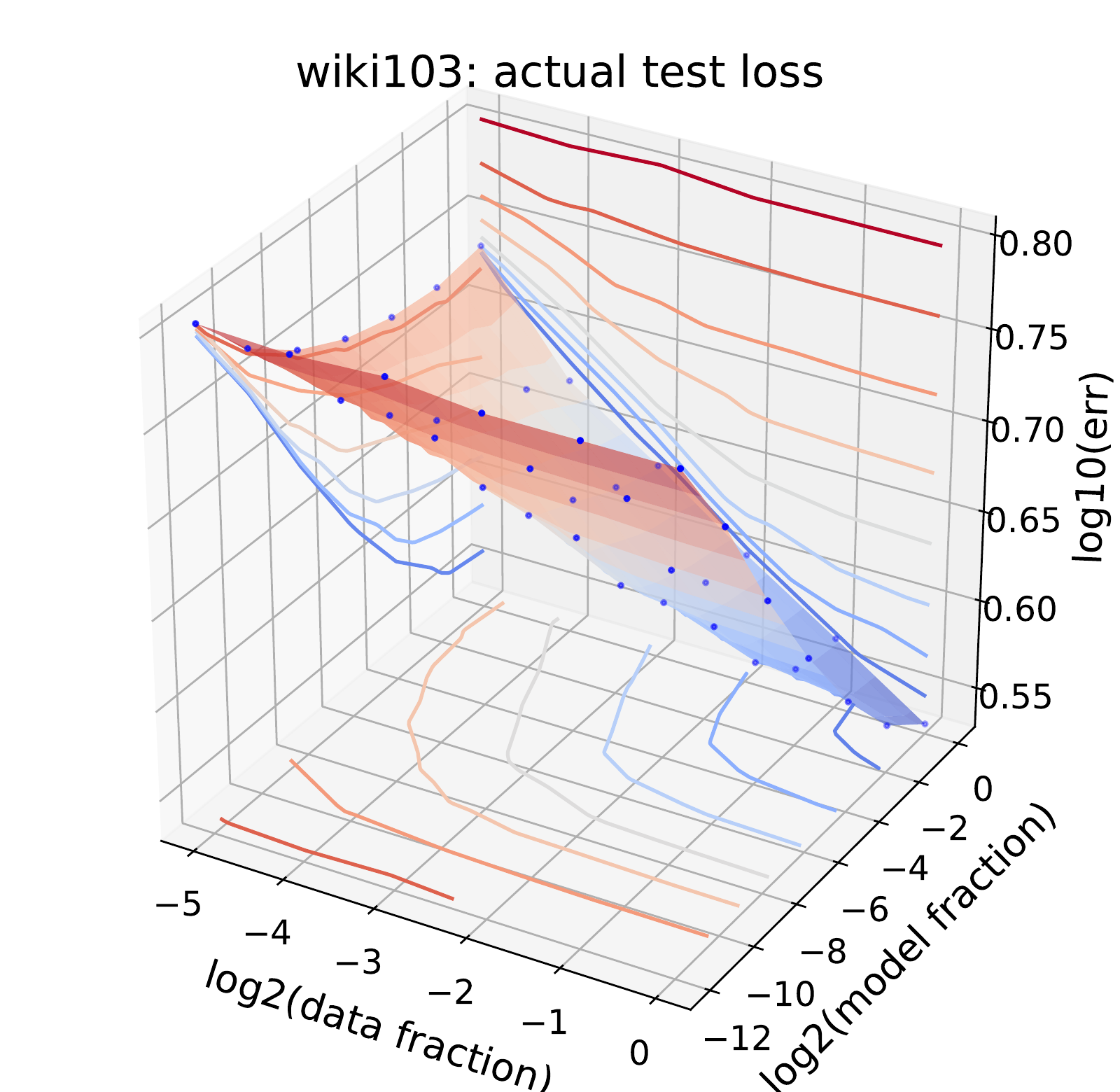}
        \caption{Wiki103 error (cross entropy) landscape.}    
        \label{fig_sub:observe_3d_wiki103}
    \end{subfigure}
    \hfill 
    \begin{subfigure}[b]{0.45\linewidth}   
        \centering 
        \includegraphics[width=\linewidth
        ,trim={0 0 0 1.3cm}
        ,clip]{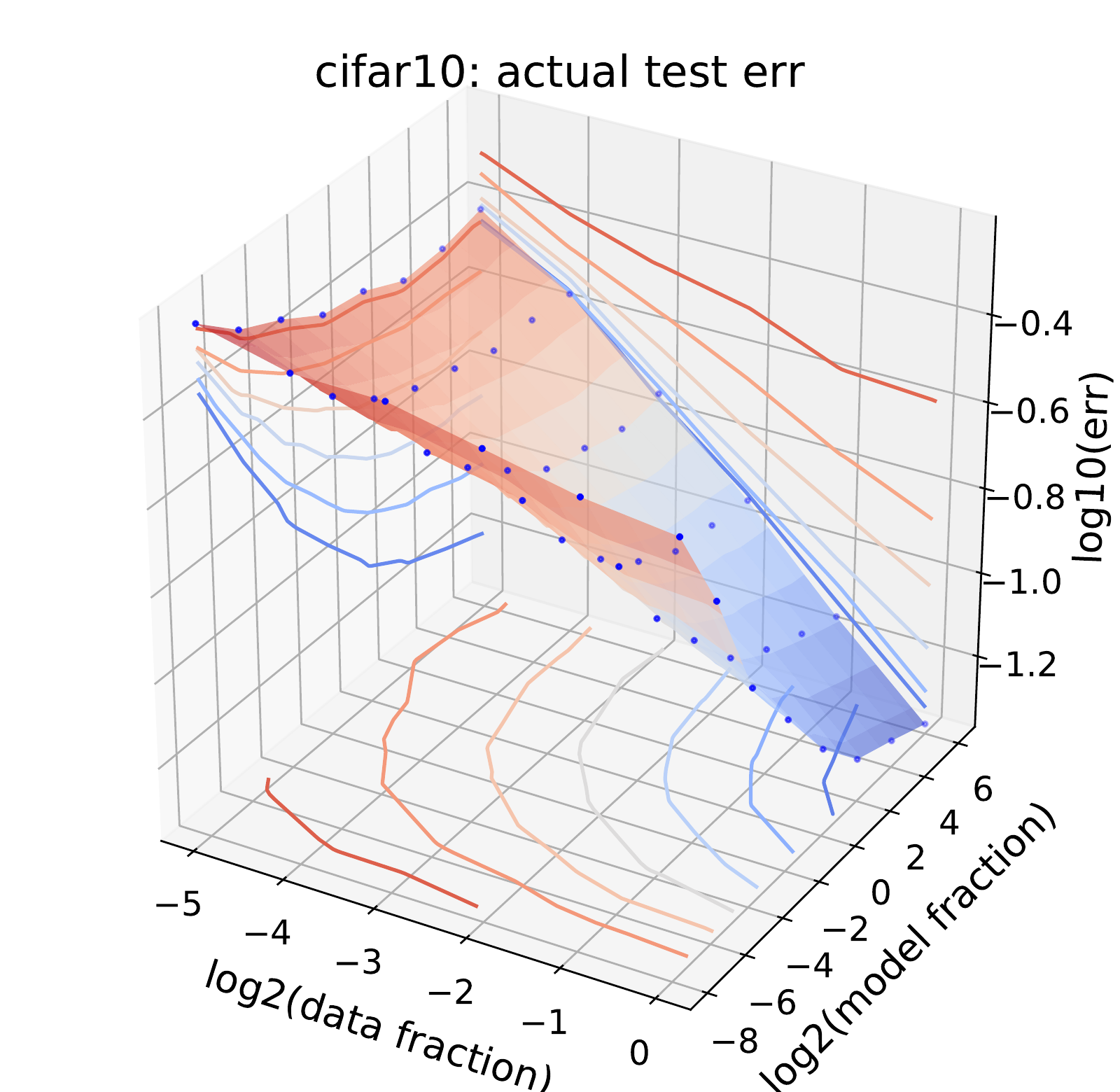}
        \caption{CIFAR10 error (top1) landscape.}    
        \label{fig_sub:observe_3d_cifar10_depth=44}
    \end{subfigure}
    \caption{ 
    Error landscapes in log-log-log scale. Each point (blue dot) is the error resulting from training with a model/data configuration $m,n$. The surface is a linear interpolation between the points, which is then projected on the $(m,\epsilon)$, $(n,\epsilon)$ and $(m,n)$ planes. 
    See Appendix~\ref{app:landscape} for  details. 
    } 
    \label{fig:landscape-3d}
\end{figure}

\begin{figure}[t]
    \centering
    \begin{subfigure}[b]{0.475\textwidth}
        \centering
        \includegraphics[width=\linewidth]{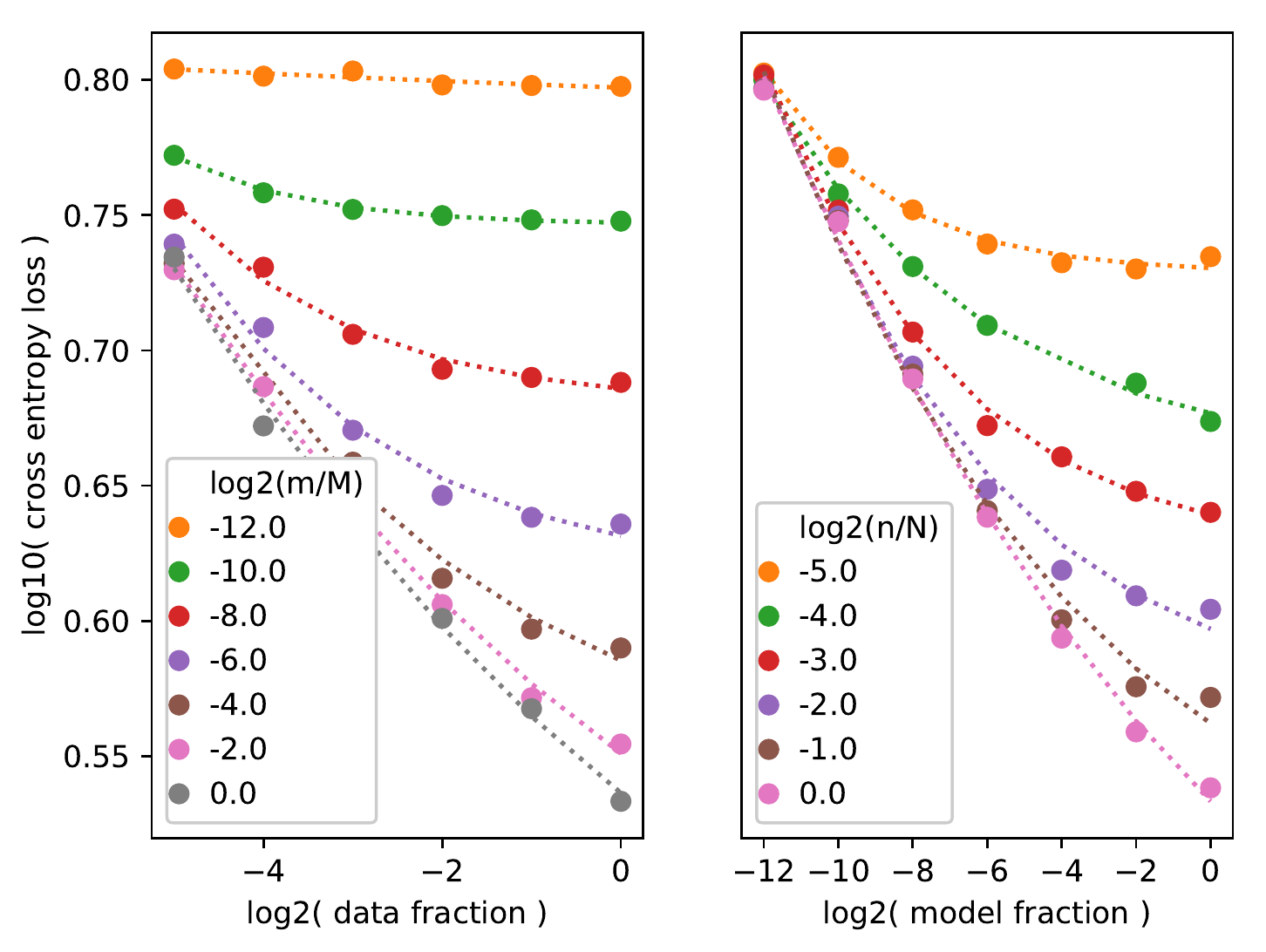}
        \caption{ Wiki103 cross entropy vs.\ data and model size. }
        \label{fig_sub:observe_2d_wiki103}
    \end{subfigure}
     \hfill
    \begin{subfigure}[b]{0.475\linewidth}  
        \centering 
        \includegraphics[width=\linewidth]{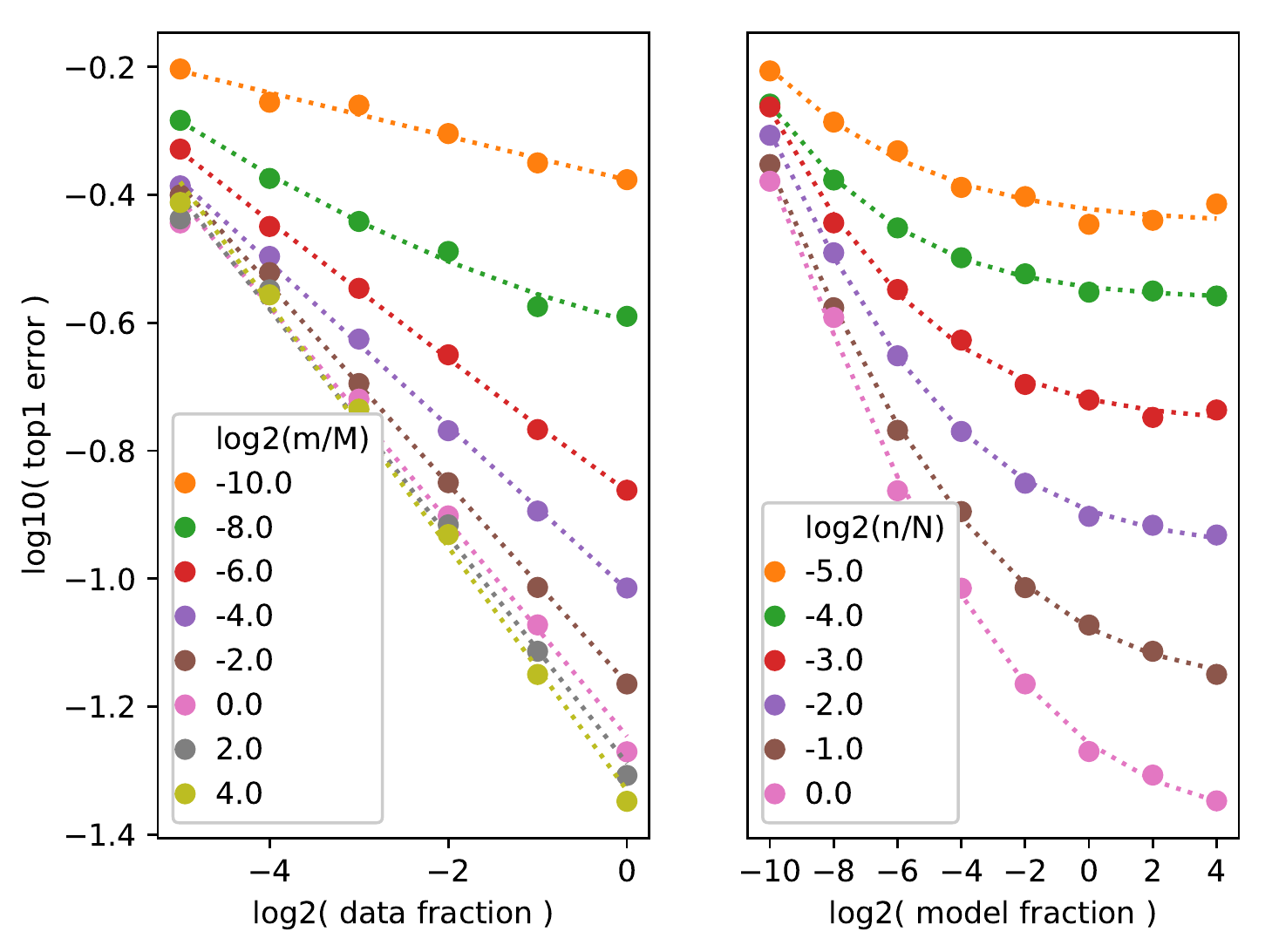}
        \caption{CIFAR10 top1 error vs.\ data and model size.}
        \label{fig_sub:observe_2d_cifar10}
    \end{subfigure}
\caption{Error  vs.\  data size (left part of each subfigure) and model size (right part) for Wiki103 and CIFAR10.  
Solid dots are measurements, dashed lines are best fit to saturating power-law.}
\label{fig:landscape-2d}    
\end{figure}

\section{Observations on the Error Landscape}
\label{sec:observations}

\Twofigref{fig_sub:observe_3d_wiki103}{fig_sub:observe_3d_cifar10_depth=44} respectively show an example test error landscape for width scaling of Transformer-XL on WikiText-103 and WRN-44-16 on CIFAR10. 

Various additional such landscapes are found in appendix \ref{app:landscape}, showing largely consistent patterns. 
Examining the error landscapes yields the following observations:
%

\begin{enumerate}[leftmargin=*,label=O\arabic*]
    \item \textbf{Model Scaling} 
    \begin{enumerate}[label*=.\arabic*,leftmargin=15pt]
        \item For a given dataset size, scaling up the model results in an initial decrease in test error, which then saturates to a level determined by the dataset size.\footnote{At some point error increase ensues; this point differs between datasets.  
        , see  Appendix \ref{app:landscape} for examples.}
         This behavior has been noted by \citet{tan2019efficientnet} across varied model scaling methods, although they have not engaged with the dependency on dataset size.
        \item The rate of error decrease with model size appears well approximated by a power-law. 
        
        \medskip 
        These two observations together can be summarized as the following relation: 
        \begin{equation}
            \epsilon(m,n) \approx b(n)m^{-\beta(n)} +c_m(n) \label{eq:error_of_m}
        \end{equation}
        where $b, \beta, c_m$  may depend on the data size $n$, s.t.\ as $m$ grows, $\epsilon \rightarrow c_m$. 
        Example fits to this form (allowing $b, \beta, c_m$ to be fit per $n$) are seen in \figref{fig_sub:observe_2d_wiki103} (right) and \figref{fig_sub:observe_2d_cifar10} (right). 
    \end{enumerate}
    \item \textbf{Data scaling} 
    \begin{enumerate}[label*=.\arabic*,leftmargin=15pt]
        \item For a given model size, scaling up the dataset results in an initial increase in performance, which then saturates to a level determined by the model size. 
        \item The rate of error decrease with dataset size appears well approximated by a power-law.
        \citet{hestness2017deep} 
        also noted a similar relationship, but did not functionally tie the saturation level to the dataset size. 
        
        \medskip 
        These two observations together can be summarized as the following relation: 
        \begin{equation}
            \epsilon(m,n) \approx a(m)n^{-\alpha(m)} +c_n(m)  \label{eq:error_of_n}
        \end{equation}
        where $a, \alpha, c_n$  may depend on the model size $m$, s.t.\ as $n$ grows, $\epsilon \rightarrow c_n$. Example fits to this form (allowing $a, \alpha, c_n$ to be fit per $m$) are seen in \figref{fig_sub:observe_2d_wiki103} (left) and \figref{fig_sub:observe_2d_cifar10} (left).
    \end{enumerate}
    \item \textbf{Joint properties }
    \label{obs:joint}
    The behavior of the error when scaling model size while holding data size fixed, and vice versa, extends to the entire error landscape in a well-behaved manner, such that the manifold $\epsilon(m,n)$ is smooth everywhere as a function of both model and data scales. 
\end{enumerate}

\section{Functional Approximation of the Generalization Error} \label{sec:function}
\subsection{Criteria}
\label{sec:criteria}

Motivated by the above observations, we now consider a functional approximation for the error landscape.  
In particular, 
let us consider function families meeting the following criteria which augment and restrict our observations:%
%
\begin{enumerate}[label=C\arabic*,topsep=1pt,parsep=0pt]
    \item As \textbf{either} model or dataset size goes to zero, the expected performance is equivalent to a random-guess error level $\epsilon_0$.\footnote{Best guess when $m \rightarrow 0$ ($\epsilon_{0n}$) or $n \rightarrow 0$ ($\epsilon_{m0}$) need not coincide, but can, e.g., in a balanced dataset.} 
    \label{crt:init_rand_guess}
    \item For a given dataset size, scaling up the model will result in an initial increase in performance, which will then saturate, taking the form in \eqref{eq:error_of_m}.
    \label{crt:scaling_m}
    \item For a given model size, scaling up the dataset will result in an initial increase in performance, which will then saturate, taking the form in \eqref{eq:error_of_n}. 
    \label{crt:scaling_n}
    \item There exists an irreducible error $\epsilon_\infty$, intrinsic to the dataset. 
    \label{crt:final_sat}
    \item The function must be smooth everywhere and monotonic non-increasing in terms of model and data size (observation~\ref{obs:joint}). 
    \label{crt:joint}
\end{enumerate}
While there are many possible function families meeting the above criteria, below we propose a simple function family for our evaluation. We do not claim that this is in fact the true underlying dependency, but rather that it serves as a good approximation of the error landscape---consistent with these criteria.

\subsection{Proposed Function Family}

As a first insightful step, consider the implications of satisfying  \ref{crt:scaling_m} and \ref{crt:scaling_n} \textit{simultaneously}. 
By examining the limiting behavior as $m$ or $n$ grow,  we have:
\begin{align*} 
  &\qquad \text{As $m$ grows large:}  & c_m(n) &\approx a(m)n^{-\alpha(m)}+c_n(m) \\ 
  &\qquad \text{As $n$ grows large:} & c_n(m) &\approx b(n)m^{-\beta(n)}+c_m(n) 
\end{align*}
Thus, a consistent form satisfying  \ref{crt:scaling_m} and \ref{crt:scaling_n} simultaneously is:
\begin{equation} \label{eq:consitent_form}
    \epsilon(m,n) \approx a(m)n^{-\alpha(m)} + b(n)m^{-\beta(n)} + c_\infty 
\end{equation}
where $c_\infty$ is a constant not dependent on either $m$ or $n$.

Let us now examine the simplified case where $a,b,\alpha,\beta$ are constant: 
\begin{equation} \label{eq:norm_err} 
    \tilde{\epsilon}(m,n) = an^{-\alpha} + bm^{-\beta} + c_\infty  
\end{equation}
where  $\alpha \ge 0$ and $\beta \ge 0$ control the \textit{global} rate at which error decreases with data and model size, respectively, $a>0$ and $b>0$ are a form of unit conversion between data and model sizes and error, and $c_\infty>0$ is the asymptotic lower value attainable.
This function is a special case of \eqref{eq:consitent_form} and meets criteria \ref{crt:scaling_m} and \ref{crt:scaling_n} by construction. Importantly \ref{crt:final_sat} and \ref{crt:joint} are also met. 

However, by giving up the dependence of $a,b,\alpha,\beta$ on $m,n$, this function does not meet criterion \ref{crt:init_rand_guess}.
We thus need to  model the transition from the initial random-guess level to the power-law region.
We propose to parameterize the transition using the following envelope (complex) function:
\begin{equation} \label{eq:envelope}
    \hat{\epsilon}(m,n) = \epsilon_0 \left\Vert \frac{ \tilde{\epsilon}(m,n)}{\tilde{\epsilon}(m,n)-i\eta} \right\Vert = \epsilon_0 \left\Vert \frac{an^{-\alpha} + bm^{-\beta} + c_\infty  }{an^{-\alpha} + bm^{-\beta} + c_\infty -i\eta} \right\Vert 
\end{equation}
where $i = \sqrt{-1}$. 
Here the simple pole at $ \eta$ controls the transition point from the initial random-guess level $\epsilon_0$ as $(m,n)$ increase. As $(m,n)$ grow,  $\tilde{\epsilon}\rightarrow c_\infty$ and the final irreducible error $\epsilon_\infty\triangleq \epsilon_0c_\infty\eta^{-1}$ is approached.
The random-guess error, $\epsilon_0$, is a known parameter determined by dataset statistics (e.g, $(N_{classes}-1) / N_{classes}$ for a balanced  dataset). 
Note that due to our choice of rational envelope, we can  divide by a constant the form in  \eqref{eq:norm_err}. Without loss of generality, let us choose $a=1$.

Note that while the forms in equations \ref{eq:consitent_form} and \ref{eq:norm_err} are well motivated, the approach taken for modeling the transition is solely a convenience one. In fact,   the transition(s) as function of $m$ and $n$ may be captured in the functional forms of $a,b,\alpha,\beta$ or another envelope mechanism. 
We leave a more refined investigation of the nature of the transitions to future work. 

\section{Error landscape estimation - fit results} \label{sec:fit}

\begin{figure}[t]
\centering
\begin{subfigure}[t]{0.4\textwidth}
    \centering 
        \includegraphics[width=\linewidth,trim={0 0 0 0.7cm},clip]{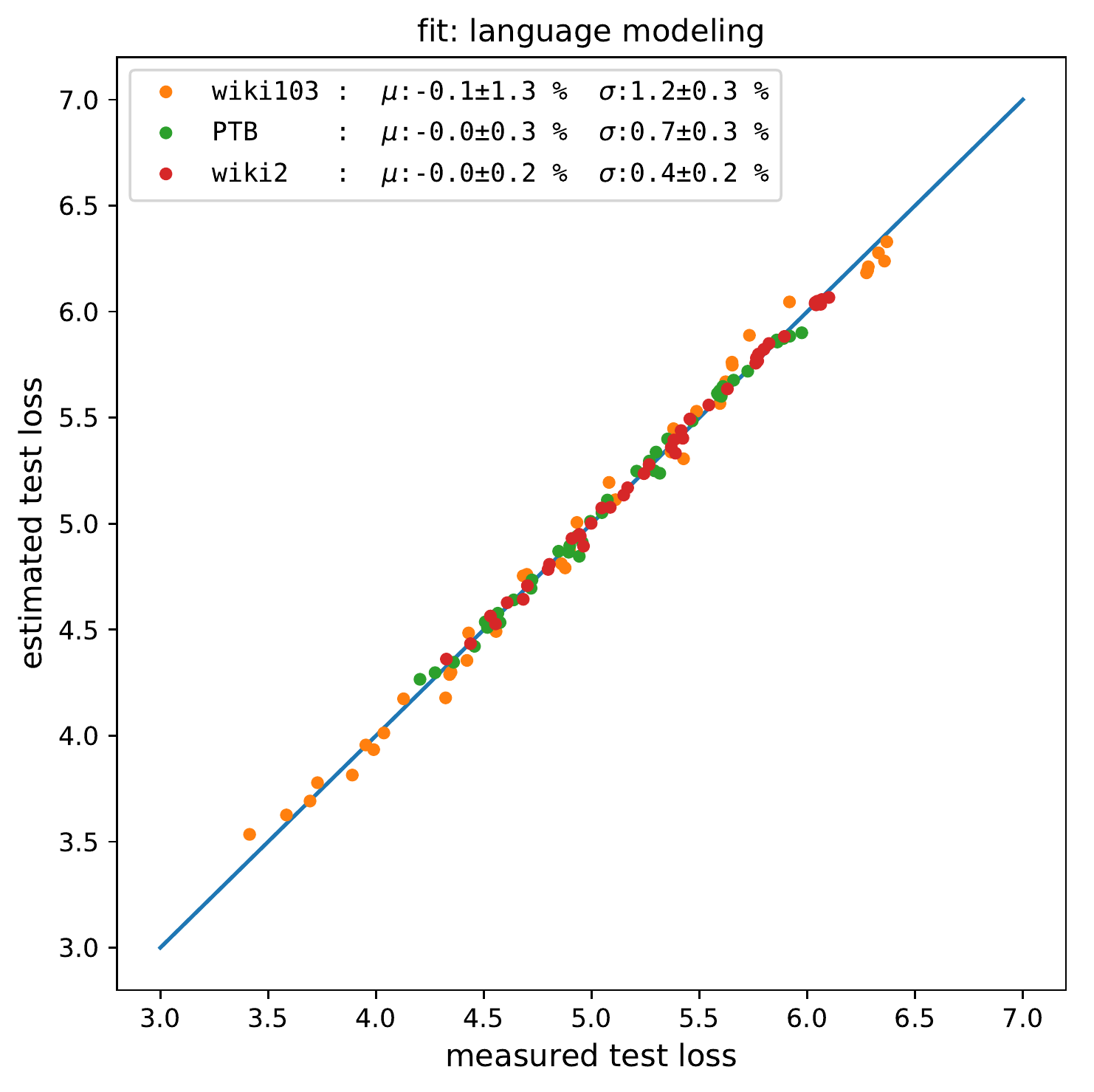}
    \caption{ Estimated  vs.\ actual cross-entropy loss for various language modeling datasets. 
    }
    \label{fig:fit-language}
\end{subfigure} \hspace{3em} 
\begin{subfigure}[t]{0.4\textwidth}
    \centering 
        \includegraphics[width=\linewidth,trim={0 0 0 0.7cm},clip]{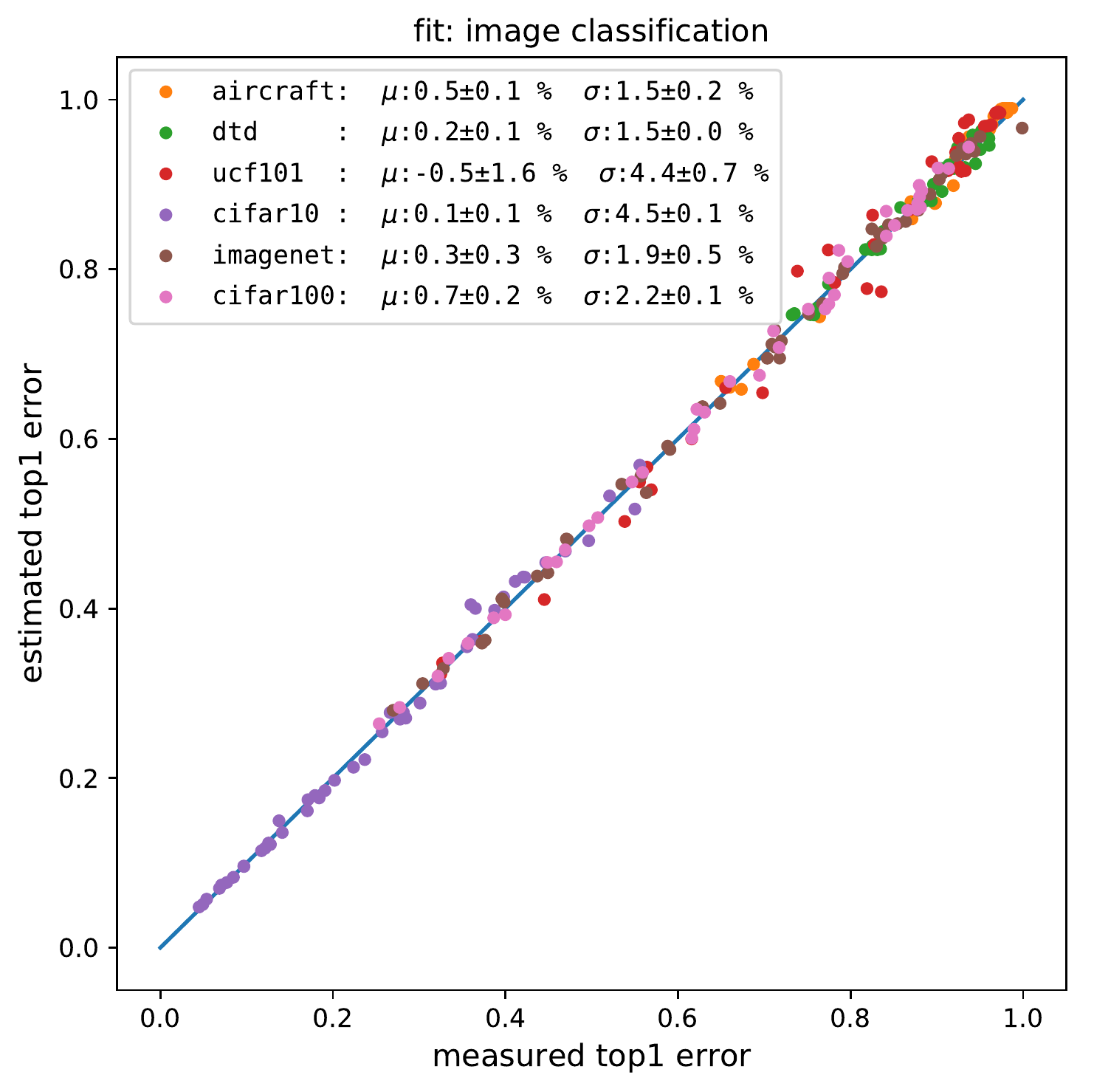}
    \caption{Estimated  vs.\  actual test error for various image classification datasets. 
    }
    \label{fig:fit-vision}
\end{subfigure}
\caption{Error estimation results, using 10-fold cross-validation
on all configurations in each dataset. 
For reference, in blue is the identity line.
The legend shows mean $\mu$ and standard deviation  $\sigma$ of the divergence $\delta$ ($\pm$ one std). 
See Appendix \ref{app:landscape} for the actual and estimated landscapes in each dataset.
}
\label{fig:fit}
\end{figure} 

We wish to empirically estimate the quality of the proposed functional parameterization as a fit to the true error landscape. 
Let $\hat{\epsilon}(n,m ; \vtheta)$ be the parametric function family (\eqref{eq:envelope}) approximating the error landscape  $\epsilon \left(n,m \right)$, where $\vtheta = \{\alpha,\beta,b,c_\infty,\eta\}$.\footnote{
For image classification, we set $\epsilon_0 = (N_{classes}-1) / N_{classes}$ (the balanced dataset case). For language modeling, we estimate $\epsilon_0$ as another parameter, such that $\vtheta = \{\alpha,\beta,b,c_\infty,\eta,\epsilon_0\}$ in this case. }
Define the divergence $\delta(n,m;\vtheta)$ as  the  relative difference between the estimated error $\hat{\epsilon}(m,n;\vtheta)$ and the true error $\epsilon(m,n)$:
\begin{equation*}
    \delta(n,m;\vtheta) \triangleq \frac{\hat{\epsilon}(m,n;\vtheta)-\epsilon(m,n)}{\epsilon(m,n)} 
\end{equation*}
We fit a least squares regression model to find the best parameters minimizing the divergence.
In this section, we fit the function 
using 10-fold cross-validation across all model/data configurations $m , n$ (see Table \ref{tab:stats}) and evaluate the fit quality.  (In the next section, we perform \emph{extrapolation} experiments, from seen to unseen points.) 
We perform the fit separately for each dataset and evaluate its quality by 
the mean $\mu$ and standard deviation $\sigma$ of the 
divergence $\delta$ over all points $(m,n)$. 
See Appendix~\ref{app:fit-exp} for experimental details.
%
%

As \figref{fig:fit} shows,  estimated test accuracy is highly correlated with  actual test accuracy for various datasets, with worst-case values $\mu<1\%$ and $\sigma<5\%$ . Note that the number of free parameters 
is small ($|\vtheta|\leq 6$) compared to the number of  points (\mbox{42--49} model-data configurations),
demonstrating the appropriateness of the proposed function 
for modeling the complex error landscape. 


\section{A Probe into Depth Scaling} \label{sec:width_and_depth_scaling}

\begin{figure}[t]
\centering
\begin{subfigure}[t]{0.35\textwidth}
    \centering 
        \includegraphics[width=\linewidth,trim={0 0 0 1.3cm},clip]{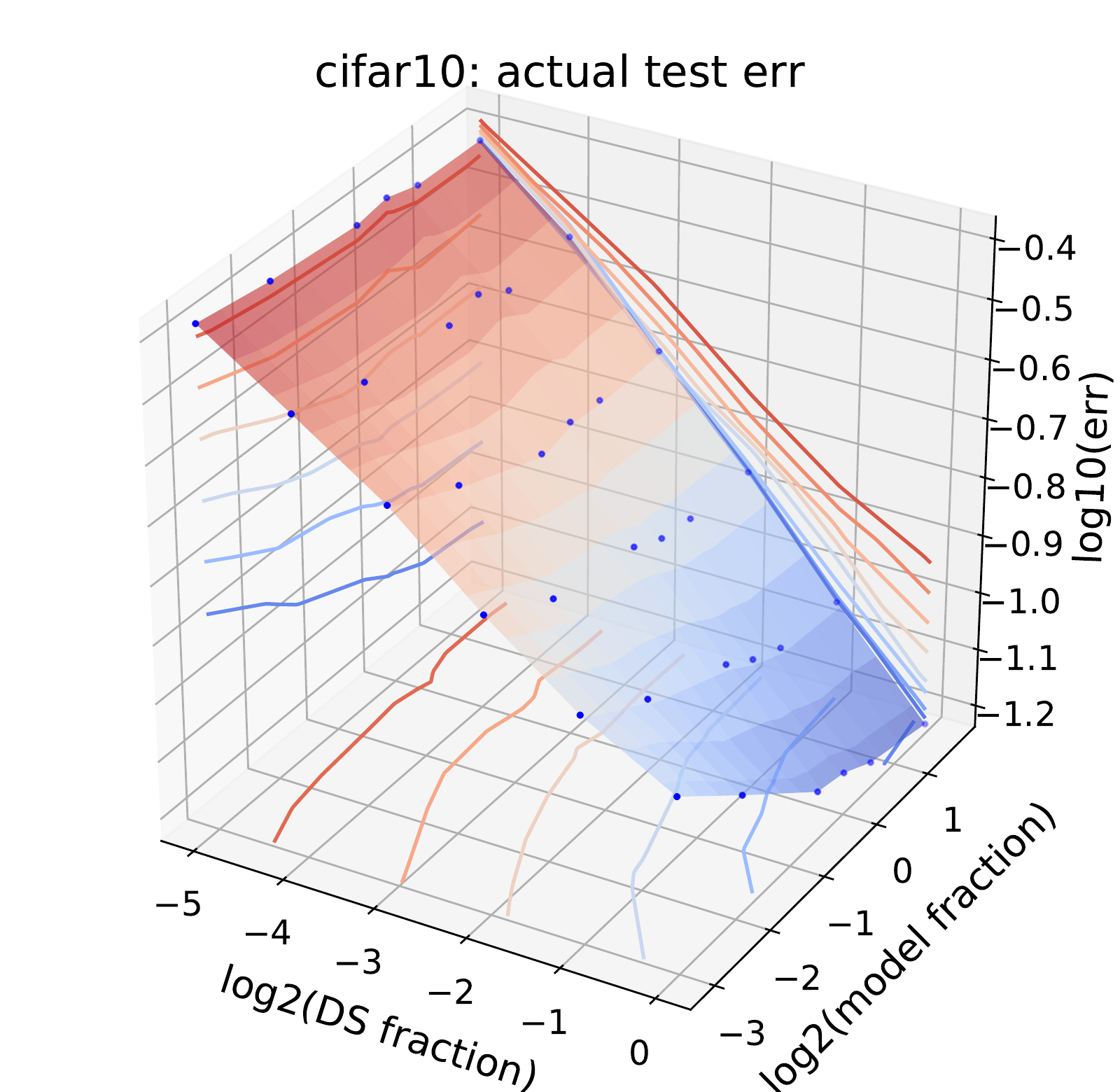}
    \caption{Error landscape when scaling depth (at constant baseline width).}
    \label{fig:cifar10-depth}
\end{subfigure} \hspace{5pt} 
\begin{subfigure}[t]{0.29\textwidth}
    \centering 
        \includegraphics[width=\linewidth,trim={0 0 0 0.9cm},clip]{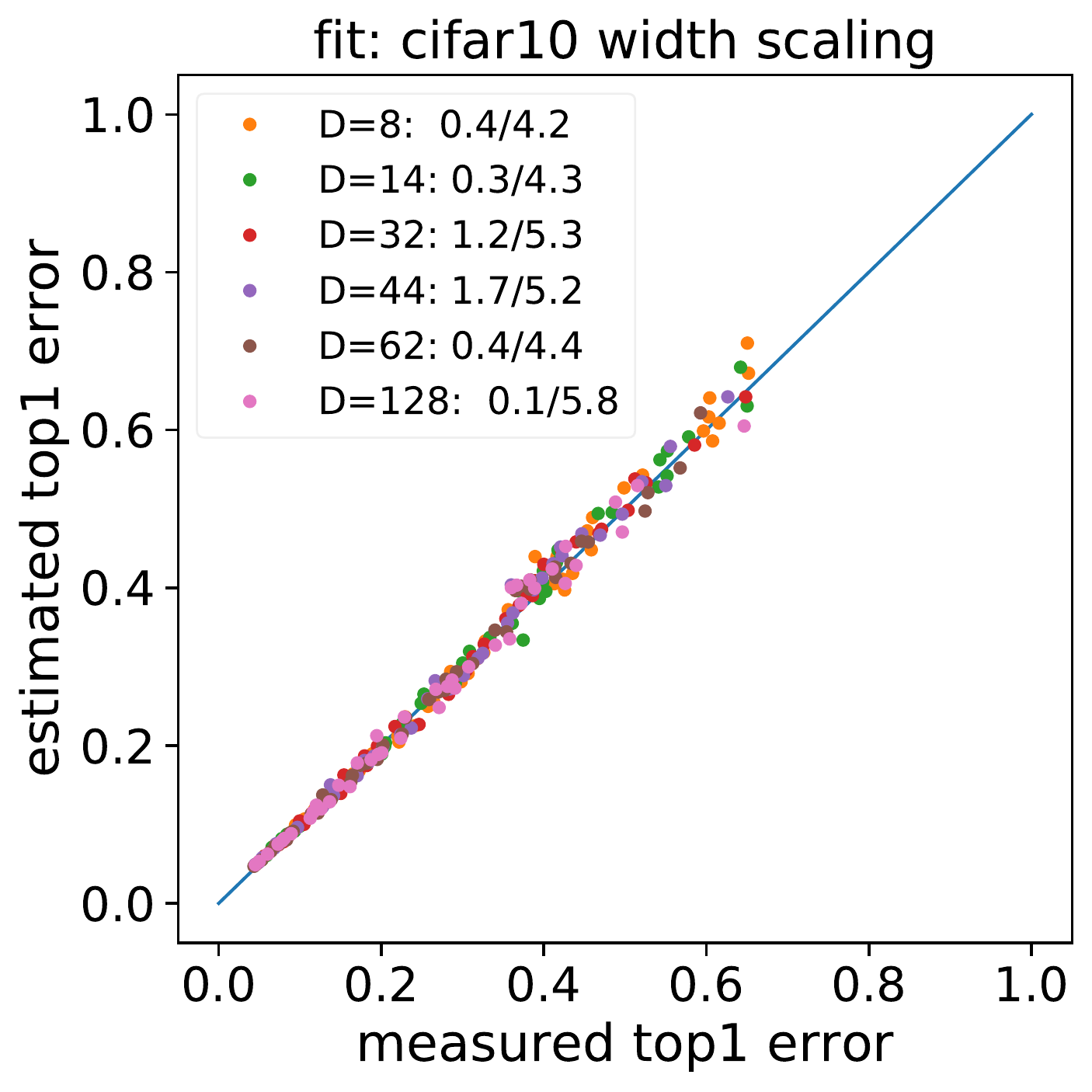} 
    \caption{Width scaling fit at different constant depths (D). }
    \label{fig:fit-cifar10-width}
\end{subfigure} \hspace{5pt}
\begin{subfigure}[t]{0.29\textwidth}
    \centering 
        \includegraphics[width=\linewidth,trim={0 0 0 0.9cm},clip]{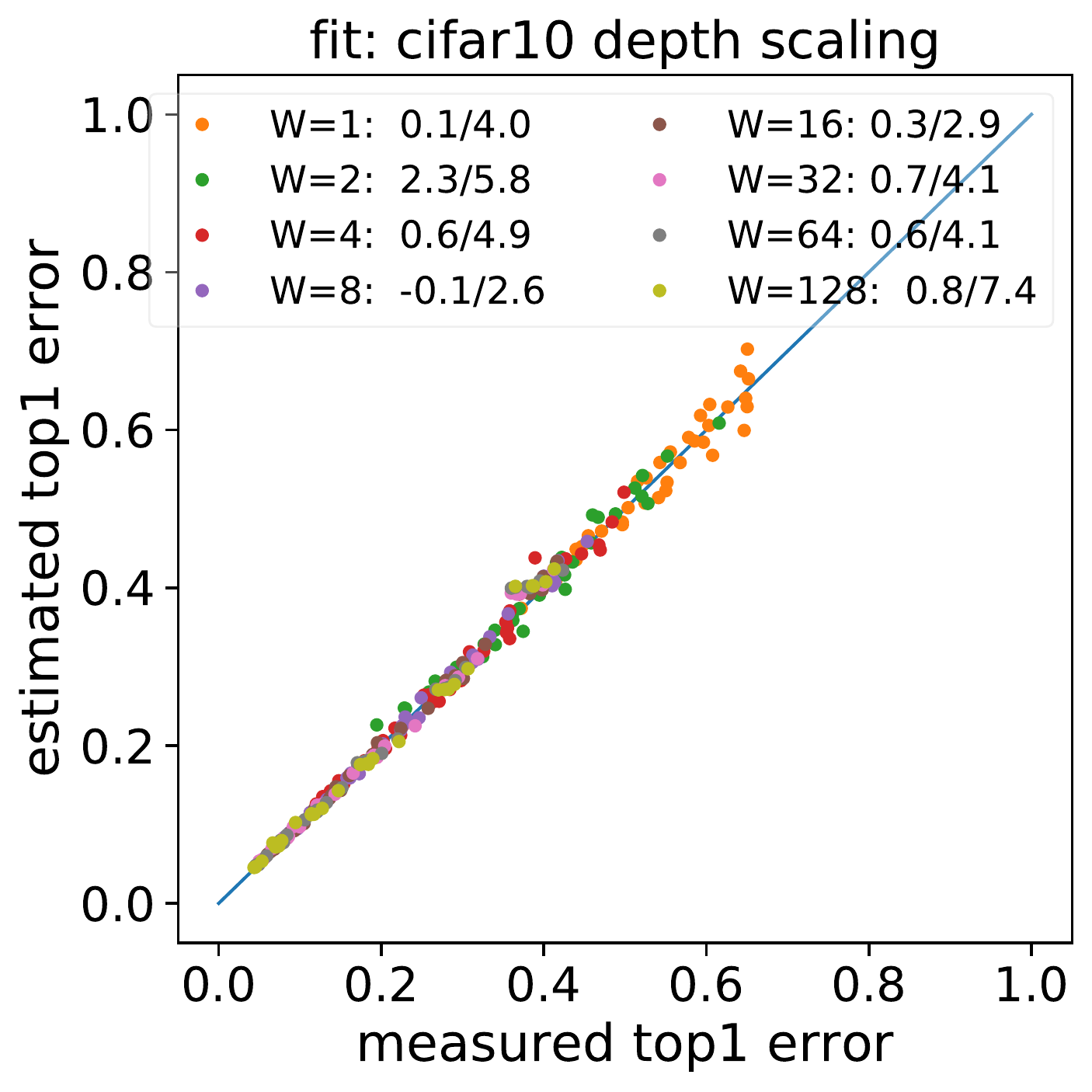}
    \caption{Depth scaling fit at different constant widths (W). }
    \label{fig:fit-cifar10-depth}
\end{subfigure}
\caption{Error landscape estimation results on CIFAR10 for width and depth scaling, showing small and comparable fit errors in both cases. Numbers in legends denote mean/variance of the estimation divergence. }
\label{fig:fit_cifar_width_depth}
\end{figure}

Here we verify that our results extend to another canonical scaling policy, namely depth scaling. 
\Figref{fig:cifar10-depth} shows the error landscape with depth scaling on  CIFAR10, exhibiting the same characteristics as width scaling. 
\Twofigref{fig:fit-cifar10-width}{fig:fit-cifar10-depth} show error landscape estimation results for both cases of width and depth scaling,
exhibiting small and comparable fit errors  (confidence intervals $<3\%$).

Since the difference in approximation quality is effectively indistinguishable  when scaling   depth or width orthogonally, we expect compound scaling to adhere to the same functional form. Indeed, we verified this on the publicly available (model scaling only) results for EfficientNet \citep{tan2019efficientnet}.

More fundamentally, these results demonstrate that indeed the error is not dependent on the number of parameters per se, but rather on the structural degrees of freedom - i.e. width and depth - seperately.

\clearpage


\section{On the Variability  of Optimizers and Architectures}
\label{sec:arch_optim_var}



Our study covers a deliberate variety of architectures (ResNet, WRN, LSTM, Transformer) and optimizers (Adam, SGD variants), following standard implementations in the literature as recommended for each dataset/model setting; 
see Appendix~\ref{app:data-models}. 
\begin{figure}[t]
  \centering
    \includegraphics[width=0.5\linewidth,trim={0 0 0 0.7cm},clip]{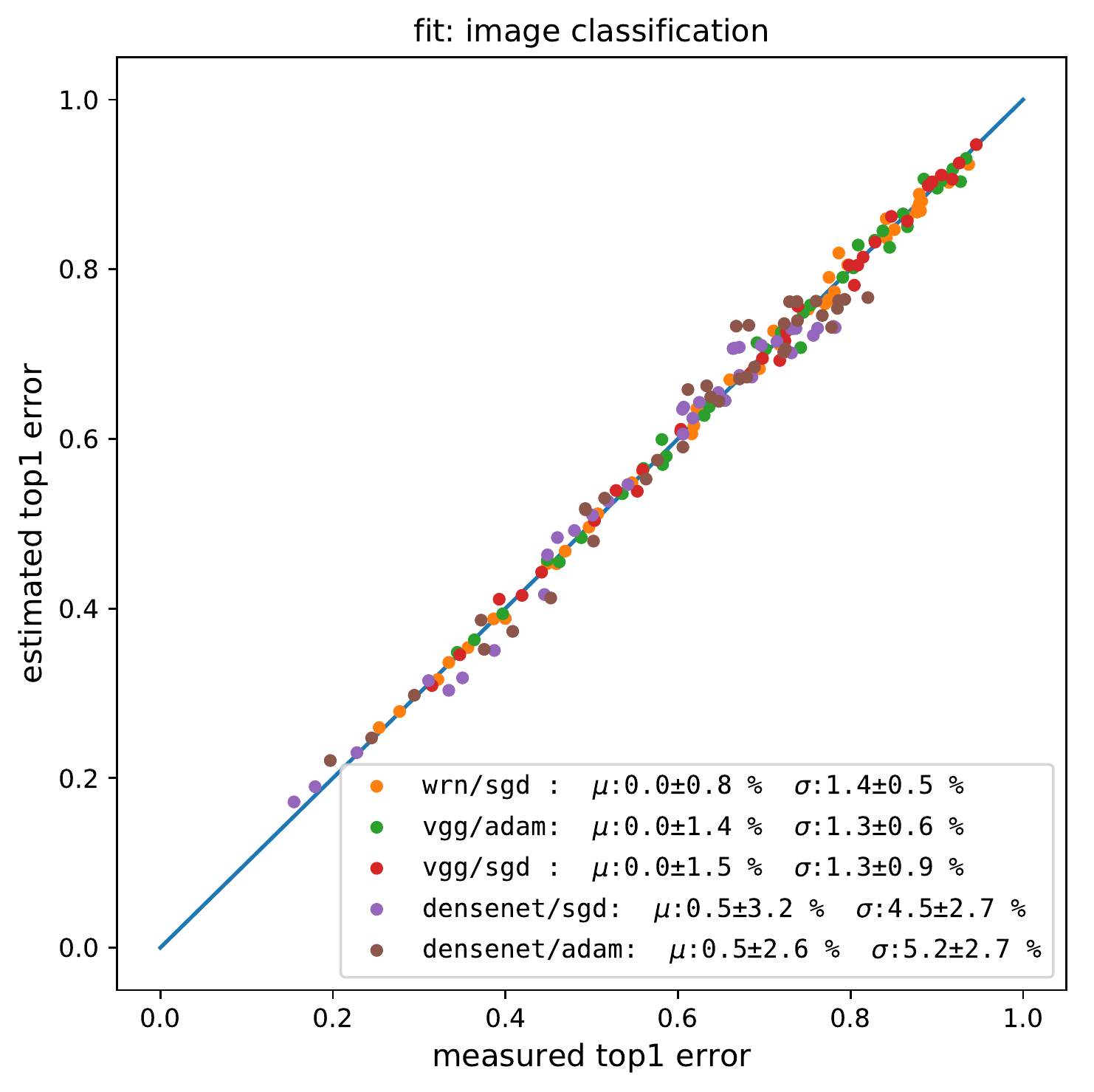}
    \caption{CIFAR100 Error estimation results with three architectures (WRN, VGG,  DenseNet) and two optimizers (SGD, Adam). 
    }
    \label{fig:fit-cifar100_arch_optim}
\end{figure}
However, the model/optimizer settings differ in multiple aspects across the different tasks
, rendering the comparison of, say, different optimizers, challenging.
In this section we verify that the functional form holds when varying the optimizer and/or the architecture on the same task, namely image classification on CIFAR100.

In addition to the previously examined setting of WRN with SGD, we add four more settings: two
 well known architectures (VGG and DenseNet), each trained with both SGD and Adam optimizers. 
 See Appendix \ref{app:data-models} for  experimental details. 
Figure~\ref{fig:fit-cifar100_arch_optim} exhibits consistent, accurate, fit values across all architecture/optimizer settings, with mean divergence of $\mu <1\%$ (std: $\sigma <
 6\%$; confidence intervals $<4\%$).

\clearpage 
\section{Extrapolation results} \label{sec:extrapolation}

\begin{figure}[t]
\centering
\begin{subfigure}[b]{0.23\textwidth}
\centering
\label{fig_sub:extrap_array}
\includegraphics[width=\linewidth,trim={0 0 0 0.7cm},clip]{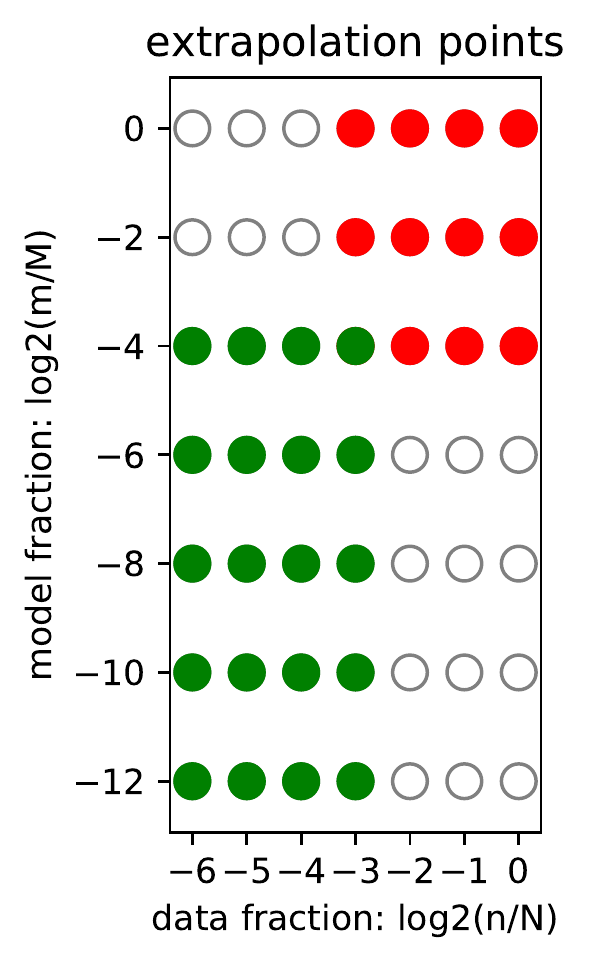}
\caption{Illustration. }
\label{fig:extrapolation-array}
\end{subfigure}
\begin{subfigure}[b]{0.37\textwidth}
\centering
\label{fig_sub:extrp_imgnet_m16n8}
\includegraphics[width=\linewidth,trim={0 0 0 0.7cm},clip]{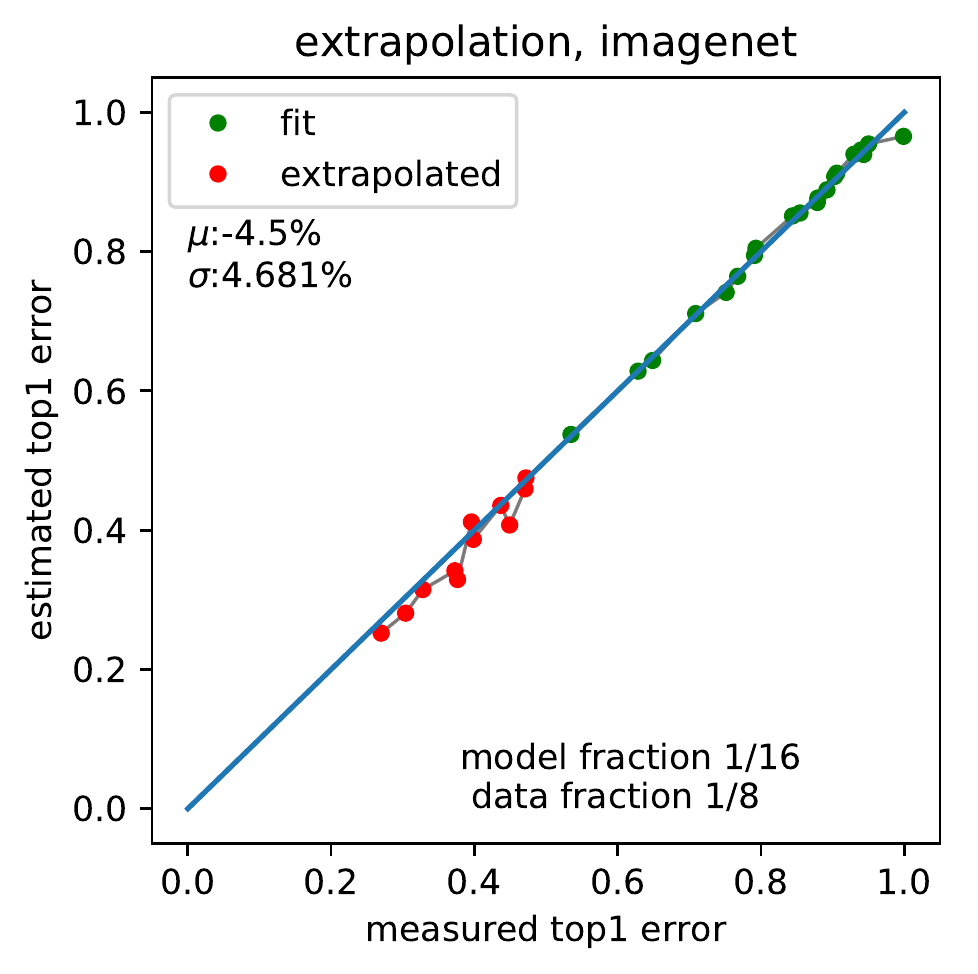}
\caption{Extrapolation on ImageNet}
\label{fig:extrapolation-single-vision}
\end{subfigure}
\begin{subfigure}[b]{0.37\textwidth}
\centering
\label{fig_sub:extrp_wiki103_m16n8}
\includegraphics[width=\linewidth,trim={0 0 0 0.7cm},clip]{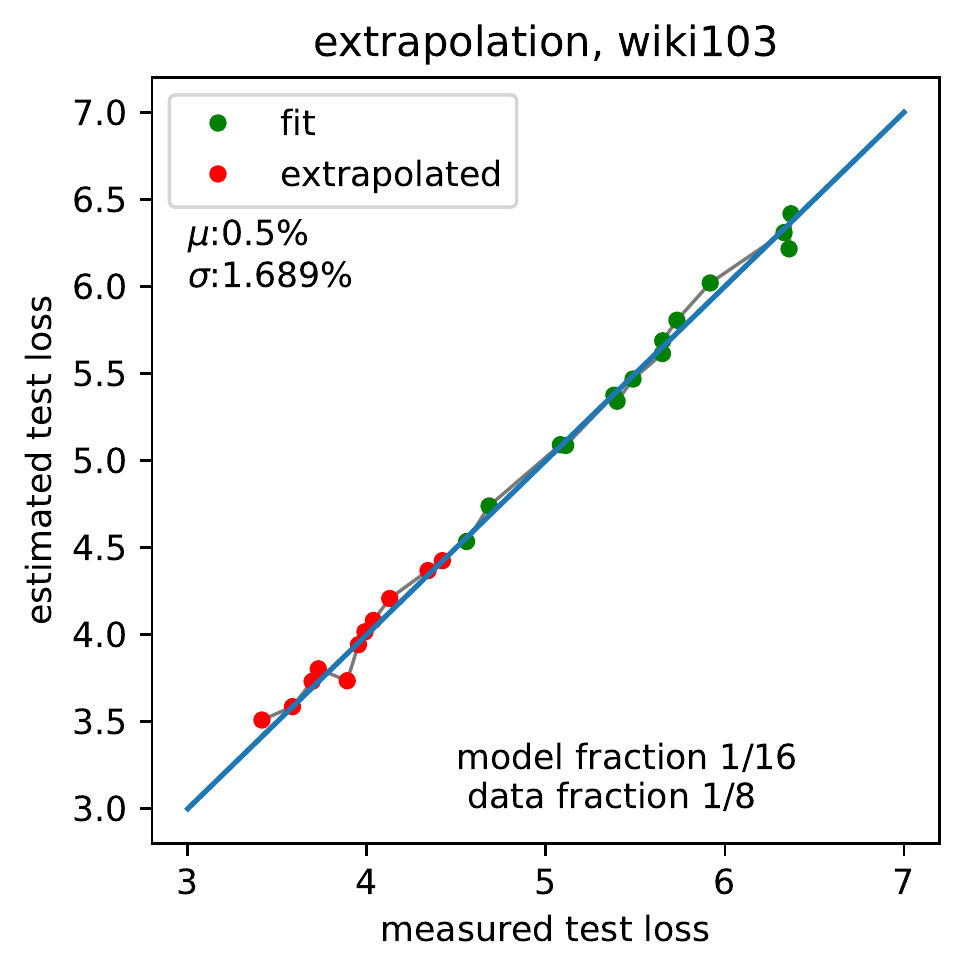}
\caption{Extrapolation on WikiText-103.}
\label{fig:extrapolation-single-language}
\end{subfigure}
\caption{Extrapolation results. (a) Illustration of the extrapolation setup, where we fit on a subset of the points (in green) and predict on larger points (in red). (b) and (c) show example results on one configuration in two benchmark datasets.
Comprehensive results are given in Appendix~\ref{app:extrapolations}. }
\label{fig:extrapolation}
\end{figure}

In this section, we evaluate the ability of our functional approximation to extrapolate beyond seen model/data configurations. The primary question we ask is: can we predict the error of a large model/data configuration from the errors of smaller-scale model/data configurations? To do this, we fit the least squares regression on a subset of the configurations and predict the error on larger, unseen configurations. 
More formally, let $(m_i, n_j)$ denote a given model/data configuration. We first estimate parameters $\vtheta_{ij}$ by fitting the function in \eqref{eq:envelope} on all points of at most that size ($m \leq m_i, n \leq n_j$).  
Then we predict the  error $\epsilon(m,n)$ in all points corresponding to larger configurations ($m > m_i, n > n_j$) using estimated $\vtheta_{ij}$.  
Finally, we measure the divergence $\delta(m,n)$ between the estimated  error and the 
actual  error at all larger configurations. 
This process is illustrated in \figref{fig:extrapolation-array}.

\Figref{fig:extrapolation-single-vision} shows the results of one such extrapolation experiment, on ImageNet. In this case, we have fit the functional form on all configurations of model size $m \le m_i = M/16 $ and data size $n \le n_j = N/8$, and predicted the error on all larger configurations. 
As the figure shows, the extrapolation is highly accurate, with a mean divergence of $\mu=4.5\%$ (std: $\sigma=4.7\%$). \Figref{fig:extrapolation-single-language} reports a similar experiment on WikiText-103. Here, again, we see very good extrapolation, with a mean divergence of $\mu=0.5\%$ (std: $\sigma=1.7\%$).
Note that each extrapolation is run 10 times with different random initializations of $\vtheta_{ij}$ in the least squares with negligible effect on the prediction. 

In practice, we may be interested in  extrapolation quality with different subsets of configurations. 
Appendix \ref{app:extrapolations} provides detailed extrapolation results on multiple subsets of configurations, for both vision and language datasets. 
Generally,  the extrapolation performs well once not ill-posed, which may be caused by lack of signal in the region of the initial ``random-guess'' level, or in degenerate  cases like having fewer measurements than the number of free parameters in $\vtheta$. 

\section{Discussion} \label{sec:discussion}

In this work, through insights gained by the joint examination of the dependencies of generalization error on both model and data size, we arrive at criteria for functions consistent with the form of the generalization error under a given scaling policy. We consider one such function and find it to be in very good agreement with the actual behavior of the error landscape. Indeed, the agreement is strong enough  that \textit{extrapolation} 
from small to large scale  becomes feasible: the function  predicts the behavior of the generalization error in practice for the practical case of scaling models and  data.
We discuss several example implications of knowing such a functional form. 

\paragraph{Small-scale network development:} 
    At the core of small fidelity searches is the notion of \textit{performance rank} comparison between models. However, small scale and large scale ranks are not assured to be consistent.
    If indeed a functional form such as empirically found in this work 
    holds very generally, then in contrast, one can safely assess \textit{scaling rank} between models at small scale, with the assurance that it remains consistent.
    This suggests that one would be well served by searching over scaling policies; a pertinent example of such a success is \cite{tan2019efficientnet}. The functional form also explains the limitation of small-scale search: once reaching the random-guess  error level, where the sensitivity to scaling vanishes,  the informativeness of ranking diminishes. Finally, the functional form allows direct usage of differentiable methods for NAS.

\paragraph{Principled design:}
Knowing the error landscape function facilitates reasoning about the choice of $(m,n)$ attaining a specified error level. In other words, for any given error level, one can solve Eq.~\ref{eq:envelope} for $m,n$ based on small-scale measurements. Thus, one can quantitatively answer design questions regarding the expected (in particular, large-scale) relations between $m$, $n$, and $\epsilon$.    
 In fact, Eq.~\ref{eq:envelope} provides direct ansewrs to questions such as "how much data would one require to reach a prescribed performance level?" or "how big a model would be needed?" 
Imposing constraints is also straightforward. For instance, consider the following question: "What is the maximal model size possibly needed (useful), when the data is limited in size, $n=n_{lim}$ (for a given model architecture and scaling policy)?"
%
For a fixed dataset size, model scaling eventually contributes marginally to error reduction and becomes negligible when $bm^{-\beta} \ll n_{lim}^{-\alpha}$ (Eq. \ref{eq:envelope}).
Define the relative contribution threshold $T$ as satisfying $ T = \frac{n_{lim}^{-\alpha} }{ bm_{max}^{-\beta}}$. (For example, $T=10$.) Then the maximal useful model size meeting threshold $T$ is:
\begin{equation*}
    m_{max}(T) = \left(bT\right)^{1/\beta} n_{lim}^{\alpha/\beta} 
\end{equation*}

Similarly, The maximal useful amount of data for a limited sized model $m_{lim}$ is:
\begin{equation*}
    n_{max}(T) = \left(1/bT\right)^{1/\alpha} m_{lim}^{\beta/\alpha} 
\end{equation*}


 Moreover, Eq. \ref{eq:envelope} allows for complex design trade-offs. Generally, given some design-tradeoff cost function $C(m,n,\epsilon)$, one can minimize such cost s.t.\ Eq.~\ref{eq:envelope}. 
For example, consider the case of optimizing for efficient  computation which has both practical and environmental importance \citep{schwartz2019green}. 
Since the number of FLOPs during training is $\propto m\cdot n$ (for constant epoch budget), the trade-off cost function may be formulated as $C(\textrm{FLOPS},\epsilon) = C(mn,\epsilon)$. Further, since constant error contour is very well approximated by $c = \frac{1}{n^\alpha}+\frac{b}{m^\beta}$ (Eq. \ref{eq:envelope}), dataset and models may be scaled  with optimal resource efficiency with no effect on performance by solving for:
\begin{equation*}
\argminB_{m,n} \quad   m\cdot n   \quad \qquad 
\text{s.t.} \quad  c=\frac{1}{n^\alpha}+\frac{b}{m^\beta} 
\end{equation*}
The solution gives us the optimal-computational-efficiency ratio of model to data size: $
\frac{b\beta}{\alpha}\frac{n^\alpha}{m^\beta} = 1$.

\paragraph{Limitations:}
    We have made a few simplifying assumptions in our choice of approximating function, in particular in how to model the transition from the initial random-guess  error level and the union of the random-guess level of the two scenarios (small model with large data  and large model with small data). We leave a more detailed examination of the behavior of the transitions from random-guess error levels and refinements of the functional form to  future work.

    Critically, the restrictive nature of our scaling framework (all parameters and hyperparameters described by a policy) is both a blessing and a challenge. The blessing comes in fulfilling the goal of finding simultaneously both the form of the generalization error and the full specification of the model and hyperparameters that attain it across scales. The challenge is that we have demonstrated in this work only the case of constant hyper-parameters. We conjecture that the relation between model configuration and hyperparameter choice \citep{zela2018towards} may entail the potential to formulate hyperparameter-scaling policies similar in nature to the model-scaling polices, and that these too fall under the scope of the form we find in this work. This too will be the subject of future work.

\chapter{Pruning Scaling Laws} \label{sec:Pruning}

\epigraph{\textit{``He can compress the most words into the smallest ideas of any man I ever met.''} \\ ---  Abraham Lincoln  }

\glsresetall
In the previous chapter we have demonstrated that scaling laws empirically exist tying jointly model degrees of freedom, dataset size and generalization error when considering dense models. 

Of key import in the quest for their understanding is the question of where/ when do they break. 
Examining pruning - the procedure of removing parts of the neural network - poses an interesting opportunity on three levels:
First, pruning is a very different setting from usual training, where by design there is a decoupling of the original model footprint and pruned model. It offers a chance to see if and what of the scaling laws found carries over or breaks between these settings which differ dramatically in the resultant models and a by design discrepancy in parameter count.
Second, the dependence of performance on sparsity is the subject of heightened interest in itself. For example, Is the level of sparsity in-itself important for the attainment of a specified performance in a specified budget?
Last but not least, uncovering laws governing the predictability and associated principled design of pruning methodology is of heightened practical interest. In particular, pruning is crucially important for the reduction of costs in the model inference context, which comprises the vast majority of the industry expenditures.

In this chapter - which is heavily based on our work as published in \cite{rosenfeld2020predictability} we examine this case of pruning and show what scaling laws govern it for state of the art iterative magnitude pruning methods. 

\section{Introduction}

For decades, neural network \emph{pruning}---eliminating unwanted parts of a network---has been a popular approach for reducing network sizes or the computational demands of inference \citep{optimalbraindamage,reed,han}.
In practice, pruning can reduce the parameter-counts of contemporary models by 2x \citep{bertpruning} to 5x \citep{Renda2020Comparing} with no increase in error.
More than 80 pruning techniques have been published in the past decade \citep{blalock2020state}, but, despite this enormous volume of research, there remains little guidance on important aspects of pruning.
Consider a seemingly simple question one might ask when using a particular pruning technique:

\textit{Given a family of neural networks (e.g., ResNets on ImageNet of various widths and depths), which family member should we prune (and by how much) to obtain the network with the smallest parameter-count such that error does not exceed some threshold $\epsilon_k$?}

As a first try, we could attempt to answer this question using brute force: 
we could prune every member of a network family (i.e., perform grid search over widths, depth, and pruned densities) and select the smallest pruned network that satisfies our constraint on error.
However, depending on the technique, pruning one network (let alone grid searching) could take days or weeks on expensive hardware.

If we want a more efficient alternative, we will need to make assumptions about pruned networks:
namely, that there is some \emph{structure} to the way that their error behaves.
For example, that pruning a particular network changes the error in a predictable way.
Or that changing its width or depth changes the error when pruning it in a predictable way.
We could then train a smaller number of networks, characterize this structure, and estimate the answer to our question.

In the context of standard training, Chapter \ref{sec:Dense}, such structure has been observed---and, further, codified explicitly---yielding insights and predictions in the form of scaling laws. 

We have reason to believe that such structure extends to certain pruning methods. Practically, there are already techniques that take advantage of it implicitly.
For example, \citet{cai2019once} create a single neural network architecture that can be scaled down to many different sizes;
to choose which subnetwork to deploy, \citeauthor{cai2019once} train an auxiliary, black-box neural network to predict subnetwork performance.
Although this black-box approach implies the existence of structure for this pruning method, it does not reveal this structure explicitly or make it possible to reason analytically in a fashion that could answer our research question.
More conjecturally, since under the lottery ticket hypothesis \cite{frankle2018lottery} there are subnetworks that if trained in the standard fashion converge to the same performance as if arrived by prunning the larger model, we can suspect that these subnetworks themselves (if found in advance) scale governed by the same dynamics as their dense counterparts.

Inspired by this work, we address our research question about pruning by finding a scaling law to predict the error of pruned networks.
We focus on a pruning method called \emph{iterative magnitude pruning (IMP)}, where weights with the lowest magnitudes are pruned in an unstructured fashion interspersed with re-training to recover accuracy.
This method is a standard way to prune \citep{han} that gets state-of-the-art tradeoffs between error and unstructured density \citep{gale, Renda2020Comparing}.
To the best of our knowledge, this is the first explicit scaling law that holds for pruned networks, let alone entire network families.

To formulate such a predictive scaling law, we consider the dependence of generalization error on the pruning-induced \emph{density} for networks of different depths and widths trained on different dataset sizes.
We begin by developing a functional form that accurately estimates the generalization error of a specific model as it is pruned (Section \ref{sec:single-network}).
We then account for other architectural degrees of freedom, expanding the functional form for pruning into a scaling law that jointly considers density alongside width, depth, and dataset size (Section \ref{sec:joint}).
The basis for this joint scaling law is an \emph{invariant} we uncover that describes ways that we can interchange depth, width, and pruning without affecting error.
The result is a scaling law that accurately predicts the error of pruned networks across scales.
And, now that we have established this functional form, fitting it requires only a small amount of data (Section \ref{app:interpolation}), making it efficient to use on new architectures and datasets (Appendix \ref{app:more_arch_alg}).
Finally, we use this scaling law to answer our motivating question (Section \ref{sec:conclusions}).

\section{Experimental Setup}
\label{sec:setup}
\textbf{Pruning.}
We study \emph{iterative magnitude pruning} (IMP) \citep{jankowsky, han}.
IMP prunes by removing a fraction---typically 20\%, as we do here---of individual weights with the lowest magnitudes in an unstructured fashion at the end of training.%
\footnote{We do not prune biases or BatchNorm, so pruning 20\% of weights prunes fewer than 20\% of parameters.}
We choose these weights globally throughout the network, i.e., without regard to specific layers.
We use per-weight magnitude pruning because it is generic, well-studied \citep{han}, and produces state-of-the-art tradeoffs between density and error \citep{gale, blalock2020state, Renda2020Comparing}.

Pruning weights typically increases the error of the trained network, so it is standard practice to further train after pruning to reduce error.
For IMP, we use a practice called \emph{weight rewinding} \citep{frankle2020linear, Renda2020Comparing}, in which the values of unpruned weights are \emph{rewound} to their values earlier in training (in our case, epoch 10) and the training process is repeated from there to completion.
To achieve density levels below 80\%, this process is repeated \emph{iteratively}---pruning by 20\%, rewinding, and retraining---until a desired density level is reached.
For a formal statement of this pruning algorithm, see Appendix \ref{app:pruningalg}.

\textbf{Datasets.}
In the main body of the paper, we study the image classification tasks CIFAR-10 and ImageNet.
Our scaling law predicts the error when training with the entire dataset and smaller \emph{subsamples}.
We include subsampling because it provides a cost-effective way to collect some of the data for fitting our functional form.
To subsample a dataset to a size of $n$, we randomly select $n$ of the training examples without regard to individual classes such that in expectation we preserve the original dataset distribution (we always retain the entire test set).
When performing iterative pruning, we maintain the same subsample for all pruning iterations.
We consider other datasets in Appendix \ref{app:more_arch_alg}.

\textbf{Networks.}
In the main body of the paper, we study ResNets for CIFAR-10 and ImageNet.\footnote{See Appendix \ref{app:resnets} for full details on architectures and hyperparameters. Note that CIFAR-10 and ImageNet ResNets have different architectures \citep{he2016deep}.}
We develop a scaling law that predicts the error (when pruned) of an entire \emph{family} of networks with varying widths and---in the case of the CIFAR-10 ResNets---depths.
To vary width, we multiply the number of channels in each layer by a \emph{width scaling factor}.
To vary depth of the CIFAR-10 ResNets, we vary the number of residual blocks.
We refer to a network by its depth $l$ (the number of layers in the network, not counting skip connections) and its width scaling factor $w$.
We consider other networks in Appendix \ref{app:more_arch_alg}.

\begin{table*}
\begin{center}

{\scriptsize
\begin{tabular}{@{\ }l@{\ }|@{\ }c@{\ \ }c@{\ \ }c@{\ \ }c@{\ \ }c@{\ \ }c}
\toprule
Network Family & $N_{\text{train}}$ & $N_{\text{test}}$ & Densities ($d$) & Depths ($l$) & Width Scalings ($w$) & Subsample Sizes ($n$) \\ \midrule
CIFAR-10 ResNet & 50K & 10K & $0.8^i$, $i \in \{0, \ldots, 40\}$ & 8, 14, 20, 26, 50, 98 & $2^i, i \in \{-4, \ldots, 2\}$ & $\frac{N}{i}$, $i \in \{1, 2, 4, 8, 16, 32, 64\}$ \\
ImageNet ResNet & 1.28M & 50K & $0.8^i, i \in \{0, \ldots, 30\}$ & 50 &  $2^i, i \in \{-4, \ldots, 0\}$ & $\frac{N}{i}$, $i \in \{1, 2, 4\}$ \\ 
\bottomrule
\end{tabular}
}

\end{center}
\vspace{-4.5mm}
\label{tab:dims}
\caption{The ranges of settings we consider in our experiments in the main body of the paper. We consider all densities $d \in \{0.8^i~|~i \in \mathbb{N}_0\}$ where (1) the network is not disconnected and (2) the network does better than random chance; this value varies based on the configuration $(l, w, n)$ and the random seed. Where neither of these conditions apply, we cap $i$ at 40 (CIFAR-10 ResNets) or 30 (ImageNet ResNets).
We consider all configurations of $(l, w, n)$ before which increasing depth or width of the unpruned network increases test error; configurations beyond this point are overly large for a given dataset subsample.
By this criterion, we use 152 configurations of $(l, w, n)$ for the CIFAR-10 ResNets and 15 for the ImageNet ResNets.
Taking into account all feasible densities, we use a total of 4,301 CIFAR-10 ResNet points and 274 ImageNet ResNet points.
Note that we use these configurations to find and evaluate the functional form. Once we do so, far fewer configurations are needed to fit the functional form for each setting (see Section \ref{app:interpolation}).}
\vspace{-4.5mm}
\end{table*}

\textbf{Notation and terminology.} Throughout the paper, we use the following notation and terminology:
\begin{itemize}[leftmargin=1.3em, topsep=0pt, itemsep=1pt, parsep=0.5pt, partopsep=-1pt]
    \item $\sD_N = \{  \vx_i,y_i \}_{i=1}^{N}$ is a labeled training set with $N$ examples. A \emph{subsample} of size $n$ is a subset of $\sD_N$ with $n$ examples selected uniformly at random.
    \item $l$ and $w$ are, respectively, the depth (i.e., the number of layers, excluding skip connections) and the width scaling factor of a particular network.
    \item Networks that vary by width and depth are a \emph{family}.
    \item $d$ is the \emph{density} of a pruned network (i.e., the fraction of weights that have not been pruned).
    \item $\epsilon \left(d, l, w, n\right)$ is the test error of a network with the specified density, depth, width scaling, and dataset size.
    \item $\epsilon_{np}\left(l, w, n\right) = \epsilon \left(1, l, w, n\right)$ is the test error of the unpruned network with the specified depth, width scaling, and dataset size. When clear from context, we omit $(w,l,n)$ and write $\epsilon_{np}$.
    \item $\hat{\epsilon}(\epsilon_{np}, d \mid l, w, n)$ is an estimate of the error of a pruned model for a scaling law that has been fitted to a specific network with the specified depth, width scaling, and dataset size (Section \ref{sec:single-network}).
    \item $\hat\epsilon \left(\epsilon_{np}, d, l, w, n\right)$ is an estimate of the error of a pruned model with the specified depth, width scaling, and dataset size for a scaling law that has been fitted to a network family (Section \ref{sec:joint}).
\end{itemize}
\vspace{-5pt}

\textbf{Dimensions.}
In developing our scaling laws, we vary four dimensions: dataset subsample size ($n$)
and network degrees of freedom density ($d$), network depth ($l$), and width scaling factor ($w$).
In the main body of the paper, we consider the ranges of these values as specified in Table 1.
See the caption in Table 1 for full details.
We train three replicates of each CIFAR-10 configuration with different seeds.

\section{Modeling the Error of a Pruned Network}
\label{sec:single-network}

Our goal in this section is to develop a functional form that models the error of a member of a network family as it is pruned (using IMP) based on its unpruned error $\smash{\epsilon_{np}(w,l,n)}$.
In other words, we wish to find a function $\smash{\hat\epsilon(\epsilon_{np}, d~|~l, w, n)}$ that predicts the error at each density $d$ for a network with a specific depth $l$, width scaling factor $w$, and dataset size $n$.

\begin{figure}
\vspace{-1mm}
\centering
\begin{minipage}{0.325\textwidth}
\includegraphics[width=\linewidth,trim={0.2cm 0 0.1cm 0.65cm},clip]{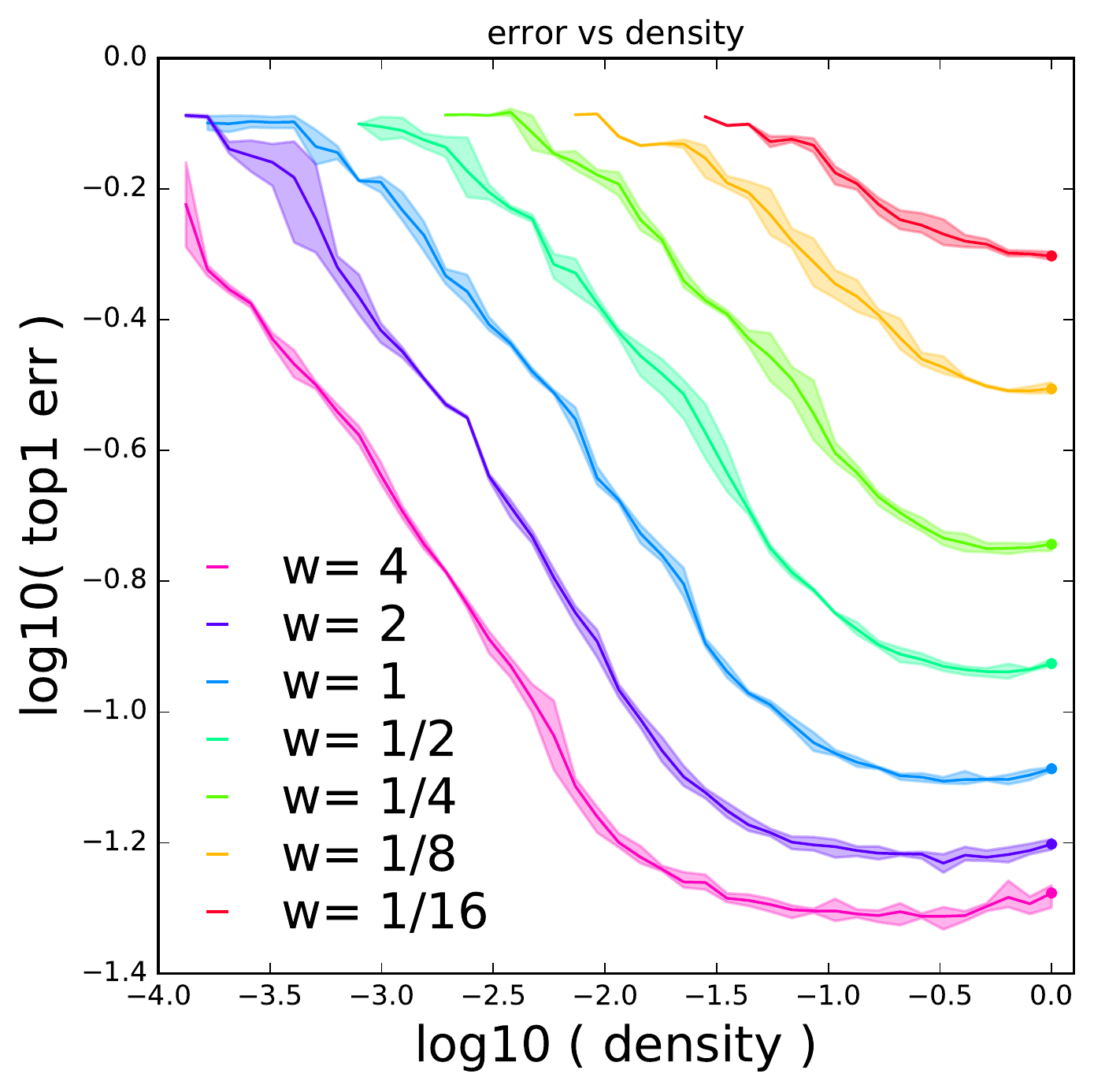}%
\end{minipage}%
\begin{minipage}{0.325\textwidth}
\includegraphics[width=\linewidth,trim={10.8cm 2.1cm 10.9cm 5.65cm},clip]{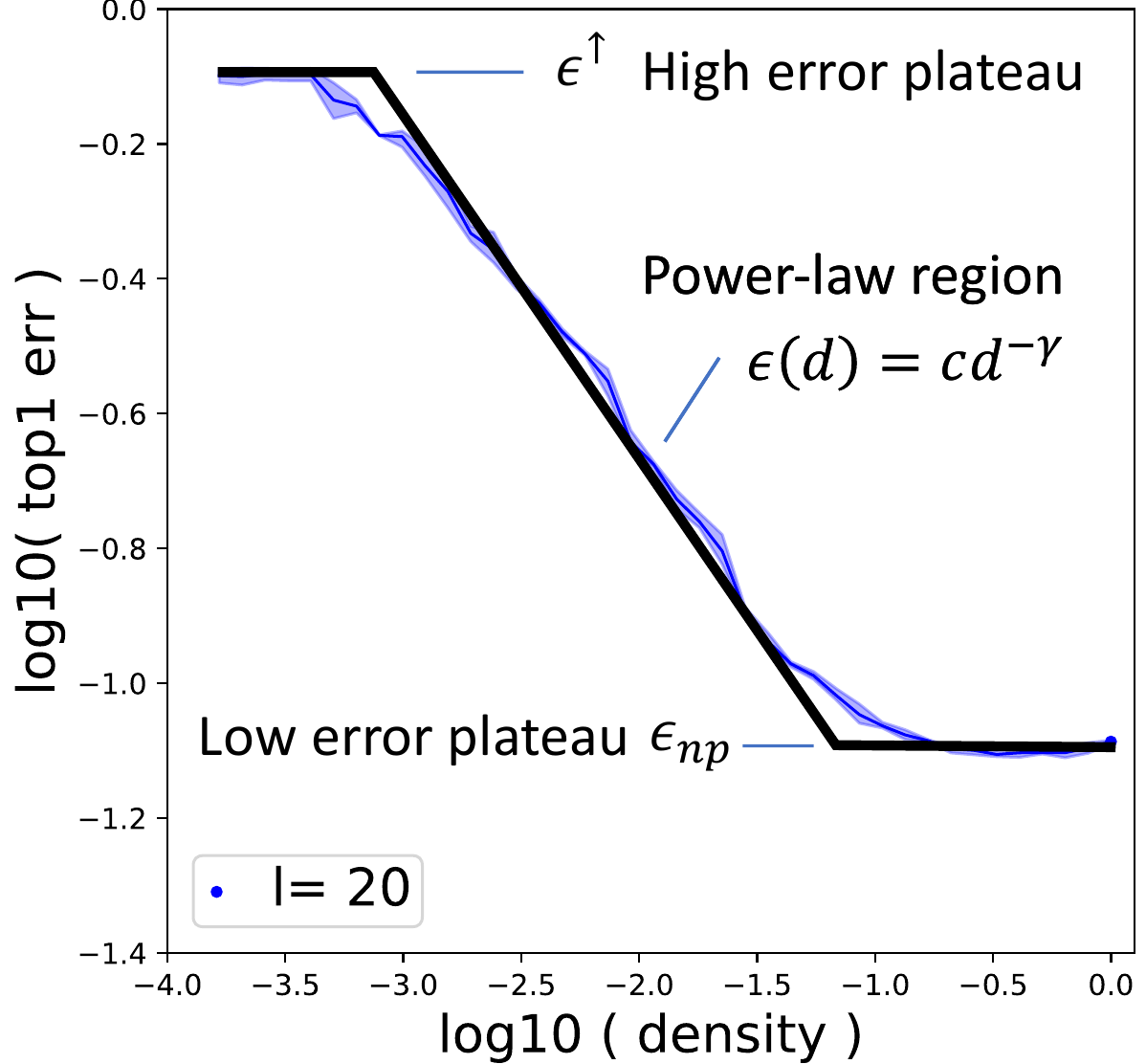}%
\end{minipage}%
\begin{minipage}{0.32\textwidth}
\includegraphics[width=\linewidth,trim={10.8cm 1.9cm 11cm 5.6cm},clip]{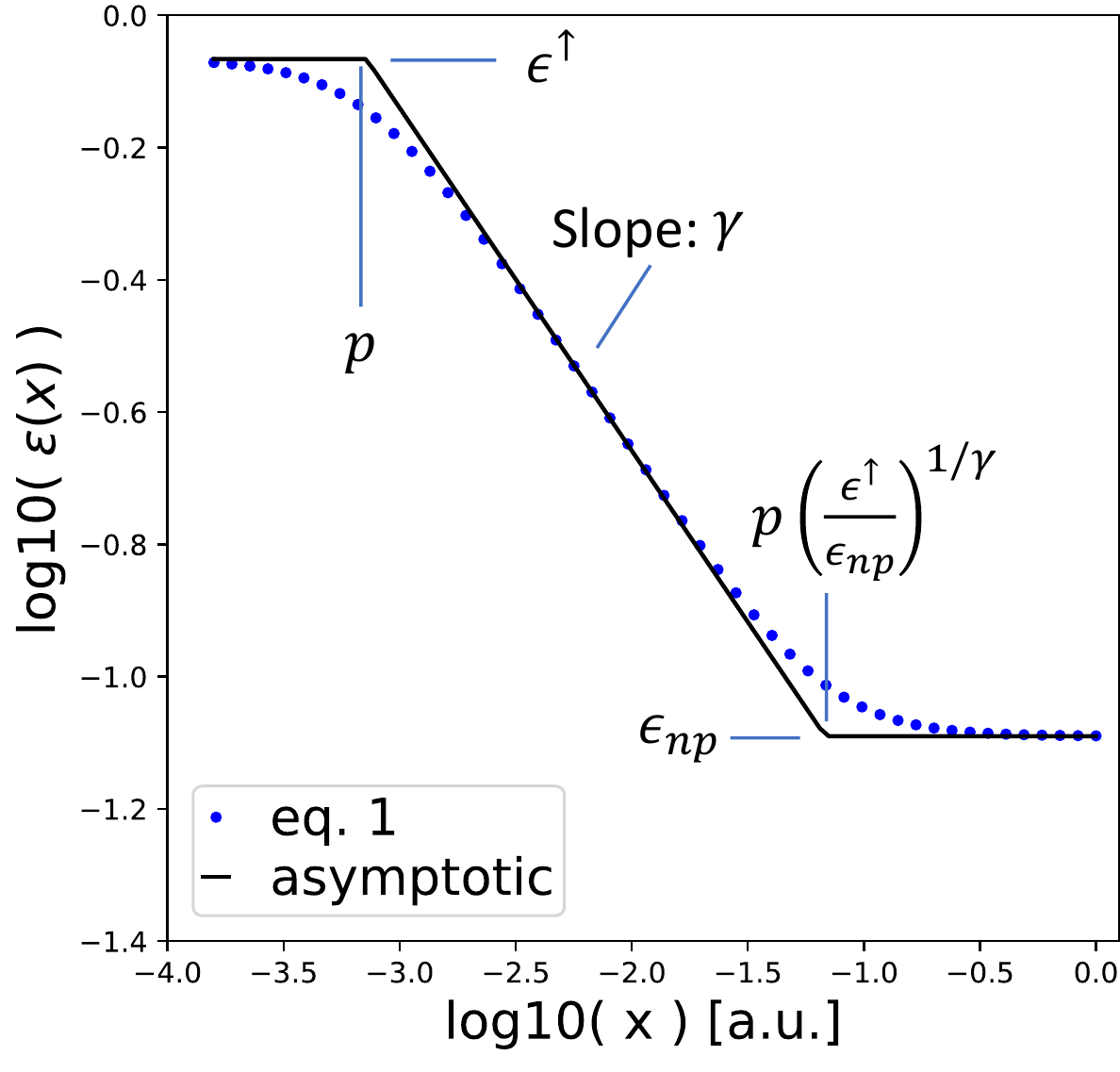}%
\end{minipage}
\vspace{-2mm}
\caption{Relationship between density and error when pruning CIFAR-10 ResNets; $w$ varies, $l=20$, $n=N$ (left).
Low-error plateau, power-law region, and high-error plateau for CIFAR-10 ResNet $l=20$, $w=1$, $n=N$ (center).
Visualizing Equation \ref{eq:prune_density} and the roles of free parameters (right).}
\vspace{-2mm}
\label{fig:typical_prune_curve}
\end{figure}

\textbf{Intuition.}
Since IMP prunes 20\% at a time, it produces pruned networks at intermediate densities $\smash{d_k = 0.8^k}$ in the process of pruning to density $\smash{d_K = 0.8^K}$.
In Figure \ref{fig:typical_prune_curve} (left), we plot the error of these pruned networks for CIFAR-10 ResNets with depth $l=20$ and different widths $w$.
All of these curves follow a similar pattern:%
\footnote{The same patterns occur for $l$ and $n$ for CIFAR-10 and $w$ and $n$ for ImageNet (see Appendix \ref{app:sec3-key-observations-alldimensions}). We focus on width for CIFAR-10 here for illustration.}

\textit{Observation 1: Low-error plateau.} The densest networks (right part of curves) have similar error to the unpruned network: $\smash{\epsilon_{np}(w)}$. We call this the \emph{low-error plateau}.

\textit{Observation 2: Power-law region.}
When pruned further, error increases linearly on the logarithmic axes of the figure.
Linear behavior on a logarithmic scale is the functional form of a \emph{power law}, where error relates to density via exponent $\gamma$ and coefficient $c$: $\smash{\epsilon(d, w) \approx cd^{-\gamma}}$.
In particular, $\gamma$ is the slope of the line on the logarithmic axes.

\textit{Observation 3: High-error plateau.} When pruned further, error again flattens; we call this the \emph{high-error plateau} and call the error of the plateau $\smash{\epsilon^\uparrow}$.

Figure \ref{fig:typical_prune_curve} (center) labels these regions for CIFAR-10 ResNet-20 ($w=1$, $n=1$) and shows an approximation of these regions that is piece-wise linear on logarithmic axes.
These observations are our starting point for developing a functional form that estimates error when pruning.

\textbf{Functional form.}
Our next task is to find a functional form that captures these observations about the relationship between density and error.
In the Chapter \ref{sec:Dense} we observe that the relationship between width and error shares the same general shape:
it has a region of lower error, a power-law region, and region of higher error.
However, this relationship is different enough from the one we observe (see Appendix \ref{app:difference_in_powerlaws}) to merit an entirely new functional form.

To develop this functional form, we note that the three regions of the curves in Figure \ref{fig:typical_prune_curve} (the low-error plateau, the power-law region, and the high-error plateau) can be described by three power laws: two plateaus with exponent zero and an intermediate region with exponent $\gamma$.
A functional family that arises frequently in systems that exhibit different power-law regions is the \emph{rational family}.
The particular family member we consider is as follows:%
\footnote{The expression $\left\Vert \frac{d-ja}{d-jb} \right\Vert^\gamma = \left(\frac{d^2+a^2}{d^2+b^2}\right)^\frac{\gamma}{2}$ meaning Eq. \ref{eq:prune_density} can be rewritten as 
    $\epsilon_{np} \left(\frac{d^2+p^2\left(\epsilon^\uparrow/\epsilon_{np}\right)^{2/\gamma}}{d^2 + p^2} \right)^{\gamma/2}$

}
\vspace{-2mm}
{
\small
\begin{equation}
\label{eq:prune_density}
    \hat{\epsilon}(\epsilon_{np}, d~|~l, w, n) = \epsilon_{np} \left\Vert \frac{d-jp\left(\frac{\epsilon^\uparrow}{\epsilon_{np}}\right)^{\frac{1}{\gamma}}}{d-j p} \right\Vert^\gamma
    \mbox{where } j = \sqrt{-1}
\end{equation}
}
\vspace{-3mm}

\begin{figure}
\vspace{-2mm}
\centering
\begin{minipage}{0.325\textwidth}
\includegraphics[width=\linewidth,trim={0 0 0 0.65cm},clip]{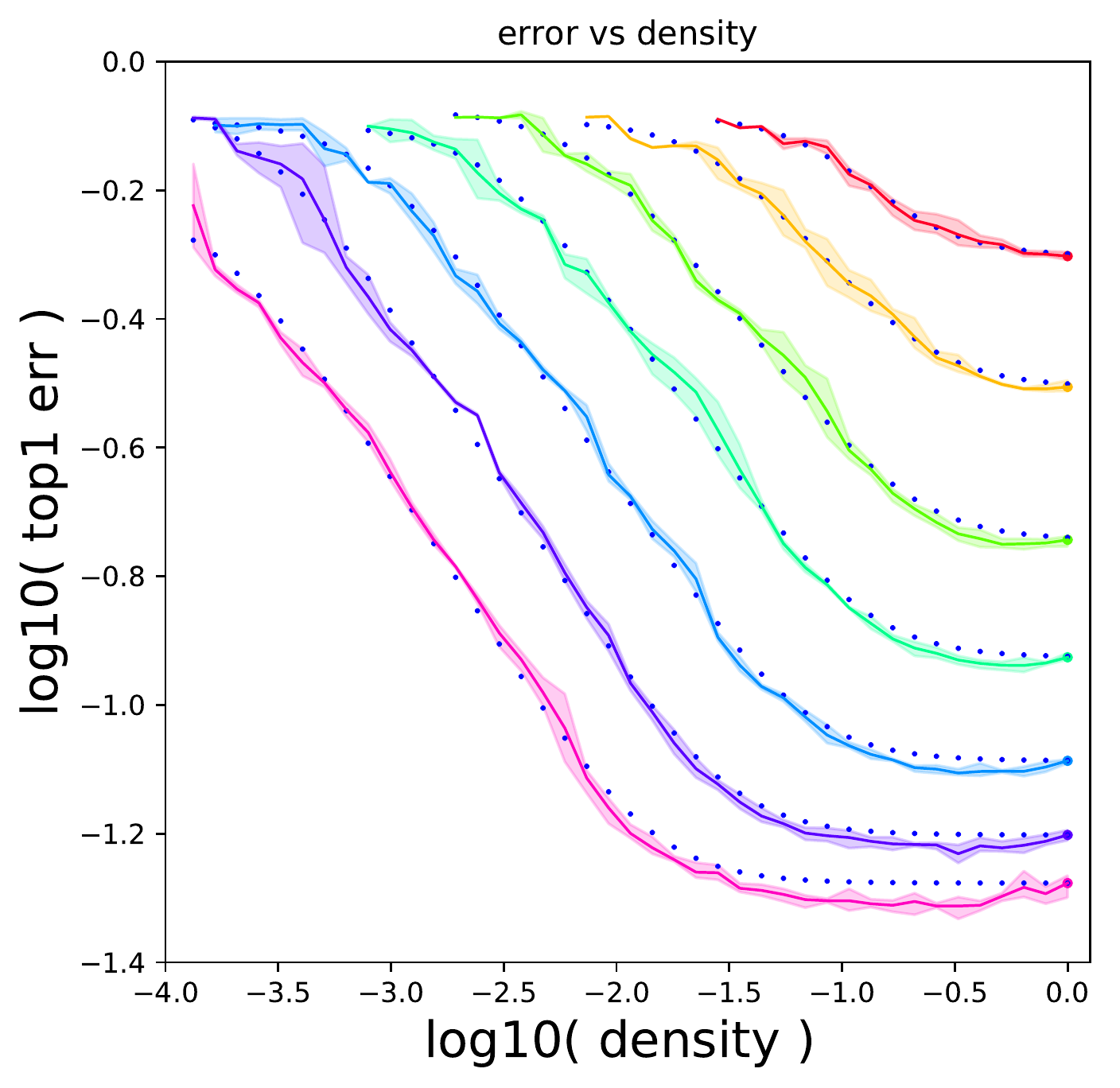}
\end{minipage}\hfil
\begin{minipage}{0.32\textwidth}
\includegraphics[width=\linewidth,trim={0 0 0 0.3cm},clip]{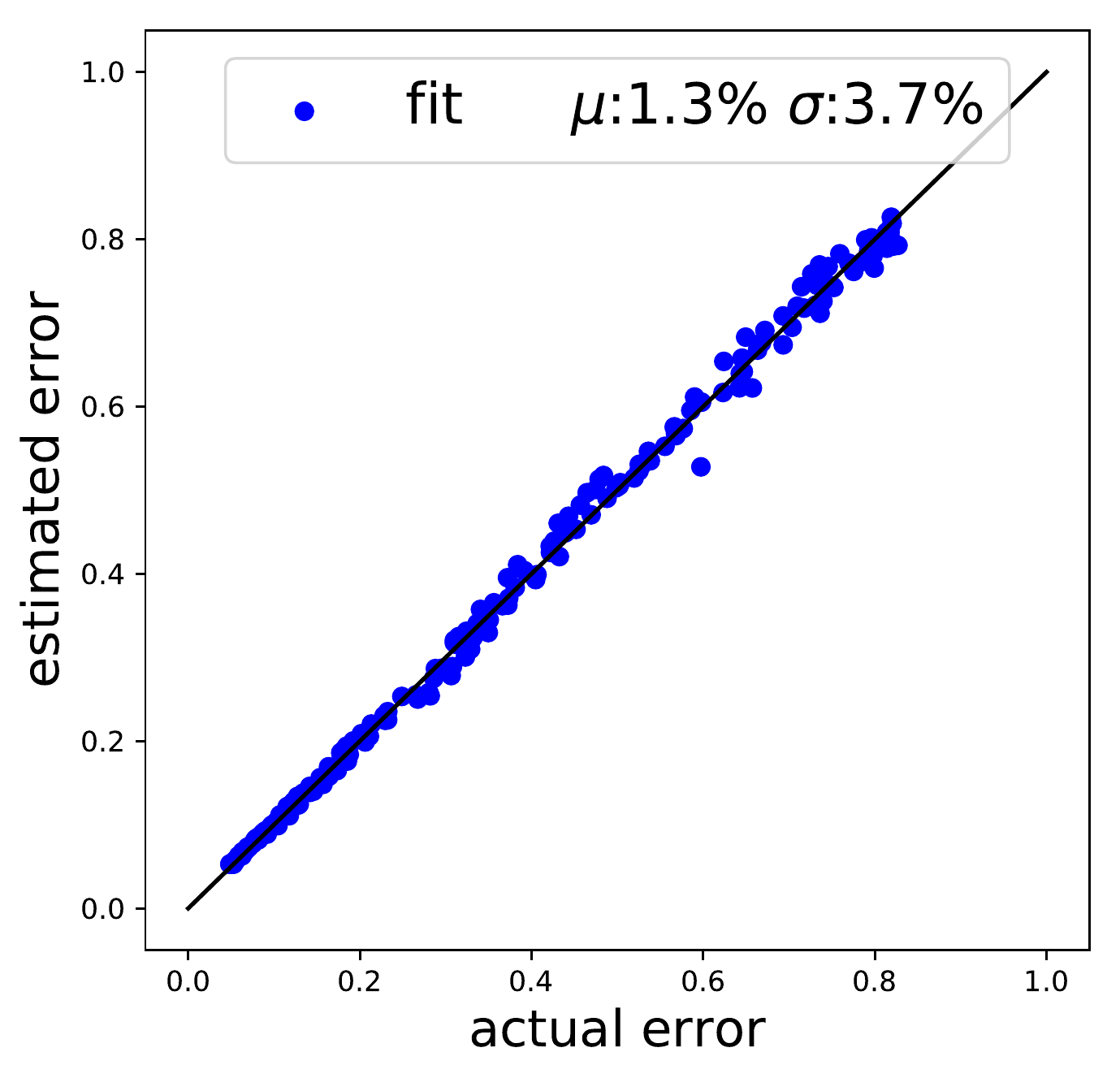}
\end{minipage}\hfil
\begin{minipage}{0.32\textwidth}
\includegraphics[width=\linewidth,trim={0 0 0 0.3cm},clip]{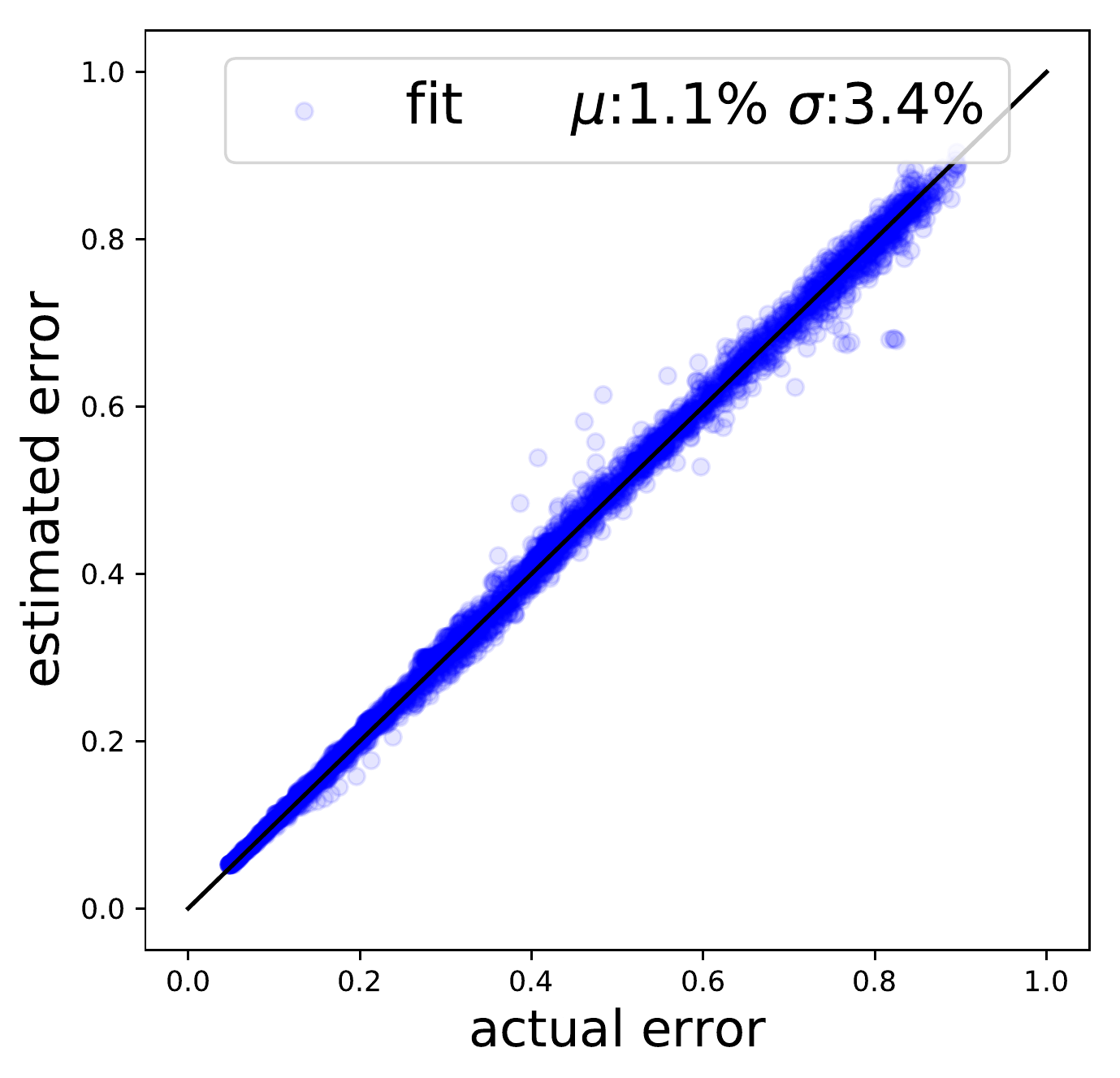}
\end{minipage}
\vspace{-1mm}
\caption{Estimated (blue dots) and actual error (solid lines) when pruning CIFAR-10 ResNets; $w$ varies, $l=20$, $n=N$ (left).
Estimated versus actual error for the same networks (center).
Estimated versus actual error for all CIFAR-10 ResNet configurations (right). }
\label{fig:fit_density}
\vspace{-4mm}
\end{figure}
This function's shape is controlled by $\smash{\epsilon_{np}}$, $\smash{\epsilon^\uparrow}$, $\smash{\gamma}$, and $p$ (visualized in Figure \ref{fig:typical_prune_curve}, right).
$\smash{\epsilon_{np}}$ and $\smash{\epsilon^\uparrow}$ are the values of the low and high-error plateaus.
$\smash{\gamma}$ is the slope of the power-law region on logarithmic axes.
$p$ controls the density where the high-error plateau transitions to the power-law region. 

\textbf{Fitting.}
To fit $\smash{\hat \epsilon(\epsilon_{np}, d~|~l, w, n)}$ to actual data $\smash{\epsilon(d, l, w, n)}$, we estimate values for the free parameters $\smash{\epsilon^\uparrow}$, $\smash{\gamma}$, and $p$ by minimizing the relative error $\delta \triangleq \frac{\hat{\epsilon}(\epsilon_{np}, d|l, w, n) - \epsilon(d, l, w, n)}{\epsilon(d, l, w, n)}$ using least squares regression. 
The fit is performed separately for each configuration $(l, w, n)$ for all densities, resulting in per-configuration estimates of $\smash{\hat\epsilon^\uparrow}$, $\smash{\hat\gamma}$, and $\smash{\hat p}$.
 
\textbf{Evaluating fit.}
For a qualitative view,\footnote{Error is a 4-dimensional function, so we can only qualitatively examine 2D projections. All such projections are in Appendix \ref{app:more_fits}.} we plot the actual error\footnote{We compute the error as the mean across three replicates with different random seeds and dataset subsamples.} $\epsilon(d, l, w, n)$ and the estimated error $\hat \epsilon(\epsilon_{np}, d~|~l, w, n)$ as a function density for CIFAR-10 ResNets of varying widths (Figure \ref{fig:fit_density}, left). 
Our estimated error appears to closely follow the actual error.
The most noticeable deviations occur at large densities, where the error decreases slightly on the low-error plateau whereas we treat it as flat (see Section \ref{sec:design}).

Quantitatively, we measure the extent to which estimated error departs from the actual error using the mean $\mu$ and standard deviation $\sigma$ of the relative deviation $\delta$.
Figure \ref{fig:fit_density} (center) compares the estimated and actual errors for the networks in Figure \ref{fig:fit_density} (left);
Figure \ref{fig:fit_density} (right) shows the same comparison for  all configurations of $l$, $w$, and $n$ on CIFAR-10 and the more than 4,000 pruned ResNets that result.
The relative deviation on all configurations has mean $\mu<2\%$ and standard deviation $\sigma<4\%$; this means that, if the actual error is $10\%$, the estimated error is $9.8 \pm 0.4\%$ ($\hat{\epsilon} = (1-\delta) \epsilon $).

\section{Jointly Modeling Error For All Dimensions}
\label{sec:joint}

In Section \ref{sec:single-network}, we found a functional form $\hat\epsilon(\epsilon_{np}, d~|~l, w, n)$ (Eq. \ref{eq:prune_density}) that accurately predicts the error when pruning a \emph{specific} member of a network family (with depth $l$ and width $w$) trained with a dataset of size $n$.
The parameters governing Equation \ref{eq:prune_density} ($\smash{\epsilon^\uparrow}$, $p$, and $\smash{\gamma}$ ) varied between and depended on the specific configuration of $l$, $w$, $n$.
However, we are interested in a single \emph{joint} scaling law $\hat\epsilon(\epsilon_{np}, d, l, w, n)$ that, given the unpruned network error $\epsilon_{np}(l, w, n)$, accurately predicts error across \emph{all} dimensions: all members of a network family that vary in depth and width, all densities, and all dataset sizes.
Importantly, the parameters of this scaling law must be constants as a function of all dimensions. In this section, we develop this joint scaling law.

\begin{figure}
\vspace{-1mm}
\centering
\begin{minipage}{.3\textwidth}
\includegraphics[width=\linewidth,trim={0 0 0 0.7cm},clip]{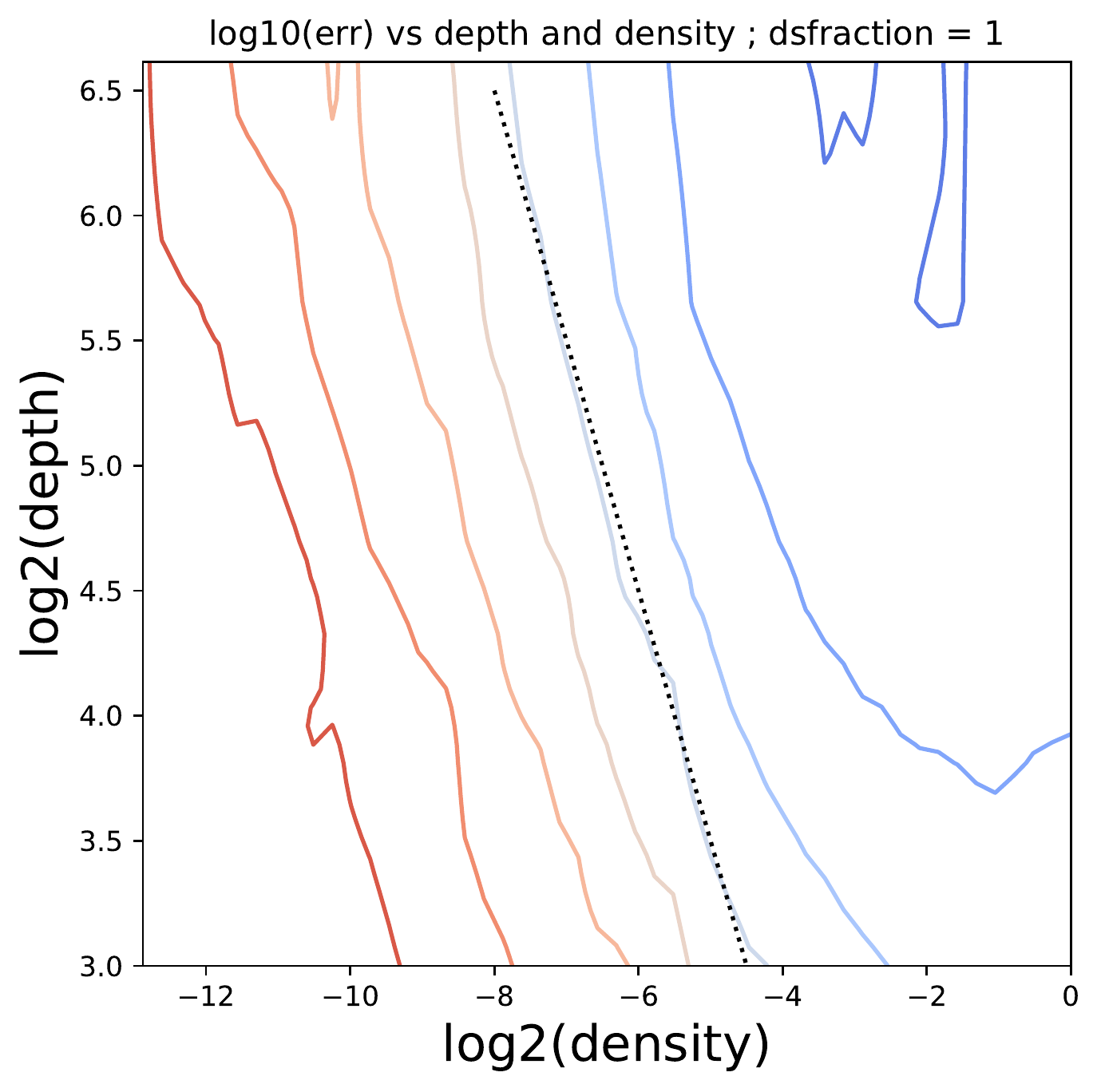}
\end{minipage}%
\begin{minipage}{0.3\textwidth}
\includegraphics[width=\linewidth,trim={0 0 0 0.7cm},clip]{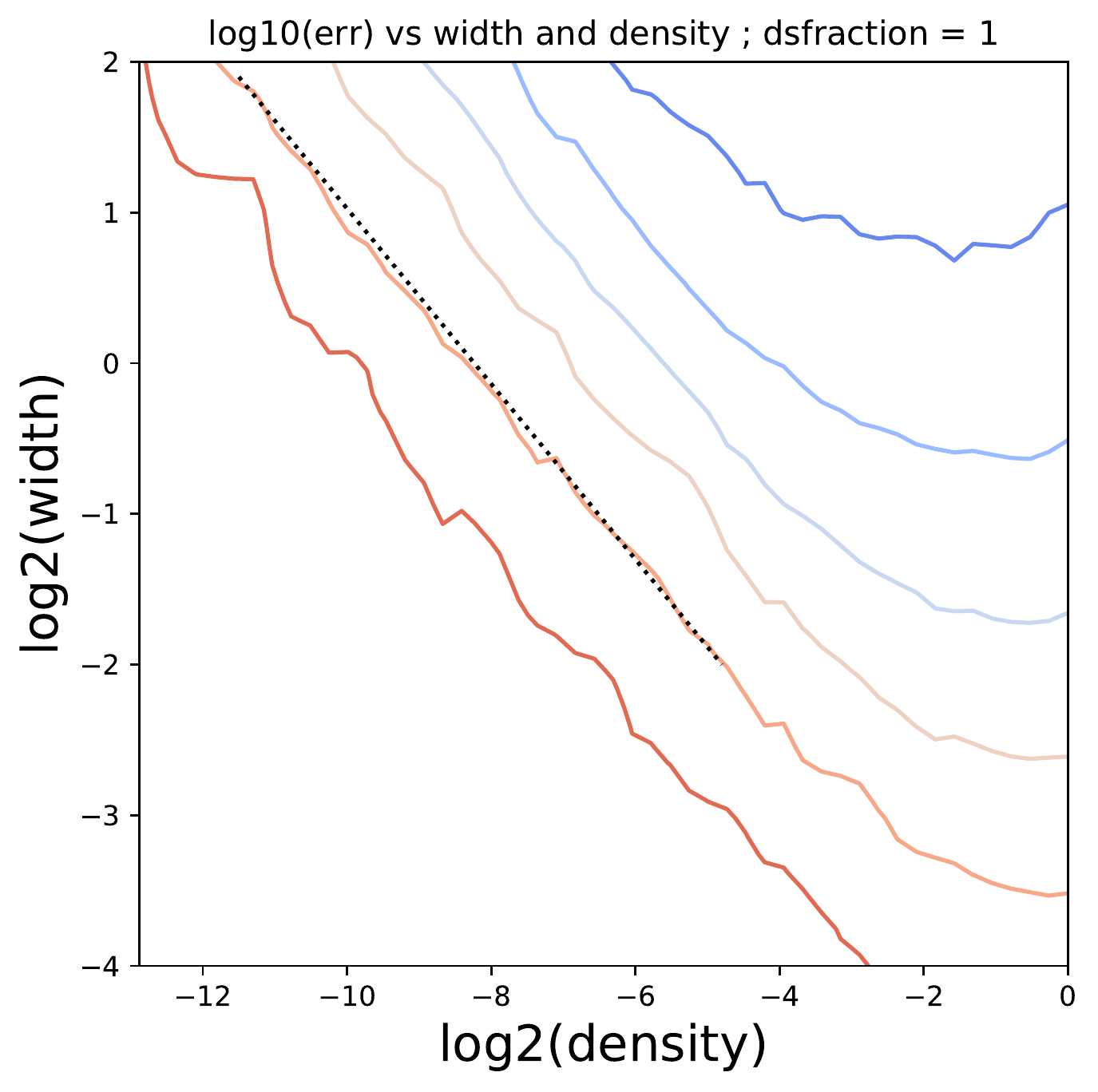}
\end{minipage}
\vspace{-1mm}
\caption{Projections of $\epsilon(d, l, w, n)$ onto two-dimensional planes for the CIFAR-10 ResNets, showing contours of constant error. For low enough densities, the contours have linear slopes on the logarithmic axes---depicted by a reference black-dotted line. The density/depth plane (left).  The density/width plane (right).}
\vspace{-4mm}
\label{fig:basic_form}
\end{figure}

\textbf{Intuition: the error-preserving invariant.}
Our desired scaling law $\hat\epsilon(\epsilon_{np}, d, l, w, n)$ will be a four-dimensional function of $d$, $w$, $l$, and $n$. 
To develop an intuition for the relationship between these inputs, we study the interdependence between density and depth or width by examining two-dimensional projections of the actual error $\epsilon(d, l, w, n)$ in Figure \ref{fig:basic_form}.
These plots display contours of constant error as density and depth or width vary. 

Consider the projection onto the plane of density and depth (Figure \ref{fig:basic_form}, left).
The constant-error contours are linear except for in the densest networks, meaning each contour traces a power-law relationship between $d$ and $l$.
In other words, we can describe all combinations of densities and widths that produce error $\epsilon_v$ using $l^\phi d = v$, where $v$ is a constant at which network error is $\epsilon_v$ and $\phi$ is the slope of the contour on the logarithmic axes.
The contours of density and width also have this pattern (Figure \ref{fig:basic_form}, right), meaning we can describe a similar relationship $w^\psi d = v'$.
Finally, we can combine these observations about depth and width into the expression $l^\phi w^\psi d = v''$.

We refer to the expression $l^\phi w^\psi d$ as the \emph{error-preserving invariant}, and we denote it $m^*$.
This invariant captures the observation that there exist many interchangeable combinations of depth, width, and density that achieve the same error and tells us which combinations do so. For example, networks of vastly different densities reach the same error if we vary $l$ and $w$ according to the invariant.

\textbf{Functional form.}
The invariant allows us to convert the functional form $\hat\epsilon(\epsilon_{np}, d~|~l, w, n)$ for a specific $l$, $w$, and $n$ from Section \ref{sec:single-network} into a joint functional form $\hat\epsilon(\epsilon_{np}, d, l, w, n)$ for all $l$, $w$, and $n$.
Rewriting the definition of the invariant, $d = \frac{m^*}{l^\phi w^\psi}$.
We can substitute this for $d$ in the functional form from Section \ref{sec:single-network}.
Finally, by rewriting $p$ as $\frac{p'}{l^\phi w^\psi}$ and canceling, we arrive at the following expression:
\vspace{-2mm}
{
\small
\begin{equation}
\label{eq:intermediate_state}
    \hat{\epsilon}(\epsilon_{np}, d~|~l, w, n) = \epsilon_{np} \left\Vert \frac{m^*-jp'\left(\frac{\epsilon^\uparrow}{\epsilon_{np}}\right)^{\frac{1}{\gamma}}}{m^*-j p'} \right\Vert^\gamma
    = \epsilon_{np} \left\Vert \frac{l^\phi w^\psi d-jp'\left(\frac{\epsilon^\uparrow}{\epsilon_{np}}\right)^{\frac{1}{\gamma}}}{l^\phi w^\psi d-j p'} \right\Vert^\gamma
    = \hat{\epsilon}(\epsilon_{np}, d, l, w, n)
\end{equation}
}
\vspace{-3mm}

which is the joint functional form $\smash{\hat{\epsilon}(\epsilon_{np}, d, l, w, n)}$ of all four dimensions $d$, $l$, $w$, and $n$.
Critically, for this to be a useful joint form, the free parameters $\smash{e^\uparrow, p', \text{ and } \gamma}$ must be constants shared across all possible values of $d$, $l$, $w$, and $n$. We will assume this is the case and directly quantify how well this assumption holds in the evaluation section below.

\begin{figure}
\vspace{-2mm}
\centering 
\begin{minipage}{0.33\textwidth}
\includegraphics[width=\linewidth,trim={0cm 0 0 0.6cm},clip]{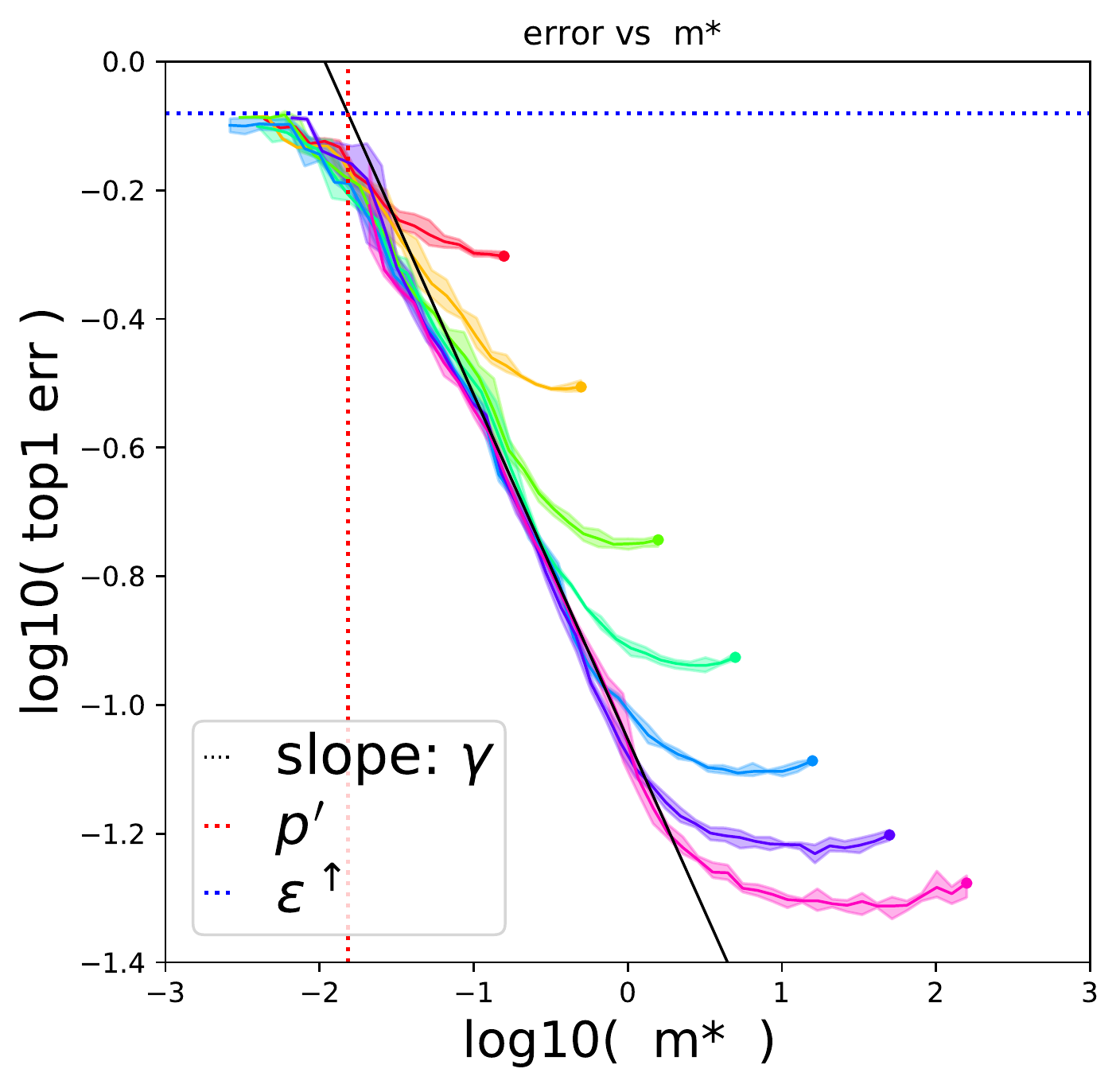}
\end{minipage}\hfil 
\begin{minipage}{0.28\textwidth}
  \includegraphics[width=\linewidth,trim={2cm 0 0 0.6cm},clip]{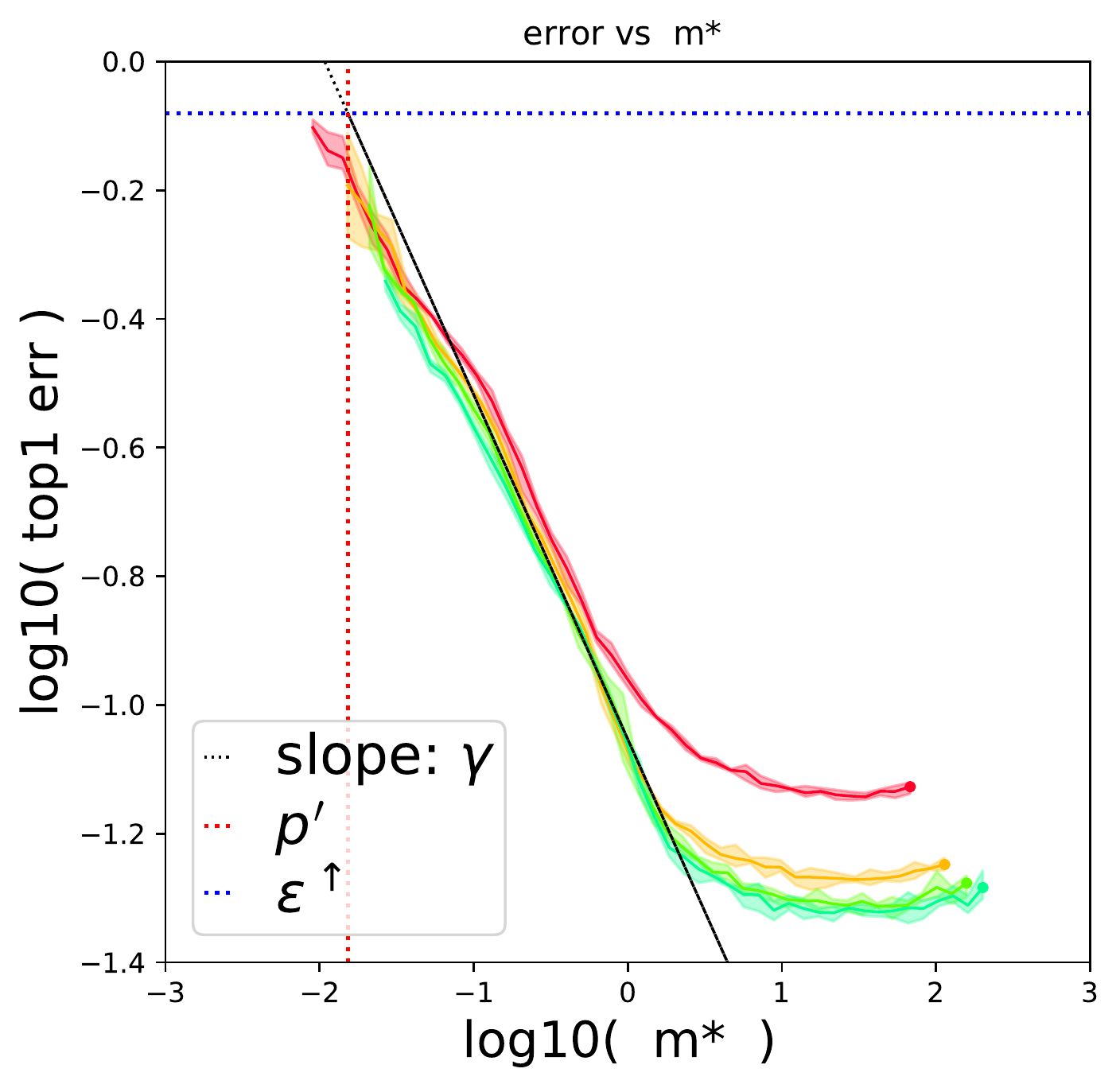}
\end{minipage}\hfil 
\begin{minipage}{0.28\textwidth}
  \includegraphics[width=\linewidth,trim={2cm 0 0 0.6cm},clip]{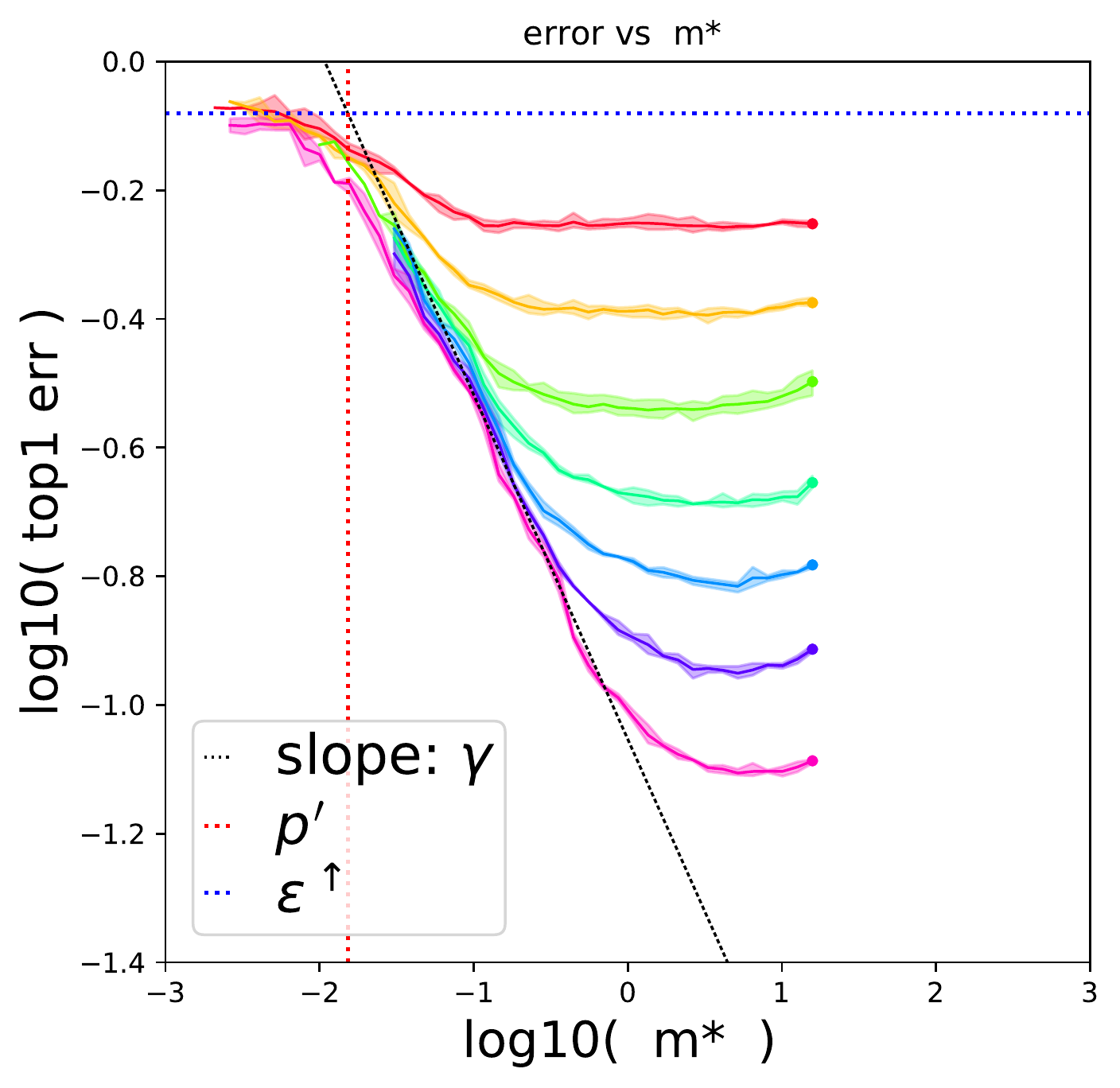}
\end{minipage}
\caption{Relationship between $m^*$ and error when pruning CIFAR-10 ResNets and varying $w$ (left, $l=20$, $n=N$),  $l$ (center, $w=1$, $n=N$), $n$ (right, $l=20$, $w=1$). We annotate $\gamma$, $\epsilon^\uparrow$, and $p'$; they qualitatively appear to take on similar values in all cases, an observation that we use to inform the design of our joint scaling law. }
\label{fig:cifar_joint_observations}
\end{figure}

For qualitative intuition as to why this is a reasonable assumption, consider the relationship between $m^*$ and the test error of pruned networks as we vary depth, width, and dataset size (Figure \ref{fig:cifar_joint_observations}).
Across all projections, the annotated $\smash{e^\uparrow}$ (error of the high-error plateau), $\gamma$ (slope of the power-law region) and $\smash{p'}$ (value of $\smash{m^*}$ where the high-error plateau transitions to the power-law region) appear the same.

The preceding discussion addresses how we handle $l$, $w$, and $d$ in our joint scaling law.
We address dataset size $n$ in Eq. \ref{eq:intermediate_state} implicitly through the way that it affects $\epsilon_{np}$, and we validate that this is a reasonable choice through the evaluation below.
We retain the explicit form $\hat\epsilon(...,n)$ to stress that the lack of explicit dependency on $n$ is non-trivial and was not known prior to our work.

\textbf{Fitting.}
To fit $\hat \epsilon(\epsilon_{np}, d, l, w, n)$ to the actual data $\epsilon(d, l, w, n)$, we estimate values for the free parameters $\smash{\epsilon^\uparrow, \gamma, p', \phi \text{ and } \psi}$ by minimizing the relative error $\delta \triangleq \frac{\hat{\epsilon}(\epsilon_{np}, d, l, w, n) - \epsilon(d, l, w, n)}{\epsilon(d, l, w, n)}$ using least squares regression. The fit is performed jointly over all configurations of $d, l, w,$ and $n$, resulting in joint estimates of  $\smash{\hat\epsilon^\uparrow, \hat\gamma, \hat p, \hat\phi, \text{ and } \hat\psi}$.
One can also perform a partial fit for a subset of dimensions (e.g., just $d$, $l$, and $n$) by omitting $\smash{\phi}$ and/or $\smash{\psi}$ (see Appendix \ref{app:more_fits}).

\begin{figure}

    \centering
        \begin{minipage}{0.315\textwidth}
            \centering
            \includegraphics[width=\linewidth,trim={0.2cm 0 0 0.3cm},clip]{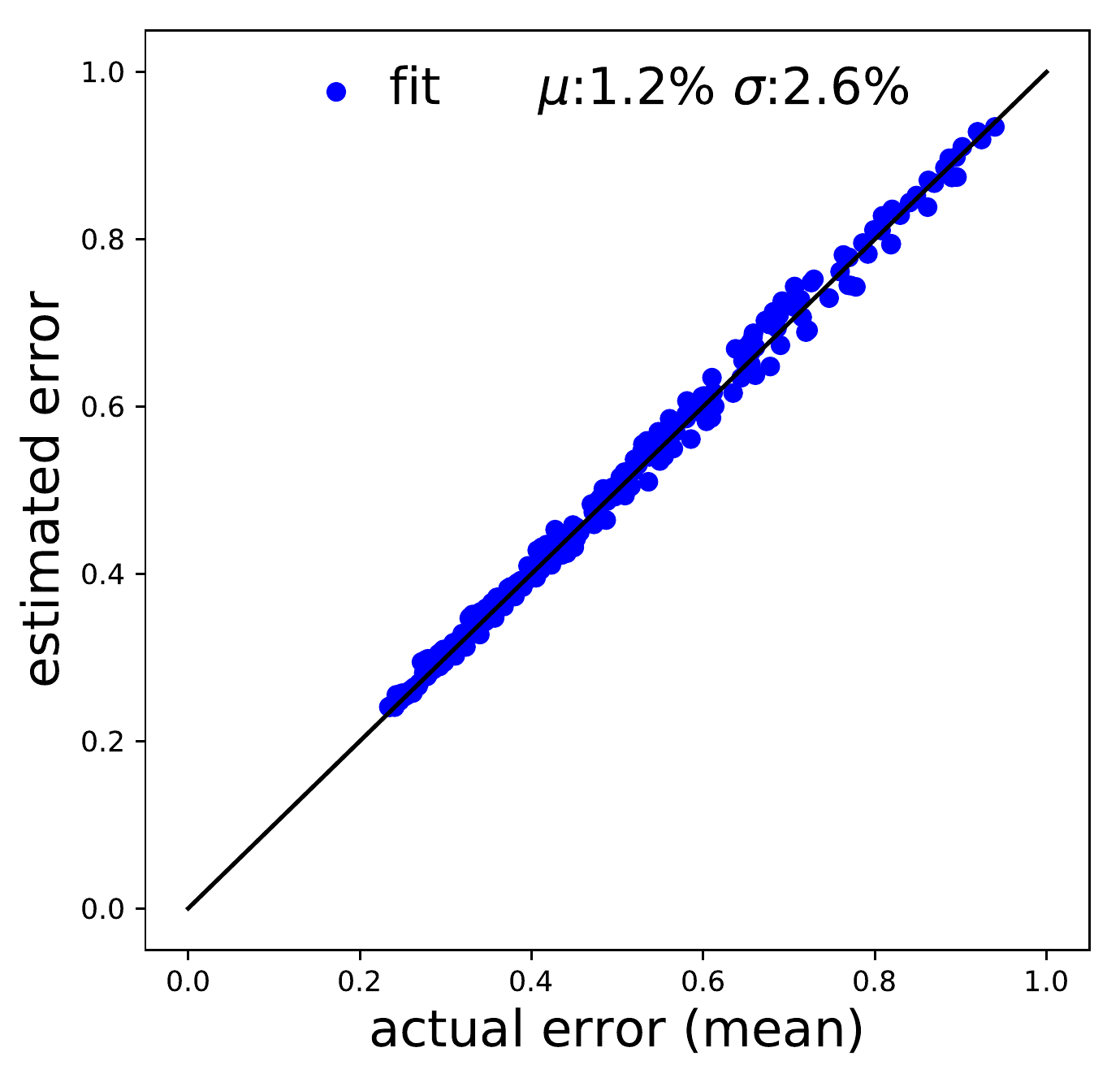}
            \label{fig:fit_imagent}
        \end{minipage}%
        \begin{minipage}{0.315\textwidth}
            \centering
            \includegraphics[width=\linewidth,trim={0.2cm 0 0 0.3cm},clip]{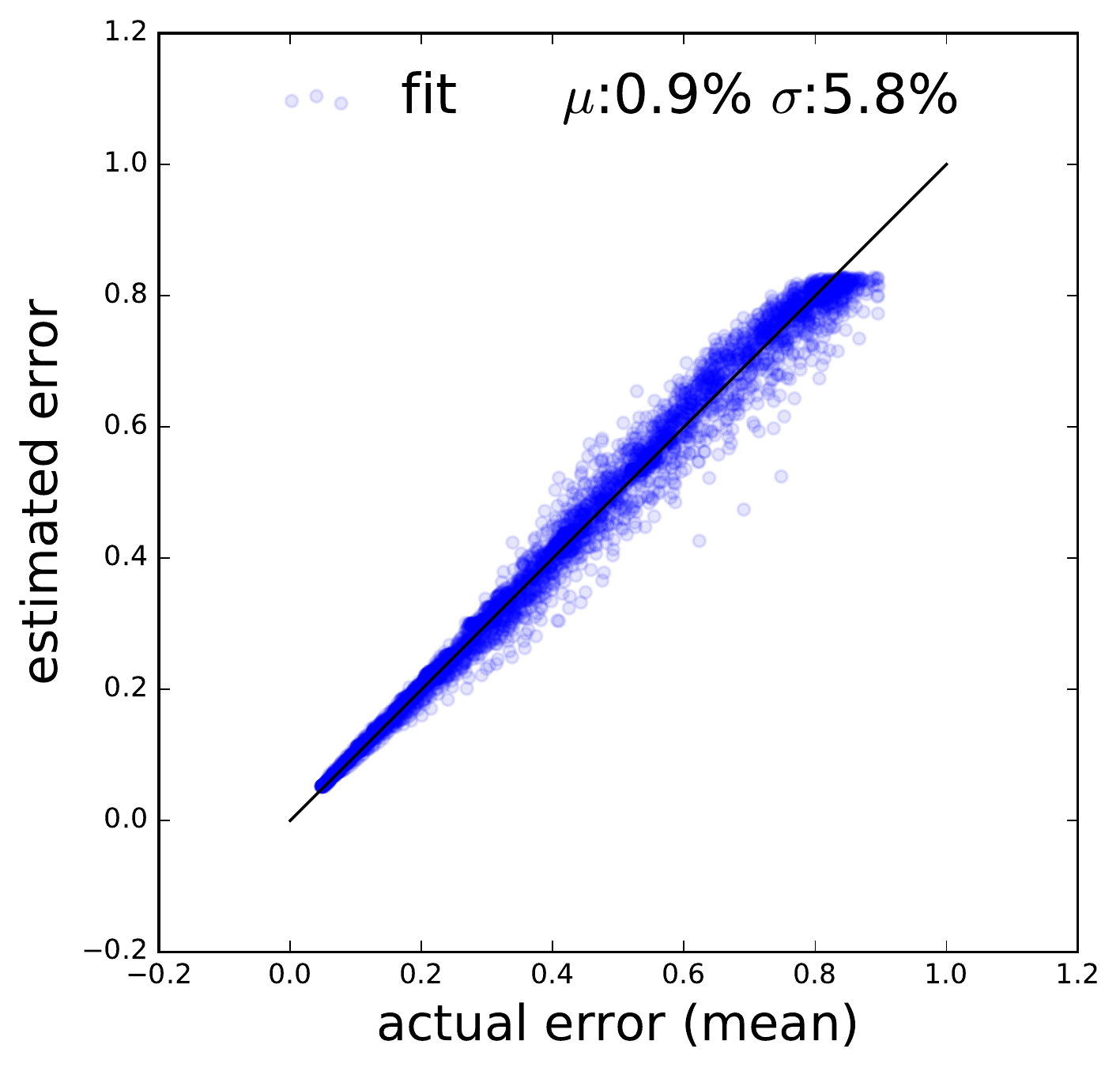}
            \label{fig:fit_corr_all}
        \end{minipage}%
        \begin{minipage}{0.315\textwidth}
            \centering
            \includegraphics[width=\linewidth,trim={0.2cm 0 0 0.3cm},clip]{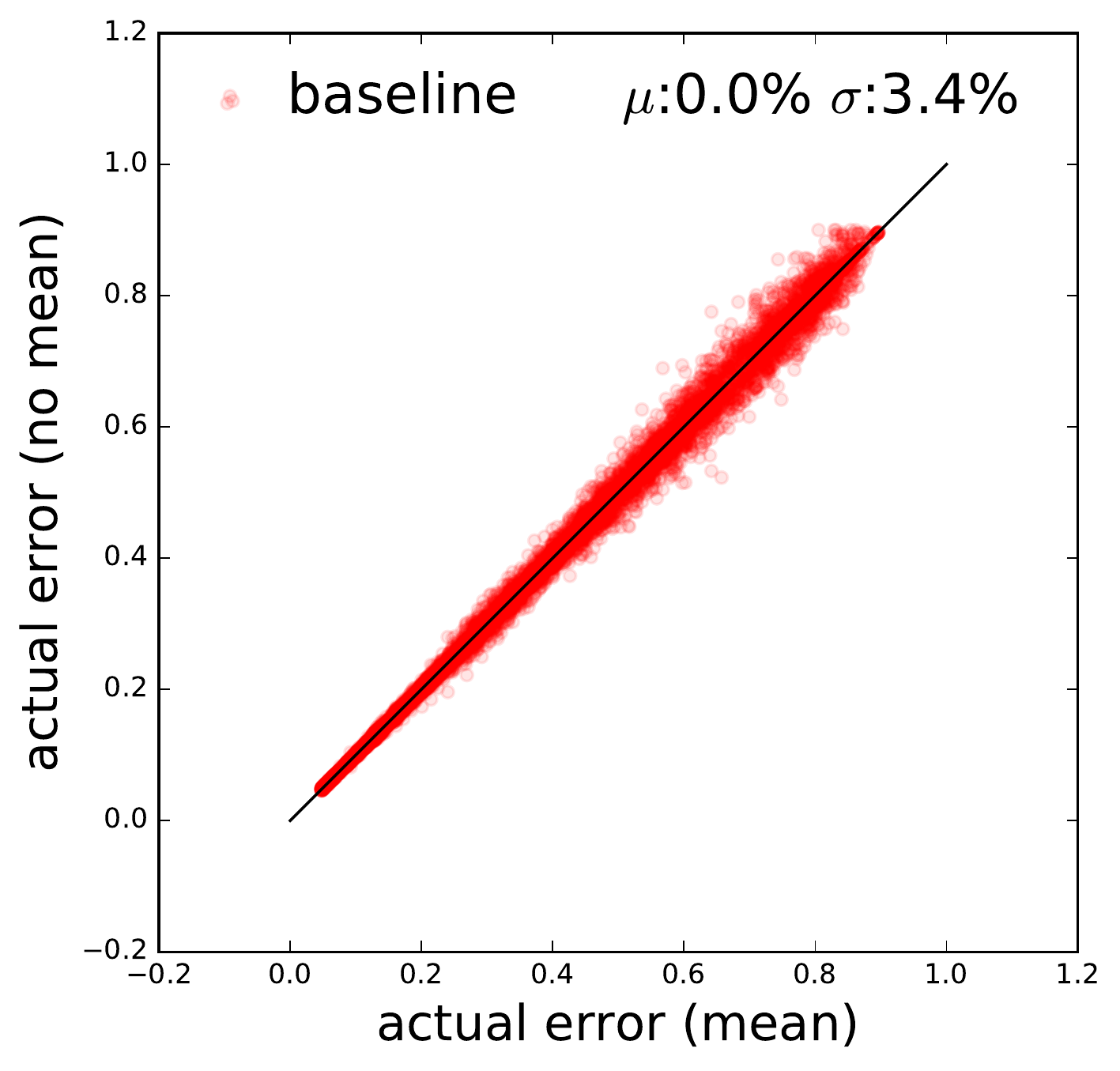}
            \label{fig:cifar_baseline_joint}
        \end{minipage}

\caption{Estimated versus mean actual error for all configurations $(d,w,n)$ for ImageNet (left) and $(d,w,l,n)$ for CIFAR-10 (center). 
The variation in error when running the same experiment on CIFAR-10 three times with different random seeds (right).}
\label{fig:joint_fit1}
\end{figure}

\textbf{Evaluating fit.}
In Figure \ref{fig:joint_fit1}, we plot the actual error $\epsilon(d, l, w, n)$ and the estimated error $\smash{\hat \epsilon(\epsilon_{np}, d, l, w, n)}$ for
the CIFAR-10 ResNets (all widths, depths and dataset sizes)
and ImageNet ResNets (all widths and dataset sizes for depth 50).
As in Section \ref{sec:single-network}, our estimated error appears to closely follow the actual error. 
Deviations arise mainly at high densities where error decreases below $\epsilon_{np}$ and low densities approaching high error saturation.

We again quantify the fit of the estimated error using the mean $\mu$ and standard deviation $\sigma$ of the relative deviation $\delta$.
The relative deviation on the joint scaling laws for the CIFAR-10 and ImageNet networks has a mean $\mu<2\%$ and standard deviation of $\sigma<6\%$.

To contextualize these results, Figure \ref{fig:joint_fit1} (right) quantifies the variation in error we see over multiple replicates of the CIFAR-10 experiments due to using different random seeds.
It plots the minimum, maximum, and mean errors across the three replicates we ran.%
\footnote{We only ran a single replicate of the ImageNet experiments due to the significant cost of collecting data.}
The variation across trials has a standard deviation of $\sigma=3.4\%$, sizeable relative to the estimation error of $\sigma=5.8\%$ for the joint scaling law. This indicates that a significant portion of our error may stem from measurement noise.

The functional form has just five parameters and obtains an accurate fit on 4,301 points on CIFAR-10 and 274 points on ImageNet, suggesting it is a good approximation.
In Appendix \ref{app:more_arch_alg}, we show that it achieves a similarly good fit for additional architectures and datasets.
In Section \ref{app:interpolation}, we show that, although we use a large number of points to develop and evaluate our functional form here, it is possible to get a good fit with far fewer points and the fit has low sensitivity to the choice of points.

\setlength{\tabcolsep}{0pt}
\renewcommand{\arraystretch}{0.5}

\section{Sensitivity of Fit to Number of Points}
\label{app:interpolation}

In Section \ref{sec:joint}, we showed that our scaling law was accurate when we fit it on all of the available data.
Now that we possess the functional form and know that it can accurately model the behavior of IMP, we study the amount of data necessary to obtain a stable,\footnote{Stability is defined as a small change in output relative to a change in input. The requirement here is that a change in choice of points leads to a small expected change in estimation accuracy.} accurate fit.
This question is especially relevant when the functional form is applied to new settings---new networks, datasets, or pruning algorithms---and we must collect new data to do so.
The functional form has only five parameters, suggesting that few experiments will be necessary to obtain an accurate fit.

\textbf{Experiments.}
To evaluate the effect of the number of points on the stability and accuracy of the fit, we randomly sample varying numbers of points, fit the scaling law to those points, and evaluate the quality of the fit over all points.
We sample these points in two ways.

\textit{Experiment 1.} Randomly sample $T$ networks $(w, l, n, d)$.
This experiment evaluates the stability and accuracy of the fit when naively varying the number of points.

\textit{Experiment 2.} Randomly sample $T$ network configurations $(w, l, n)$ and include all densities $d$ for each configuration. This experiment captures the specific use case of IMP, where obtaining data at density $d$ requires obtaining all densities $d' > d$. As such, we anticipate that data will be obtained by iteratively pruning a small number of configurations $(w, l, n)$ to low density.

\begin{figure}
\centering
\begin{minipage}{0.4\textwidth}
    \includegraphics[width=\linewidth]{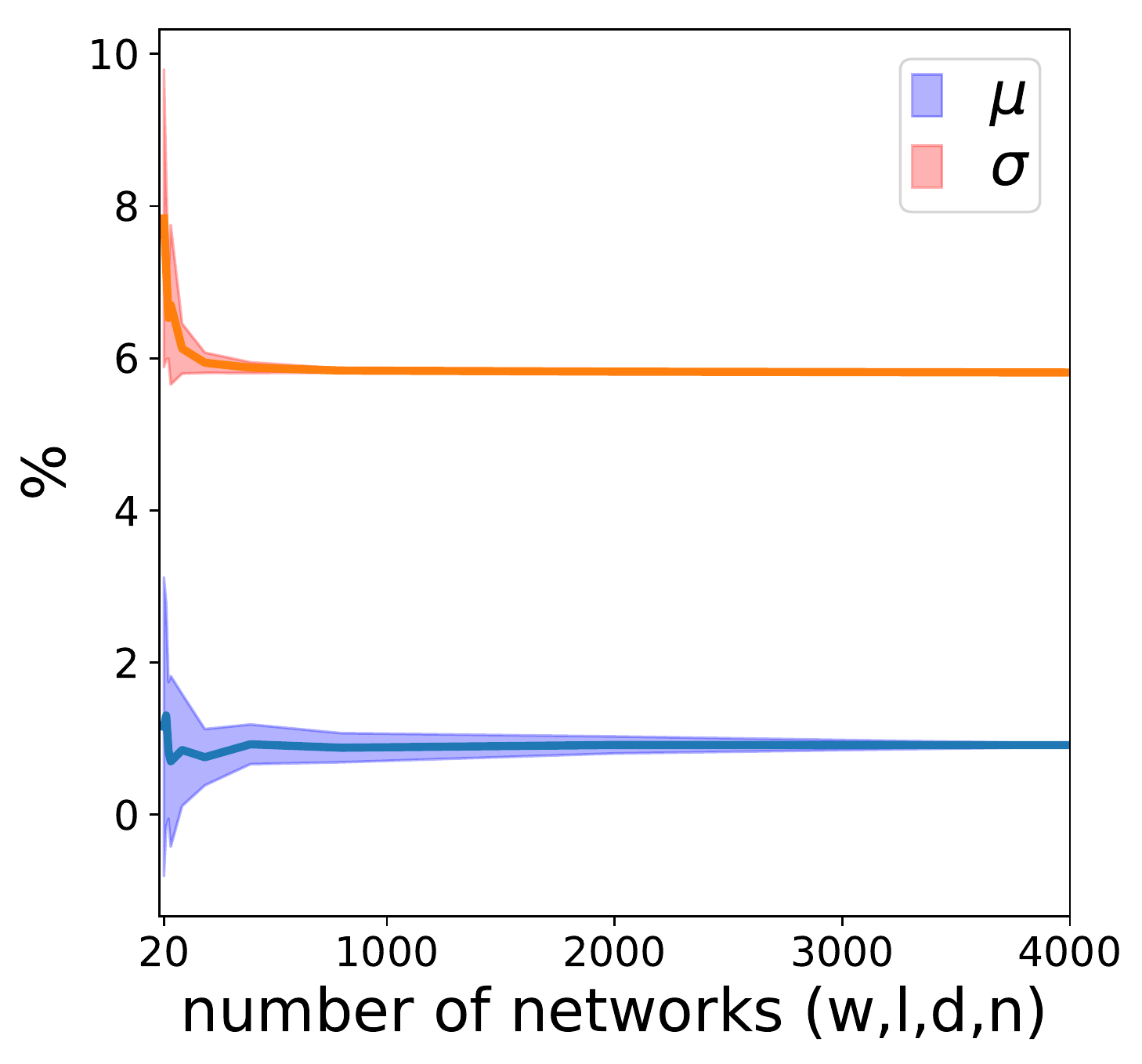}
\end{minipage}%
\begin{minipage}{0.4\textwidth}
    \includegraphics[width=\linewidth]{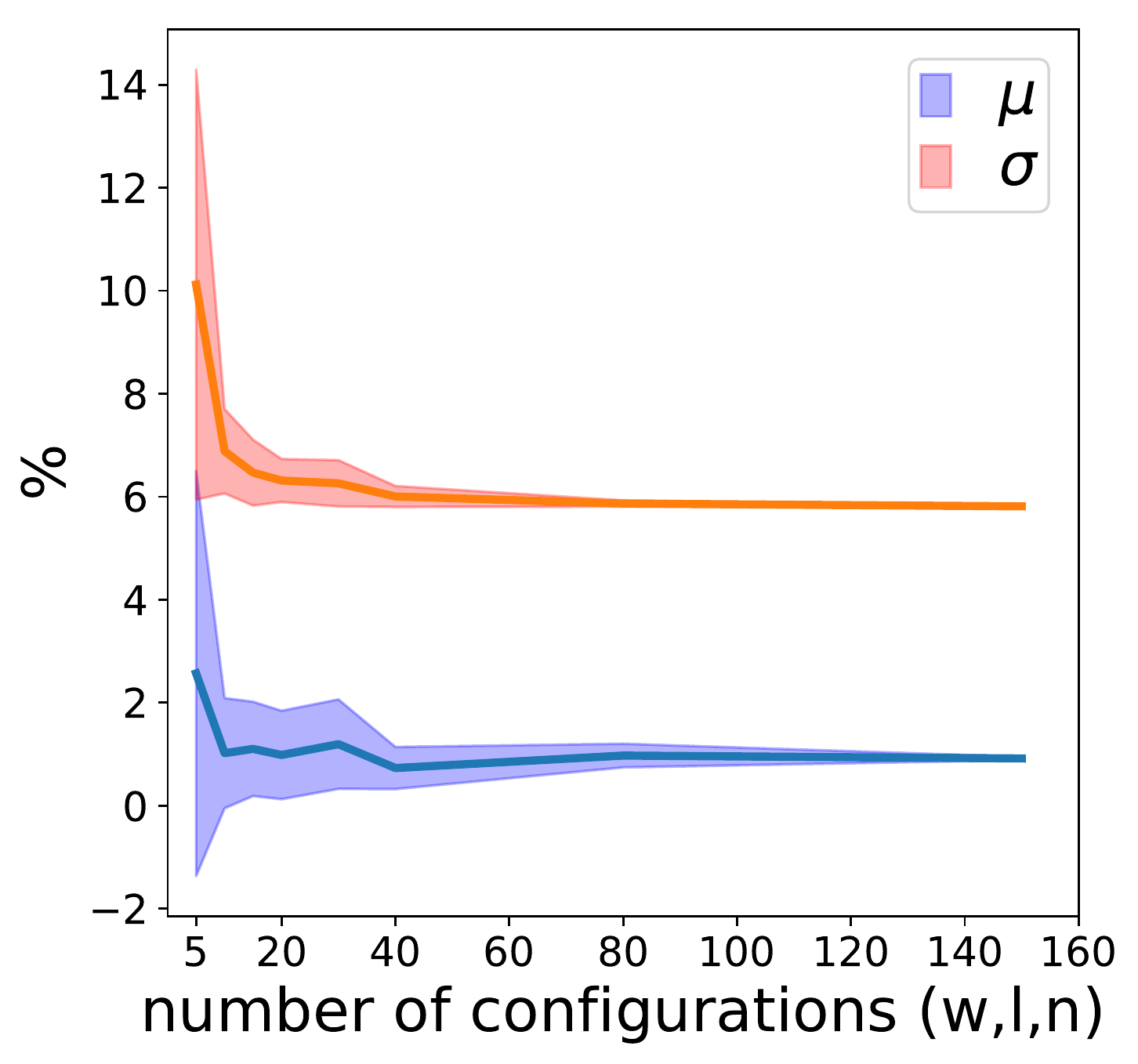}
\end{minipage}

\caption{The effect of the number of points used to fit our scaling law (on the CIFAR-10 ResNets) on $\mu$ and $\sigma$. Left: experiment 1 from Section \ref{app:interpolation} (random points $w$, $l$, $d$, $n$). Right: experiment 2 from Section \ref{app:interpolation} (random configurations $w$, $l$, $n$ and all densities).}
\label{fig:stability1}
\end{figure}

\textbf{Results.}
We perform each experiment for many different values of $T$ on the CIFAR-10 ResNets pruned with IMP.
We repeat the experiment at each value of $T$ 30 times with different samples of points and report the mean and standard deviation of $\mu$ and $\sigma$ for the fit.
Experiments 1 and 2 respectively appear in Figure \ref{fig:stability1} left and right.
The shaded areas represent one standard deviation from the mean in each direction.
On Experiment 1, when just 40 networks $(w, l, d, n)$ are available, the standard deviation on both $\mu$ and $\sigma$ is just one percentage point.
On Experiment 2, when just 15 random configurations of $(w, l, n)$ are available at all densities, we similarly achieve standard deviation below 1\%.
In both cases, as the number of networks increases, the standard deviation decreases further.

These results show that, now that our scaling law is known, it is possible to obtain an accurate (and stable) estimation using far less data than we used to evaluate the quality of the fit in Section \ref{sec:joint}.
This implies that translating our scaling law to new settings will be far less data-intensive than developing and evaluating it in the first place.
Moreover, the results in this section reflect a particularly naive way of selecting points: doing so randomly; we made no effort to ensure that, for example, the networks represented a diverse range of widths, depths, dataset sizes, and densities. 
By selecting these networks in a strategic way, it may be possible to further reduce the number of networks necessary to obtain a similarly accurate fit.


\section{Considerations in Selecting a Functional Form}
\label{sec:design}

We have shown that our proposed functional form $\hat\epsilon(d, l, w, n)$ accurately approximates the error when pruning families of neural networks.
In this section, we discuss some of the key criteria that led us to select this particular functional form.
We intend this section to provide insight into our choices in the context of the broader design space and to highlight opportunities for further refinement.

\begin{figure}
\centering
\begin{minipage}{.31\linewidth}
    \includegraphics[width=\linewidth,trim={0 0 0 0.65cm},clip]{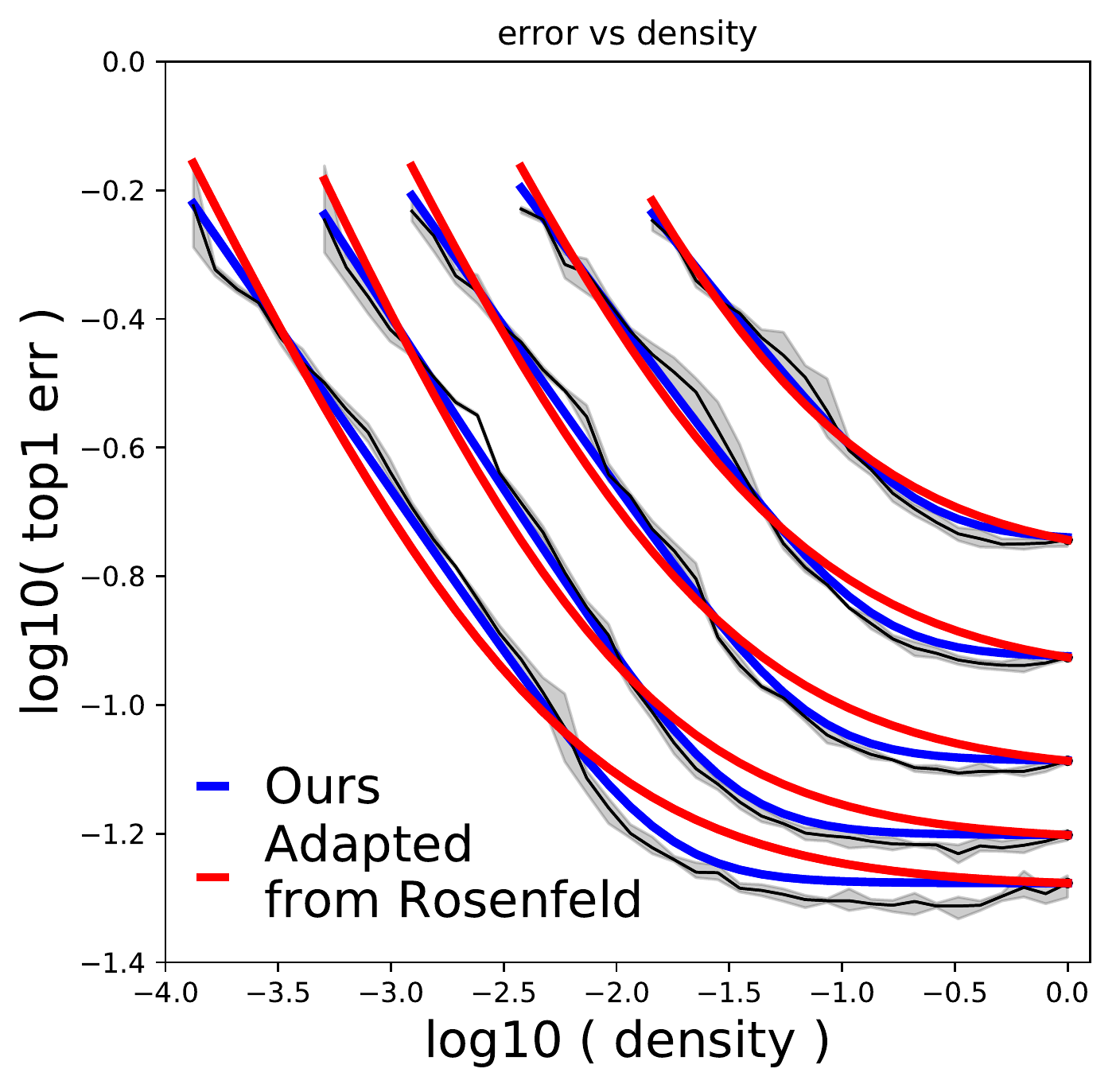}
\end{minipage}
\begin{minipage}{.31\linewidth}
    \includegraphics[width=\linewidth,trim={10.5cm 1.9cm 11cm 5.6cm},clip]{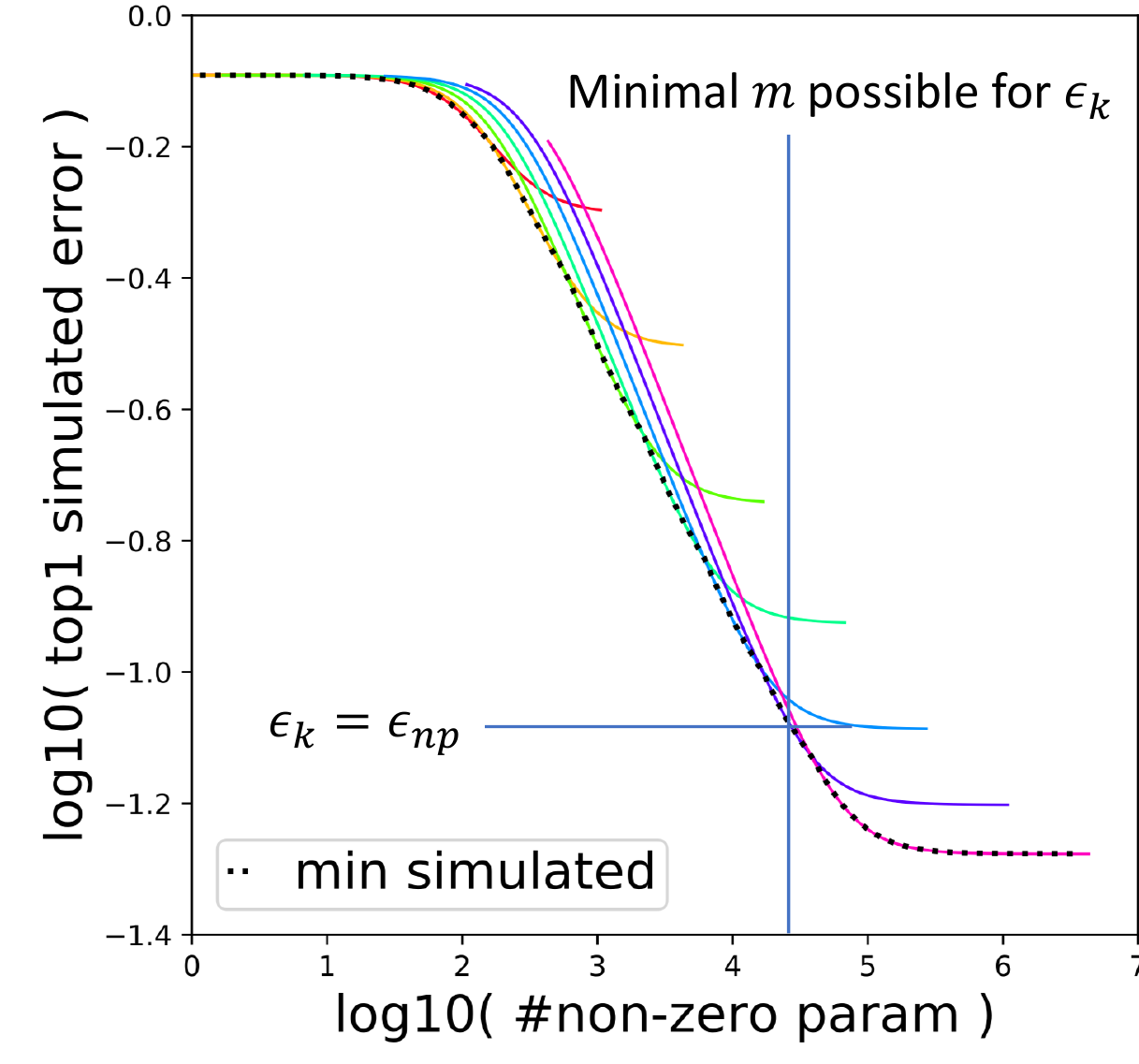}
\end{minipage}
\begin{minipage}{.31\linewidth}
    \includegraphics[width=\linewidth,trim={10.5cm 1.9cm 11cm 5.6cm},clip]{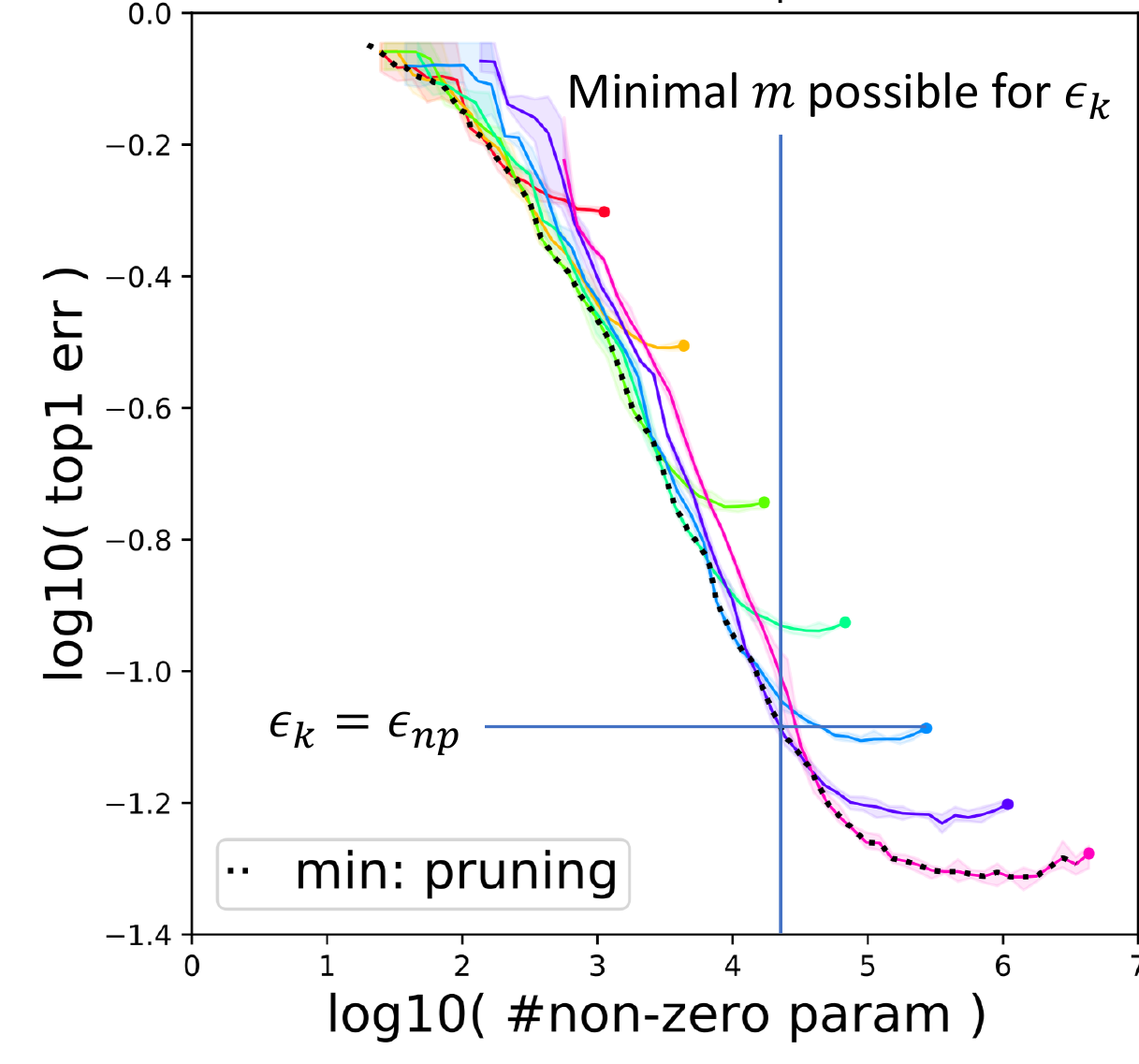}
\end{minipage}
    \caption{Estimated error as width varies for the CIFAR-10 ResNets  (left).
    Actual error as width varies for the CIFAR-10 ResNets (right).
    The dotted black line is the minimal number of parameters necessary to reach each error $\epsilon_k$ among all of the pruned networks.
    Reaching this point requires starting with a particular lower-error network (purple) and pruning until error increases to $\epsilon_k$.
    Starting too large (pink) will miss this point.
    }
    \label{fig:misc}
\end{figure}

\textbf{Criterion 1: Transitions.}
In Section \ref{sec:single-network}, we observe that, when pruning a neural network, error has a low-error plateau, a power-law region, and a high-error plateau.
Between these regions are \emph{transitions} where error varies smoothly from one region to the next.
Matching the shape of these transitions was a key consideration for selecting our function family.
To illustrate the importance of properly fitting the transitions, Figure \ref{fig:misc} (left) shows two possible functional families for fitting the relationship between density and error for CIFAR-10 ResNets.
Actual error is in gray, and the functional form from Section \ref{sec:single-network} is in blue.
In red is the fit for a functional form adapted from the one used in Chapter \ref{sec:Dense} to model the relationship between width and error.
The difference between these functional families is the way they model transitions, and the one we choose in this paper better models the transitions in our setting.
For further discussion of this comparison, see Appendix \ref{app:difference_in_powerlaws}.

\textbf{Criterion 2: A small number of interpretable parameters.}
Selecting a functional form is not merely a curve-fitting exercise.
We seek the underlying structure that governs the relationships between $d, l, w, n$, and error in a manner akin to a law of physics.
As such, our functional form should have a small number of parameters that are \emph{interpretable}.
In our functional form (Equation \ref{eq:intermediate_state}), each parameter has a clear meaning. 
The parameters 
$\epsilon^\uparrow$, $p'$, and $\gamma$ control the high-error plateau, the transition to the power-law region, and the slope of the power-law region.
$\phi$ and $\psi$ control the interchangeability of width and depth with density.
We approximate error over multiple orders of magnitude and over 4,000 configurations of ResNet-20 on CIFAR-10 with just five parameters, indicating we have distilled key information about the behavior of pruning into our functional form.

\textbf{Sources of systemic error and limitations of our approximation form.}
By seeking to minimize the number of parameters in our functional form, we leave some phenomena unmodeled.
In particular, there are two phenomena we have chosen \emph{not} to model that introduce systemic error.
First, the low-error plateau is not a plateau.
Error often improves slightly at high densities before returning to $\epsilon_{np}$ during the transition to the power-law region.
Our model treats the region as flat and treats error as monotonically increasing as density decreases.
This source of error accounts for a bias of $\sim1\%$ relative error in our estimation (Appendix \ref{app:magic-one-percent}).
Second, we model both transitions (between the power-law region and each plateau) with a single shape and the same transition rate.
If we treated each transition separately and used higher-order terms in the rational form, we believe that we could reduce some of the residual error in our estimation at the cost of additional complexity.

\section{Discussion}
\label{sec:conclusions}

Our main contribution is a functional form $\hat \epsilon(\epsilon_{np}, d, l, w, n)$ that accurately predicts the error when pruning members of a network family using IMP.
There are several broader implications of our ability to characterize pruning in this way.
The mere existence of this functional form means there is indeed structure to the way pruning affects error. 
Although prior work \citep{cai2019once} has implicitly relied on the existence of structure for a different pruning method, we are the first to explicitly describe such structure.
This functional form enables a framework in which we can reason conceptually and analytically about pruning.
In doing so, we can make new observations about pruning that are non-obvious or costly to exhaustively demonstrate empirically.
For example, recall our motivating question:

\textit{Given a family of neural networks, which should we prune (and by how much) to obtain the network with the smallest parameter-count such that its error does not exceed some threshold $\epsilon_k$?}

This is an optimization problem: find the configuration of $d$, $l$, and $w$ that minimizes parameter-count $m$ subject to an error constraint: $\smash{\argminB_{w,l,d} m \text{   s.t.   } \hat{\epsilon}=\epsilon_k}$. For ResNets, the parameter-count $m$ is proportional to $(dlw^2)$.%
\footnote{Increasing the depth linearly increases the number of parameters, but increasing the width quadratically increases the number of convolutional filters and thereby the parameter-count.}
Hence, this yields the following optimization problem:
\vspace{-3mm}

$$\smash{l,w,d = \argminB_{l,w,d} lw^2d \text{~~~~~s.t.~~~~~} \epsilon_{np} \left\Vert  \left[l^\phi w^\psi d-jp'(\epsilon^\uparrow/\epsilon_{np})^{1/\gamma}\right]\cdot\left[l^\phi w^\psi d-j p'\right]^{-1} \right\Vert^\gamma=\epsilon_k}$$

This optimization problem is solvable directly without running any further experiments.

Studying this optimization problem reveals a useful insight about---in this case---the CIFAR-10 ResNets.
In the pruning literature, it is typical to report the minimum density where the pruned network matches the error $\epsilon_{np}(l, w)$ of the unpruned network  \citep{han}.
However, our scaling law suggests  this is not the smallest model to achieve error $\epsilon_{np}(l, w)$.
Instead, it is better to train a larger network with depth $l'$ and width $w'$ and prune until error reaches $\epsilon_{np}(l, w)$, despite the fact that error will be higher than $\epsilon_{np}(l', w')$.
This analytic result parallels and extends the findings of \citet{li2020train} on NLP tasks.
However, unlike \citeauthor{li2020train}, our scaling law suggests starting too large is detrimental for the CIFAR-10 ResNets, leading to a higher parameter-count at error $\epsilon_k$.

Figure \ref{fig:misc} (left) illustrates this behavior concretely: it shows the error predicted by our scaling law for CIFAR-10 ResNets with varying widths.
The dotted black line shows the minimal parameter-count at which we predict it is possible to achieve each error.
Importantly, none of the low-error plateaus intersect this black dotted line, meaning a model cannot be minimal until it has been pruned to the point where it increases in error.
This occurs because the transitions of our functional form are gradual.
On the other hand, if we start with a model that is too large, it will no longer be on the black line when it has been pruned to the point where its error reaches $\epsilon_{np}(l, w)$; this behavior occurs because error decreases as a function of the invariant $m^*$ rather than the parameter-count $m$ and because $m \not \propto m^*$.
In Figure \ref{fig:misc} (right), we plot the same information from the actual CIFAR-10 data and see the same phenomena occur in practice.
The difference between the estimated and actual optimal parameter count is no more than 25\%.

Looking ahead, there are many directions for future work.
Further studying sources of systematic error (transition shape and error improvements on the low-error plateau) is a promising avenue for making it possible to \emph{extrapolate} from small-scale settings to large-scale settings (see Appendix \ref{app:more_extrapolations} for a forward-looking discussion). 
Furthermore, while we focus on CIFAR-10 and ImageNet ResNets in the main body, it is important to understand the generality of our functional form for other networks and tasks (see Appendix \ref{app:more_arch_alg}).
Finally, now that we have described the structure of the error of IMP-pruned networks, it will be valuable to study the nature of scaling laws that capture the behavior of the plethora of other pruning methods that achieve different tradeoffs between parameter-count and error.

\chapter{On the Origins of the Scaling Laws } \label{sec:towards_origin}

\epigraph{\textit{``Kepler's laws, although not rigidly true, are sufficiently near to the truth to have led to the discovery of the law of attraction of the bodies of the solar system. ''} \\ --- Isaac Newton, The Principia: Mathematical Principles of Natural Philosophy  }

\glsresetall

\section{Introduction}

We have shown that laws govern the error and loss dependency in both the normal training and pruning settings. As such, these have, already, applicative value --- as they chart the path towards principled design and quantitative reasoning about tradeoffs. 
By analogy, we are now poised to make predictions, armed with kepler's laws.


However, we would like to know what gives rise to these, phenomenological, laws. What are the theoretical origins governing them?
If you will, we wish now to graduate from Kepler's observations, to Newton's laws.

\section{An Approximation Process Intuition}
A recurring hallmark of the scaling laws observed in practice is that they are very well approximated by power-laws before plateauing to a lower error limit. Importantly, it does not seem to originate due to properties of data (e.g. long tail), but rather govern all cases tested.

It is common to view NN as mappings, functions from input to output, and to consider how closely they can approximate the true underlying function assumed to give rise to a given task. Crucially, we have existence results in the form of universal approximation theorems for virtually all modern NN building blocks.

A marginally decreasing error power-law is a hallmark of an approximation process, so we consider, what possible mechanisms could govern the residual error and scale with model and data. 

\section{Possible Origins of the Approximation Error}

Let us consider, using the same terminology introduced in Chapter \ref{sec:Dense}, the dependence of the error on the number of samples $n$ and number of parameters $m$. We recognize three potential categories of error sources (Figure \ref{fig:sources}):




\begin{figure}[t]
\centering
\begin{subfigure}[t]{0.3\textwidth}
    \centering 
        \includegraphics[width=\linewidth,trim={6cm 0cm 7.5cm 0cm},clip]{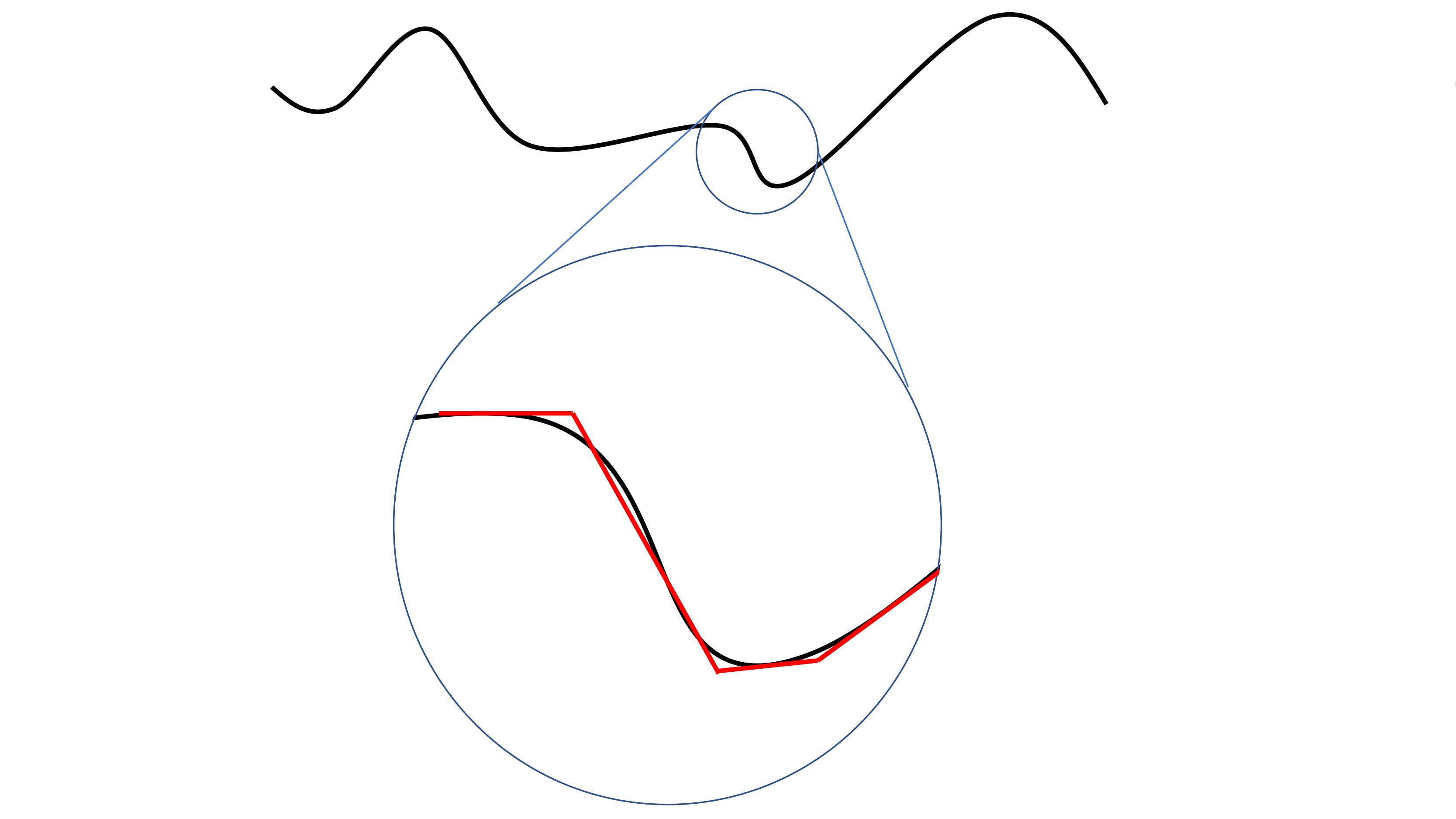}%
    \caption{Realizability error}
    \label{fig:source_realize}
\end{subfigure} \hspace{5pt} 
\begin{subfigure}[t]{0.3\textwidth}
    \centering 
        \includegraphics[width=\linewidth,trim={6cm 0cm 7.5cm 0cm},clip]{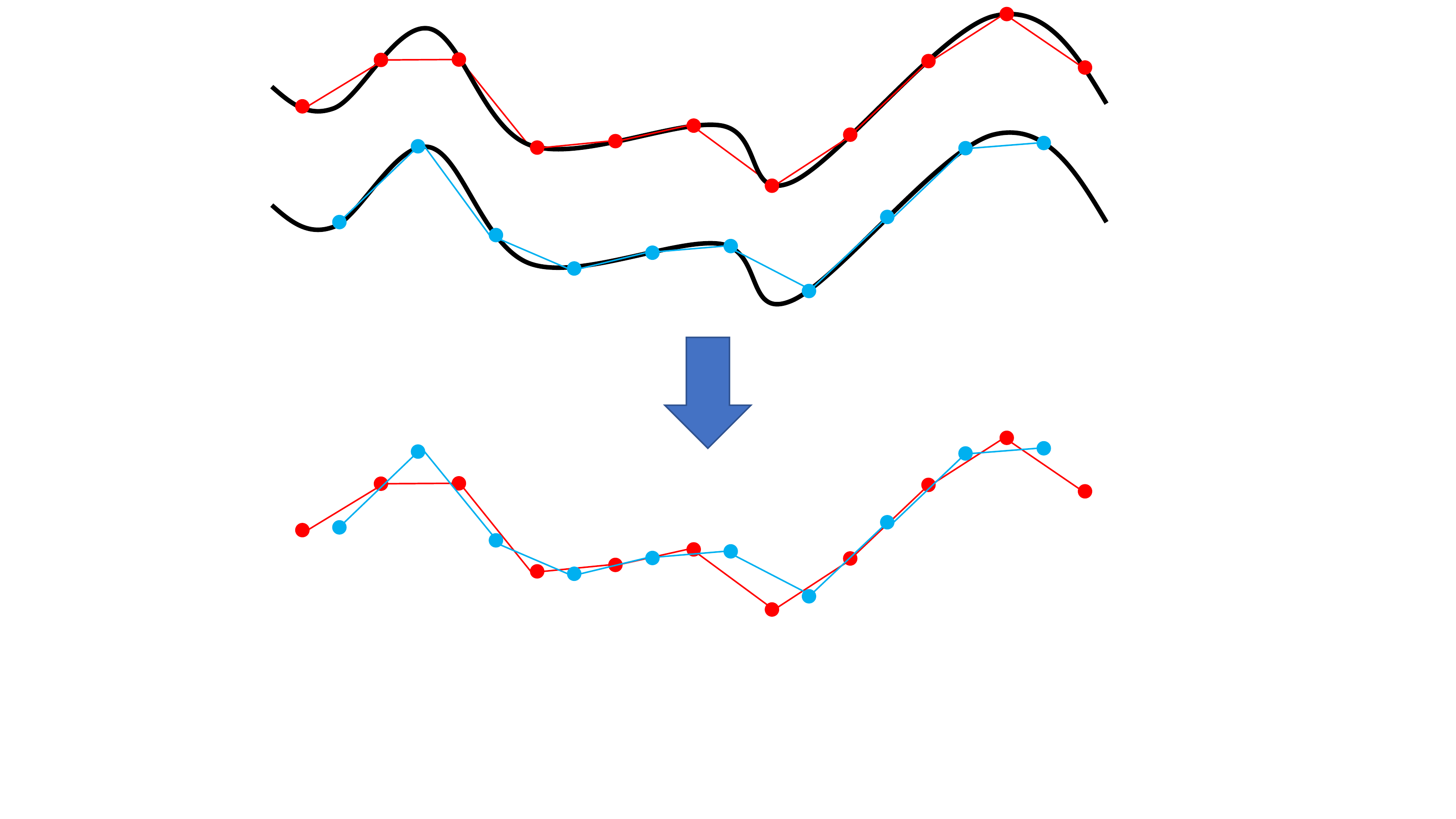}%
    \caption{Uncertainty error }
    \label{fig:source_uncertain}
\end{subfigure} \hspace{5pt}
\begin{subfigure}[t]{0.3\textwidth}
    \centering 
        \includegraphics[width=\linewidth,trim={6cm 0cm 7.5cm 0cm},clip]{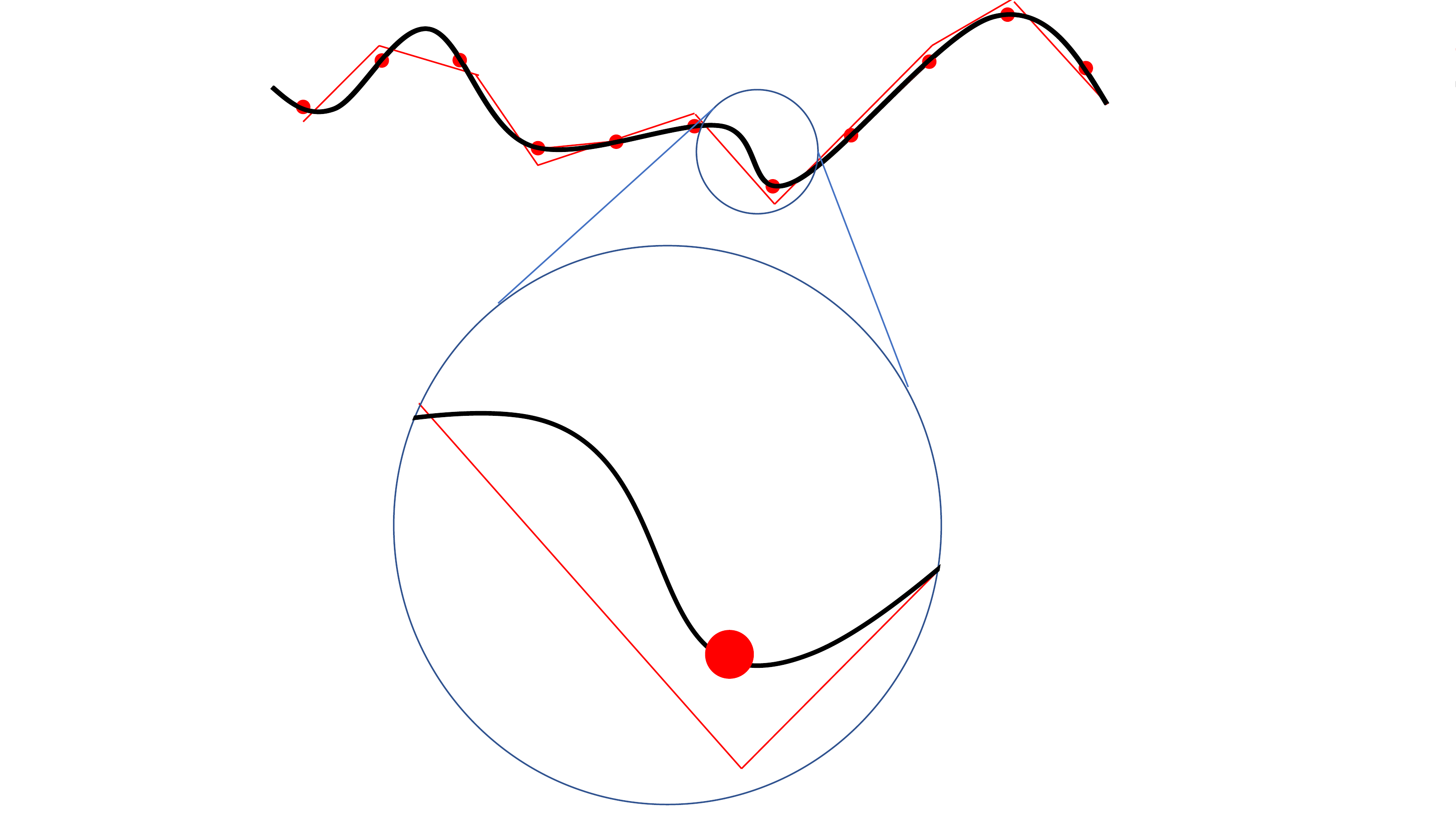}%
    \caption{Learning deficiency error}
    \label{fig:source_defiecient}
\end{subfigure}
\vspace{-5pt}
\caption{Error sources: (a) \textit{Realizability error} stems by definition from a mismatch between the underlaying function and the model hypothesis class, such that any finite size model cannot --- even upon reaching a loss global minimum --- perfectly approximate the underlaying function. Shown is a depiction where the underlying function (black) is smooth while the model (red) is piece wise linear. (b) \textit{Uncertainty error} stems from the stochastic nature of the data sampling process, even when one assumes that a dataset instantiation uniquely determines the model. Shown are two linear interpolators (red, blue) which differ only due to difference in samples (illustrated by a constant offset). (c) \textit{Learning deficiency error} is by definition the error associated by not perfectly interpolating the training set --- depicted for the linear interpolator case by a gap between data sample and model (red).}
\label{fig:sources}
\end{figure}

\begin{itemize}
    \item \textbf{Realizability} (lack thereof): if the underlying function being approximated is outside of the model hypothesis class, then even for infinite data, and assuming convergence to a global, unique, minimum there is by definition a gap between the model and target function for any finite model. This situation is depicted (for the infinite data case) in Figure \ref{fig:source_realize}.
    
    Assuming convergence to a global mininum, in the non-realizable case, the error diminishes with model size (so long as the data is larger than the model size) as additional degrees of freedom in the model allow for closer approximation. Classically, the nature of such a scaling is a power-law (e.g. fitting a harmonic function with a polynomial base --- a Taylor expansion). 
    
    \item \textbf{Uncertainty}: irrespective of whether the underlying function is realizable by the model, for any finite number of samples (even for an infinite model) there is a freedom in the attained model. This freedom arises as a result of the random set of samples drawn and potentially also a freedom in determining how to interpolate through a given set of samples (in the case of non-unique global minima). Alternatively put --- for different random draws of the data, we expect to achieve different models. This is true even if under SGD, for any given draw, a unique global minimum is attained. Thus there is a plurality of perfectly interpolating models which differ in their (generally non-zero) deviation from the true function. This situation is depicted in Figure \ref{fig:source_uncertain}.
    
    The expectation of the difference between models over different draws of the training set is identical to the expected generalization error (consider that the test set is but an additional draw). This difference shrinks with the number of samples until it vanishes at the limit of infinite data.
    
    \item \textbf{learning deficiency}: Consider the ideal case, where there is infinitely many noiseless data and the model is infinitely large. If a learning algorithm interpolates through all the data, then by definition it has reached the true function. Conversely, if it failed to fit the data it is defined as deficiency in the learning procedure.
    This situation is depicted (without loss of generality, for the infinite data case, with highlighted samples out of the sample continuum) in Figure \ref{fig:source_defiecient}.
    
\end{itemize}

    We note that noise also plays a role in the error attained: All above sources apply also in the noise free case --- i.e. when the underlying function is deterministic. However, in the presence of noise, there are two classical effects under this approximative view:
    (1) an ultimate, Bayes error, lower limit on the attainable error.
    (2) a, scalable with data, effect on attained approximation. This mechanism is akin to averaging in classical settings.

\section{The Realizable Case}
\epigraph{\textit{``In so far as a scientific statement speaks about reality, it must be falsifiable: and in so far as it is not falsifiable, it does not speak about reality.''} \\ ---  Karl R. Popper, The Logic of Scientific Discovery }

\begin{figure*}[t]
  \centering
    \includegraphics[width=0.8\linewidth,trim={3cm 0cm 3cm 0cm},clip]{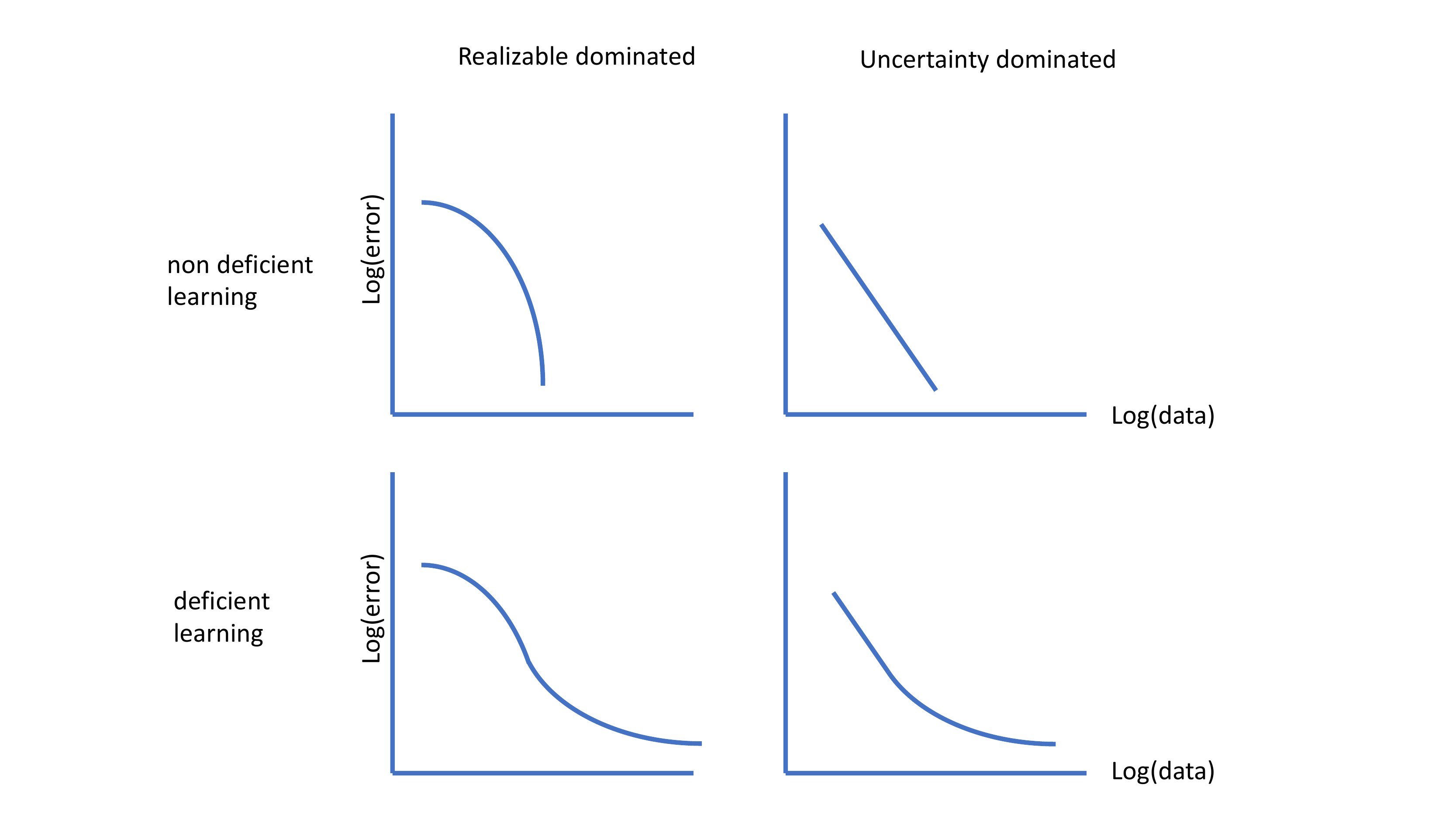}
    \caption{Error dominance mechanism expected qualitative effect on error curves in the realizable case}
    \label{fig:dominant_errors}

\end{figure*}

\glsresetall
In this section we ask, do the scaling laws hold in the realizable (and noiseless) case.
This is an informative setting to query as it allows us to gain insight into all three error sources above (while synthetically eliminating noise as a source of error). Consider what different combinations of dominant error sources may imply on the error behavior in this case (see Figure \ref{fig:dominant_errors}): (1) The absence of both uncertainty and deficiency error sources (as well as the realizability --- by construction) would result in an error function (as a function of data) characterized by an abrupt vanishing of the error, once a threshold number of samples is reached (exceeding the underlying function degrees of freedom). (2) In the presence of uncertainty error and absence of deficiency error, we expect a scaling with data which asymptotically approaches zero as discussed above --- as the marginal benefit of any additional sample diminishes. (3) Finally, in the presence of deficiency errors, we may expect a plateau in the error as learning breaks down.

These possibilities are depicted in Figure \ref{fig:dominant_errors}

\subsection{Experimental Setup}
In order to examine the realizable case we construct a teacher-student setup, where the student architecturally contains the teacher architecture. 
We are further interested in exploring this case for natural data, such that minimal questions regarding the applicability of these findings arise and to establish a strong comparison to our previous results.

\begin{figure*}[t]
  \centering
    \includegraphics[width=1\linewidth,trim={0cm 0cm 0cm 0cm},clip]{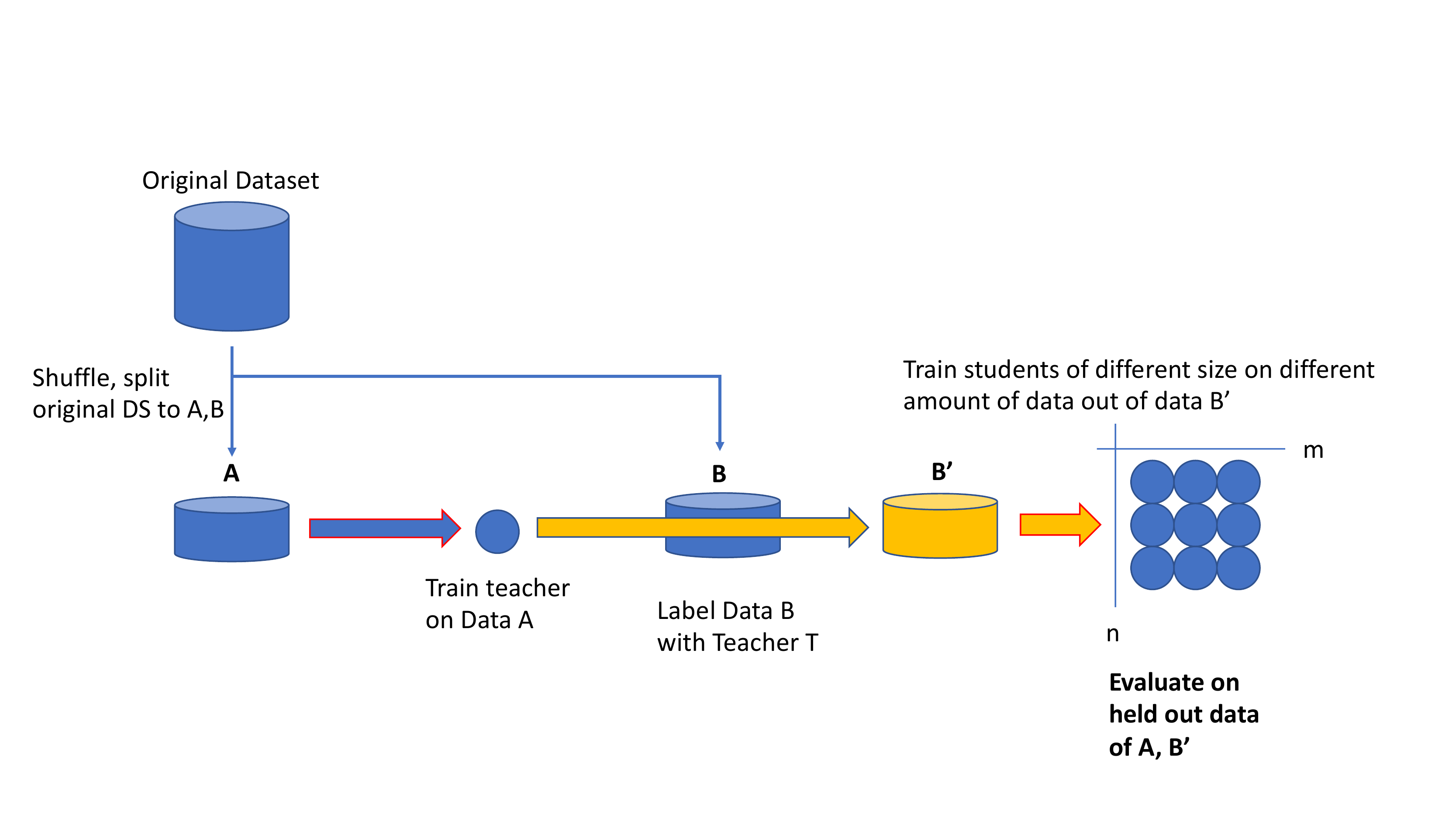}
    \caption{CIFAR10 teacher student setup. CIFAR10 is split into two disjoint datasets $A,B$. The teacher is trained on $A$ and labels $B$, producing $B'$. Students of different sizes are then trained on different subsets of dataset $B'$ }
    \label{fig:realize_setup}

\end{figure*}

In order to do so, we train a (small) teacher on a subset of CIFAR10 --- let us denote this original dataset as dataset $A$.
We then allow that teacher to label a disjoint subset (dataset $B$) of the CIFAR10 dataset arriving at a teacher labeled dataset: dataset $B'$. We train students of differing sizes on different subsets of $B'$ and evaluate, similar to chapter \ref{sec:Dense}, the error scaling. We use the same WRN model configurations as in chapter \ref{sec:Dense}. The students are ranged over width fractions $\frac{1}{16},\dots,2$, where the teacher is taken to be width fraction $1/4$, such that it is $8$ times as narrow (contains $\times 64$ less parameters) than the largest student. This procedure is depicted in Figure \ref{fig:realize_setup}

Through the above setup we effectively re-create the CIFAR10 experiment from Chapter \ref{sec:Dense}, but now where the data is labeled by a noiseless teacher which is assured to be realizable by the (student) model. The effect of the generalization error (student error on the teacher labeled data) with dataset size is accomplished by subsampling B' and the effect of model size is, similarly, measured through the variation of the student width.

\subsection{Results}
\begin{figure}
\centering
\begin{minipage}{0.35\textwidth}
    \includegraphics[width=\linewidth]{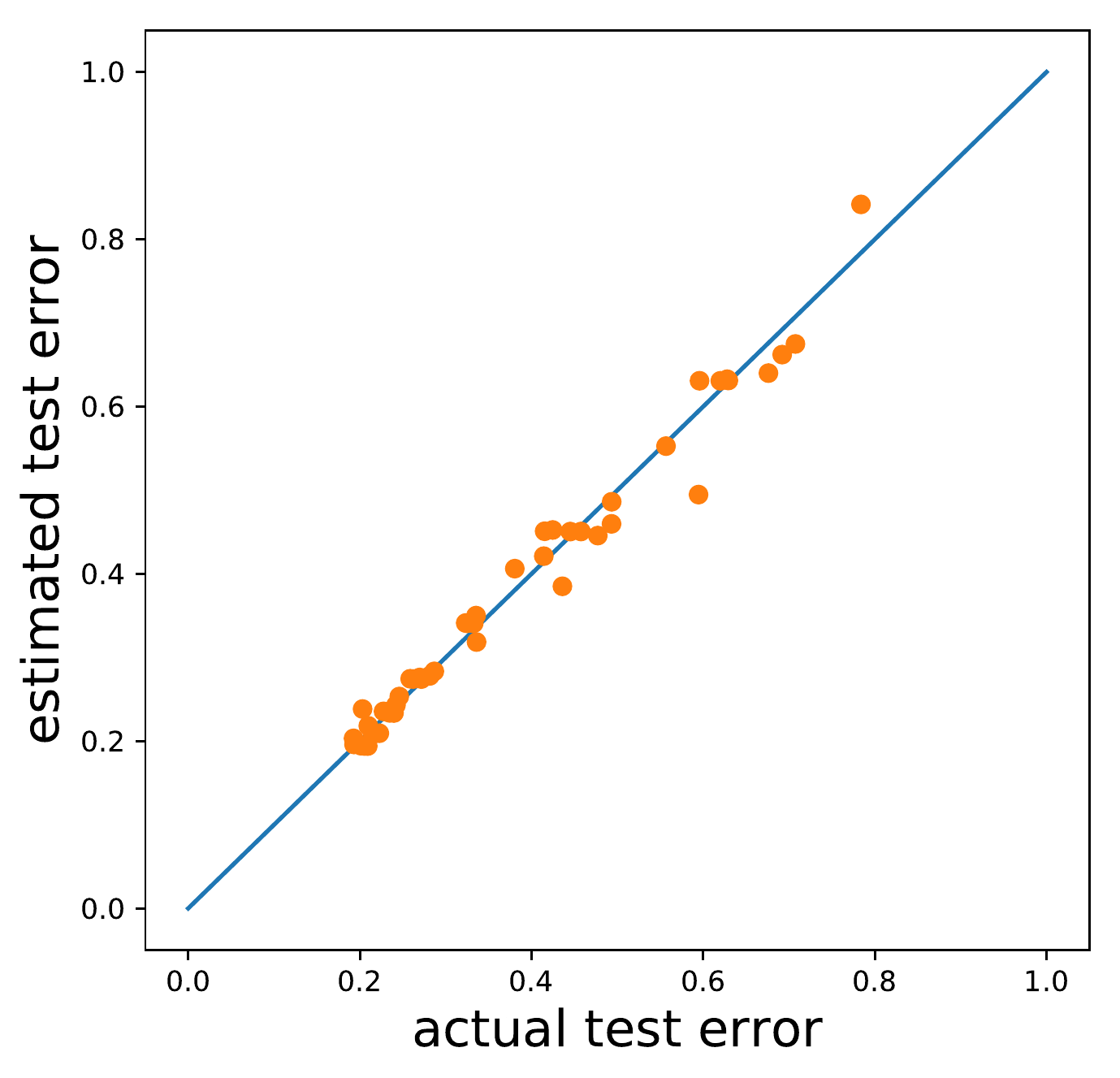}
\end{minipage}%
\begin{minipage}{0.45\textwidth}
    \includegraphics[width=1\linewidth,trim={0cm 0cm 0cm 1.4cm},clip]{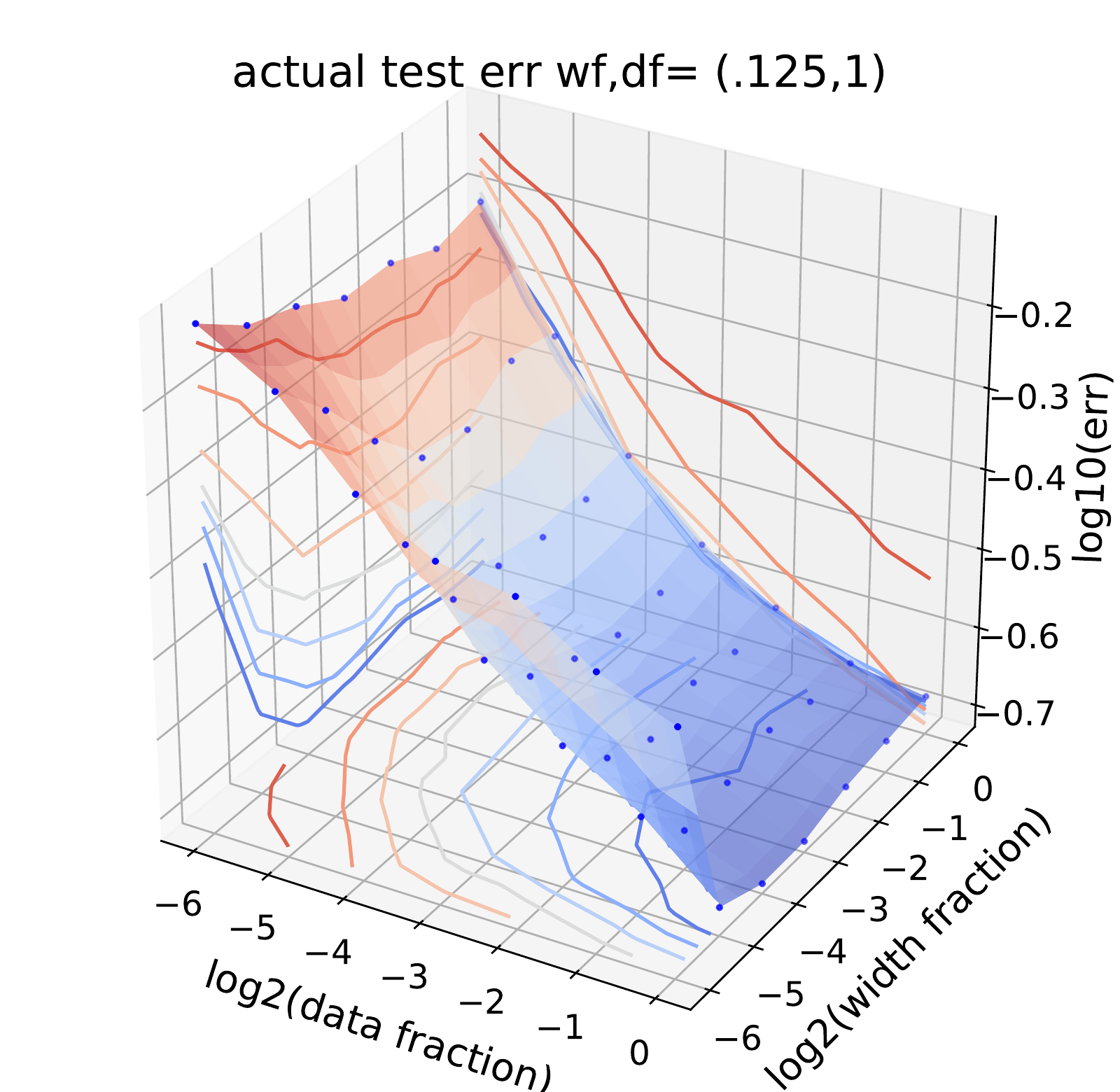}
    
\end{minipage}

\caption{Experimental results of the teacher-student experiment for CIFAR10. The teacher has a width factor of $\frac{1}{8}$ relative to the largest student, such that the three smallest students can not generally realize the teacher, while the four largest ones can. (right) The error landscape of the students on the teacher labeled data. Notably the error plateaus prior to student reaching teacher size. The error also plateaus as a function of data.(left) error agreement with the functional form of the scaling law found in Chapter \ref{sec:Dense}}
\vspace{-3mm}
\label{fig:realize_surface}
\end{figure}

Figure \ref{fig:realize_surface} shows the resultant error landscape on the teacher labeled data as a function of the student and dataset size. 

We observe that these results are well approximated by the same functional form from Chapter \ref{sec:Dense}. Quantitatively Figure \ref{fig:realize_surface} (left) shows the estimated error resultant from a fit using Equation \ref{eq:envelope}.

Specifically we observe:

\textbf{Observation/insight 1: } There is a characteristic region where a power-law approximately holds. Recall, this characteristic behavior as a region appearing as a straight line in the log-log scale. Notably, as a function of data, both smaller and equal-or-larger than teacher students exhibit this dependency.

\textbf{Observation/insight 2: } The error exhibits no threshold data size phenomena upon which it vanishes. 

\textbf{Observation/insight 3:} Further, a plateauing behavior characteristic of the scaling laws observed in Chapter \ref{sec:Dense} is observed. The plateauing is in terms of both student size and amount of data --- as both are increased.

Observations 1,2,3 combined, allow us to gain insight into the error sources for this case.
The presence of a (well approximated) power-law dependence, under normal circumstances can be attributed to a potential model-mismatch (i.e. a (lack) of realizability). But in this case, that source of error is by construction excluded for all students equal or larger than the teacher. This leaves us with an interpretation of the power-law as arising by the uncertainty error source (or another source which we may have failed to account for). In figure \ref{fig:sources} we are situated at the right column. Further, by design, this experiment is noiseless, meaning that the observed plateau cannot be attributed to a dataset Bayes error, but rather resultant from a learning/optimization failure. We find ourselves in the lower right quadrant of Figure \ref{fig:sources}.

It is worth stressing that the global minimum in the realizable case (corresponding to the left hand side of Figure \ref{fig:sources}) gives a vanishing error at a finite (dataset size) threshold, in contrast to the interpretation of a power-law asymptotically diminishing error --- even at the absence of a plateauing of the error. As such, there is a gap between the global minimum and the error in practice. This gap is dominated , in this case, by the uncertainty error until dominated by the learning deficiency.

We have focused here on the context of the realizable case revisiting the CIFAR10 dataset and models shown to exhibit the scaling laws in Chapter \ref{sec:Dense}. We have reasons to suspect that this picture of error-dominant sources extends to the general case. We have already seen that the scaling laws themselves, with the associated power-law dependencies, hold in a wide array of circumstances in Chapter \ref{sec:Dense} and \ref{sec:Pruning}. In those cases, without loss of generality, both the realizability error source and uncertainty error source may be at play. However, for arbitrarily large models, realizability error becomes arbitrarily small. In other words, we suspect that the uncertainty error source as the (power-law) dominated error source in practice, until plateauing. 


\section{Posterior Work}




Posterior work to the work done here has focused primarily on the power-law region and has taken a similar approach in exploring a classical approximation explanation. \cite{Sharma2020ANS} demonstrate the connection between the data manifold intrinsic dimension corroborating this view and importantly , following work \cite{Bahri2021ExplainingNS} further takes this intuitive notion and adds much rigor and breadth, formulating this approximation theory and showing it to hold over four regimes depending on the relative amount of data and width of model. Critical to this approximative theory is an explanation of the nature of what we have defined an uncertainty error source. 

However, the source of the learning deficiency exhibited here of noiseless natural data, remains unexplained within the context of the above theory and likely violates its basic assumptions of perfect interpolants.
Interestingly and consistently, experiments conducted in \cite{Sharma2020ANS} with synthetic teacher student random feature models also exhibit plateauing effects.

\section{Discussion}
We have intuited an approximation view of the scaling laws and identified three mechanisms of potential error contribution. For a constructed teacher-student setup, we have ruled out a (lack of) realizabilty error mechanism as the dominant error source governing the scaling laws and identified the power-law region as resulting from an uncertainty mechanism associated with the potential variations between models under different sample draws (and potential freedoms under a given draw). This view is consistent and was later corroborated and expanded in theoretical work that followed our work.

However, we have also found an error source which is outside the scope of that theory --- a learning deficiency mechanism. This mechanism is most interesting from a practical perspective, because it is where the error practically plateaus that the performance limit lay. 

We stress that the results are obtained under the realizable case --- where both the model contains the true function (is equivalent to an infinite model in term from an approximation perspective) and there is no noise. In other words --- it seems that deep learning is not, in practice, and through the eyes of the above evolving approximative theory view, noise limited nor realizability (model mismatch) limited.



This is quite embarrassing (and encouraging), implying that deep learning in its current form may actually be far from the lower limits of attainable error, where the lower limit would be realizable-error and noise (when present) dominated. 

These findings point not only to a fundamental gap but also chart a path to how one may close it. This is the subject of the next, more conjectural chapter.










\chapter{Towards Nyquist Learners} 
\label{sec:towards_nyquist}

\epigraph{\textit{``Information is the resolution of uncertainty. ''} \\ --- Claude Shannon  }

\glsresetall

\section{Recap}
\begin{figure*}[t]
  \centering
    \includegraphics[width=1\linewidth,trim={1.85cm 5.5cm 1.5cm 3cm},clip]{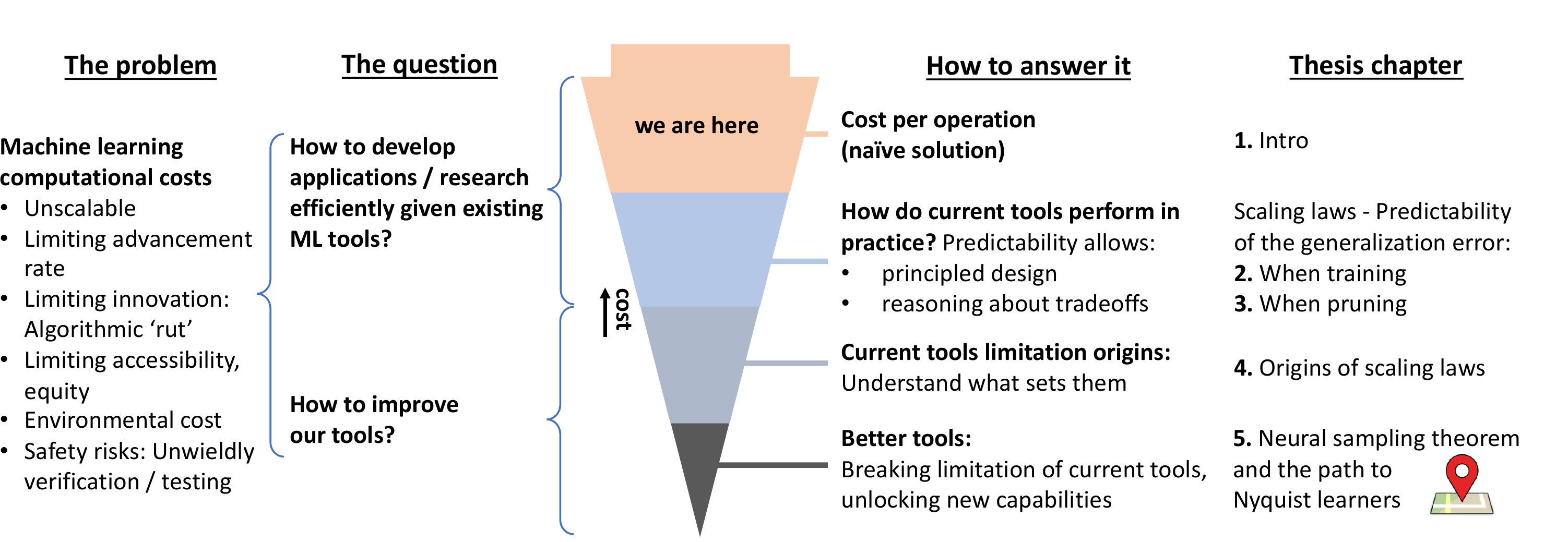}

    \caption{Thesis roadmap}
    \label{fig:roadmap_conc}

\end{figure*}

Recall our overarching organizing path depicted in Figure \ref{fig:roadmap_conc} and brought here again for convenience. 

In Chapter \ref{sec:Dense} and \ref{sec:Pruning} we have shown that power-law scaling laws govern training and pruning respectively. 
These two scenarios in tandem contribute to the practical settings of development and deployment --- i.e. the training of models for a specific business application and then the optimization of models for deployment to be called upon during inference (the dominant use case as evaluated by frequency or industry expenditures). 
The notion of predictability as embodied in the scaling laws found is necessary for principled design and reasoning about trade-offs through the eyes of a well understood performance envelope as we discuss in both cases.
In Chapter \ref{sec:towards_origin} we delineated the possible sources of error giving rise to the scaling laws, and showed that two dominant sources are at play (and not a third --- realizability / hypothesis class mismatch) through the construction and analysis of the realizable case. Importantly these two sources are an uncertainty error source and a learning deficiency error source. 

In what follows we consider a conjectural path to eliminate the dominant error source of uncertainty drawing on additional tools from the classical intersection of information theory and signal processing. 

\section{A Neural Sampling Conjecture}
Let us circle back to our depiction of the uncertainty intuitive view and consider, what would be needed in order to eliminate it? Intuitively, could we reach a situation where the approximation (model) does not depend on the particular data draw? Further, could we reach a situation, where irrespective of data draw, the difference between the model and true function being approximated could practically vanish?

In the classical context of signal processing we have the celebrated Nyquist–Shannon sampling theorem \cite{1697831} (first discovered by Whittaker \cite{whittaker_e_t_1915_1428702}).
The theorem specifies that there is a minimal sample rate sufficient and required for the perfect reconstruction of a band limited function. In particular, this minimal frequency is twice the bandwitdh of the function. Famously, the reconstructing kernel is the \textit{Sinc} function in the Cartesian basis case.
This sampling theorem has been shown to hold also in expectation by Landau \cite{10.1007/BF02395039} when the samples are drawn in random, provided that the spectral occupancy locations are known --- i.e. the non-uniform sampling case. Further, it is also known that perfect reconstruction is possible for a bandlimited function without knowledge of the location of the occupied spectral components --- with an additional factor of 2 required in the average sampling rate \cite{Mishali_2009}. 

By extension, more generally, over a given manifold, a spectrally (as defined with respect to some manifold spanning basis) limited function defined above that manifold, is similarly suspected to be perfectly reconstructable if sampled at a finite Nyquist frequency.


We thus arrive at a \textit{Neural Sampling Conjecture} (informal):

\textit{Assume data samples drawn iid from a compact manifold $x\in \mathcal{M}$, with targets (e.g. labels) given by a spectrally limited (the analog to the 1d bandwidth limited) function $y=g(x)$.
We conjecture that for a sufficiently large, finite, dataset of size $n_{Nyquist}$ perfect reconstruction of $g$ is possible over the entire manifold; And that, furthermore, the size of the dataset is given such that it creates an average sampling frequency greater or equal than the Nyquist frequency associated with the spectral bandwidth of $g$.
}


The basic insight is that over a compact manifold, the average sampling rate grows with the size of the dataset and that the key ingredient --- conjectured necessary and sufficient for perfect reconstruction (or equivalently,  the elimination of the uncertainty error source) --- is the spectral limiting of the generating function $g$. 

\section{The Bandlimited Data Hypothesis and the Existence of Nyquist Learners }
We hypothesize that natural problems can be well described as arising from band limited functions over compact (data) manifolds.

Let us dwell on the intuition for this hypothesis. We intuitively do not expect the function $g(x)$ describing the targets associated with data samples $x$, to vary wildly between adjacent samples. This smoothness requirement was already implied in Chapter \ref{sec:towards_origin} and explicitly in the derivations of \cite{Bahri2021ExplainingNS}. However, intuitively for natural tasks, we expect a stronger condition to hold, namely that $g$ be bandlimited over the manifold, as is prevalent in physical and naturally arising process. 
Perhaps, the less intuitive part of this hypothesis is the notion that within a task, the data may be presented as situated on a compact manifold. This is in fact a common assumption frequently employed but differently phrased: the assumption that the data is drawn from a distribution of finite support.
For certain tasks this is perhaps more readily envisioned, e.g. for visual tasks where the data may be viewed as samples from the continuous visual world. In the language domain, or other sequence domains (e.g. genome sequences, protein coding, etc.) this may be less intuitive as compositionality typically plays a more blatant role. However, this manifold interpretation specifically, rests on what has brought us here. Through the scaling laws, and approximative theory that explains them well and depends intimately on the manifold intrinsic dimension \cite{Sharma2020ANS,Bahri2021ExplainingNS}, we have strong evidence for this interpretation. 

To the extent that this hypothesis holds true, we have met the conditions of the above Neural Sampling Conjecture, under which perfect reconstruction becomes possible from a finite Nyquist dataset.

Under these conditions the role of function realizability becomes --- deliberately --- dominant:
Upon reaching the critical Nyquist satisfying dataset size $n_{Nyquist}$ , perfect reconstruction can, in principle, be attained in only one of two ways:
\begin{itemize}
    \item Realizable case: By a model that can realize $g$ at finite size. 
    \item Non-realizable case: By a model that cannot realize $g$ at any finite size, but then approaches it as it approaches infinity.
\end{itemize}

For any finite model in the non-realizable case, we will thus be solely model limited in our error (barring for learning deficiency). In other words, beyond the Nyquist data size, further data would not be needed (except for the averaging of noise) --- only a larger model to reduce error.


However, even if indeed the bandlimited data hypothesis holds, current learning algorithms are ill equipped to capitalize upon it. They do not make use of the stronger band limit property and --- case in point --- suffer from the uncertainty noise source which gives rise to the power-laws established in the preceding chapters. 

We thus need to modify our learning algorithms to explicitly limit the spectral bandwidth of the model over the data manifold. Such models would usher a novel situation in deep learning --- where by the Neural Sampling Conjecture, finite data would, in principle, suffice for a given ('bandlimited') task. As importantly, in such a situation, the dominant error source (at high signal to noise ratio) would be the realizability error. 

We broadly define such algorithms as Nyquist learners: A Nyquist learner is an algorithm that limits the (resultant) model spectral bandwidth.

We consider it an open, exciting, and within grasp, question as how to implement them. 

\appendix
\chapter{Appendix A: Dense Scaling further Details}
\section{Datasets and Models} \label{app:data-models}

\subsection{Image Classification}
\subsubsection{Datasets}
We evaluated our predictions on several popular image classification datasets: 
ImageNet \citep{russakovsky2015imagenet}: a large-scale recognition benchmark consisting of natural images of 1000 object categories with 1.28M training images spread roughly uniformly over the categories. It has 50K validation and 100K testing images. It has been the most popular large-scale benchmark for image classification methods for the better part of the last decade. 
CIFAR10/100 \citep{krizhevsky2009learning}: 60K natural RGB images of 10 classes (100 for CIFAR100) with a train/test split of 50K/10K. 
For each of the following datasets, we use the version collated, resized, and split into train/validation/test sets by \citet{rebuffi2017learning}. 
DTD \citep{cimpoi2014describing}: a texture database of 47 categories and 5640 images. Aircraft \citep{maji2013fine}: 10K images of 100 different aircraft classes. UCF101 \citep{soomro2012ucf101}: originally a video action recognition dataset, converted using the method of \citet{bilen2016dynamic} into a single image per video. It contains 13,320 images of 101 action classes.

\subsubsection{Models}

We experiment with four models for image classification. We use different variants of the popular ResNet architecture \citep{he2016deep} in the main experiments. 
For ImageNet we use ResNet-50 and build on the code from the PyTorch framework \citep{paszke2017automatic} to vary the model width. For all other datasets we use  WRN-44-16 \citep{wu2016wider} of varying widths, modified from the implementation of \cite{hoffer2018fix}. 



Scaling the models' width is performed by multiplying the number of channels in each convolutional layer and the width of the hidden linear layers by a constant factor and rounding to the nearest integer.
The ranges of width scales (and data scales) for the main experiments are detailed in Table \ref{tab:stats-vision}. 

In section \ref{sec:arch_optim_var}, we perform width scaling for two additional architectures, VGG16bn \citep{simonyan2014very} and DenseNet (L=40, k=32) \citep{huang2017densely}. The VGG and DenseNet models were also modified for width scaling from the implementation of \cite{hoffer2018fix}. The model scales in this case are  $4^{-k} $, $0 \leq k \leq 5$, for both VGG and DenseNEt.

Depth-scaling, in the CIFAR10 case (section \ref{sec:width_and_depth_scaling}), is performed by appending extra layers within each block.






\subsubsection{Training}
In the main experiments, training is done via SGD with a momentum of 0.9, weight decay of 1e-4 and initial learning rate of 0.1.
For ImageNet we train for 90 epochs, decreasing the learning rate by a multiplicative factor of 0.1 after and 30 and after 60 epochs. We use a batch size of 16.
For all other vision datasets we use a batch-size of 128. We begin training with a learning rate of 0.1, run for 200 epochs, and reduce by a multiplicative factor of 0.1 after 80, 120, and 160 epochs. 

For the VGG and DenseNet experiments on CIFAR100 in section~\ref{sec:arch_optim_var}, we train with both SGD and Adam optimizers. 
We train VGG for 170 epochs and Densenet for 300 epochs.
Adam hyperparameters are default, with an initial learning rate of 1e-3.
When training with SGD, we retain initial learning rate, batch size, momentum, and weight-decay, as in the main experiment (at 0.1, 128, 0.9, and 1e-4 respectively) and follow standard stepped learning rate schedules: 
For VGG, learning rate multiplicative factor of 0.1 after 80, 120, and 160 epochs; For 
DenseNet, learning rate multiplicative factor of 0.1 after 150 and 225 epochs.


\subsection{Language Modeling}

\subsubsection{Datasets}
We evaluate on several datasets commonly used for (word-level) language modeling: Penn Treebank~\citep{mikolov2010recurrent}, WikiText-2~\citep{bradbury2016quasi}, and WikiText-103~\citep{merity2016pointer}. 
The PTB is a relatively small language modeling dataset of news texts, with a vocabulary of 10K unique words and about 900K/70K/80K training/validation/test words. 
WikiText-2 is drawn from Wikipedia articles and it is both larger and richer, with a vocabulary of 33K words and 2M/210K/240K training/validation/test words. 
WikiText-103 is also based on Wikipedia, but larger still, with a vocabulary of 270K words and 100M training words (and the same validation and test sets as WikiText-2).  


\subsubsection{Models}
We experiment with two standard models for language modeling: Transformer-XL~\citep{dai-etal-2019-transformer} and AWD-LSTM~\citep{merity2018regularizing}. 
Transformer-XL is a recent language modeling architecture that is based on transformer self-attention~\citep{NIPS2017_7181}, but modified to better learn dependencies beyond a fixed length by adding a segment-level recurrence mechanism. It has achieved state-of-the-art results on multiple benchmarks. We use the official PyTorch implementation\footnote{\url{https://github.com/kimiyoung/transformer-xl}} with their base configuration: 16 layers, embedding size of 410, inner dimension of 2100 in the fully-connected layers, and 10 attention heads. Training is done with Adam. See the implementation for other details. For scaling experiments, we decimate the inner dimension. We use Transformer-XL for WikiText-103. 

AWD-LSTM is a long short-term memory~\citep{hochreiter1997long} language model with adaptive weight averaging. We use the  official implementation\footnote{\url{https://github.com/salesforce/awd-lstm-lm}}  with the recommended configuration:
3 layers, embedding size of 400, and hidden state size of 1150. Training is done with SGD. We use AWD-LSTM for PTB and WikiText-2 and follow the recommended settings for these two datasets.  
For scaling experiments, we decimate the hidden state size.

\clearpage

\section{Error Estimation Experiment}

\subsection{Experimental Details}
\label{app:fit-exp}

In the experiment described in section~\ref{sec:fit}, 
we fit a least squares regression model to find the best parameters minimizing the divergence $\delta(m,n)$ - evaluated at configurations $m,n$ as in Table 
\ref{tab:stats}:  
\begin{equation*} \vtheta^* = \argmin_\vtheta 
    \sum_{\substack{n, m}} | \delta(m,n;\vtheta) |^2
\end{equation*}
We quantify the quality of the fit by 
the mean $\mu$ and standard deviation $\sigma$ of the 
fitted divergence by performing standard 10-fold cross validation over all points $(m,n)$ with confidence intervals reported as $\pm 1$ std over the folds.

\subsection{Found Theta Values} \label{app:theta}

\npdecimalsign{.}
\nprounddigits{2}

\begin{table}[h]
\centering
\caption{Optimal values of $\vtheta$ as found by the least squres regression fitting the functional form.}
\begin{subtable}[b]{\textwidth}
\centering
\caption{Image classification (fitting top 1 error).}
\begin{tabular}{l n{2}{2} n{2}{2} n{2}{2} n{2}{2} n{2}{2} }
\toprule
 & \multicolumn{1}{c}{$\alpha$} & \multicolumn{1}{c}{$\beta$} & \multicolumn{1}{c}{$b$} & \multicolumn{1}{c}{$c_\infty$} & \multicolumn{1}{c}{$\eta$}  \\ 
 \midrule 
ImageNet & 0.75403879 & 0.61131518 & 0.75575083 & 3.62934233 & 18.50376969 \\ 
CIFAR10 & 0.655043783 & 0.534102925 & 5.87E-02 & 7.14E-14 & 19.7701518 \\ 
CIFAR100 & 0.70403326 & 0.50562759 & 0.14727227 & 0.70969734 & 6.92618391 \\ 
DTD & 0.400319211 & 1.16231333 & 4.30E-05 & 1.27E-09 & 0.846839835 \\ 
Aircraft & 1.10233368 & 0.831731092 & 3.47E-03 & 5.16E-10 & 1.12529537 \\ 
UFC101 & 0.933547255 & 0.537578077 & 4.68E-02 & 1.16E-09 & 2.98124532 \\ 
\bottomrule
\end{tabular}
\label{tab:theta-vision}
\end{subtable}
\begin{subtable}[b]{\textwidth}
\centering
\caption{Language modeling (fitting cross entropy loss).}
\begin{tabular}{l  n{2}{2} n{2}{2} n{2}{2} n{2}{2} n{2}{2} n{2}{2} }
\toprule 
 & \multicolumn{1}{c}{$\alpha$} & \multicolumn{1}{c}{$\beta$} & \multicolumn{1}{c}{$b$} & \multicolumn{1}{c}{$c_\infty$} & \multicolumn{1}{c}{$\eta$} & \multicolumn{1}{c}{$\epsilon_0$} \\
\midrule
PTB & 0.80962791 & 0.34315027 & 0.14690378 & 4.99807364 & 6.27494232 & 6.09699692 \\
WikiText-2 & 1.00822978 & 0.21667458 & 0.99145936 & 8.23497095 & 10.37612973 & 6.21205331 \\
WikiText-103 & 0.73505031 & 0.55718887 & 0.32914295 & 9.03598661 & 16.33563873 & 6.59633058 \\
\bottomrule
\end{tabular}
\label{tab:theta-language}
\end{subtable}
\label{tab:theta}
\end{table}

\clearpage

\section{Additional Error Landscape Measurements and Estimations} \label{app:landscape}

In this appendix, we provide error landscape measurements and estimations for all datasets, corresponding to the experiment in section \ref{sec:fit}. 
The results are shown in 3D graphs similar to \figref{fig:landscape-3d}. 
In each such graph, the z-axis is the logarithm of the generalization error as a function of two independent variables: the model size $m$ and the data size $n$. 
%
%

The 3D graph is deliberately portrayed in log-log-log scale, as we cover a very large range of data scales and model scales and a correspondingly wide range of errors. This view is a useful one when one wishes to evaluate both large dynamic ranges (simultaneously both very large and very small values) and is especially vivid in portraying power-law like dependencies; a power-law naturally forms a straight line in a log-log view.

%
%
%

In each figure, subfigure (a) shows  the measured error landscape is in log-log-log scale, where each point (blue dot) is the error resulting from training with a model/data configuration $m,n$. Subfigure (b) shows the best-fit estimated error landscape. 
The surface is a linear interpolation between the points, which is then projected on the model-error $(m,\epsilon)$, data-error $(n,\epsilon)$, and model-data $(m,n)$ planes. 
The contour plots on each one of these planes are the projections of the error landscape surface, and are useful in considering the behavior of the surface when holding one dimension constant. 

\begin{figure}[h] 
\centering
    \begin{subfigure}[b]{0.45\linewidth}   
        \centering 
        \includegraphics[width=\linewidth]{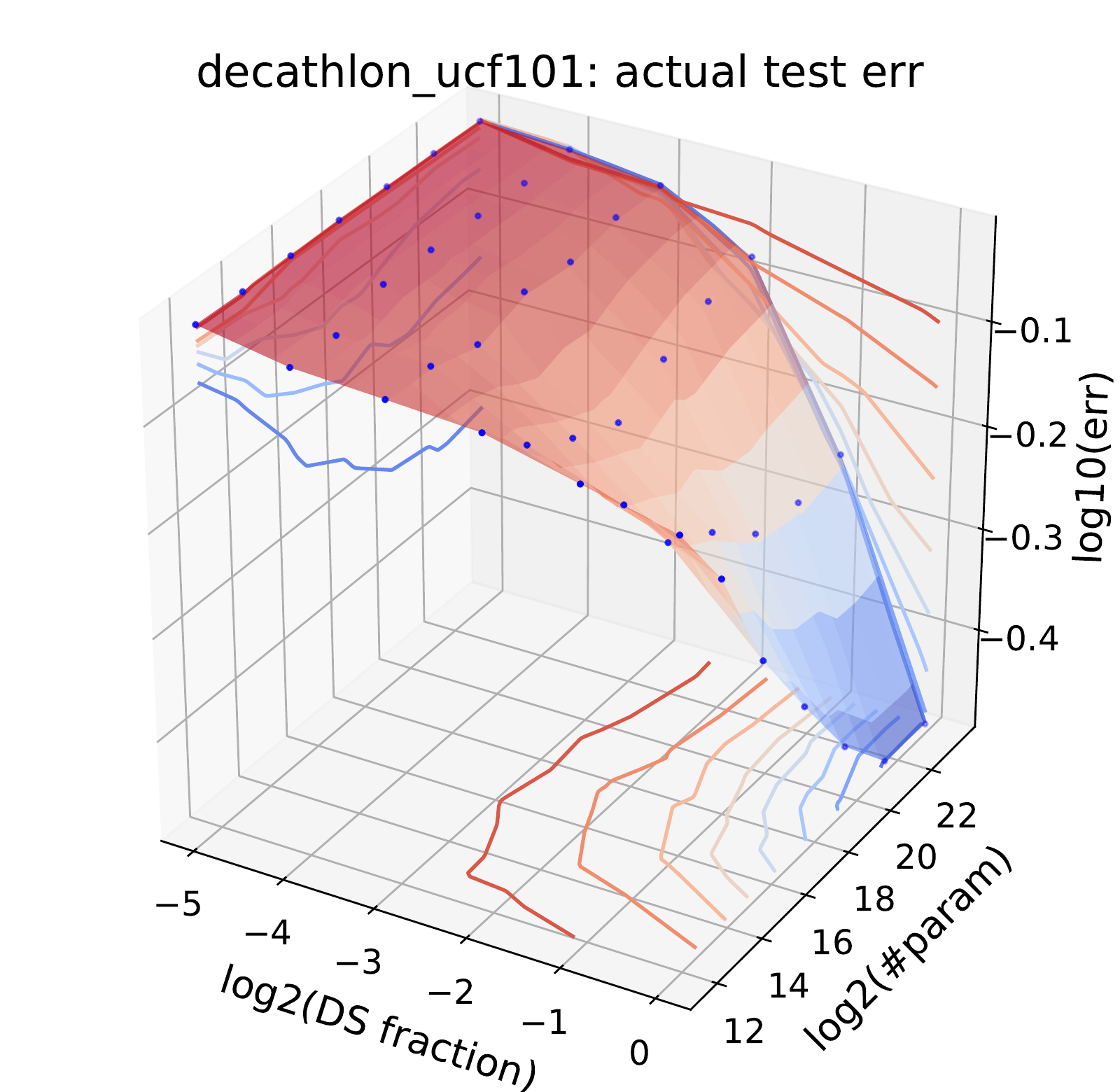}
        \caption{Actual error landscape.}    
    \end{subfigure}\hfill 
    \begin{subfigure}[b]{0.45\linewidth}   
        \centering 
        \includegraphics[width=\linewidth]{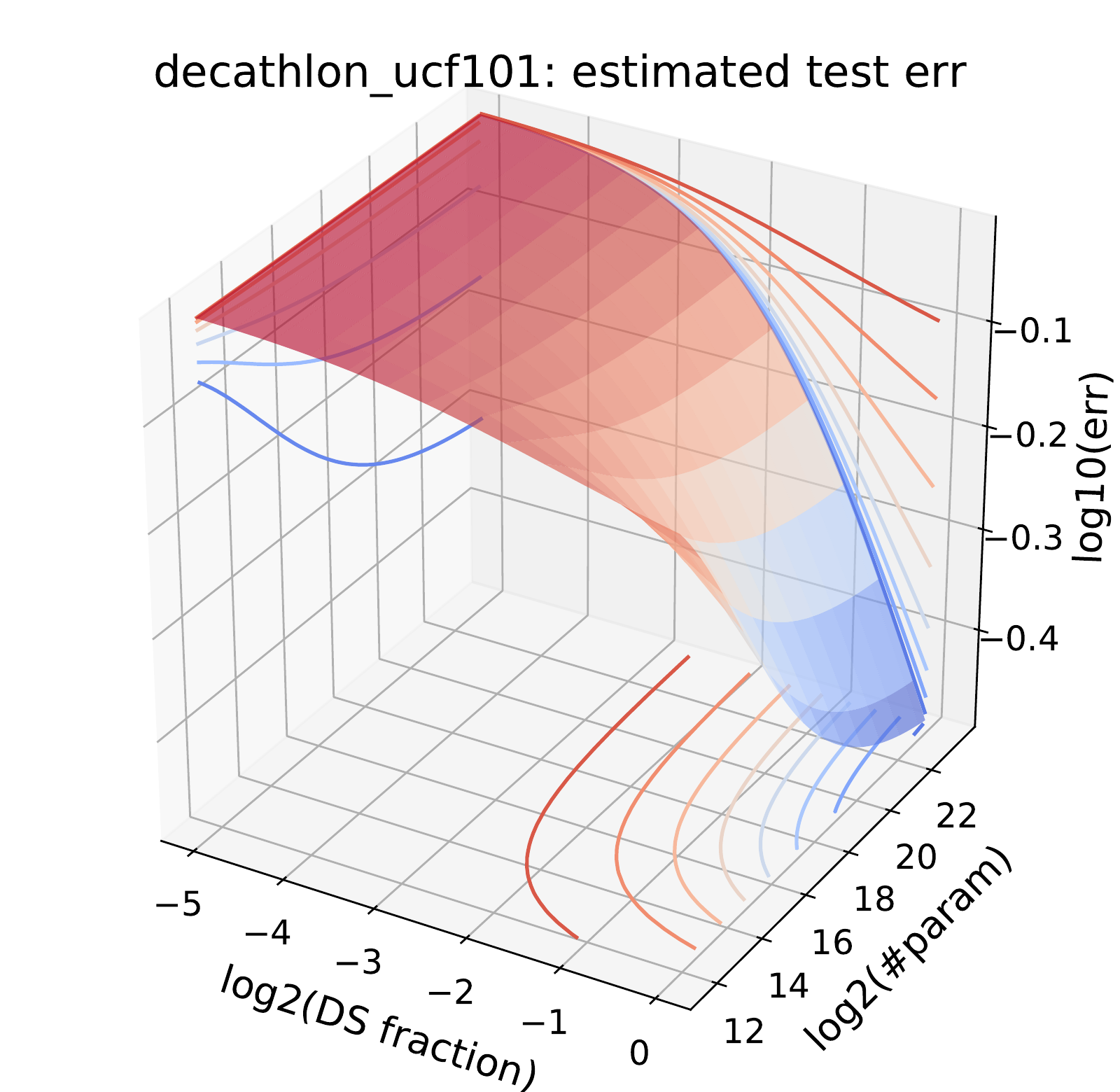}
        \caption{Estimated error landscape.}    
    \end{subfigure}
    \caption{UFC101 error landscape.} 
    \label{fig:appB_ucf101}
\end{figure}

\begin{figure}[h] 
\centering
    \begin{subfigure}[b]{0.45\linewidth}   
        \centering 
        \includegraphics[width=\linewidth]{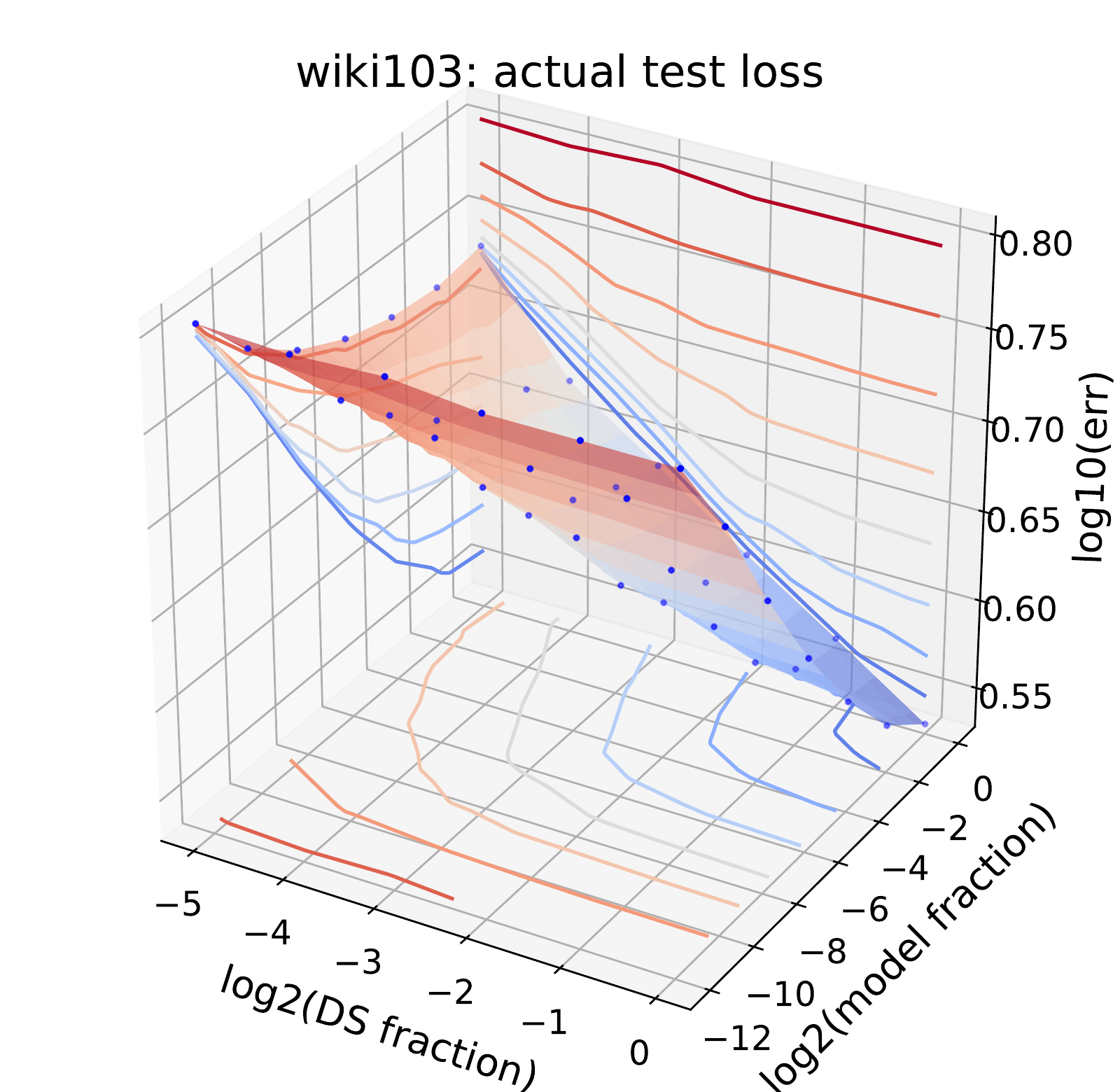}
        \caption{Actual error landscape.}    
    \end{subfigure}\hfill 
    \begin{subfigure}[b]{0.45\linewidth}   
        \centering 
        \includegraphics[width=\linewidth]{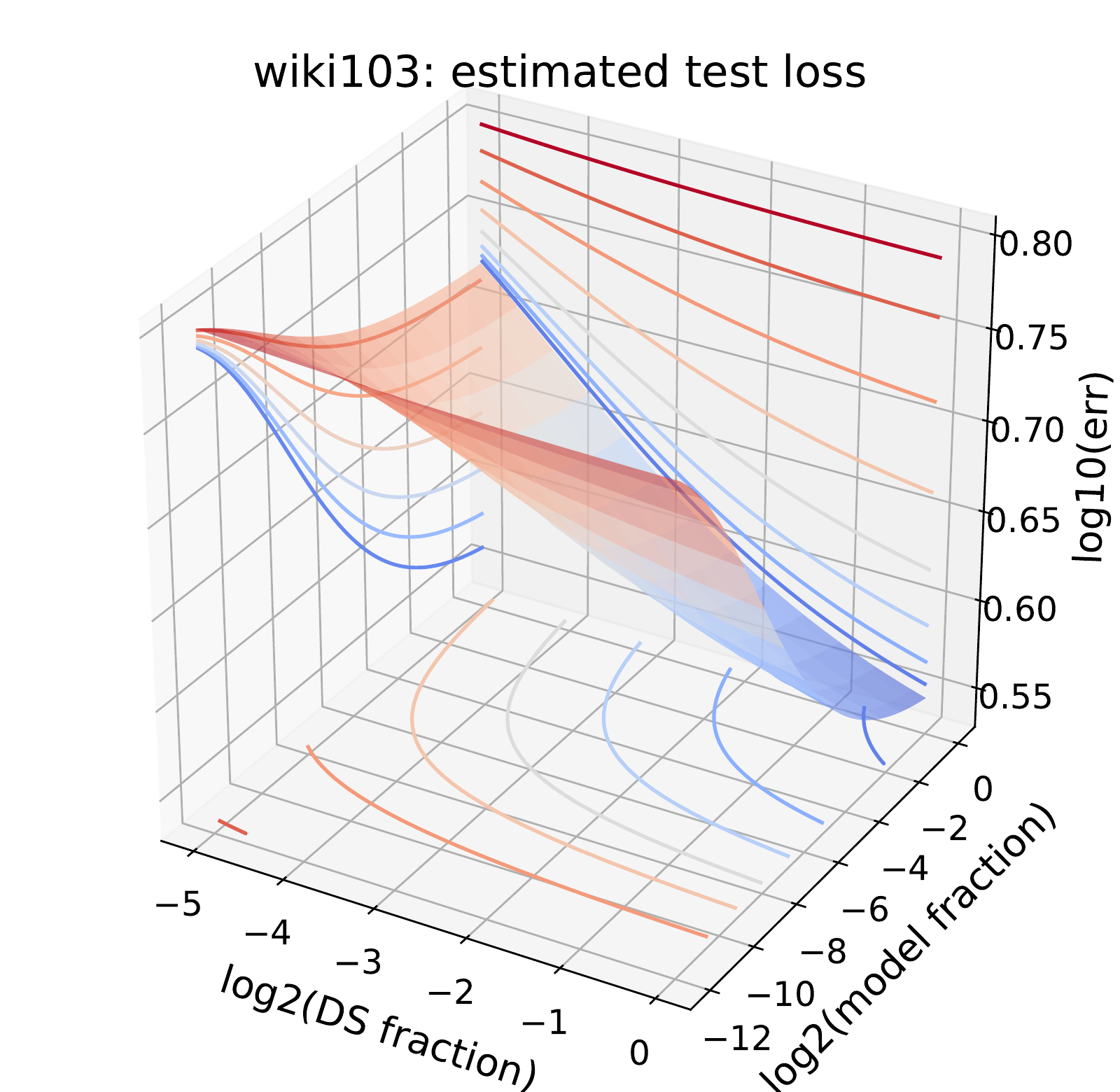}
        \caption{Estimated error landscape.}    
    \end{subfigure}
    \caption{WikiText-103 error landscape.} 
    \label{fig:appB_wiki103}
\end{figure}

We call to attention several interesting observations on the datasets explored:
\begin{itemize}
    \item As quantified rigorously in section \ref{sec:fit}, the fits perform well across error ranges. In these surfaces, one also gets qualitative sense of the fit adequacy across the wide ranges of the dataset and model scales directly. While perhaps slightly difficult to asses the surface directly, a helpful view is to consider the similarity between the projections of the actual and projected surfaces.
    \item With increasing model size, indeed typically the error does remain saturated. However, in one of our tested datasets (\figref{fig:appB_ucf101}) there was a renewed slight increase. We verify that this is indeed over-fitting, in the sense that there is no corresponding increase in the \textit{training} error.
    We note that the functional form we find can actually be used to veer clear of the $m,n$ regions where such over-fitting may occur.
    \item The simplifying approach taken by considering the random guess levels (and associated transitions) for small models or small data as identical, seems to work fairly well with some deviation apparent by examining \figref{fig:appB_wiki103}. Indeed the simplification can hold well for balanced datasets, but need not for imbalanced ones such as in the task of language modeling. Thus, a relaxation of this simplification is expected to be important conceptually and practically.
    
\end{itemize}


\begin{figure}[h] 
\centering
    \begin{subfigure}[b]{0.45\linewidth}   
        \centering 
        \includegraphics[width=\linewidth]{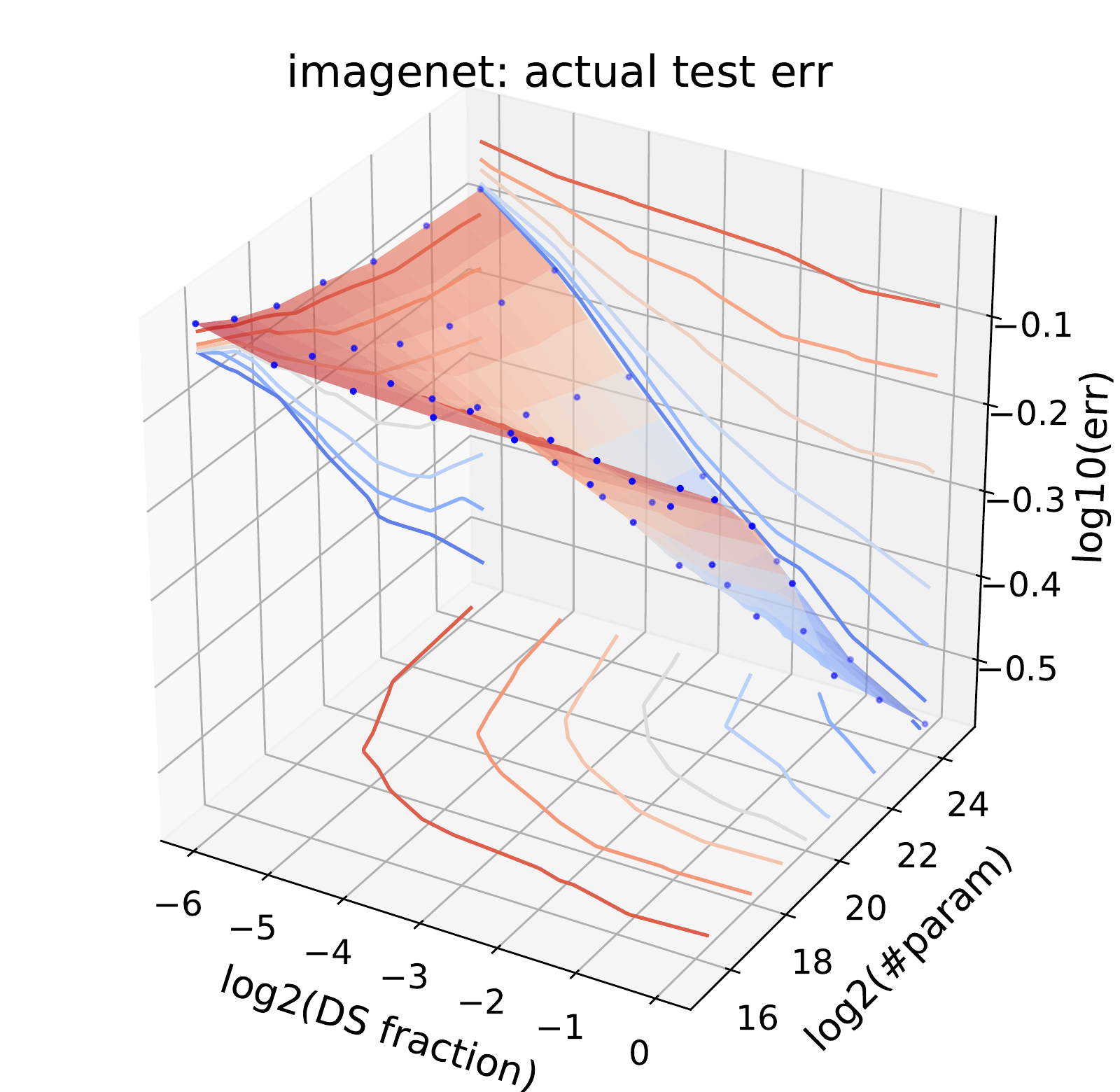}
        \caption{Actual error landscape.}    
    \end{subfigure}\hfill 
    \begin{subfigure}[b]{0.45\linewidth}   
        \centering 
        \includegraphics[width=\linewidth]{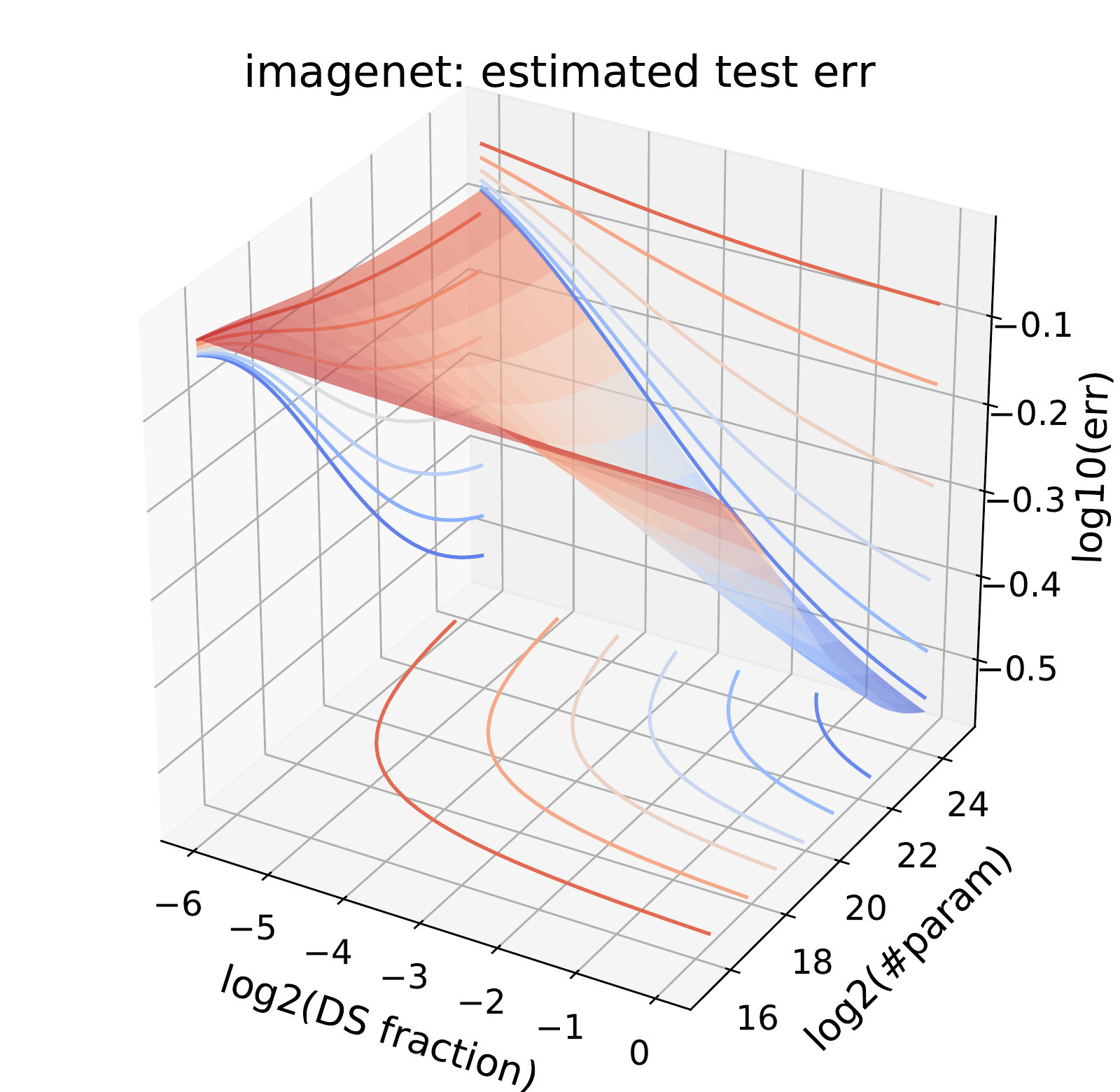}
        \caption{Estimated error landscape.}    
    \end{subfigure}
    \caption{ImageNet error landscape.} 
    \label{fig:appB_ImageNet}
\end{figure}

\begin{figure}[h] 
\centering
    \begin{subfigure}[b]{0.45\linewidth}   
        \centering 
        \includegraphics[width=\linewidth]{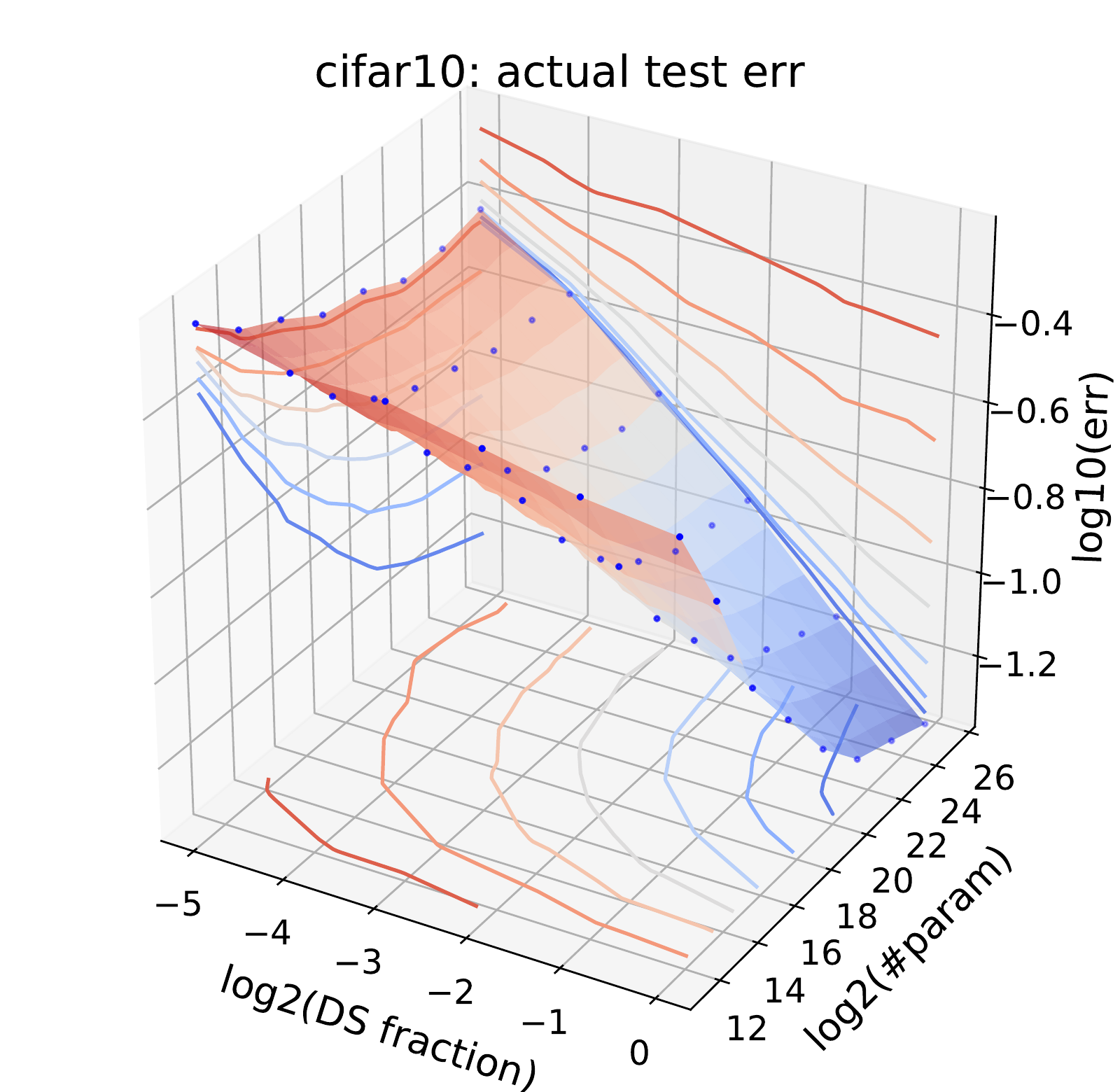}
        \caption{Actual error landscape.}    
    \end{subfigure}\hfill 
    \begin{subfigure}[b]{0.45\linewidth}   
        \centering 
        \includegraphics[width=\linewidth]{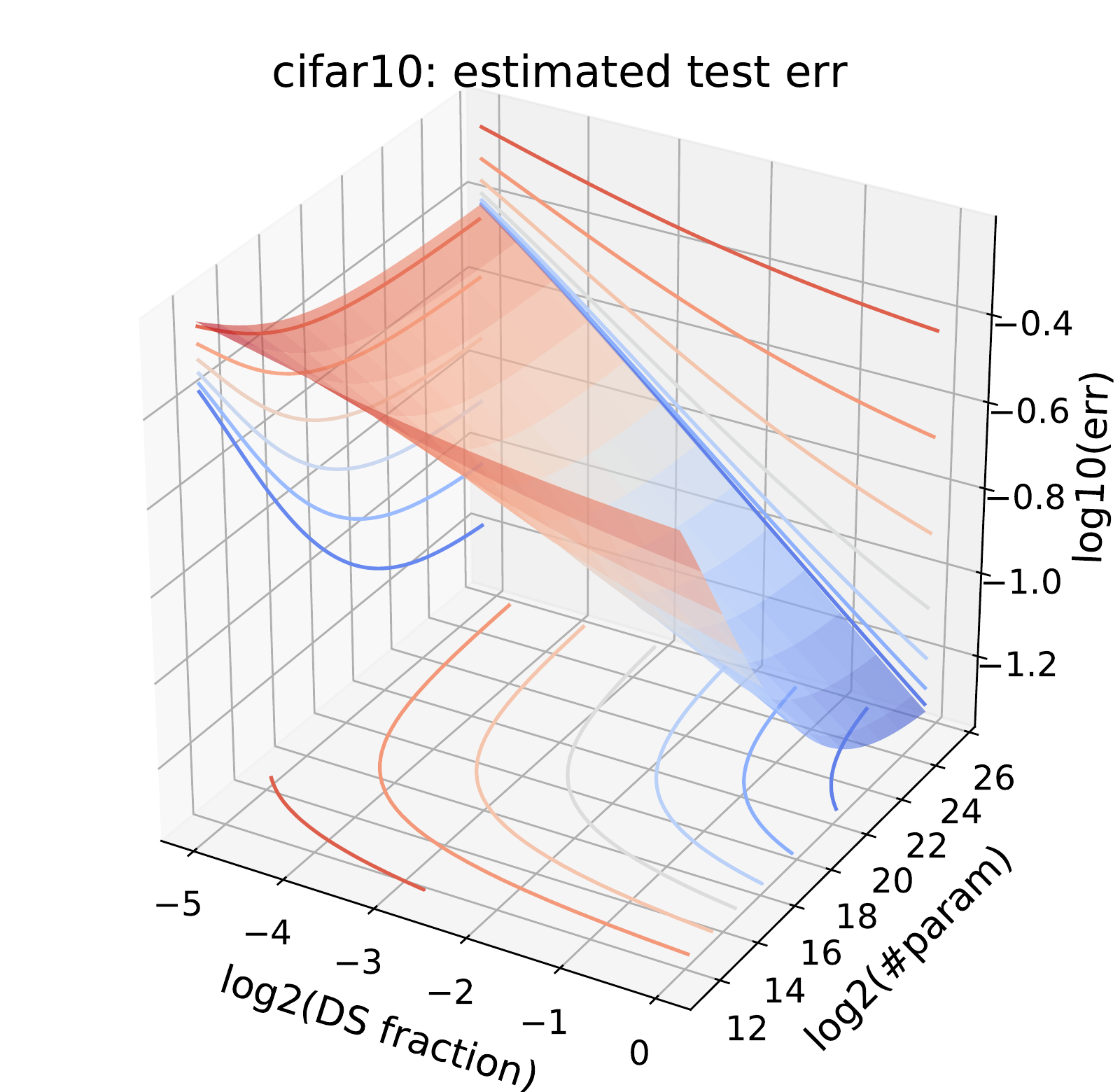}
        \caption{Estimated error landscape.}    
    \end{subfigure}
    \caption{CIFAR10 error landscape.} 
    \label{fig:appB_CIFAR10}
\end{figure}

\begin{figure}[h] 
\centering
    \begin{subfigure}[b]{0.45\linewidth}   
        \centering 
        \includegraphics[width=\linewidth]{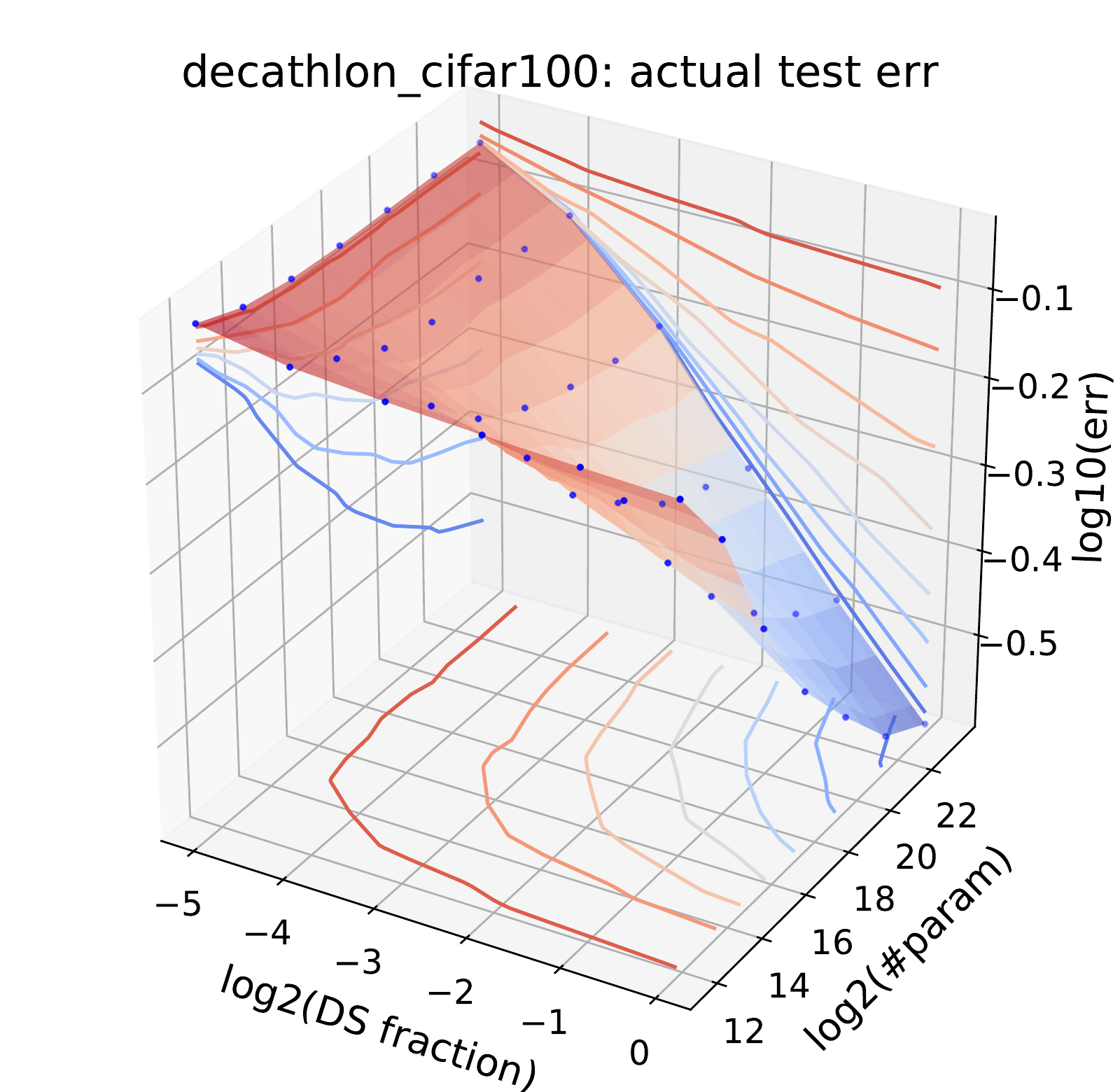}
        \caption{Actual error landscape.}    
    \end{subfigure}\hfill 
    \begin{subfigure}[b]{0.45\linewidth}   
        \centering 
        \includegraphics[width=\linewidth]{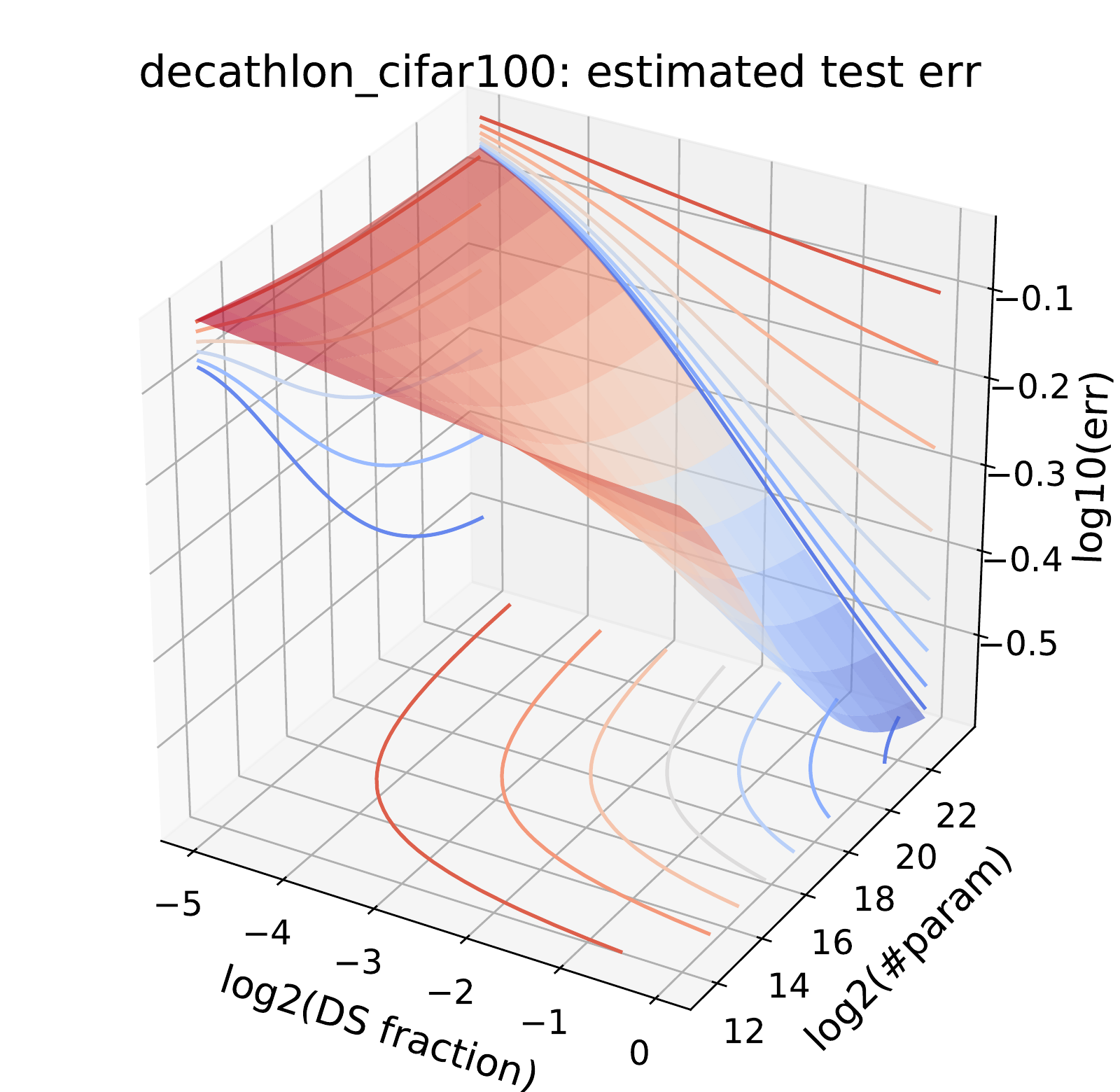}
        \caption{Estimated error landscape.}    
    \end{subfigure}
    \caption{CIFAR100 error landscape.} 
    \label{fig:appB_CIFAR100}
\end{figure}

\begin{figure}[h] 
\centering
    \begin{subfigure}[b]{0.45\linewidth}   
        \centering 
        \includegraphics[width=\linewidth]{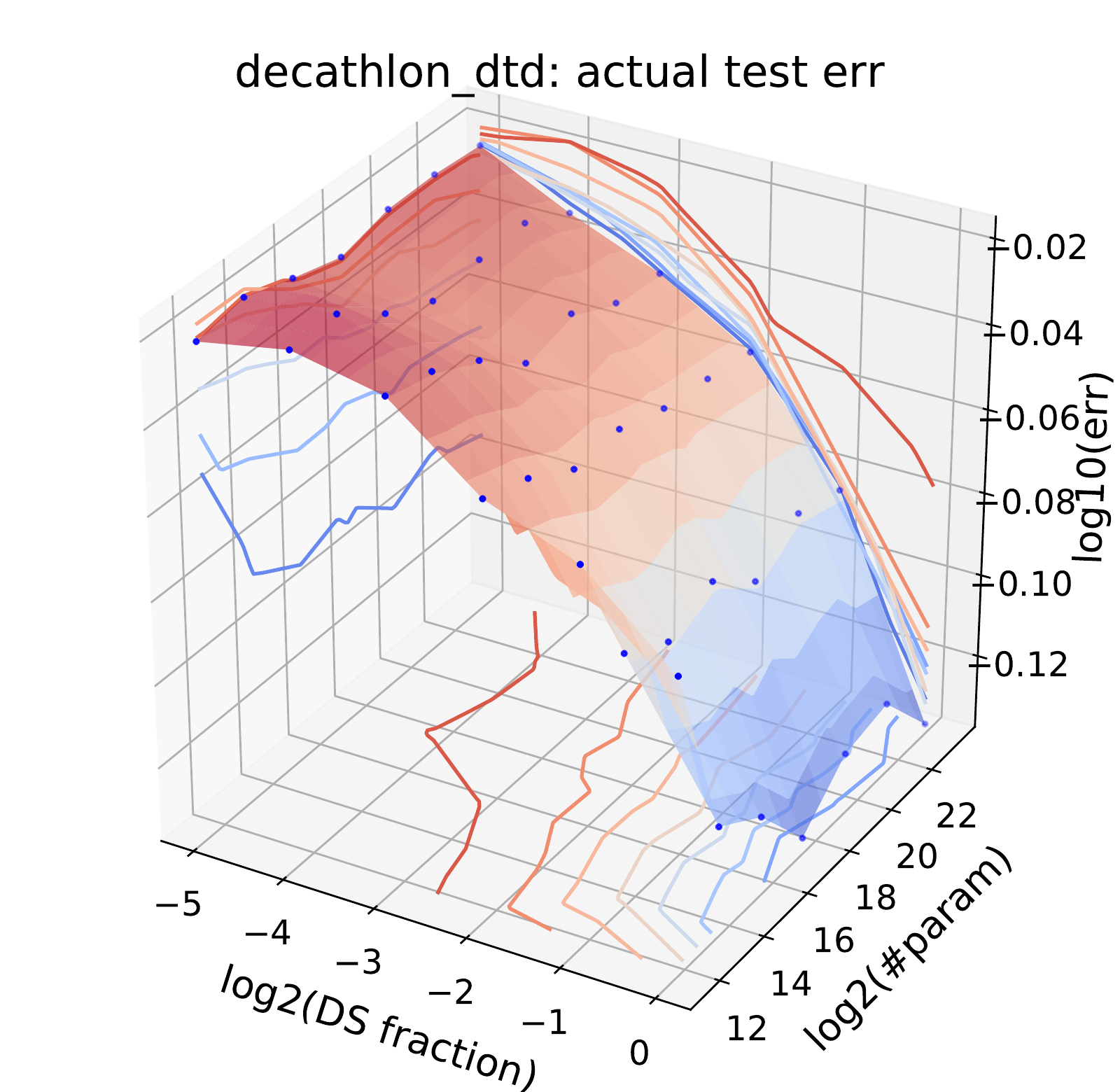}
        \caption{Actual error landscape.}    
    \end{subfigure}\hfill 
    \begin{subfigure}[b]{0.45\linewidth}   
        \centering 
        \includegraphics[width=\linewidth]{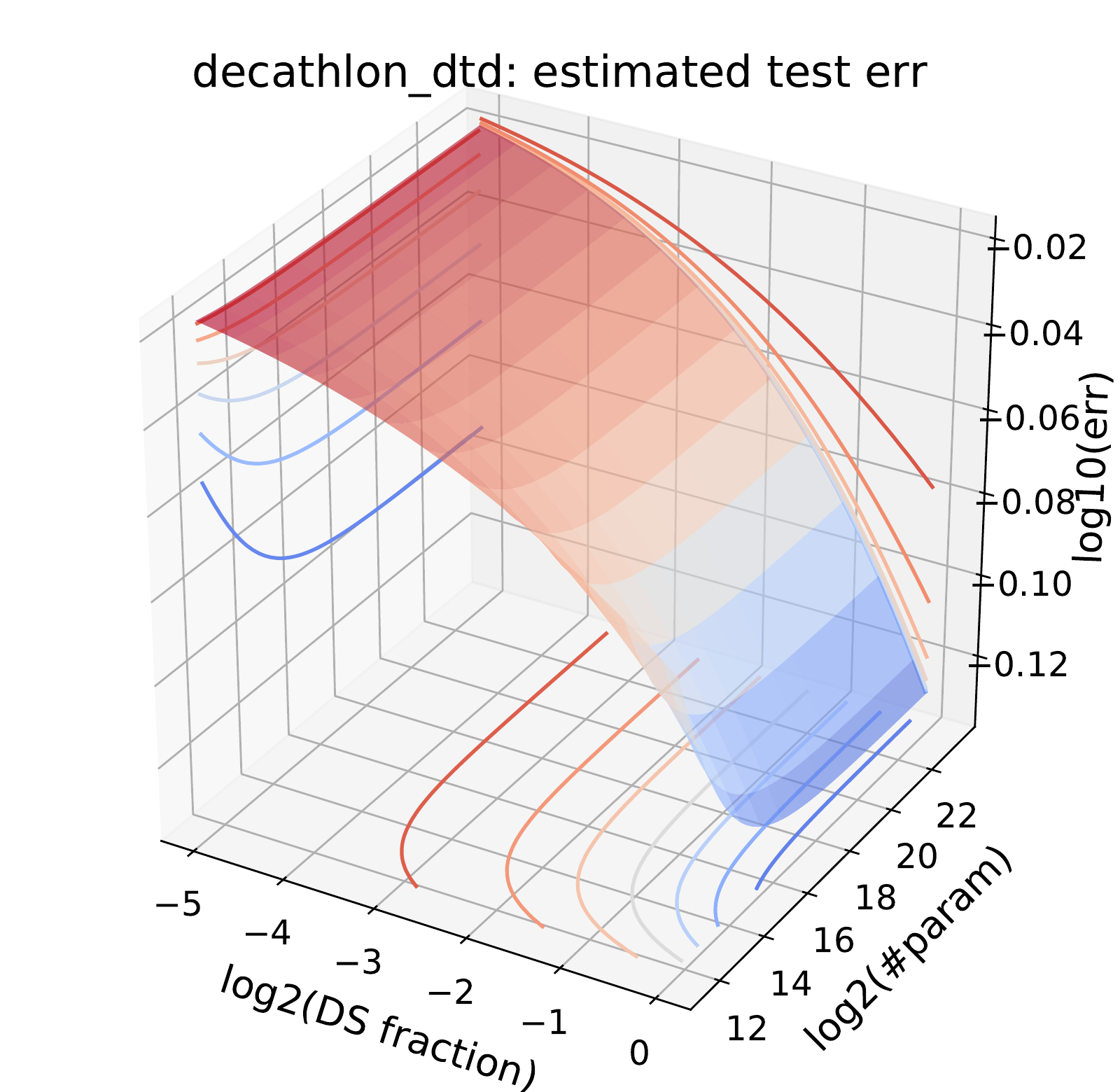}
        \caption{Estimated error landscape.}    
    \end{subfigure}
    \caption{DTD error landscape.} 
    \label{fig:appB_dtd}
\end{figure}

\begin{figure}[h] 
\centering
    \begin{subfigure}[b]{0.45\linewidth}   
        \centering 
        \includegraphics[width=\linewidth]{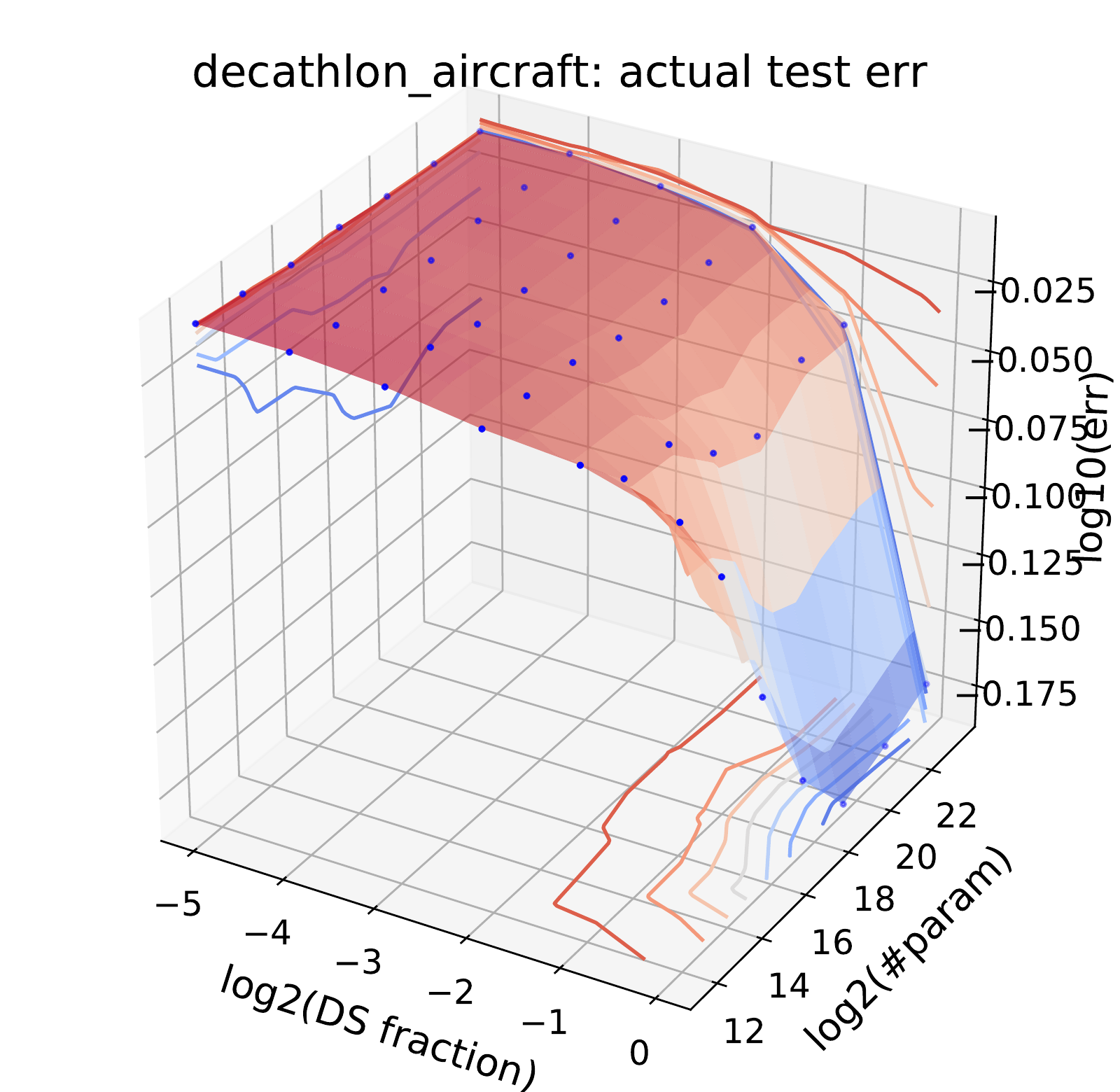}
        \caption{Actual error landscape.}    
    \end{subfigure}\hfill 
    \begin{subfigure}[b]{0.45\linewidth}   
        \centering 
        \includegraphics[width=\linewidth]{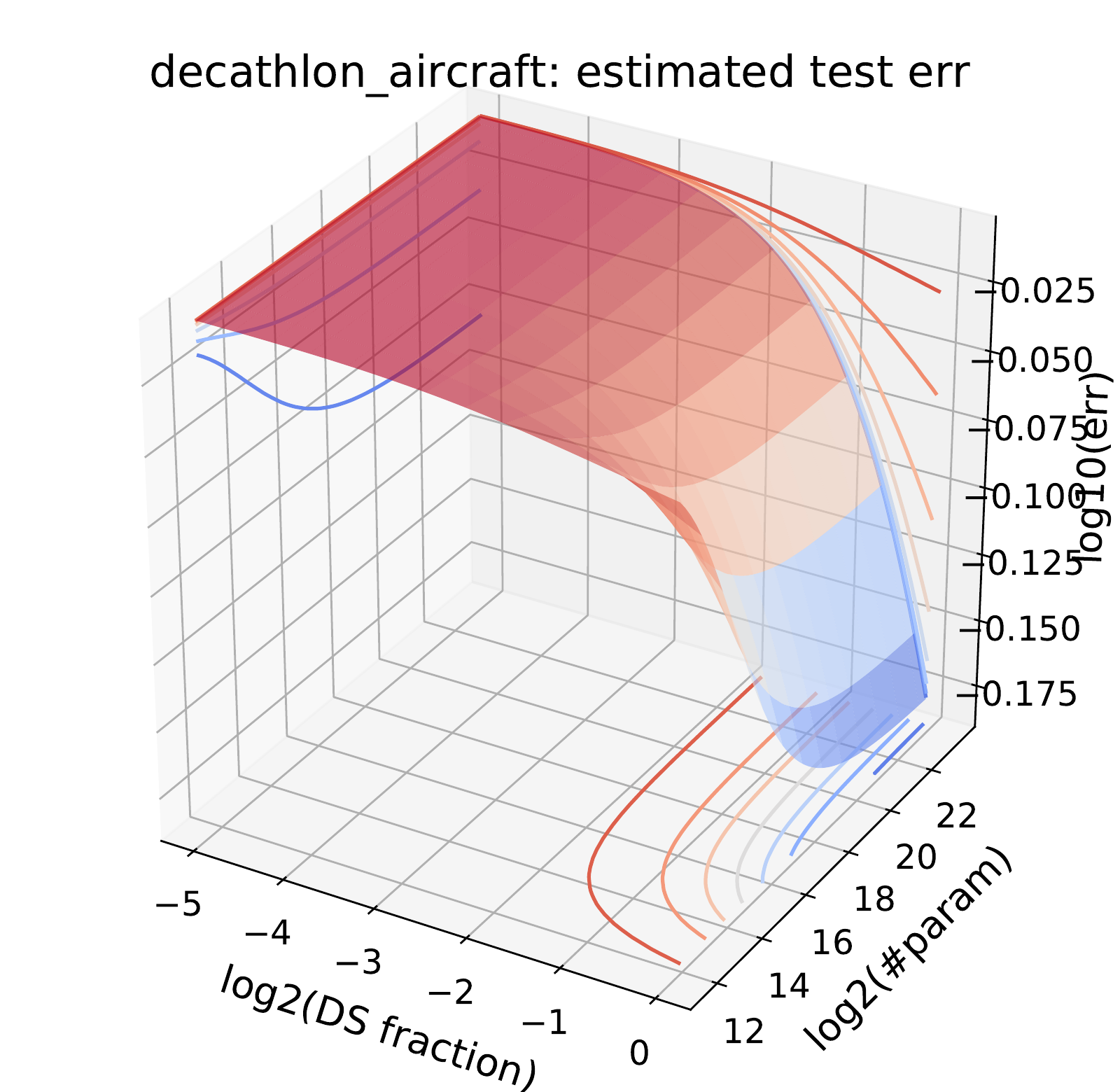}
        \caption{Estimated error landscape.}    
    \end{subfigure}
    \caption{Aircraft error landscape.} 
    \label{fig:appB_aircraft}
\end{figure}

\begin{figure}[h] 
\centering
    \begin{subfigure}[b]{0.45\linewidth}   
        \centering 
        \includegraphics[width=\linewidth]{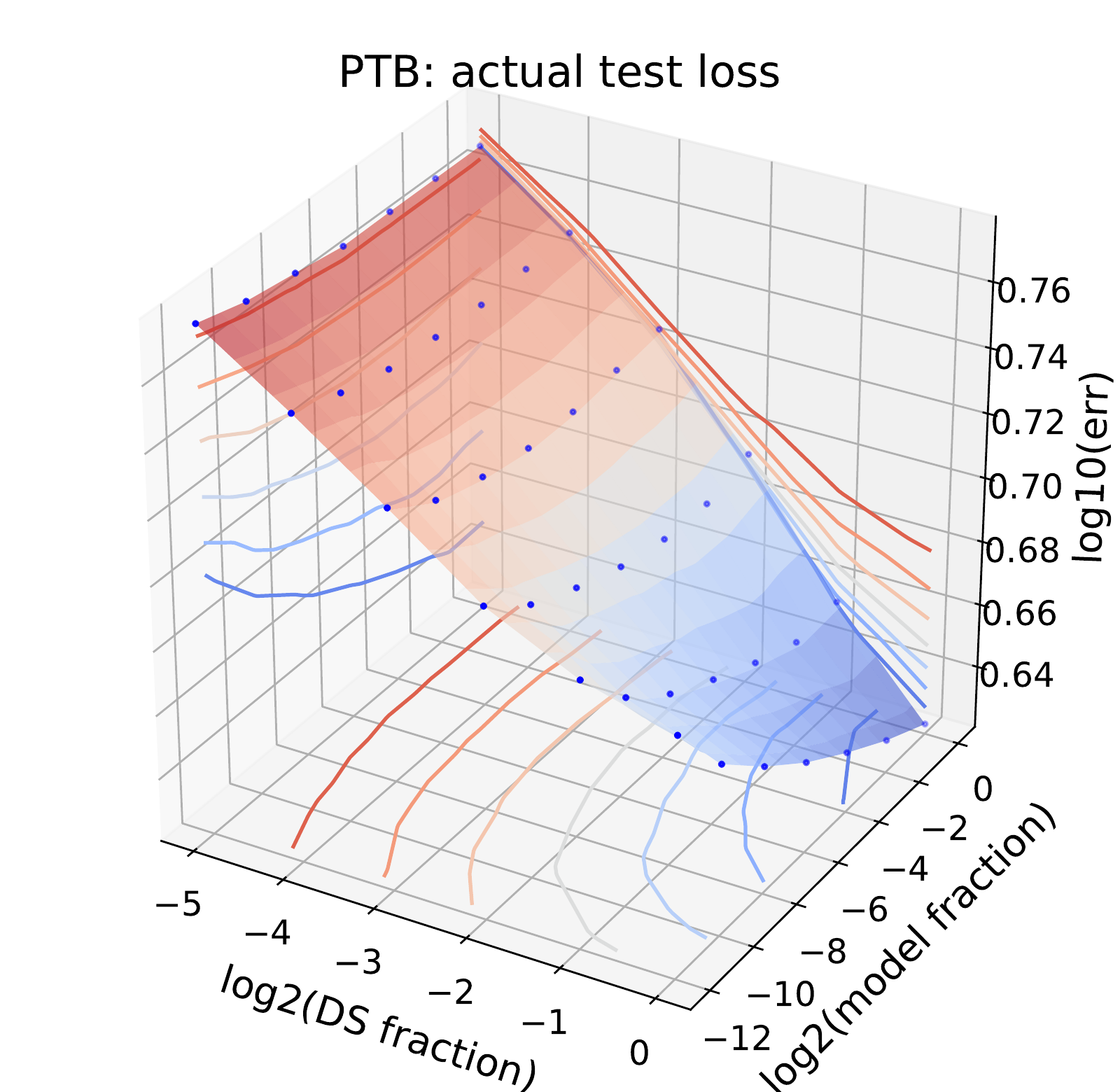}
        \caption{Actual error landscape.}    
    \end{subfigure}\hfill 
    \begin{subfigure}[b]{0.45\linewidth}   
        \centering 
        \includegraphics[width=\linewidth]{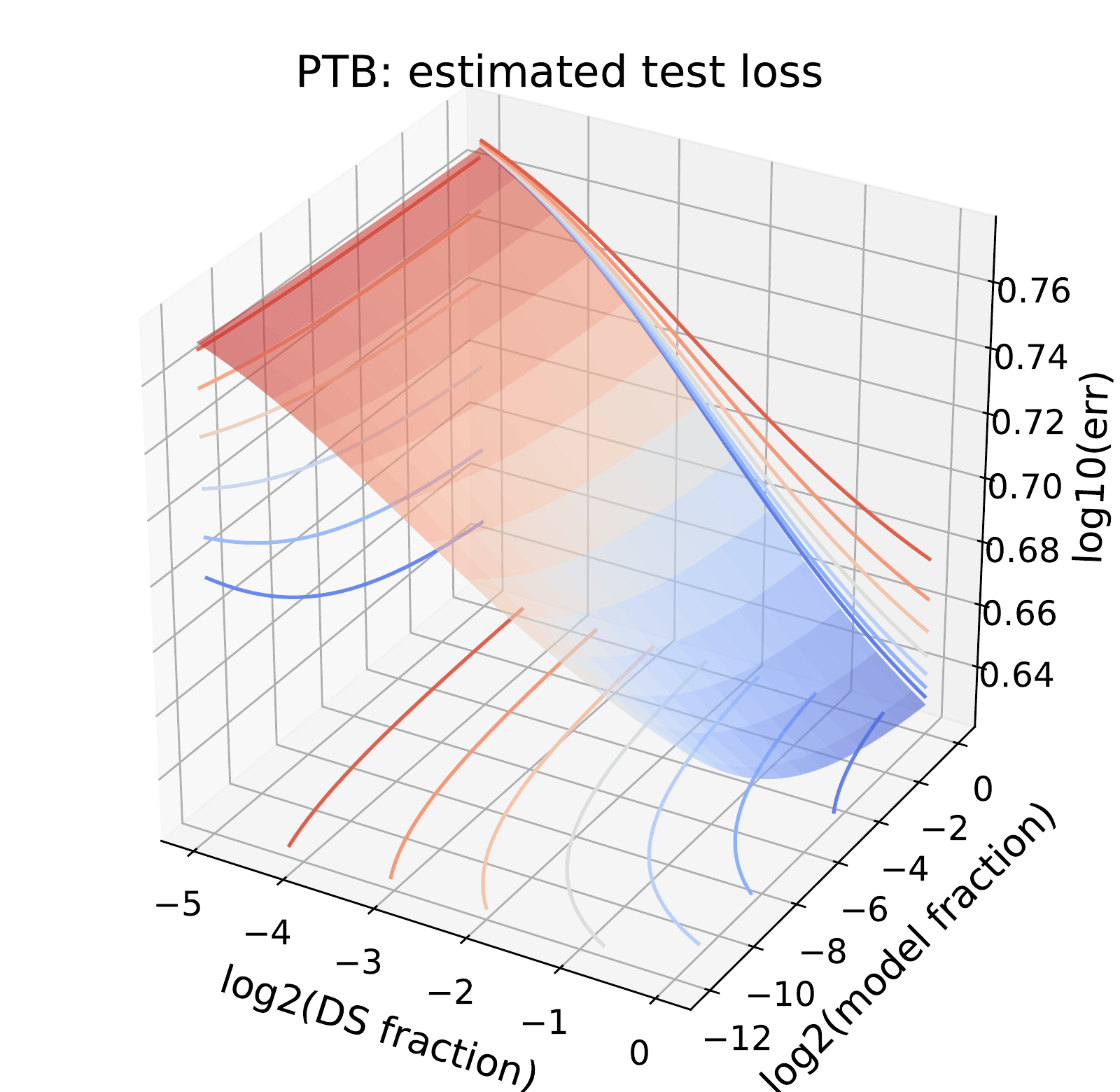}
        \caption{Estimated error landscape.}    
    \end{subfigure}
    \caption{PTB error landscape.} 
    \label{fig:appB_PTB}
\end{figure}

\begin{figure}[h] 
\centering
    \begin{subfigure}[b]{0.45\linewidth}   
        \centering 
        \includegraphics[width=\linewidth]{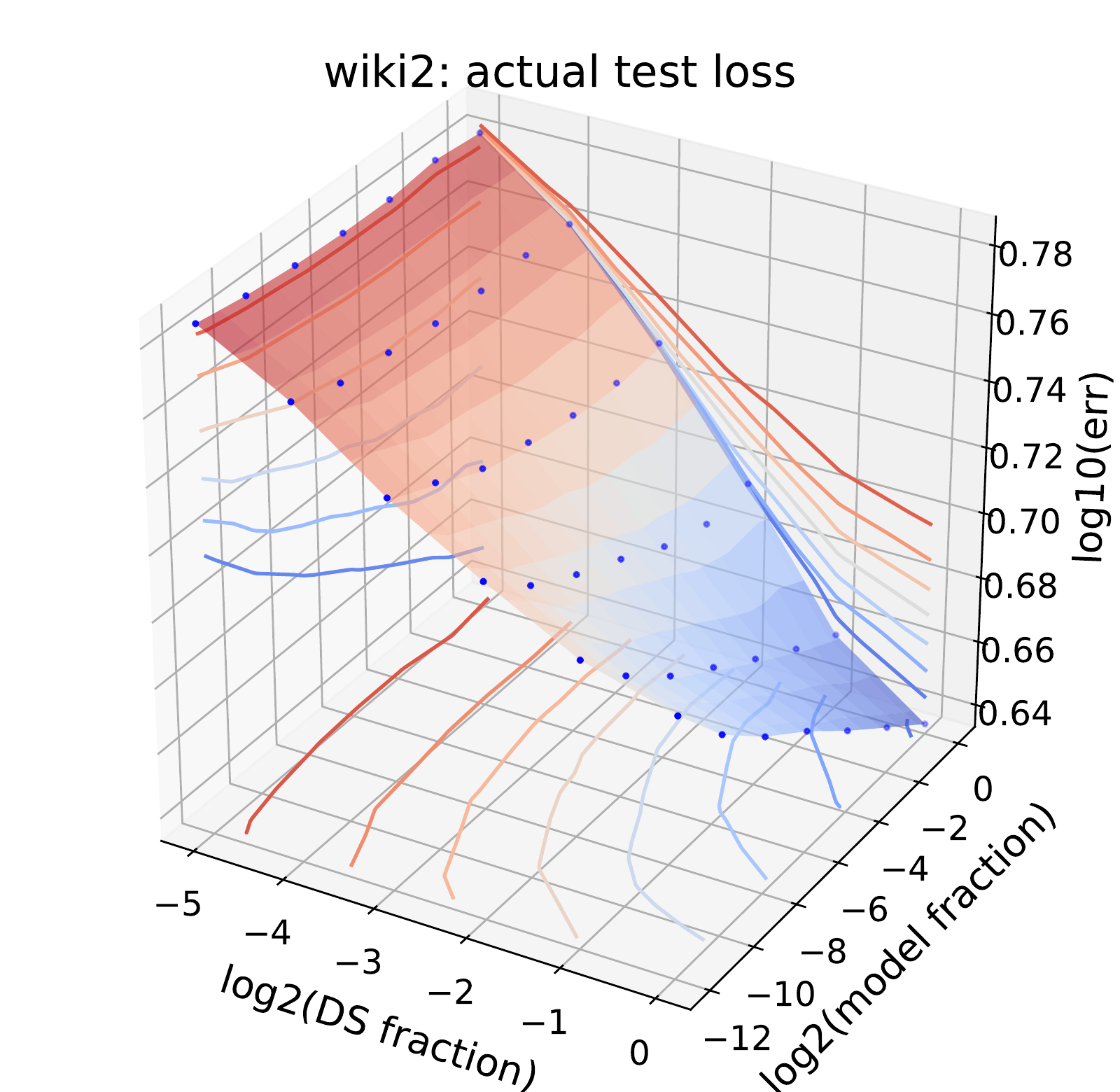}
        \caption{Actual error landscape.}    
    \end{subfigure}\hfill 
    \begin{subfigure}[b]{0.45\linewidth}   
        \centering 
        \includegraphics[width=\linewidth]{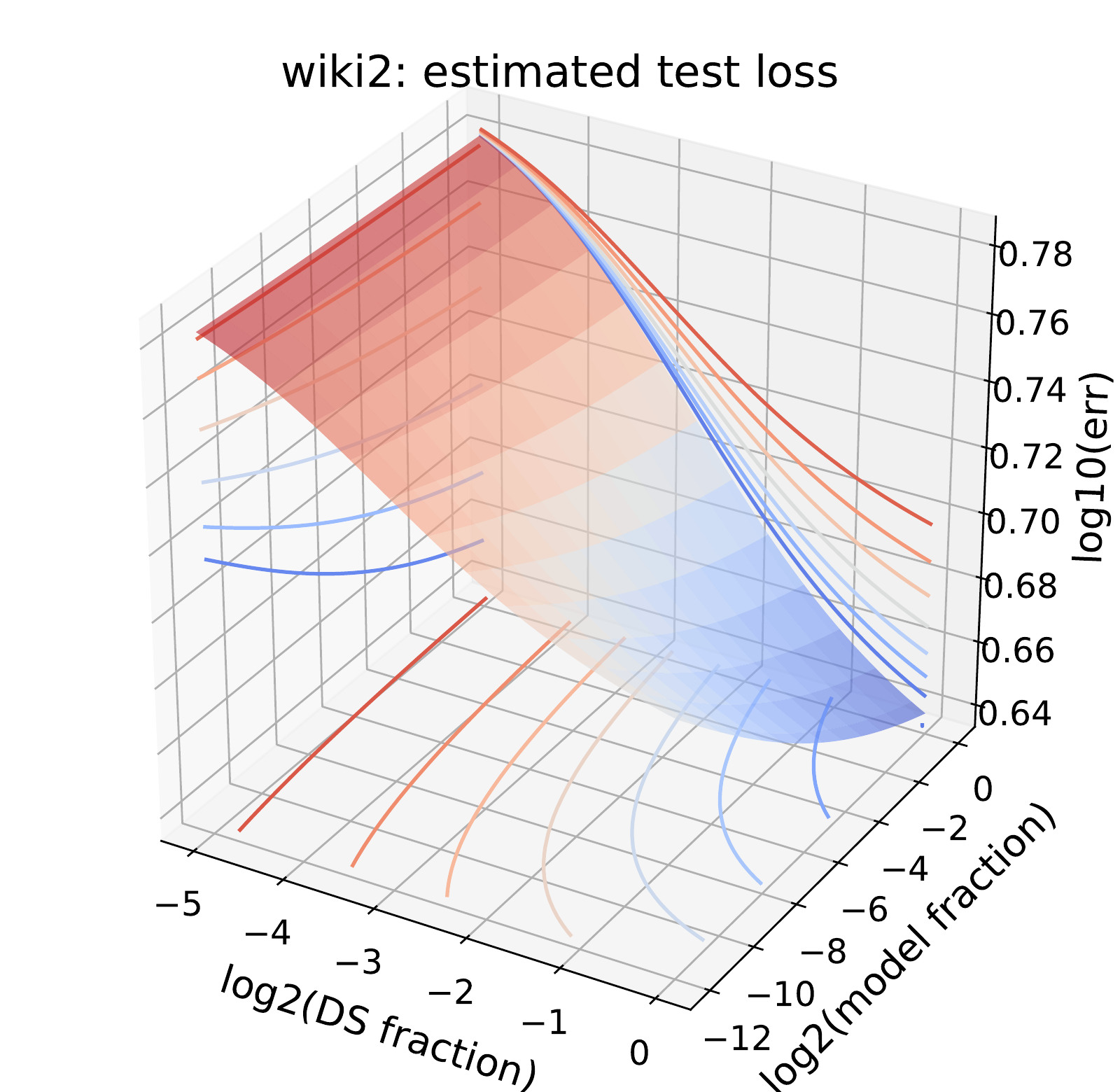}
        \caption{Estimated error landscape.}    
    \end{subfigure}
    \caption{WikiText-2 error landscape.} 
    \label{fig:appB_wiki2}
\end{figure}

\clearpage
\section{Additional Extrapolation Results} \label{app:extrapolations}

Here we provide detailed extrapolation results, for all datasets. All figures are structured in a similar way. 
Each subplot shows estimated (y-axis)  vs.\  actual error (x-axis) (0 to 1 scale on both axes). Each subplot is located at the coordinate of the maximal data and model given for the task of performing the fit to the functional form in \eqref{eq:envelope}. 

This is the point at the top-right corner of the green dots in the illustration in \figref{fig:extrapolation-array}.
The target is to find the error-landscape values for unseen, larger scales of both model and data (red points in the same illustration). 
Going from left to right in each figure indicates observed measurements of the error from models of an increasing fraction w.r.t the full size. Going from bottom-to top indicates observed measurements of the error from dataset sizes of an increasingly large fraction of the full dataset.

In each subplot, every point shows the estimated vs.\ actual error on a model-data configuration. Points that were given for fitting the function are colored in green, while unseen points that were not used are in red. 
The red points show  the estimation error vs.\  actual error when extrapolating to all larger models and data sizes. 
In each subplot,  the mean and standard deviation over all divergences $\delta$ at target points are given in text. 

Each experiment fit of the parameters was repeated 100 times, with different random initializations of $\vtheta$. The shaded bands show one standard deviation across these runs.

\begin{figure}[h]
     \centering
         \includegraphics[width=1\linewidth]{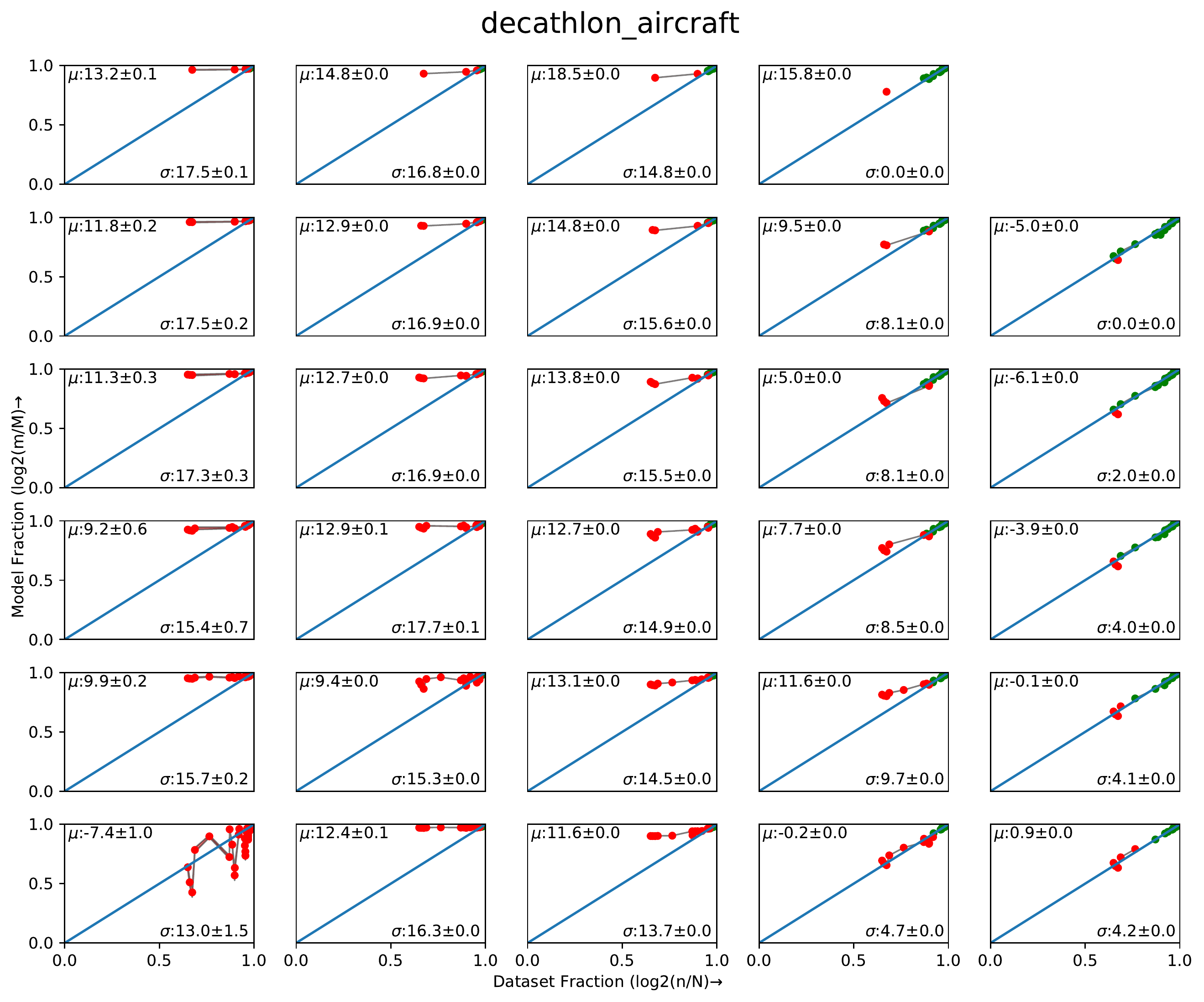}
     \caption{Aircraft extrapolation results.  } 
     \label{fig:appD_aircraft}
 \end{figure}

\begin{figure}[h]
     \centering
         \includegraphics[width=1\linewidth]{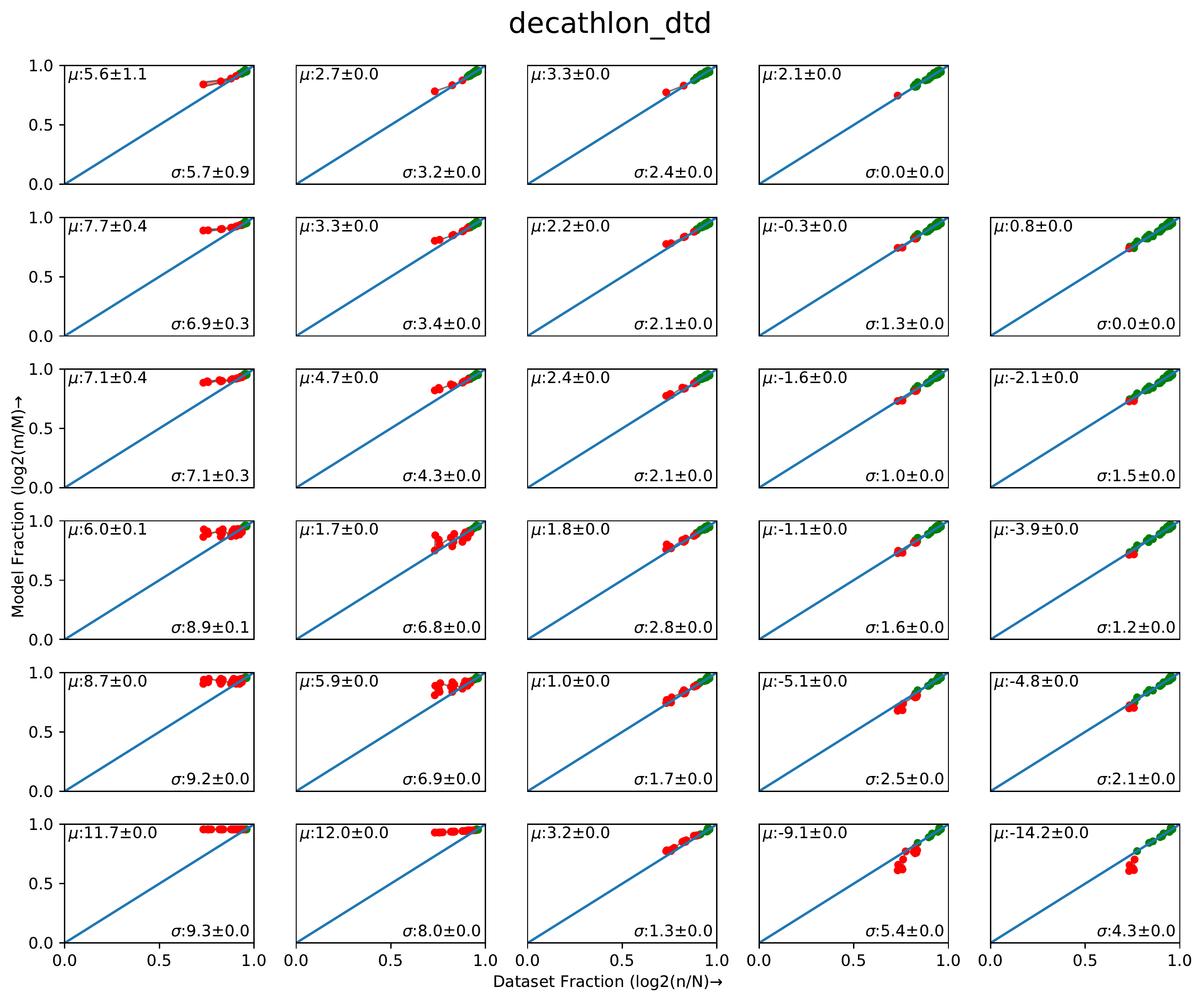}
     \caption{DTD Results} 
     \label{fig:appD_dtd}
 \end{figure}

\begin{figure}[h]
     \centering
         \includegraphics[width=1\linewidth]{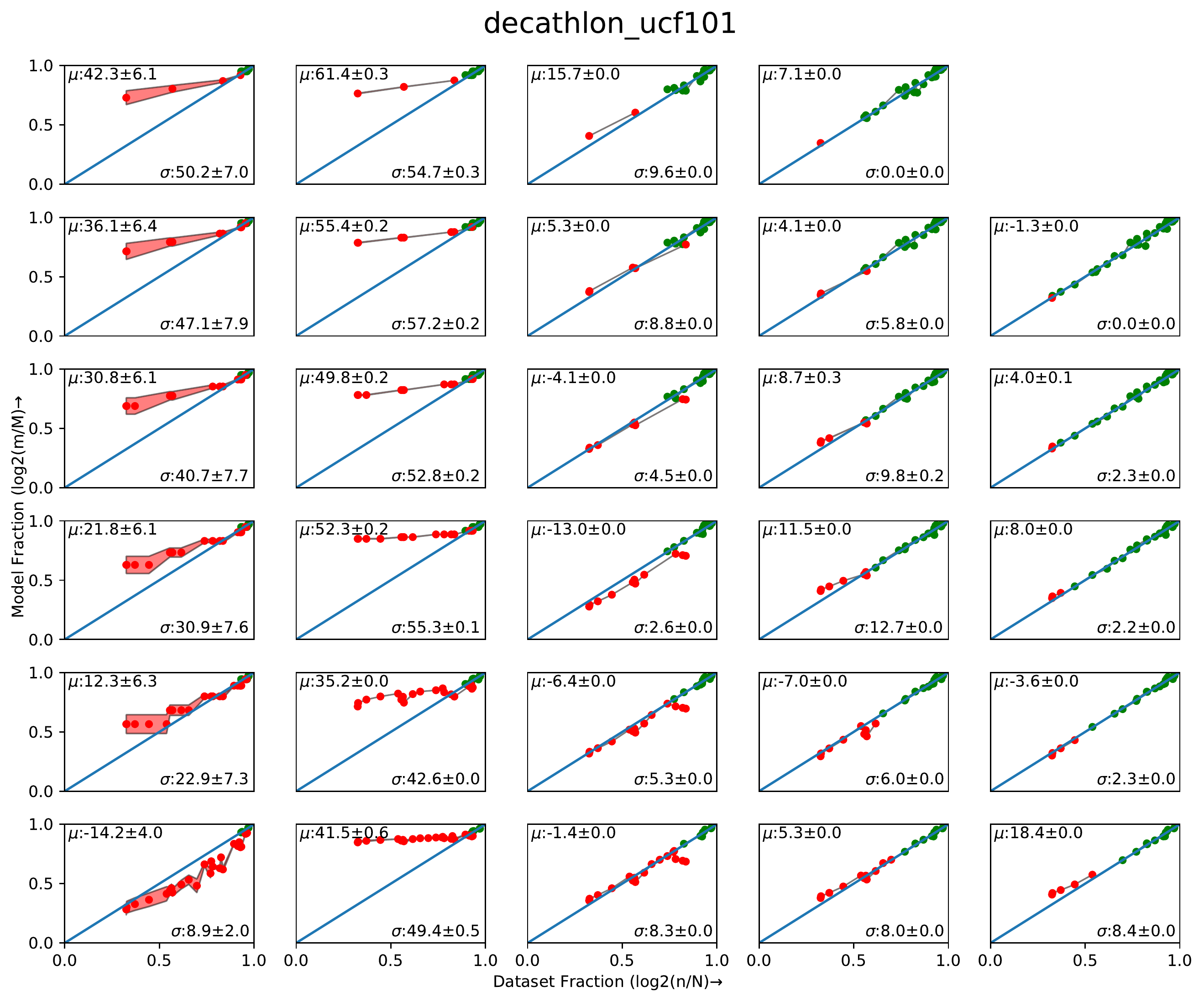}
     \caption{UCF101 extrapolation results.} 
     \label{fig:appD_ucf101}
 \end{figure}

The quality of the extrapolation is critically dependent on the signal provided in the (green) fitted points. Two limiting factors are evident by examining the figures below, which both play a role in the well-posedness of the solution:
\begin{itemize}
    \item The proximity to the initial random guess level. Only upon transitioning from the initial error plateau, does meaningful signal about the scaling rates become available. Indeed, for scales prior still in the region or close to the initial error level, one sees poor extrapolation results; see figures \ref{fig:appD_aircraft},   \ref{fig:appD_dtd}, and \ref{fig:appD_ucf101}, and the vivid origin of this phenomena by examining figures \ref{fig:appB_aircraft},   \ref{fig:appB_dtd}, and \ref{fig:appB_ucf101}.
    \item A second source of ill-posedness is tied to the number of configurations used for the estimation of $\vtheta$. Clearly, when this is small,  one cannot expect the extrapolation to be stable. In fact, at least two measurements in each scaling dimension (model/data) are needed, and no less than the number of parameters in $\vtheta$ in total. Indeed, for all the plots in this appendix, the smallest scale of $m,n$ is omitted form the graph such that the lowermost row and leftmost column span exactly two model and data scales correspondingly. Of course, there is nothing tying directly the number of points and scale of configurations measured, and one can decouple these two factors by taking closer spaced samples at small scale.
    \item When both the above factors are not limiting the measurement, one readily sees that for divergences of no more than a few percent, it is sufficient to measure model/data configurations which are far-ranged from the configurations which one wishes to extrapolate to .
\end{itemize}

 \begin{figure}[h]
     \centering
         \includegraphics[width=1\linewidth]{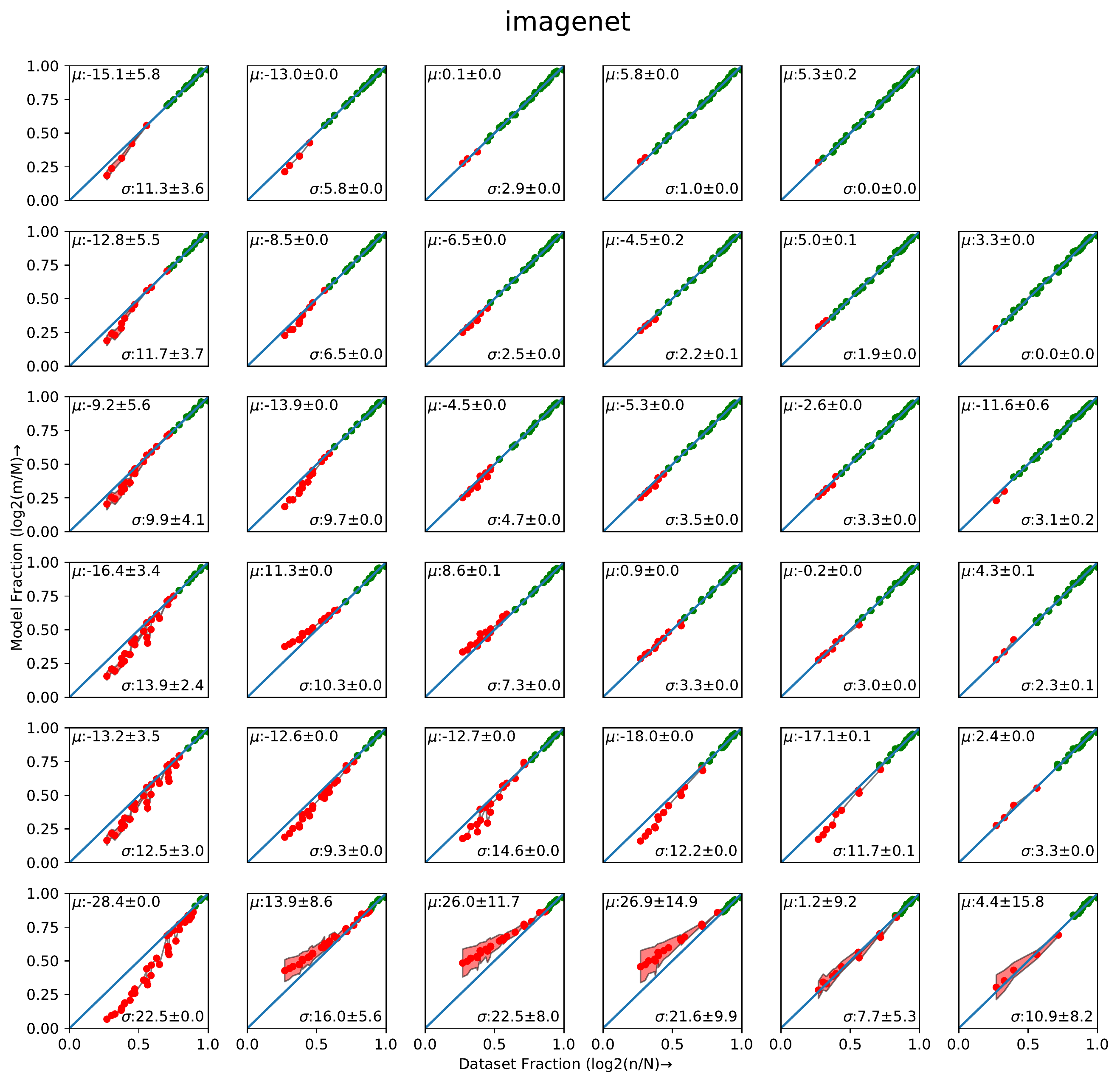}
     \caption{ImageNet extrapolation results.} 
     \label{fig:appD_imagenet}

     \label{fig1:}
 \end{figure}

  \begin{figure}[h]
     \centering
         \includegraphics[width=1\linewidth]{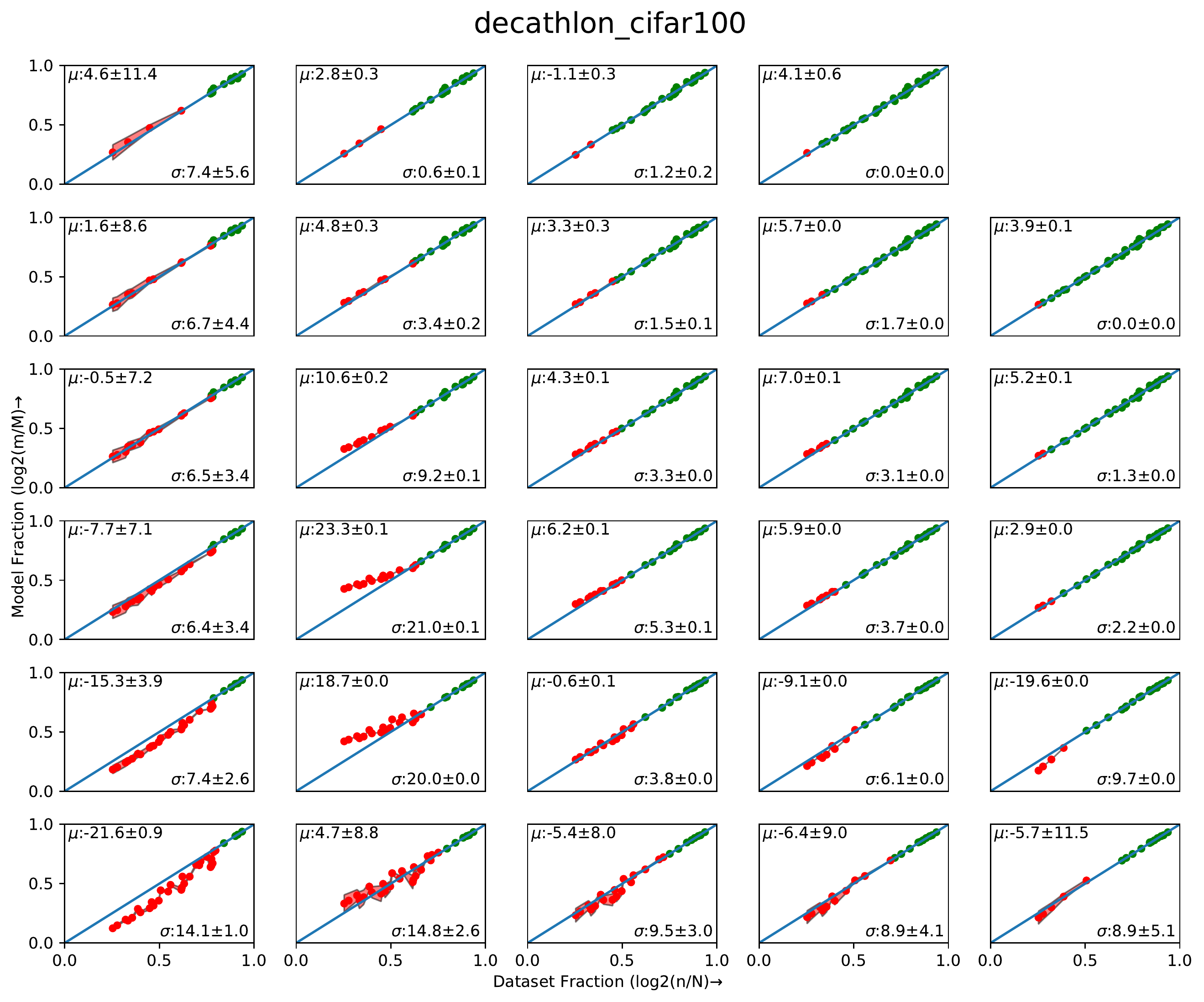}
     \caption{CIFAR100 Extrapolation Results} 
     \label{fig:appD_cifar100}
 \end{figure}

 \begin{figure}[h]
     \centering
         \includegraphics[width=1\linewidth]{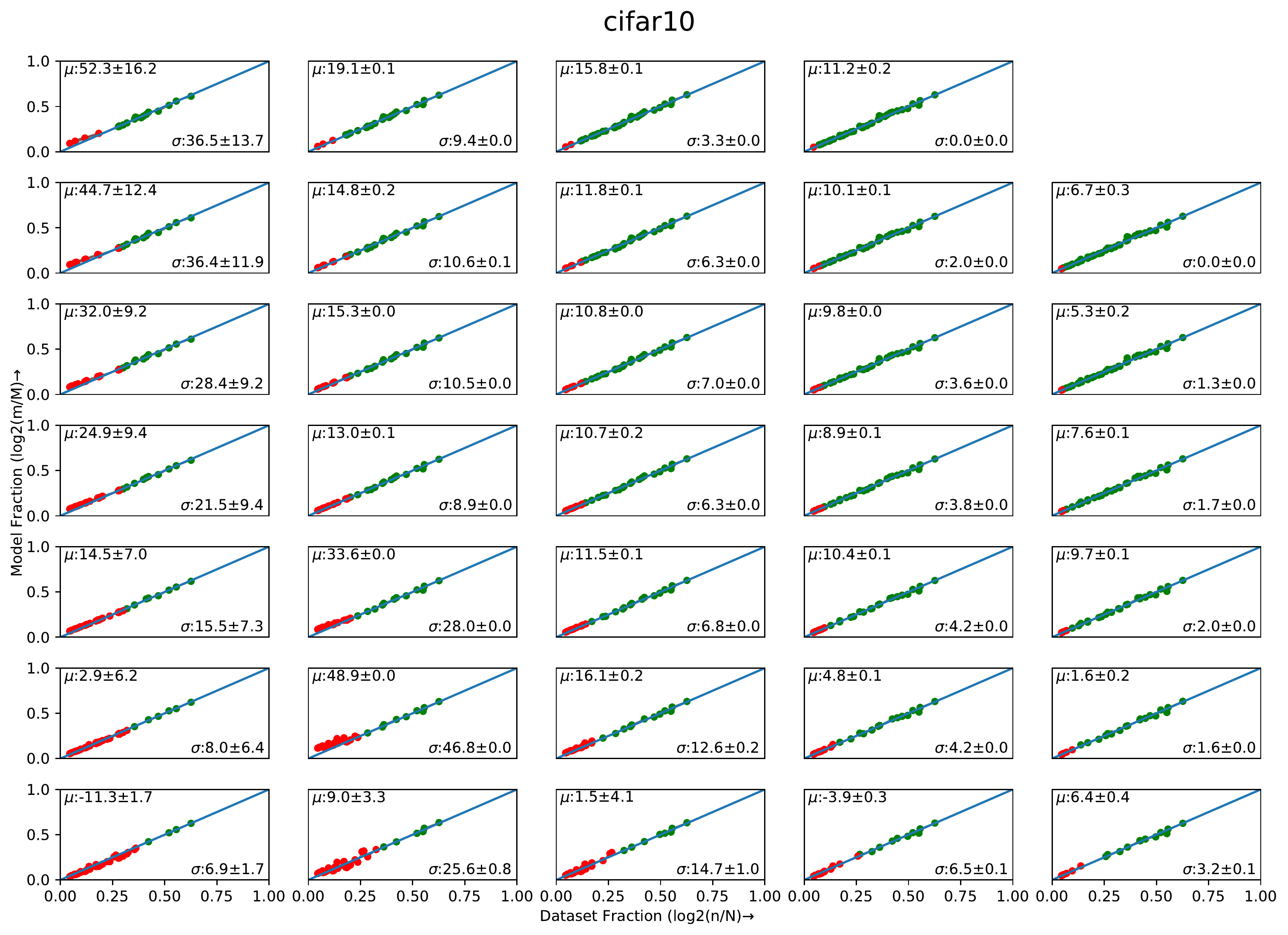}
     \caption{CIFAR10 extrapolation results.} 
     \label{fig:appD_cifar10}
 \end{figure}

 \begin{figure}[h]
     \centering
         \includegraphics[width=1\linewidth]{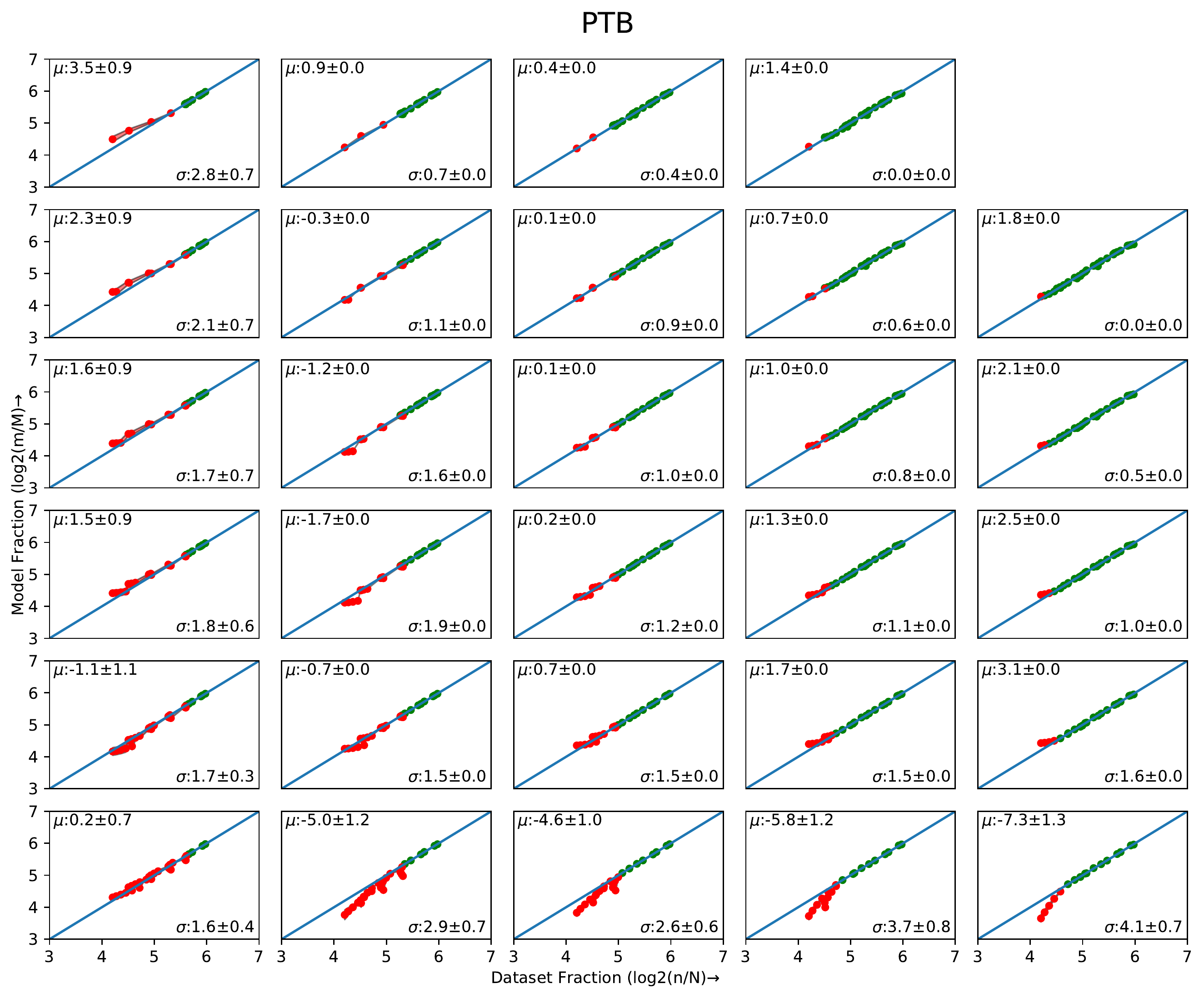}
     \caption{PTB extrapolation results.} 
     \label{fig:appD_ptb}
 \end{figure}
 
   \begin{figure}[h]
     \centering
         \includegraphics[width=1\linewidth]{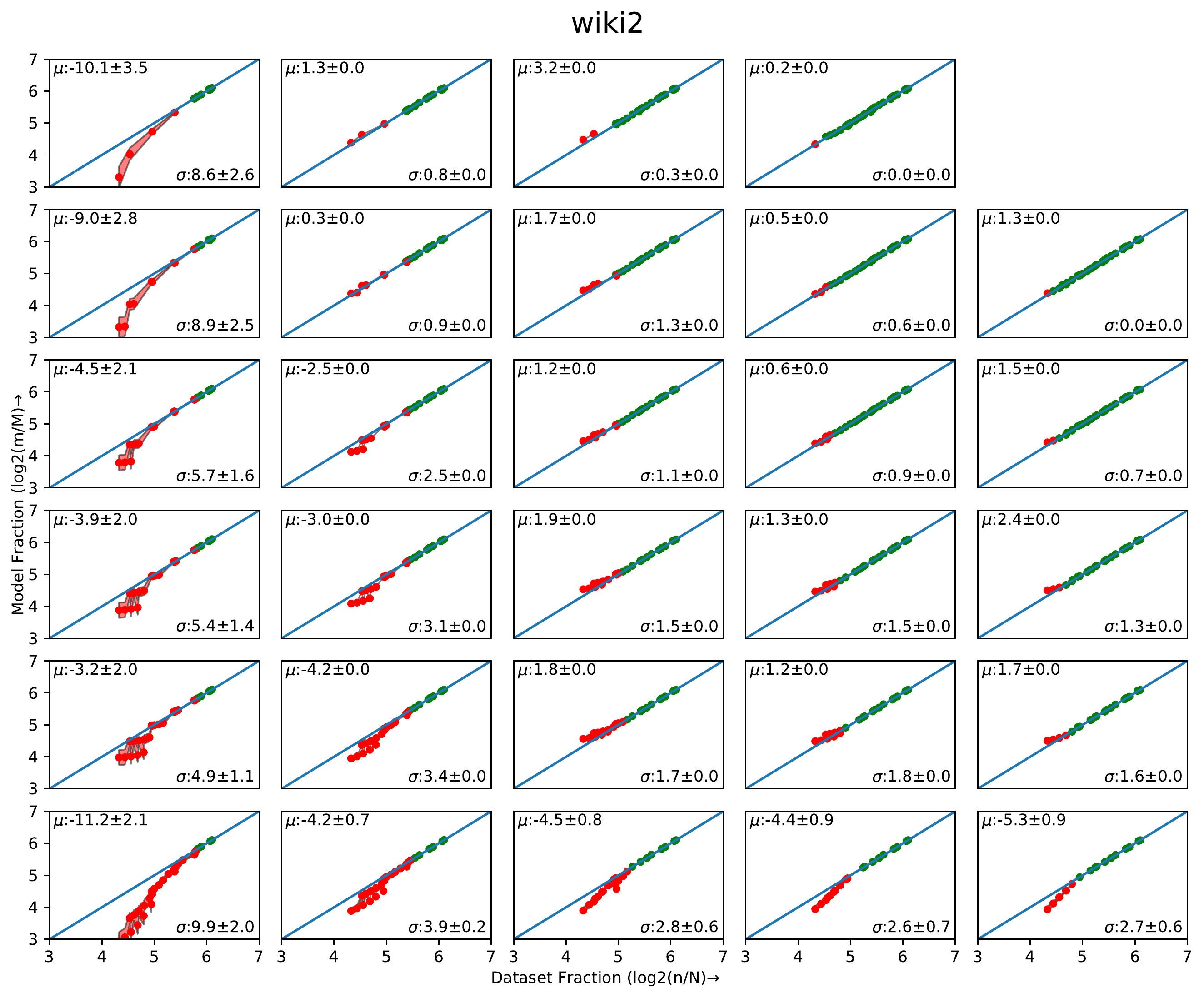}
     \caption{WikiText-2 extrapolation results.} 
     \label{fig:appD_wiki2}
 \end{figure}
 
  \begin{figure}[h]
     \centering
         \includegraphics[width=1\linewidth]{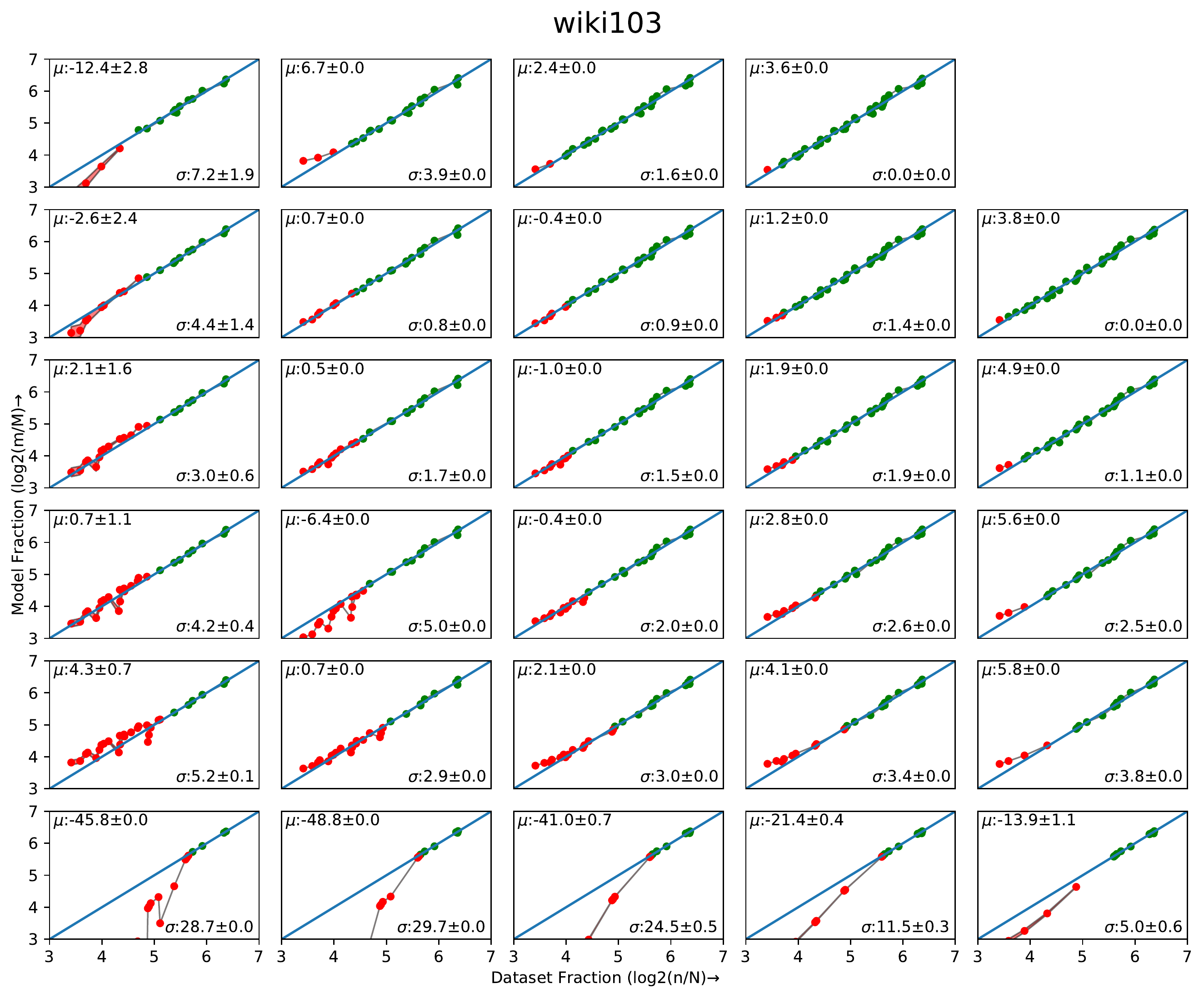}
     \caption{WikiText-103 extrapolation results.} 
     \label{fig:appD_wiki103}
 \end{figure}
 
 \clearpage






\newpage

\chapter{Appendix B: Pruning scaling further details}

\section*{Contents of Appendices}

\textbf{Appendix \ref{app:pruningalg}.} Details on the IMP pruning algorithm.

\textbf{Appendix \ref{app:resnets}.} Details on the models, datasets, and training hyperparameters.

\textbf{Appendix \ref{app:sec3-key-observations-alldimensions}.} Full data for the observations we make in Section \ref{sec:single-network}.

\textbf{Appendix \ref{app:more_fits}.} Partial fits (e.g., just width, depth, or dataset size) for the joint scaling law in Section \ref{sec:joint}.

\textbf{Appendix \ref{app:more_arch_alg}.} Demonstrating that our functional form applies to additional networks and datasets.

\textbf{Appendix \ref{app:more_extrapolations}.} A discussion of extrapolation for our scaling law.

\textbf{Appendix \ref{app:comparison-to-rosenfeld}.} A more detailed comparison between our scaling law and that of \citet{rosenfeld2020a} following up on Section \ref{sec:design}.

\textbf{Appendix \ref{app:magic-one-percent}.} How we computed the effect of error dips on our estimator in Section \ref{sec:design}.

\newpage

\section{Formal Statement of Iterative Magnitude Pruning}
\label{app:pruningalg}

{
\begin{algorithm}[H]
    \small
    \caption{Iterative Magnitude Pruning (IMP) with weight rewinding to epoch 10 and $N$ iterations.}
    \begin{algorithmic}[1]
    \State Create a neural network with randomly initialized weights $W_0 \in \mathbb{R}^d$ and initial pruning mask $m = 1^{d}$
    \State Train $W_0$ to epoch 10, resulting in weights $W_{10}$
    \For{$n \in \{1, \ldots, N\}$}
    \State Train $m \odot W_{10}$ (the element-wise product of $m$ and $W_{10}$) to final epoch $T$ and weights $m \odot W_{T, n}$
    \State Prune the 20\% of weights in $m \odot W_{T, n}$ with the lowest magnitudes. $m[i] = 0$ if $W_{T, n}[i]$ is pruned
    \EndFor
    \State Return $m$ and $W_{T, n}$ 
    \end{algorithmic}
    \label{alg:imp}
\end{algorithm}
}






\newpage
\section{Experimental Details} \label{app:resnets}

\subsection{ResNets}

We study the residual networks (ResNets) designed by \citet{he2016deep} for CIFAR-10 and ImageNet.
ResNets for CIFAR-10 are composed of an initial convolutional layer, three sets of $B$ residual blocks (each with two convolutional layers and a skip connection), and a linear output layer.
The sets of blocks have 16, 32, and 64 convolutional channels, respectively.

ResNets for ImageNet and TinyImageNet are composed of an initial convolutional layer, a max-pooling layer, four sets of residual blocks (each with three convolutional layers and a skip connection), and a linear output layer.
The sets of blocks have 64, 128, 256, and 512 convolutional channels, respectively.
On ImageNet, we use a ResNet with 50 layers. On TinyImageNet, we use a ResNet with 18 layers. Both choices are standard for these datasets.

We place batch normalization before the ReLU activations.

To vary the width of the networks, we multiply the number of convolutional channels by the width scaling factor $w$.
To vary the depth of the CIFAR-10 ResNets, we vary the value of $B$.
The depth $l$ of the network is the total number of the layers in the network, not counting skip connections.

\subsection{VGG Networks}

We study the VGG-16 variant of the VGG networks for CIFAR-10 as provided by the OpenLTH repository.\footnote{\texttt{github.com/facebookresearch/open\_lth}}
The network is divided into five sections, each of which is followed by max pooling with kernel size 2 and stride 2.
The sections contain 3x3 convolutional layers arranged as follows:

\begin{center}

\begin{tabular}{c c c}
\toprule
Section & Width & Layers \\
\midrule
1 & 64 & 2 \\
2 & 128 & 2 \\
3 & 256 & 3 \\
4 & 512 & 3 \\
5 & 512 & 3 \\
\bottomrule
\end{tabular}
\end{center}

The network has ReLU activations and batch normalization before each activation.
To vary the width of VGG-16, we multiply each of the per-segment widths by the width scaling factor $w$.

\subsection{DenseNets}

We study the densely connected residual networks (DenseNets) designed by \citet{he2016deep} for CIFAR-10.
DenseNets for CIFAR-10 are composed of an initial convolutional layer, four sets of dense blocks, and a linear output layer.
Between the sets of blocks are transition layers of 1x1 convolutions and an average pooling operation that downsamples the image by 2x.
Each block comprises a 1x1 convolution that increases the channel count by 4x and a 3x3 block that decreases it to a fixed constant size $g$; this output is then concatenated to the input of the block.
As such, if the input to the block has $n$ channels, the output of the block has $n + g$ channels.
We use DenseNet-121, which has sets of 6, 12, 24, and 16 blocks.
$g$ is set to 16 but is multiplied by the width scaling factor $w$ to modify the width.

\subsection{Training Hyperparameters}

We train CIFAR-10 and SVHN ResNets and VGG-16 for 160 epochs with a batch size of 128.
The initial learning rate is 0.1, and it drops by an order of magnitude at epochs 80 and 120.
We optimize using SGD with momentum (0.9).
We initialize with He uniform initialization.
CIFAR-10 data is augmented by normalizing, randomly flipping left and right, and randomly shifting by up to four pixels in any direction (and cropping afterwards).
SVHN data is not augmented.

We train CIFAR-10 DenseNets with the same hyperparameters but for 200 epochs (with learning rate drops at 130 and 165 epochs).
We train SVHN DenseNets with the same hyperparameters but for 100 epochs (with learning rate drops at 70 and 85 epochs).

We train ImageNet ResNets for 90 epochs with a batch size of 1024.
The initial learning rate is 0.4, and it drops by an order of magnitude at epochs 30, 60, and 80.
We perform linear learning rate warmup from 0 to 0.4 over the first 5 epochs.
We optimize using SGD with momentum (0.9).
We initialize with He uniform initialization.
Data is augmented by normalizing, randomly flipping left and right, selecting a random aspect ratio between 0.8 and 1.25, selecting a random scaling factor between 0.1 and 1.0, and cropping accordingly.

We train TinyImageNet ResNets identically except for that we train them for 200 epochs (with learning rate drops at 100 and 150 epochs) and a learning rate of 0.2.
Augmentation is identical to ImageNet.

\subsection{Dimensions}

We use the following dimensions for the additional experiments. We select configurations using the same methodology as in Table 1.

\begin{center}
{\scriptsize
\begin{tabular}{@{\ }l@{\ }|@{\ }c@{\ }|@{\ }c@{\ }|@{\ }c@{\ }|@{\ }c}
\toprule
Network Family &
    Densities ($d$) &
    Depths ($l$) &
    Width Scalings ($w$) &
    Subsample Sizes ($n$) \\ \midrule
CIFAR-10/SVHN ResNet &
    $0.8^i, i \subseteq \{0, \ldots, 40\}$ &
    $l \subseteq$ \{8, 14, 20, 26, 50, 98\} &
    $2^i, i \subseteq \{-4, \ldots, 2\}$ &
    $\frac{N}{i}$, $i \in \{1, 2, 4, 8, 16, 32, 64\}$ \\
CIFAR-10/SVHN VGG &
    $0.8^i, i \subseteq \{0, \ldots, 37\}$ &
    16 &
    $2^i, i \subseteq \{-4, \ldots, 0\}$ &
    $\frac{N}{i}$, $i \in \{1\}$ \\
CIFAR-10/SVHN DenseNet &
    $0.8^i, i \subseteq \{0, \ldots, 50\}$ &
    121 &
    $2^i, i \subseteq \{-4, \ldots, 1\}$ &
    $\frac{N}{i}$, $i \in \{1\}$ \\
TinyImageNet ResNet &
    $0.8^i, i \subseteq \{0, \ldots, 50\}$ &
    18 &
    $2^i, i \subseteq \{-6, \ldots, 0\}$ &
    $\frac{N}{i}$, $i \in \{1\}$ \\
ImageNet ResNet &
    $0.8^i, i \subseteq \{0, \ldots, 30\}$ &
    50 & 
    $2^i, i \subseteq \{-4, \ldots, 0\}$ &
    $\frac{N}{i}$, $i \in \{1, 2, 4\}$ \\ 
\bottomrule
\end{tabular}
}
\end{center}

\clearpage

\section{Full Data for Key Observations in Section \ref{sec:single-network}}
\label{app:sec3-key-observations-alldimensions}

\begin{figure}[h!]
\centering 
\begin{minipage}{0.34\textwidth}
\includegraphics[width=\linewidth,trim={0.2cm 0 0.1cm 0.65cm},clip]{figures_3/pruning_curves_widths.pdf}
\end{minipage}\hfil 
\begin{minipage}{0.295\textwidth}
  \includegraphics[width=\linewidth,trim={2cm 0 0 0.6cm},clip]{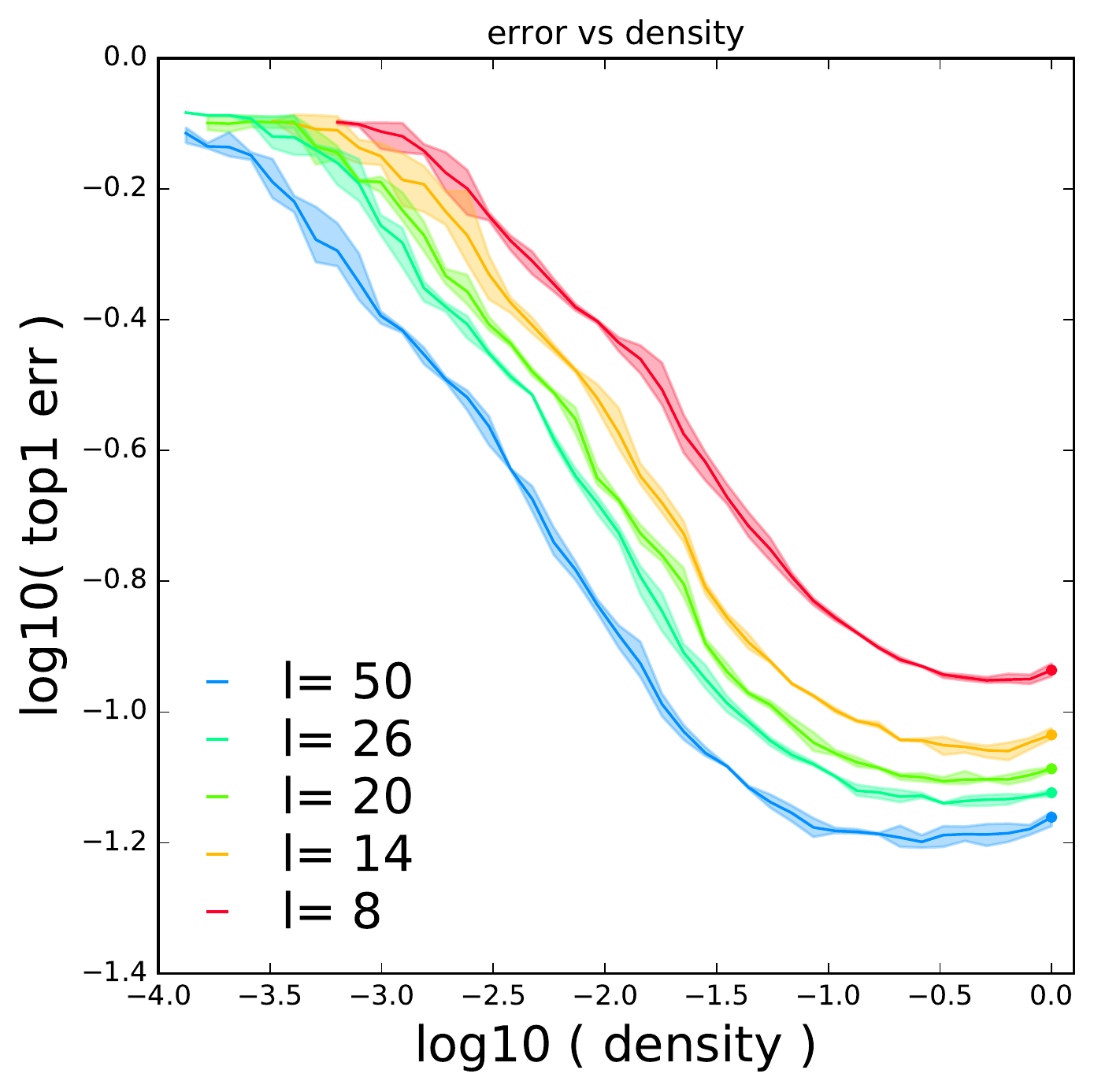}
\end{minipage}\hfil 
\begin{minipage}{0.295\textwidth}
  \includegraphics[width=\linewidth,trim={2cm 0 0 0.6cm},clip]{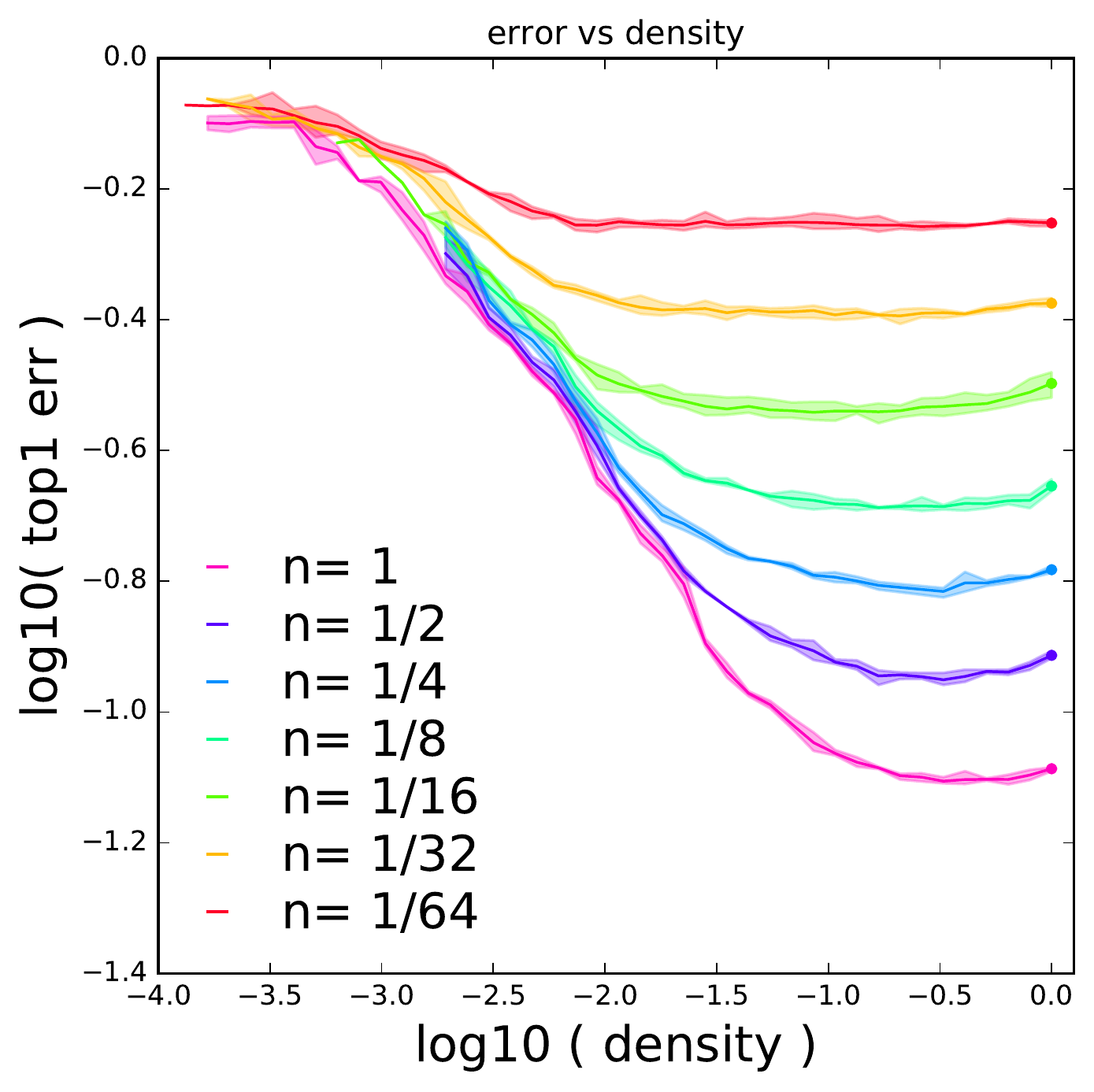}
\end{minipage}
\caption{Relationship between density and error when pruning CIFAR-10 ResNets and varying $w$ (left, $l=20$, $n=N$),  $l$ (center, $w=1$, $n=N$), $n$ (right, $l=20$, $w=1$) }
\vspace{0mm}
\label{fig:cifar_observations_all}
\end{figure}

\begin{figure}[h!]
\centering 
\begin{minipage}{0.34\textwidth}
\includegraphics[width=\linewidth,trim={0.2cm 0 0.1cm 0.65cm},clip]{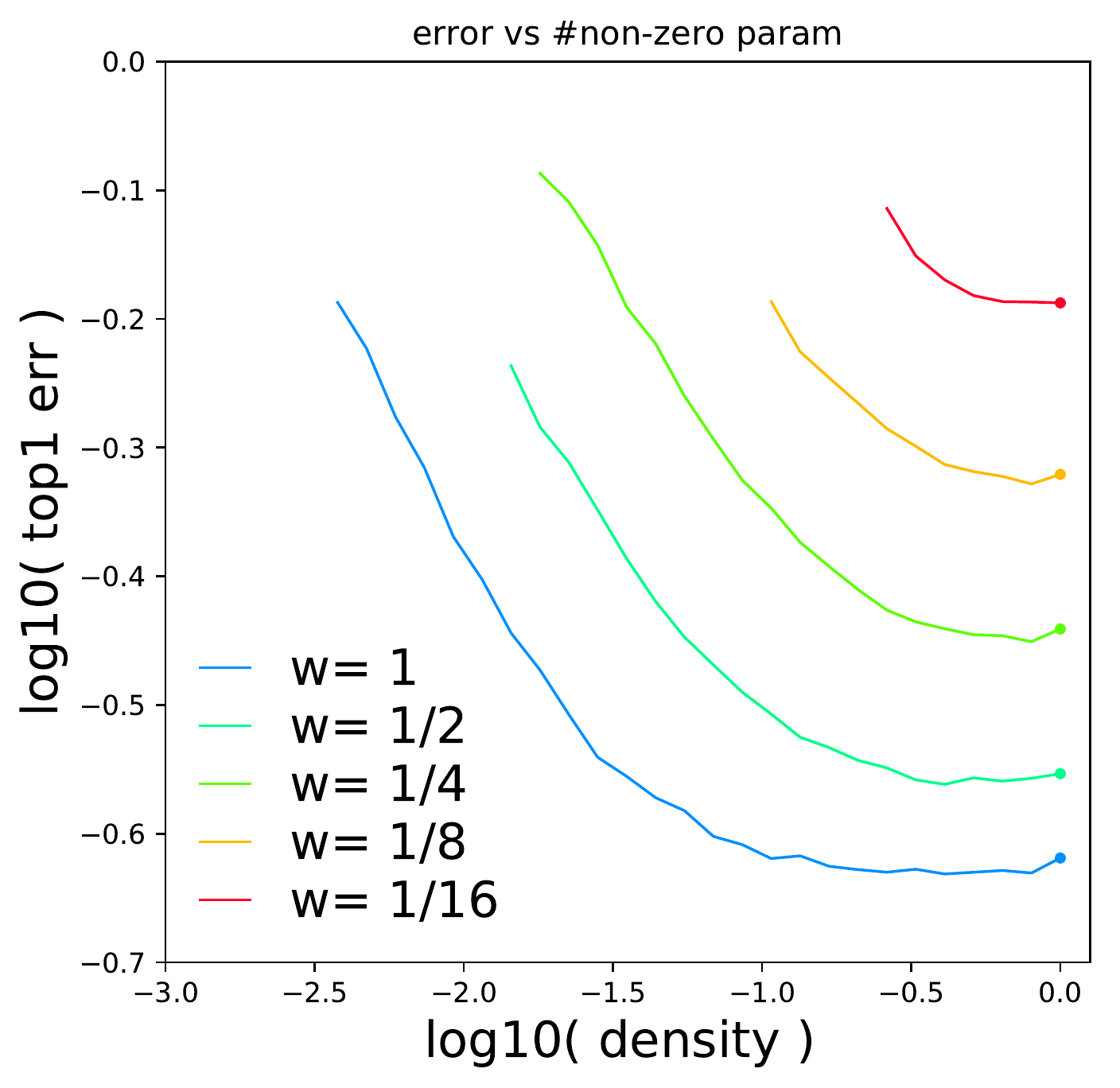}
\end{minipage}
\begin{minipage}{0.3\textwidth}
  \includegraphics[width=\linewidth,trim={2cm 0 0 0.6cm},clip]{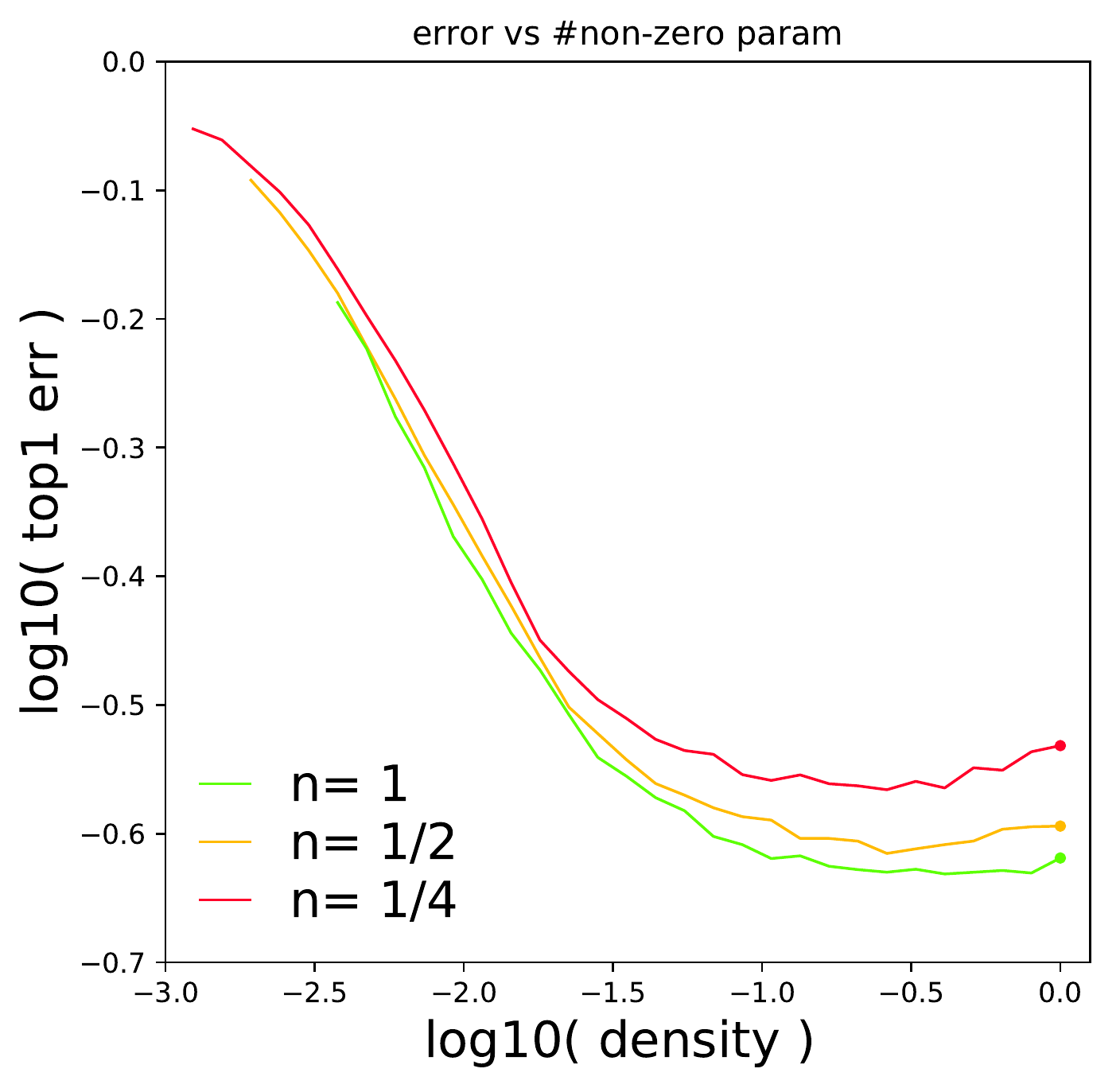}
\end{minipage}
\caption{Relationship between density and error when pruning Imagenet ResNet-50 and varying $w$ (left, $n=N$),  and $n$ (right, $w=1$) }
\vspace{0mm}
\label{fig:imagenet_observations_all}
\end{figure}

In this appendix, we show that our observations from Section \ref{sec:single-network} hold when varying all dimensions (depth, width, and dataset size) on both the CIFAR-10 and ImageNet ResNets for IMP.
Figure \ref{fig:cifar_observations_all} shows the error versus density when changing width (left) depth (center) and data (right). 
In Figure \ref{fig:imagenet_observations_all}, we similarly show the dependency of the error on density for Imagenet when varying width (left) and dataset size (right).

In Figure \ref{fig:cifar_observations_all}, we observe that all curves have a similar slope in the power-law region.
In Equation \ref{eq:prune_density}, this implies that while $\gamma$ is allowed to vary with $l$, $w$ and $n$, it is in practice approximately a constant.
Similarly, the high-error plateau $\epsilon^\uparrow$ is also shared across curves such that it too is approximately constant.
In contrast, the transition from high-error plateau to the power-law region is not constant as a function of density.
Section \ref{sec:joint} finds exactly this dependency of the transition parameter $p$.

\clearpage

\section{Partial (Projections) Fit Results for Section \ref{sec:joint}}
\label{app:more_fits}

In Section \ref{sec:joint}, we fit the error jointly as a function of all dimensions showing that Equation \ref{eq:intermediate_state} provides a good approximation to the error in practice.
In this appendix, we consider important sub-cases, such as the case when one wishes to scale only one degree of freedom while pruning. This serves both a practical scenario, but also allows for a qualitative visualization of the fit (and typical sources of error), which is otherwise difficult to perform over all dimensions jointly.
From a practical standpoint, in this case one need not estimate the parameters associated with the fixed degree of freedom.

Recall that, given the non-pruned network error $\epsilon_{np}$, all dependencies on the individual structural degrees of freedom $l,w$ are captured by the invariant $m^* \triangleq l^\phi w^\psi d$.
This means that, if one wishes to estimate the error while pruning when holding width fixed, we need not estimate $\psi$.
Similarly if depth is held constant, we need not estimate $\phi$.

\begin{figure}[h!]
         \begin{minipage}{0.34\textwidth}
            \includegraphics[width=\linewidth,trim={0.2 0 0 0.65cm},clip]{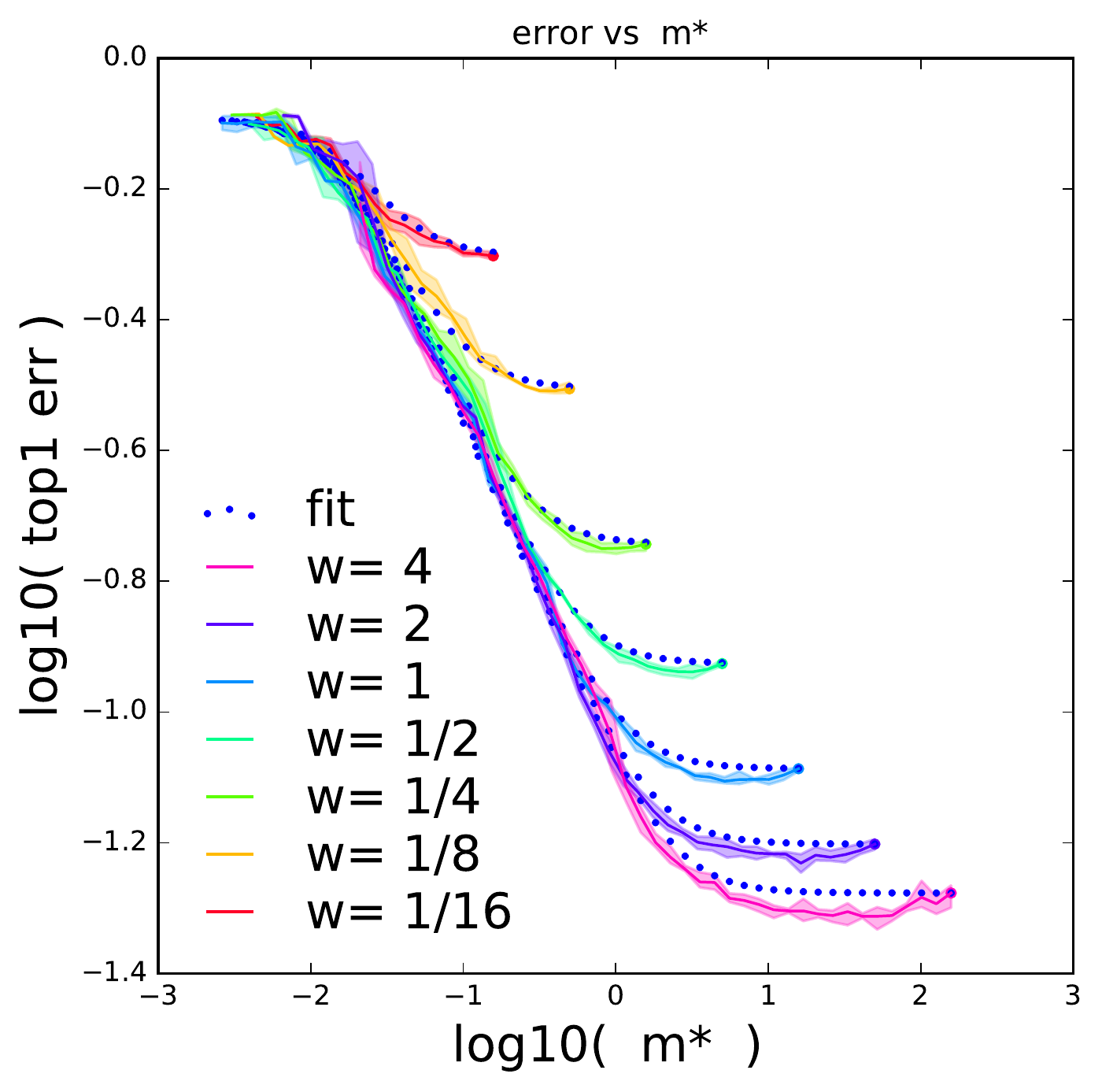}\llap{\makebox[2.15cm][l]{\raisebox{2.45cm}{\includegraphics[scale=0.17,trim={1.9cm 1.5cm 0.3cm 0.3cm},clip]{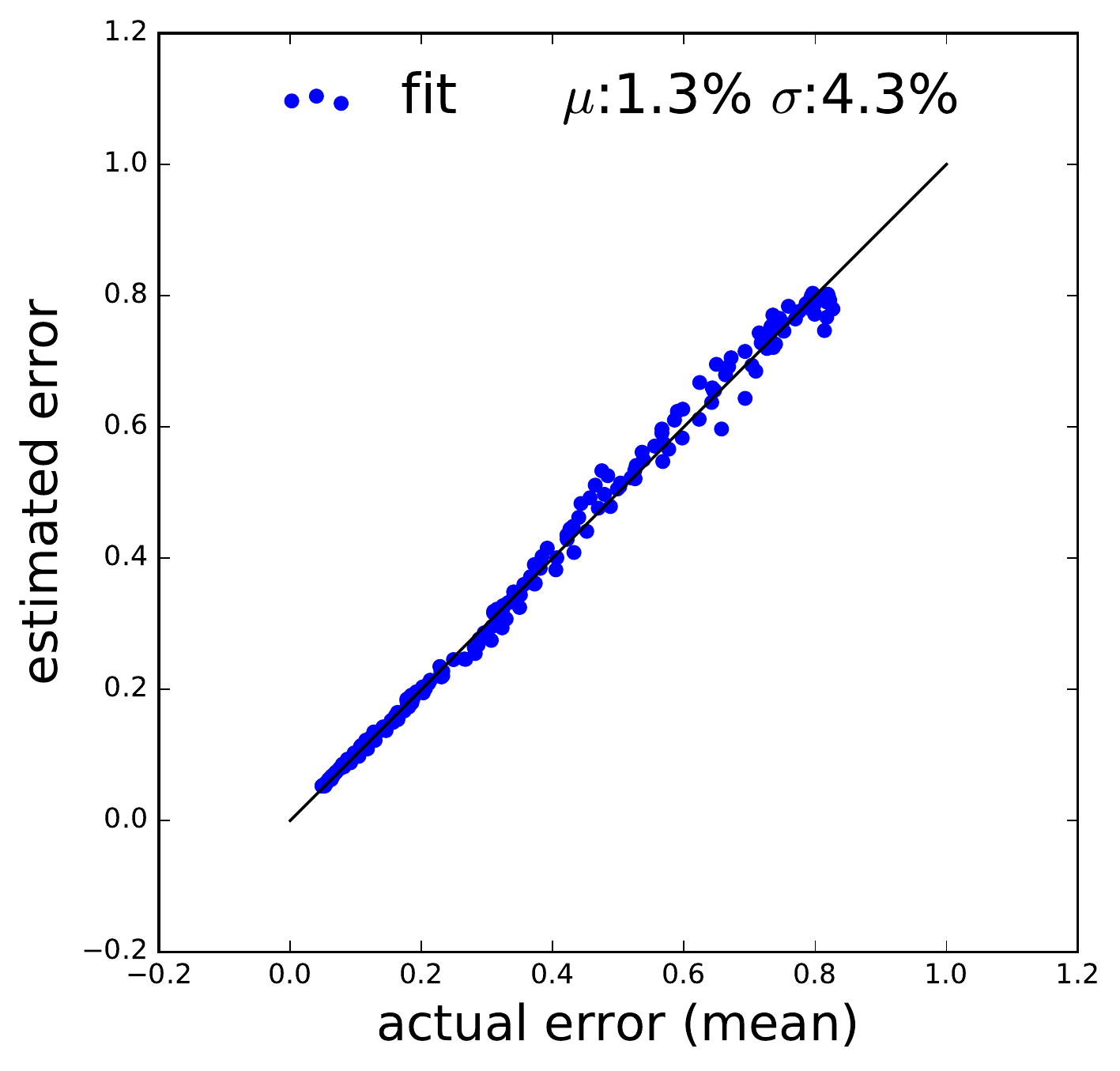}}}}
        \end{minipage}%
        \begin{minipage}{0.292\textwidth}
            \includegraphics[width=\linewidth,trim={2cm 0 0 0.65cm},clip]{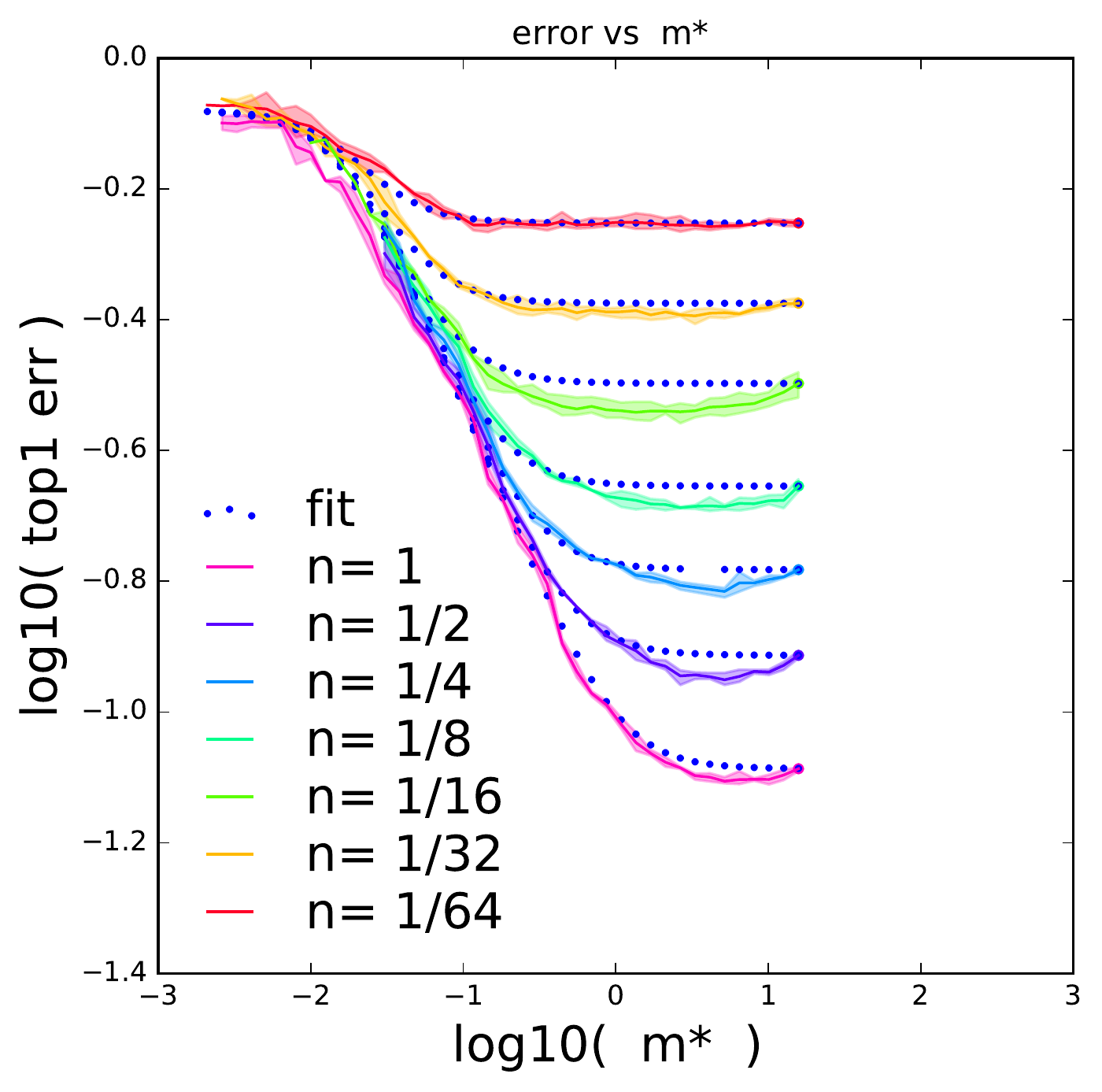}\llap{\makebox[2.25cm]{\raisebox{2.45cm}{\transparent{0.7}\includegraphics[scale=0.17,trim={1.9cm 1.5cm 0.3cm 0.3cm},clip]{figures_3/fit_corr_True,False,False.pdf}}}}
        \end{minipage}%
        \begin{minipage}{0.292\textwidth}
            \includegraphics[width=\linewidth,trim={2cm 0 0 0.65cm},clip]{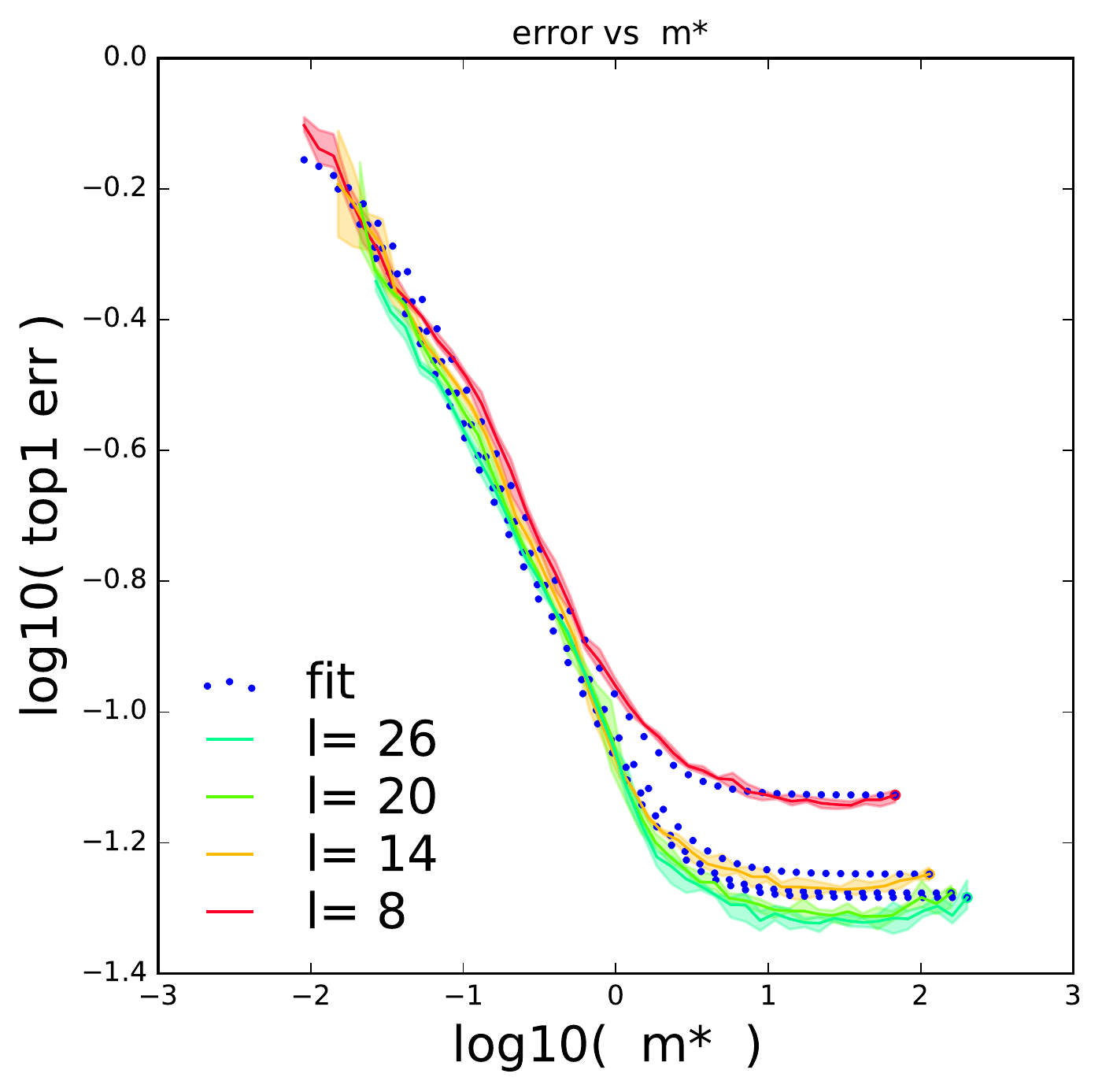}\llap{\makebox[2.25cm]{\raisebox{2.45cm}{\includegraphics[scale=0.17,trim={1.9cm 1.5cm 0.3cm 0.3cm},clip]{figures_3/fit_corr_True,False,False.pdf}}}}
        \end{minipage}
        
         \begin{minipage}{0.338\textwidth}
            \includegraphics[width=\linewidth,trim={0.2 0 0.3cm 0.65cm},clip]{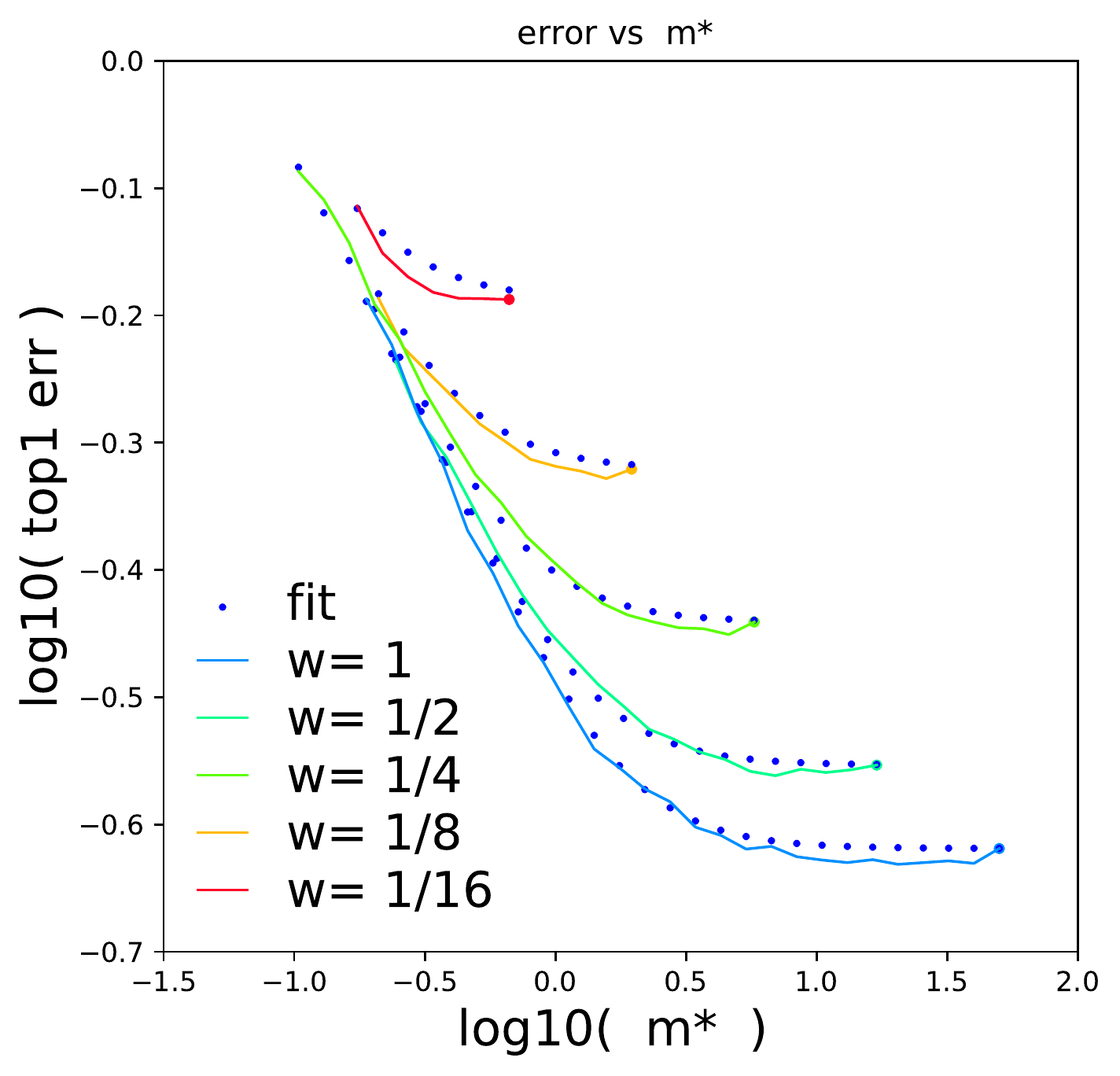}\llap{\makebox[2.1cm][l]{\raisebox{2.39cm}{\transparent{0.8}\includegraphics[scale=0.17,trim={1.8cm 1.5cm 0.3cm 0.3cm},clip]{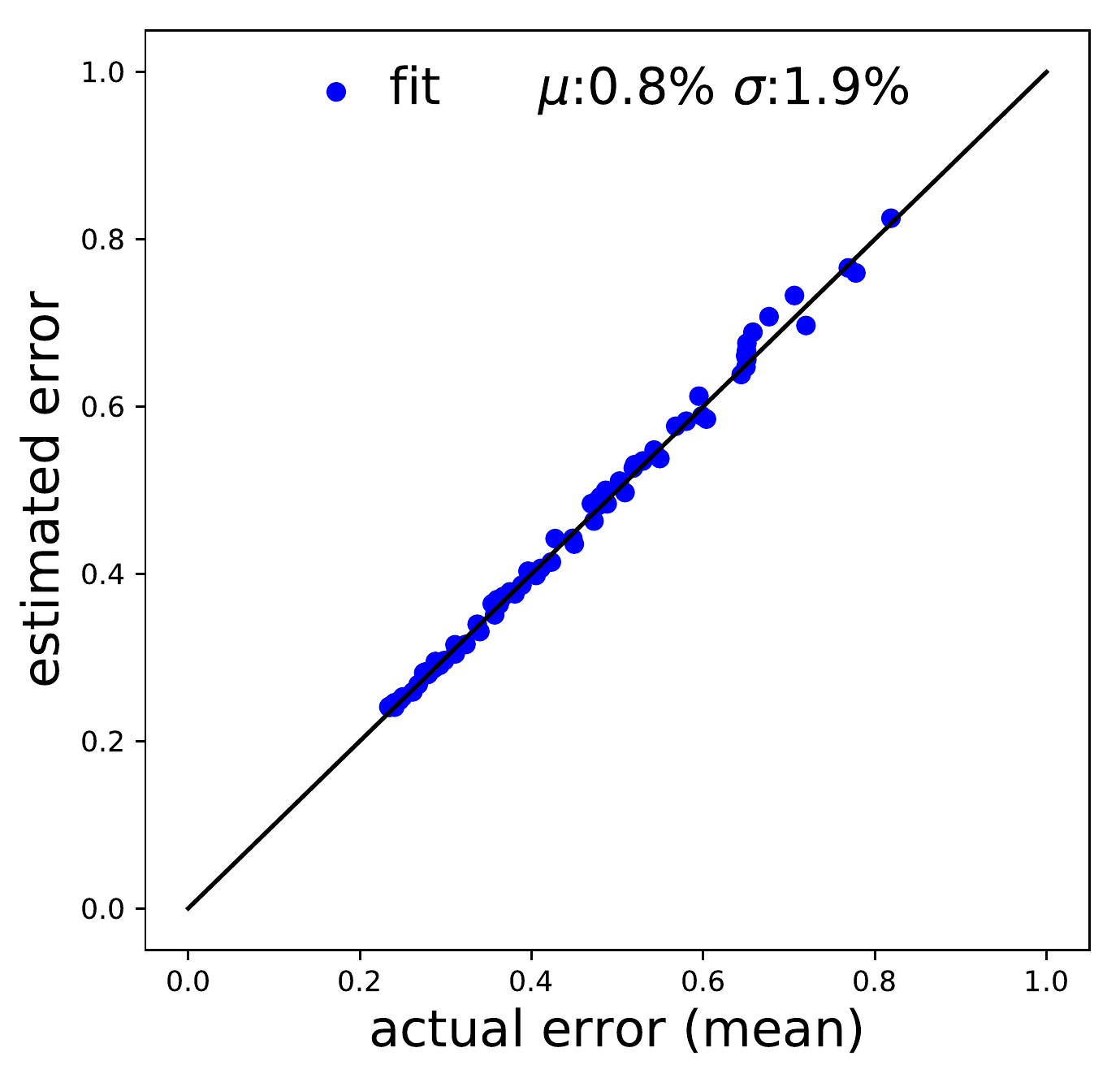}}}}
        \end{minipage}
        \begin{minipage}{0.3\textwidth}
            \includegraphics[width=\linewidth,trim={1.9cm 0 0 0.65cm},clip]{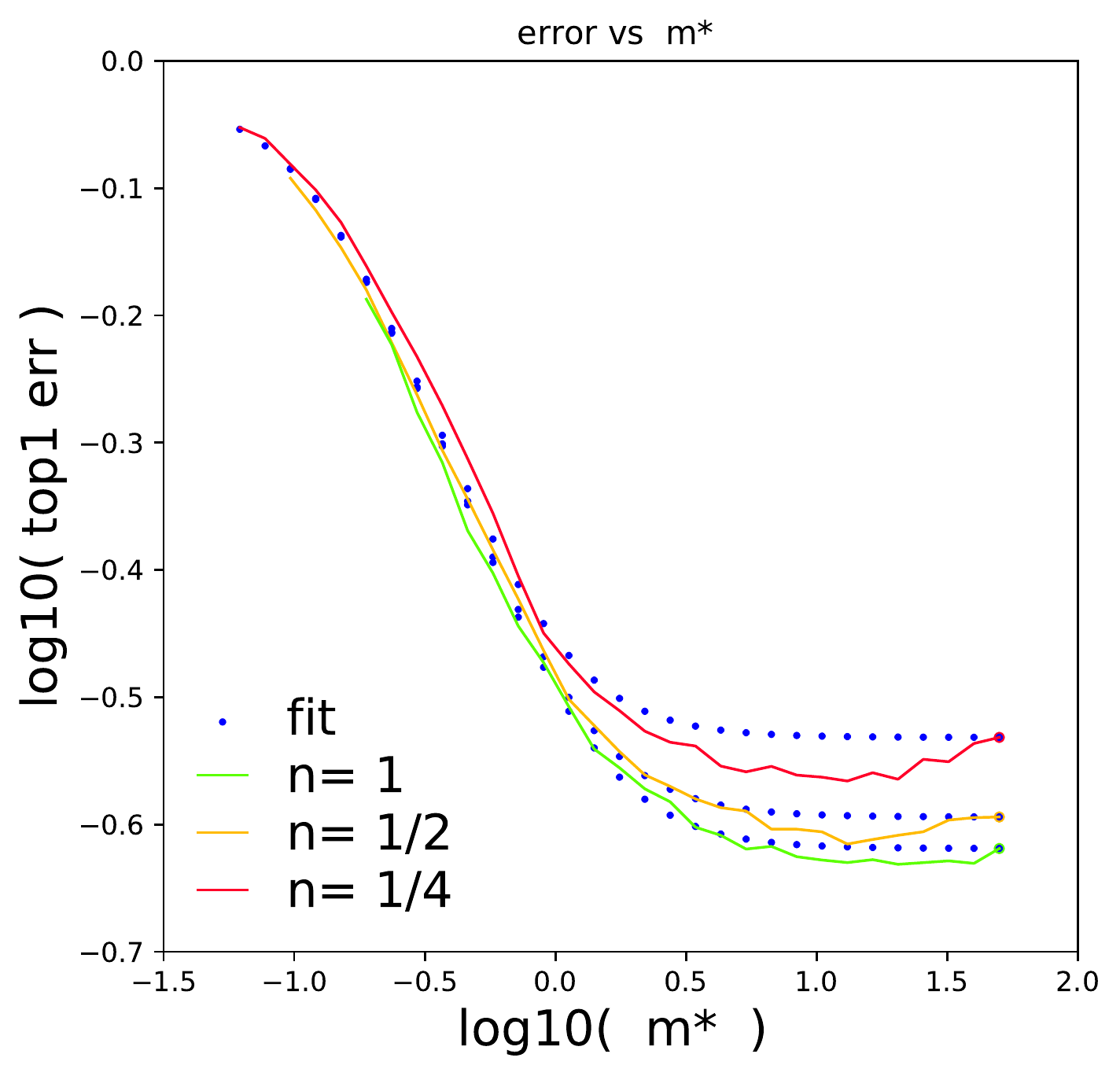}\llap{\makebox[2.2cm][l]{\raisebox{2.39cm}{\includegraphics[scale=0.17,trim={1.8cm 1.5cm 0.3cm 0.3cm},clip]{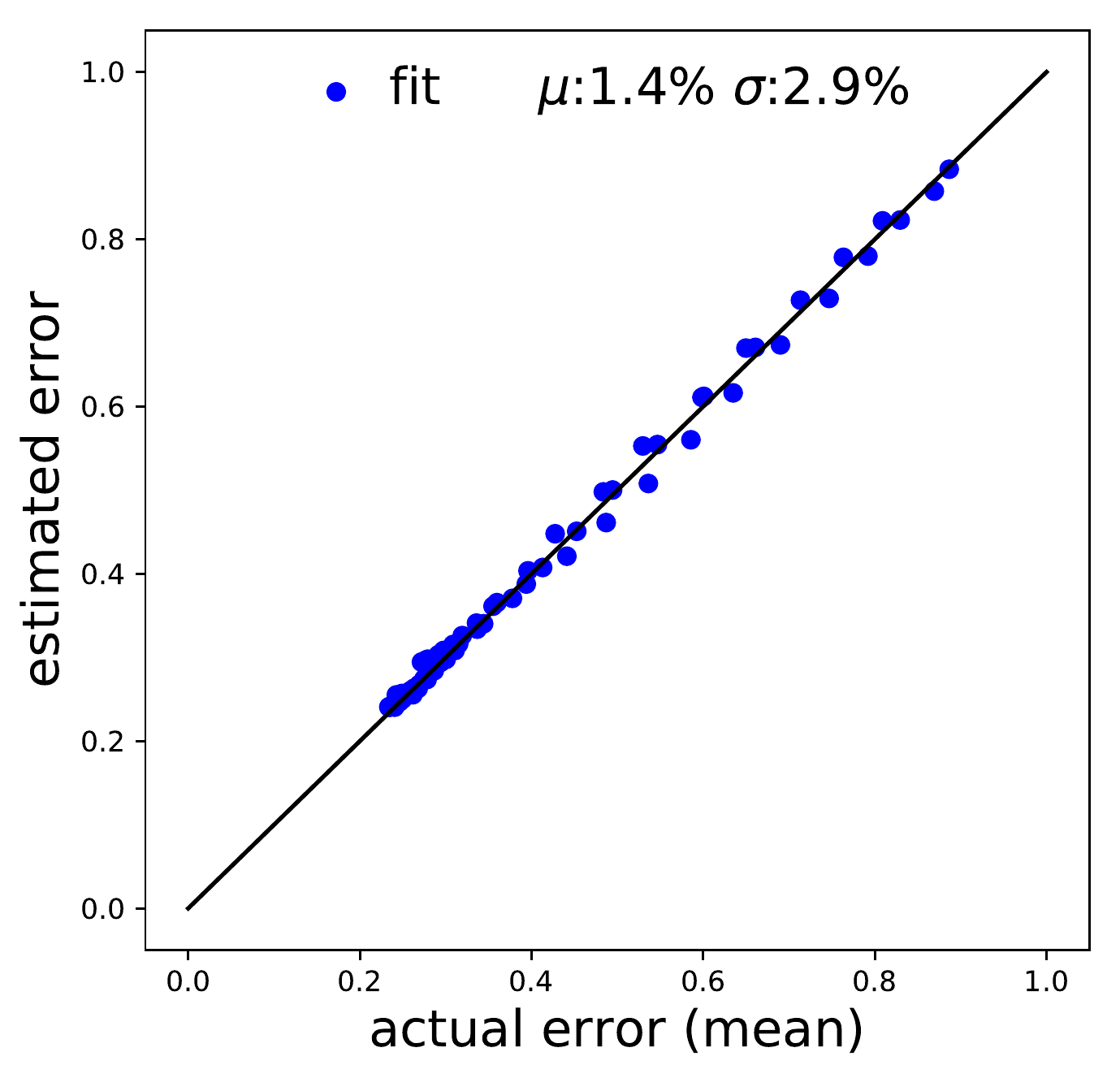}}}}
        \end{minipage}
\vspace{-5pt}
\caption{Top row: CIFAR-10. Bottom row: ImageNet. Left: varying width. Center: varying dataset size. Right: varying depth. Lines are the actual error and dots are the estimated error.}
\label{fig:cifar_partial_fit}
\end{figure}

Figure \ref{fig:cifar_partial_fit} shows these partial fits.
Shown from left to right are the fits done while pruning and varying width, depth and data respectively. Correspondingly, these fits omit separately $\psi$ or $\phi$ or omit both when depth nor width are scaled.
The fits were performed with all available density points for each dimension. For CIFAR-10: 7 widths, 224 points for the width partial fit; 7 dataset fractions, 240 points for the data partial fit; 4 depths, 164 points for the depth partial fit. For ImageNet: 5 widths, 83 points for the width partial fit; 3 dataset fractions, 86 points for the data partial fit.

This exercise, apart from its practical implications, highlights the fact that there are in effect two groups of parameters comprising the estimation. The first are the parameters $\epsilon^\uparrow$, $\gamma$ and $p'$ which control the dependency as a function of density (or more generally, as a function of the invariant). The second are $\phi$ and $\psi$ which are properties of the architectural degrees of freedom captured by the invariant. 
Moreover, within the first group of parameters $\epsilon^\uparrow$, $\gamma$, can be isolated and found from a single pruning curve, as they are not a function of $l,w,n$.

\newpage

\section{Additional Architectures and Datasets}
\label{app:more_arch_alg}

In this appendix, we show that our functional form applies to additional pairs of networks and datasets: (CIFAR-10 ResNet, SVHN), (VGG-16, CIFAR-10), (VGG-16, SVHN), (DenseNet-121, CIFAR-10), (DenseNet-121, SVHN), (ImageNet ResNet-18, TinyImageNet). 

In general, we obtain good fits on CIFAR-10 and TinyImageNet.
On SVHN, fits are worse but the networks suffer from high measurement error (i.e., accuracy varies greatly between multiple runs at the same density) at low densities; nevertheless, fits are often better than measurement error because they average out some of the error.

\begin{figure}[h!]
\centering
\begin{minipage}{0.3\textwidth}
    \includegraphics[width=\linewidth,trim={0.2 0 0 0.65cm},clip]{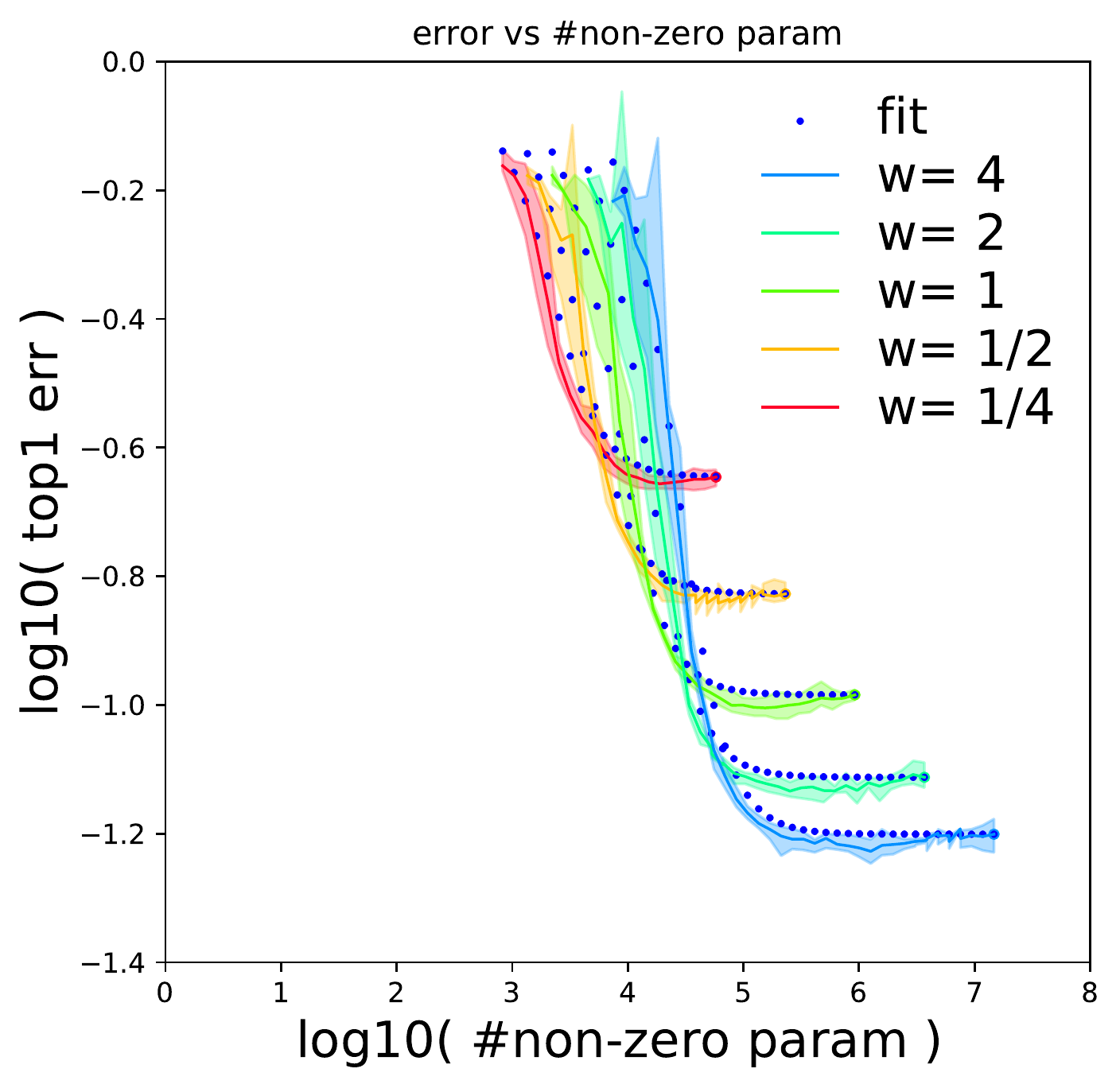}
\end{minipage}
\begin{minipage}{0.3\textwidth}
    \includegraphics[width=\linewidth]{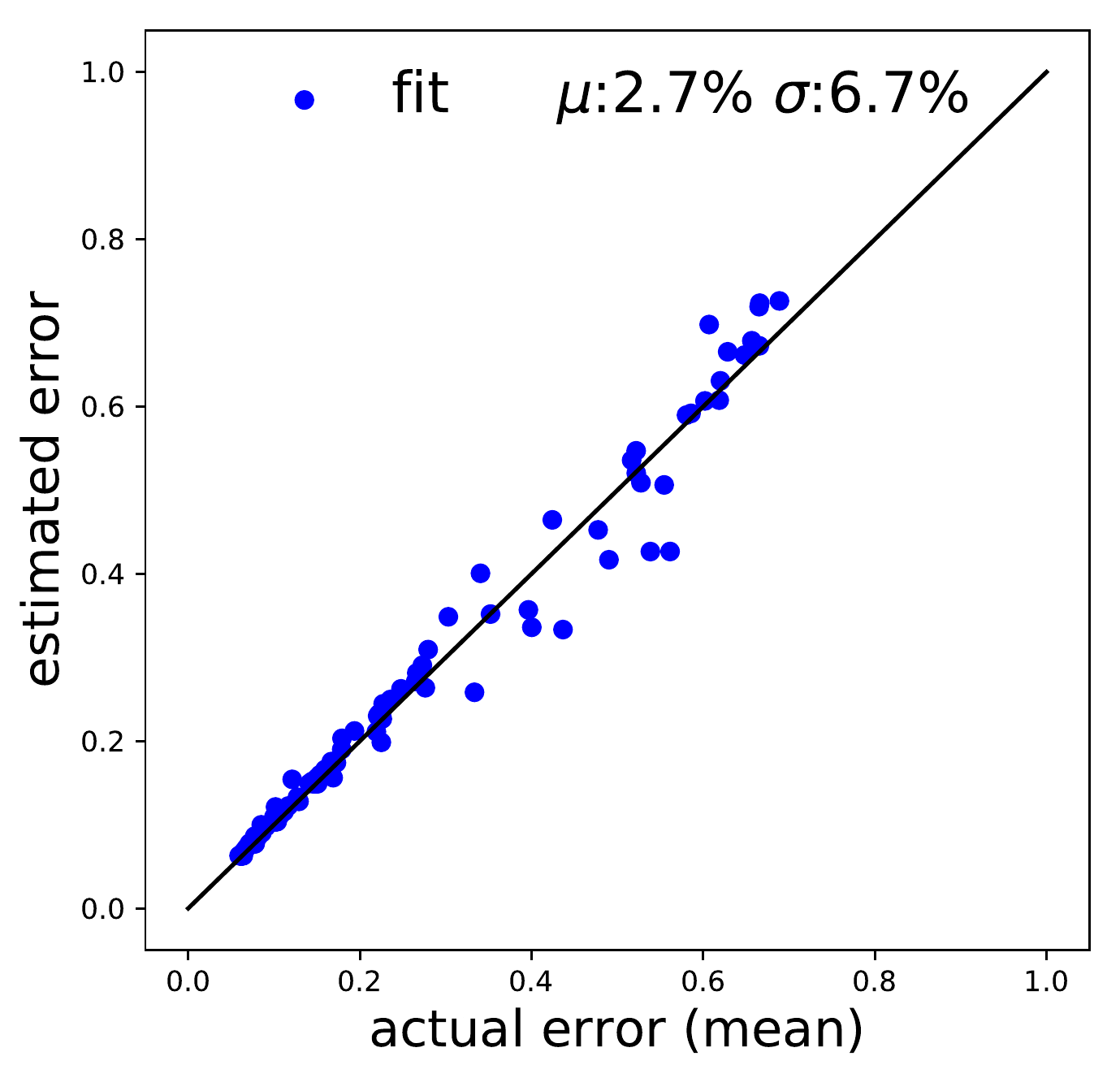}
\end{minipage}
\begin{minipage}{0.3\textwidth}
    \includegraphics[width=\linewidth]{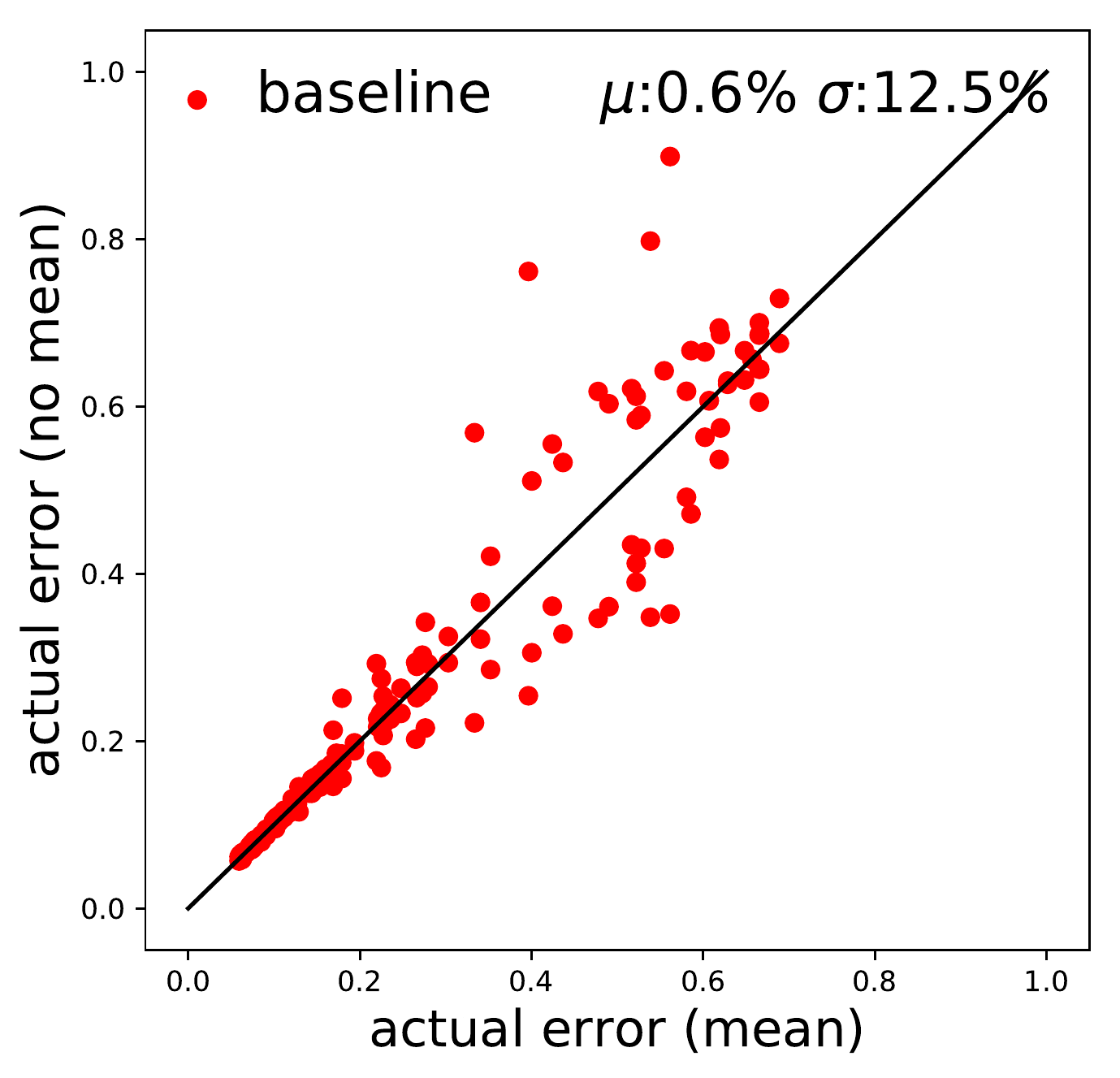}
\end{minipage}
\caption{Fit for VGG-16 on CIFAR-10 with IMP pruning.}
\vspace{0mm}
\label{fig:vgg-imp}
\end{figure}

\begin{figure}[h!]
\centering
\begin{minipage}{0.3\textwidth}
    \includegraphics[width=\linewidth,trim={0.2 0 0 0.65cm},clip]{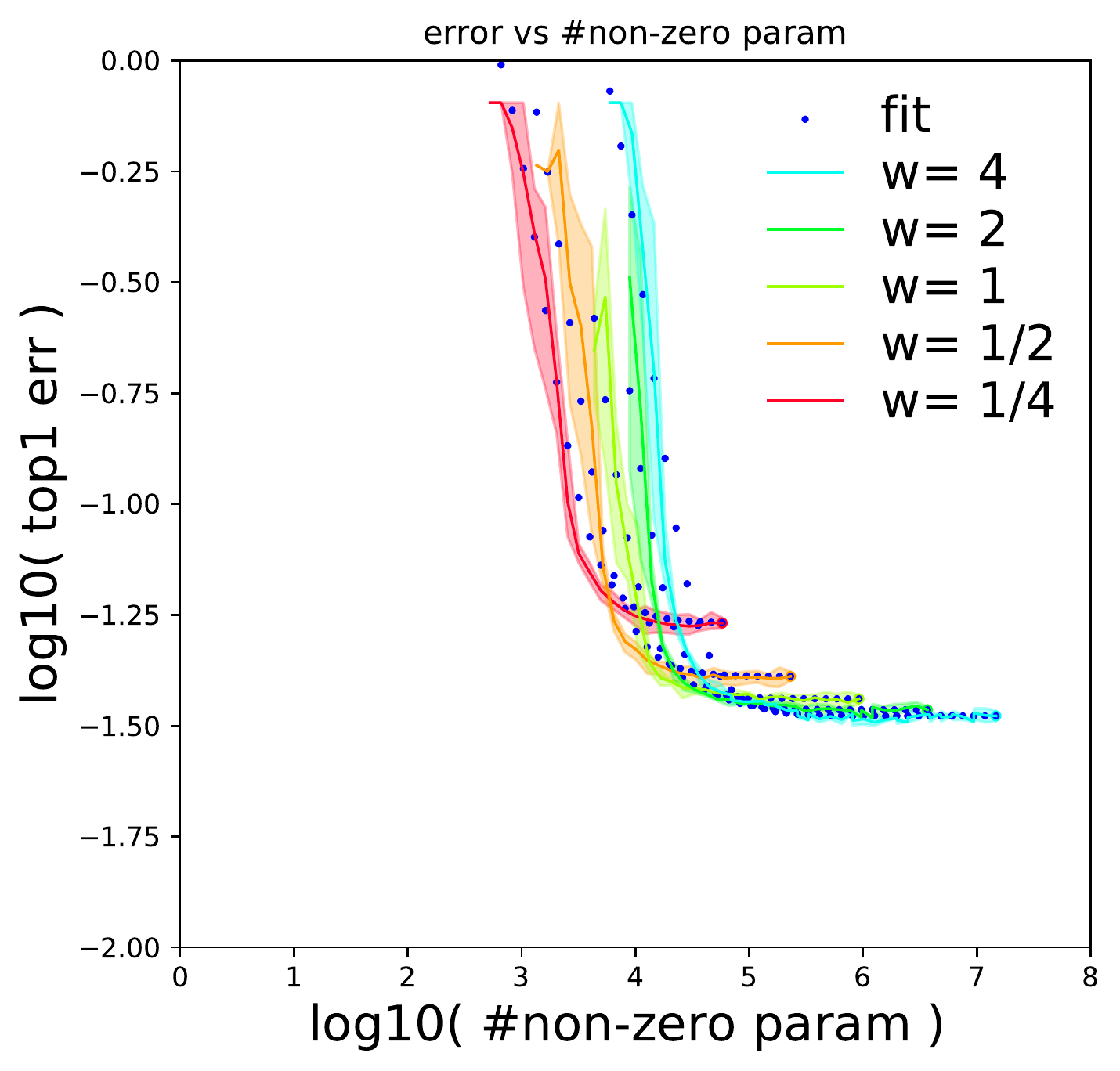}
\end{minipage}
\begin{minipage}{0.3\textwidth}
    \includegraphics[width=\linewidth]{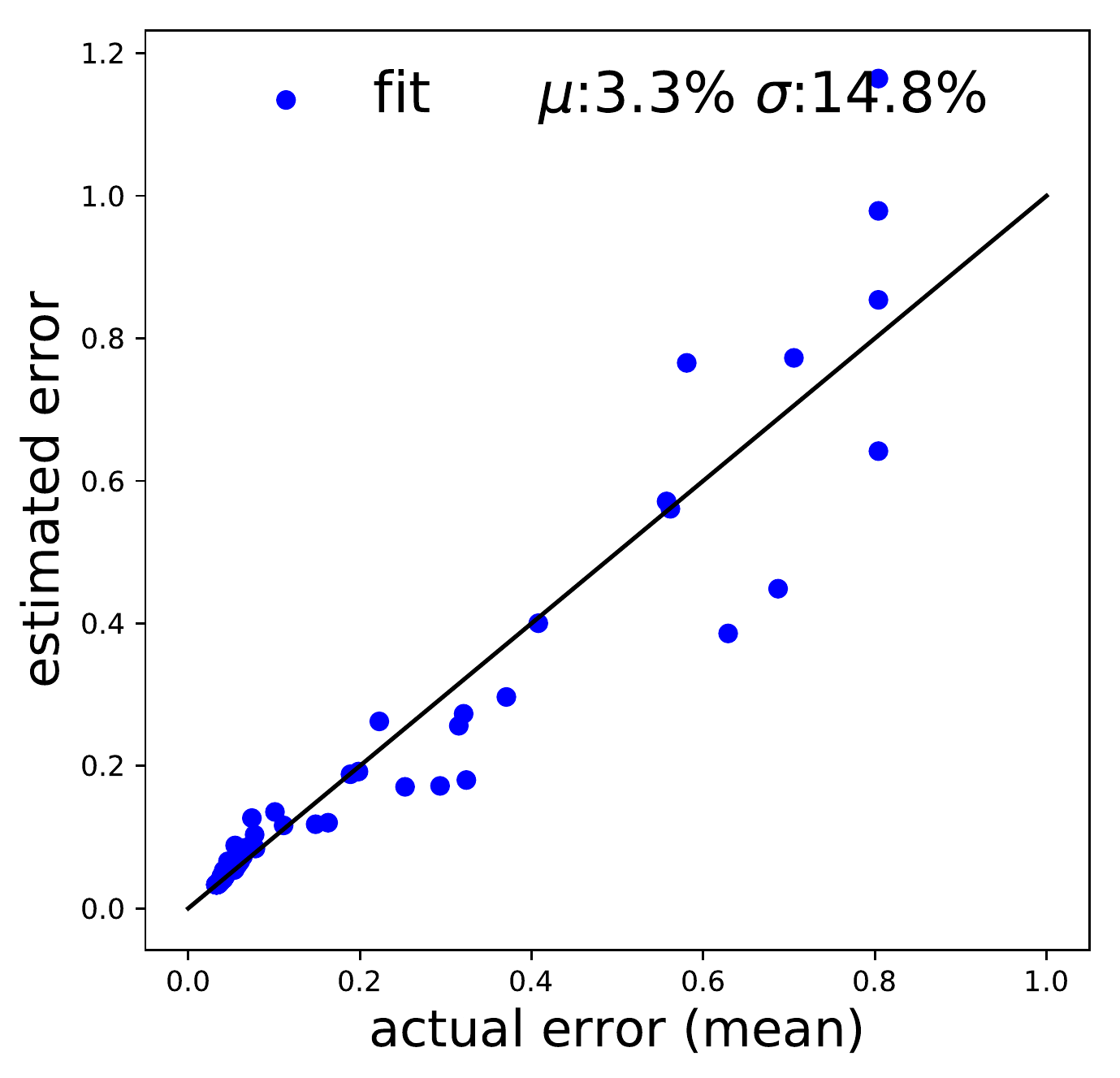}
\end{minipage}
\begin{minipage}{0.3\textwidth}
    \includegraphics[width=\linewidth]{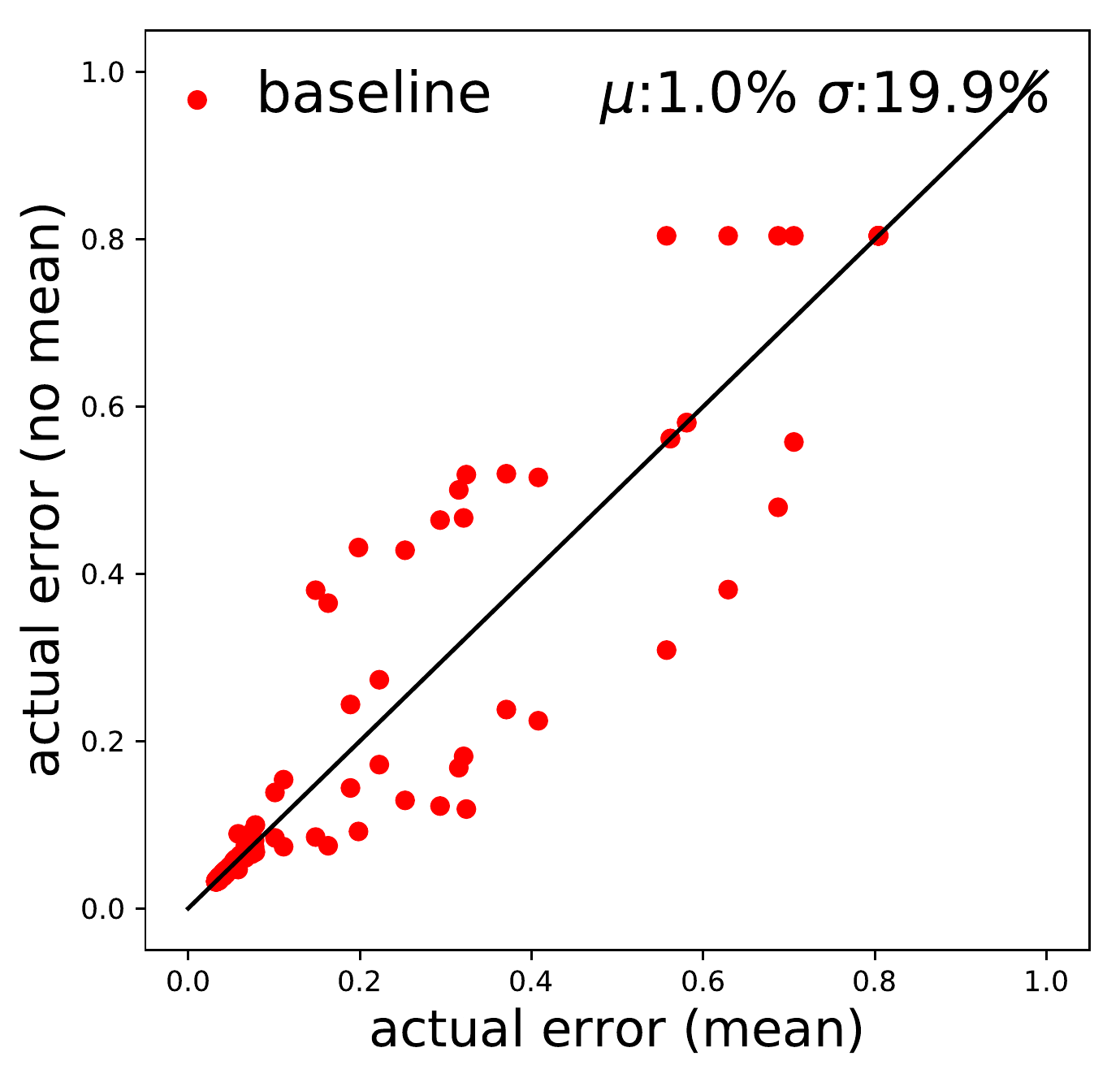}
\end{minipage}
\caption{Fit for VGG-16 on SVHN with IMP pruning.}
\vspace{0mm}
\label{fig:vgg-svhn}
\end{figure}

\begin{figure}[h!]
\centering
\begin{minipage}{0.3\textwidth}
    \includegraphics[width=\linewidth,trim={0.2 0 0 0.65cm},clip]{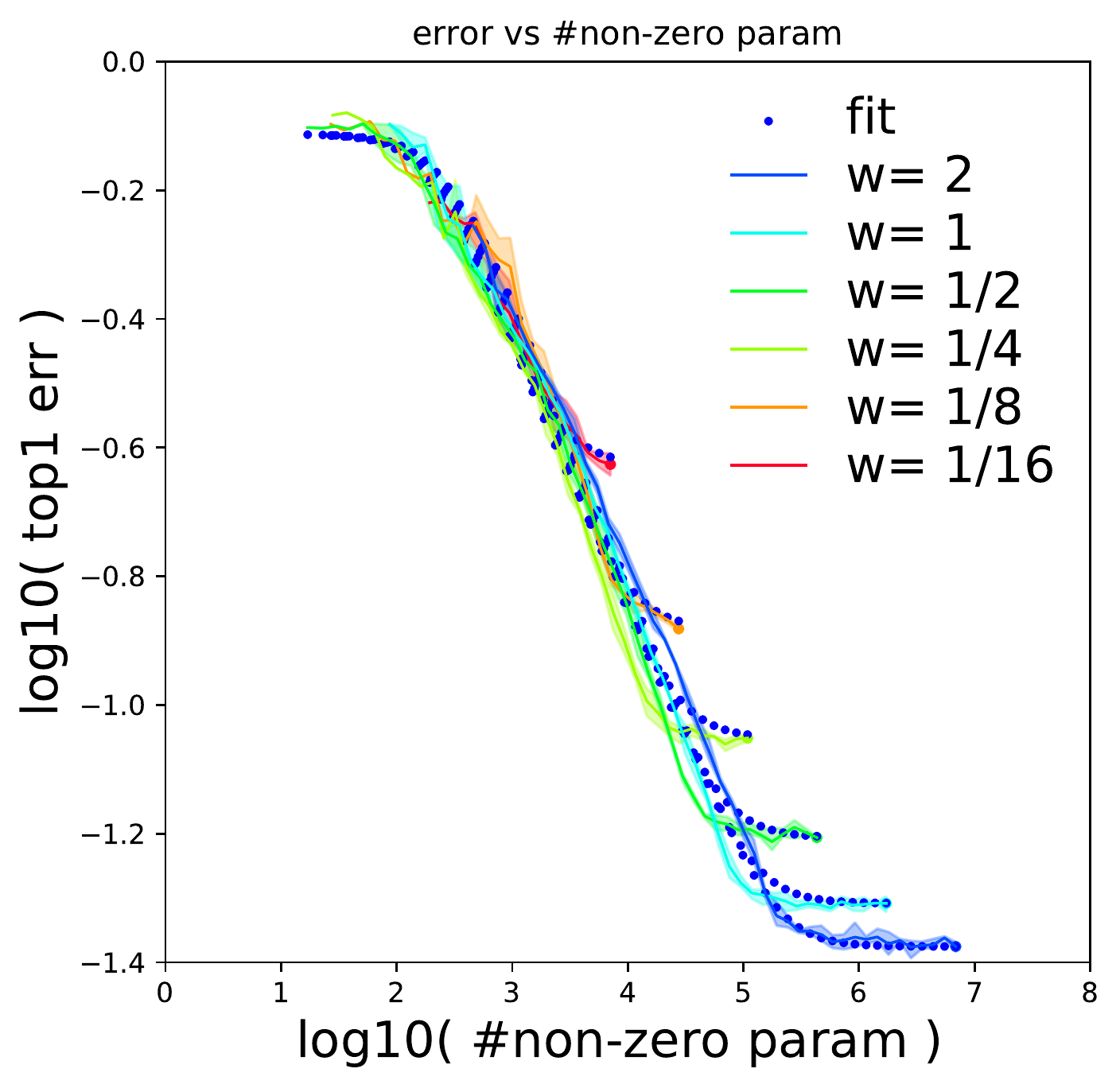}
\end{minipage}
\begin{minipage}{0.3\textwidth}
    \includegraphics[width=\linewidth]{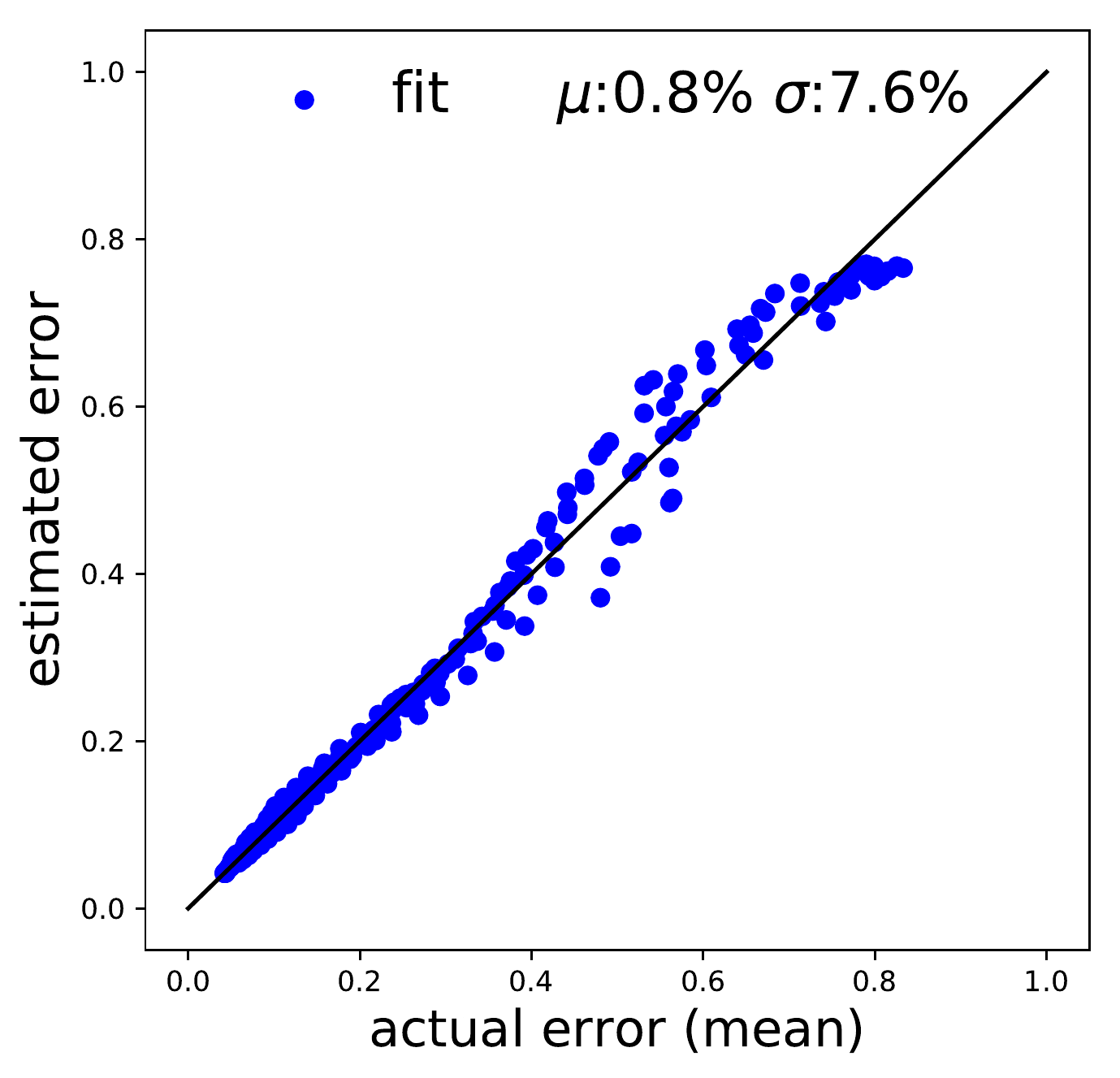}
\end{minipage}
\begin{minipage}{0.3\textwidth}
    \includegraphics[width=\linewidth]{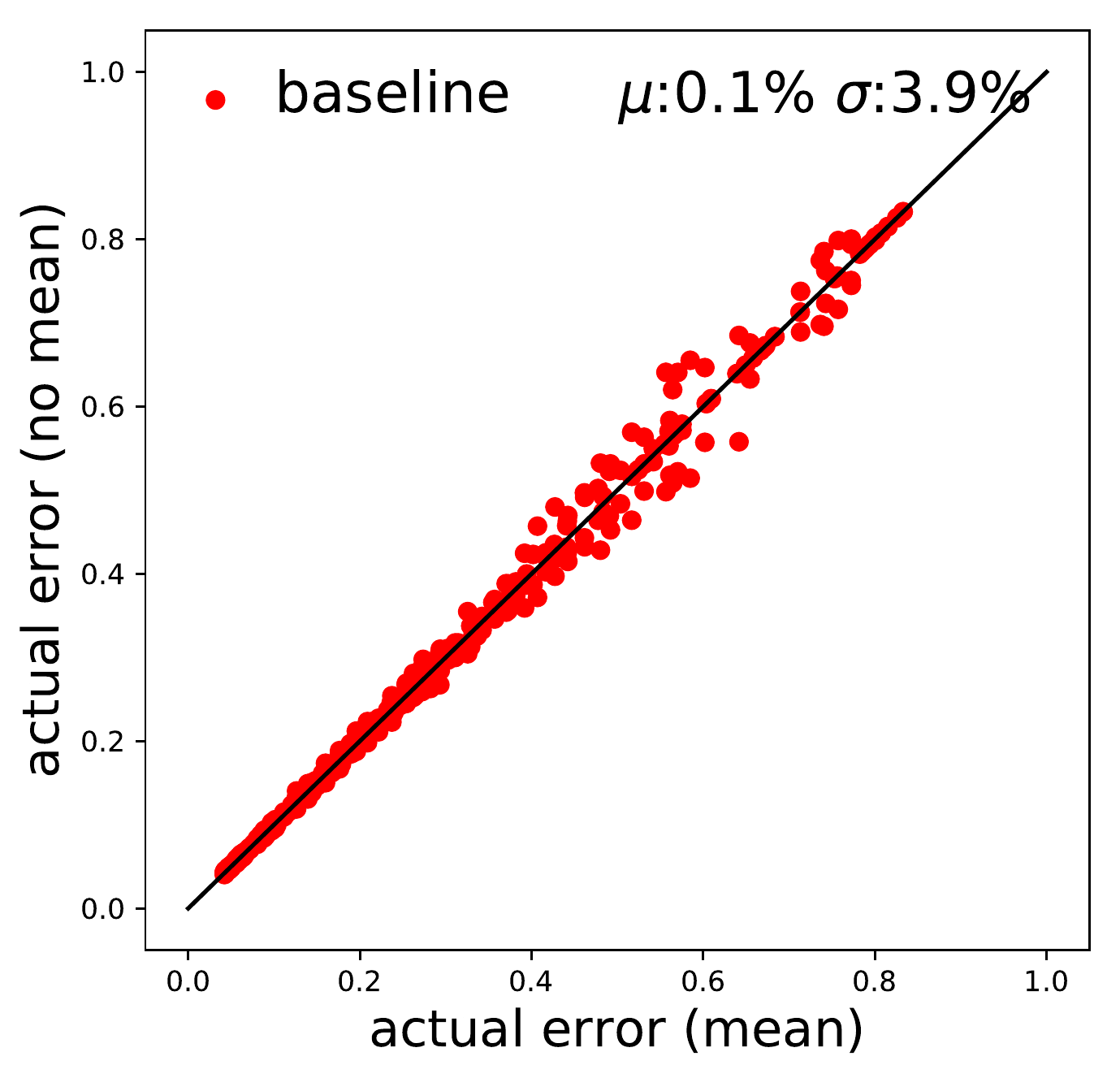}
\end{minipage}
\caption{Fit for DenseNet-121 on CIFAR-10 with IMP pruning.}
\vspace{0mm}
\label{fig:densenet-cifar}
\end{figure}

\begin{figure}[h!]
\centering
\begin{minipage}{0.3\textwidth}
    \includegraphics[width=\linewidth,trim={0.2 0 0 0.65cm},clip]{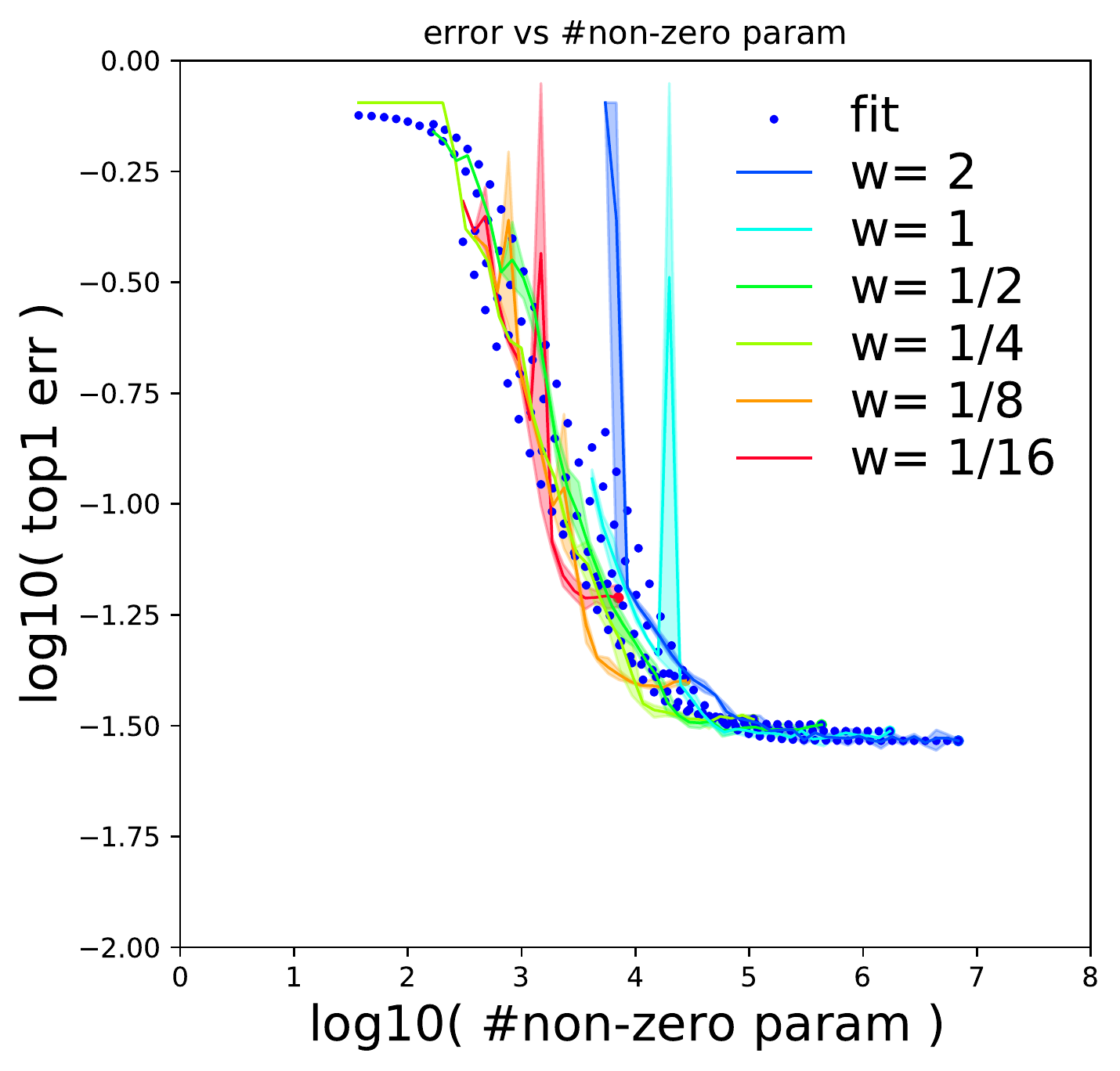}
\end{minipage}
\begin{minipage}{0.3\textwidth}
    \includegraphics[width=\linewidth]{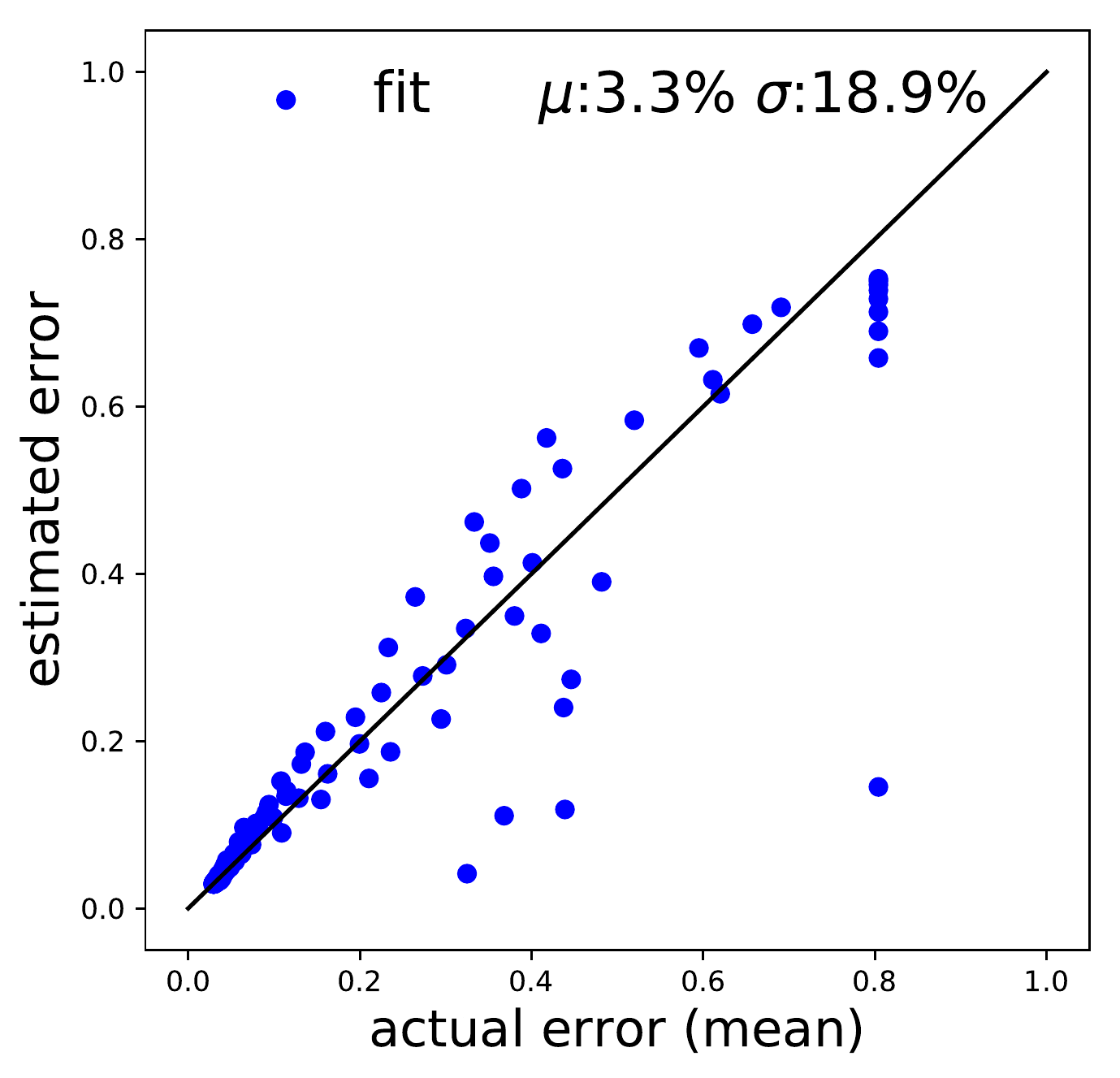}
\end{minipage}
\begin{minipage}{0.3\textwidth}
    \includegraphics[width=\linewidth]{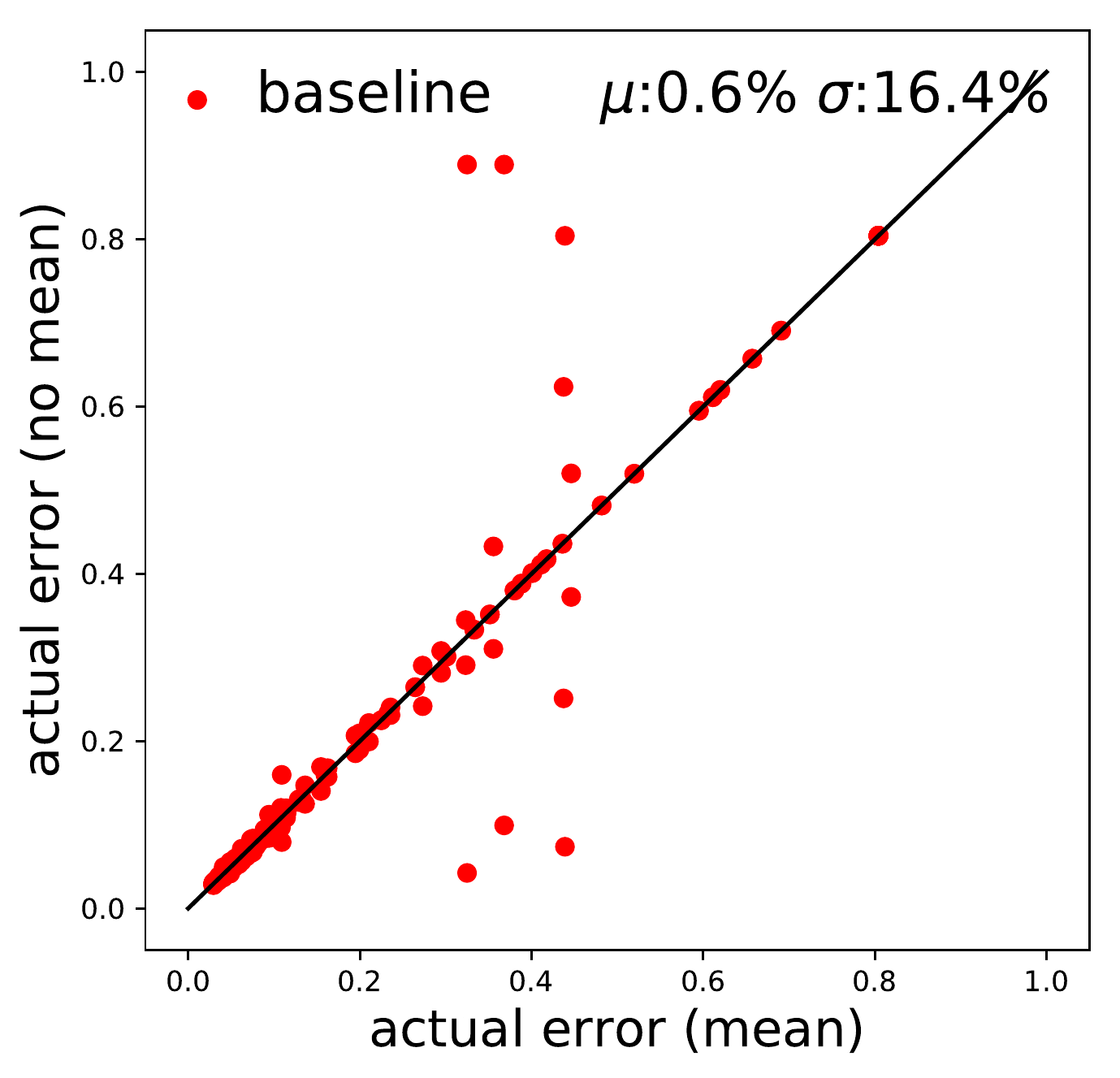}
\end{minipage}
\caption{Fit for DenseNet-121 on SVHN with IMP pruning.}
\vspace{0mm}
\label{fig:densenet-svhn}
\end{figure}

\begin{figure}[h!]
\centering
\begin{minipage}{0.3\textwidth}
    \includegraphics[width=\linewidth,trim={0.2 0 0 0.65cm},clip]{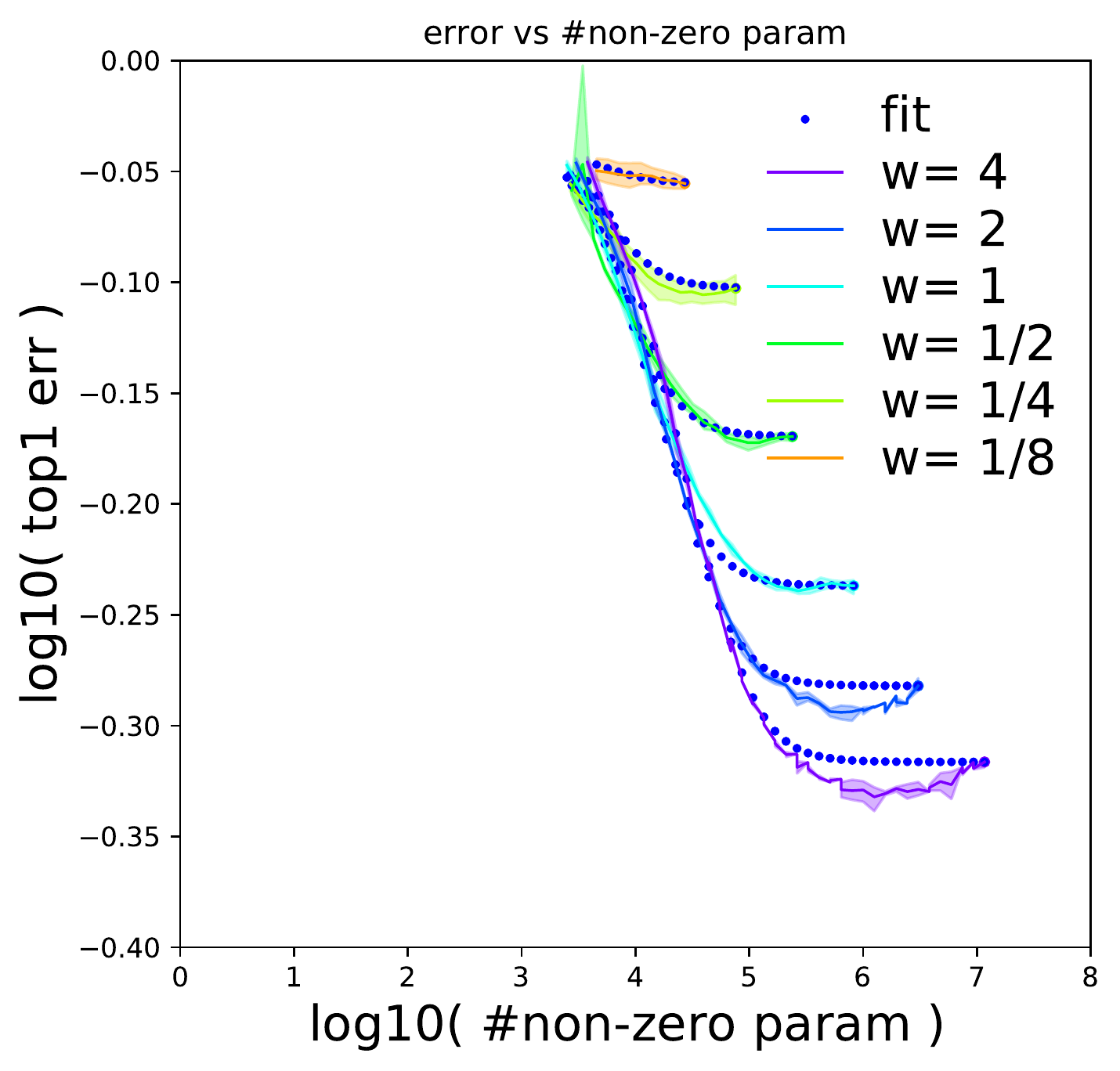}
\end{minipage}
\begin{minipage}{0.3\textwidth}
    \includegraphics[width=\linewidth]{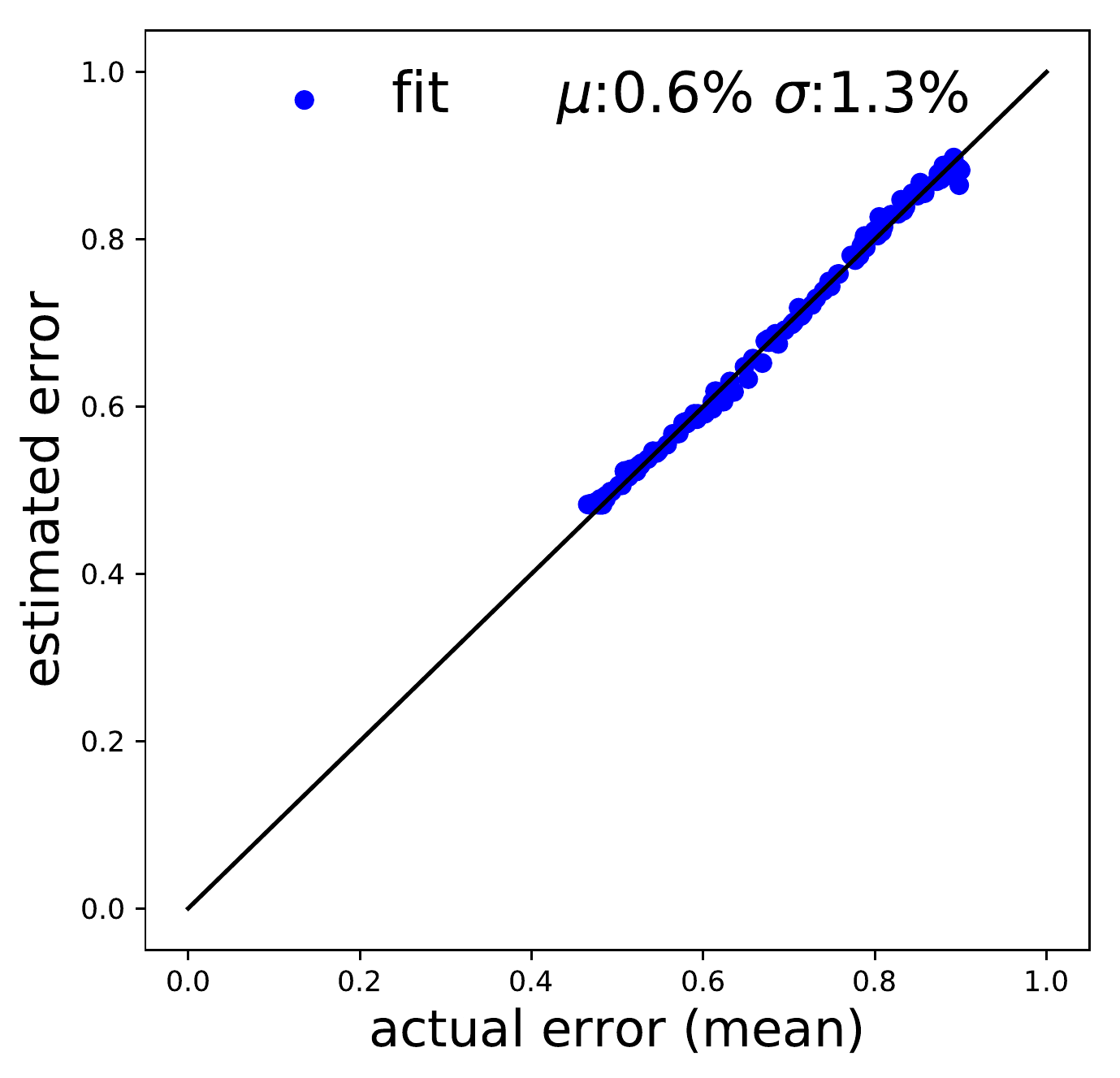}
\end{minipage}
\begin{minipage}{0.3\textwidth}
    \includegraphics[width=\linewidth]{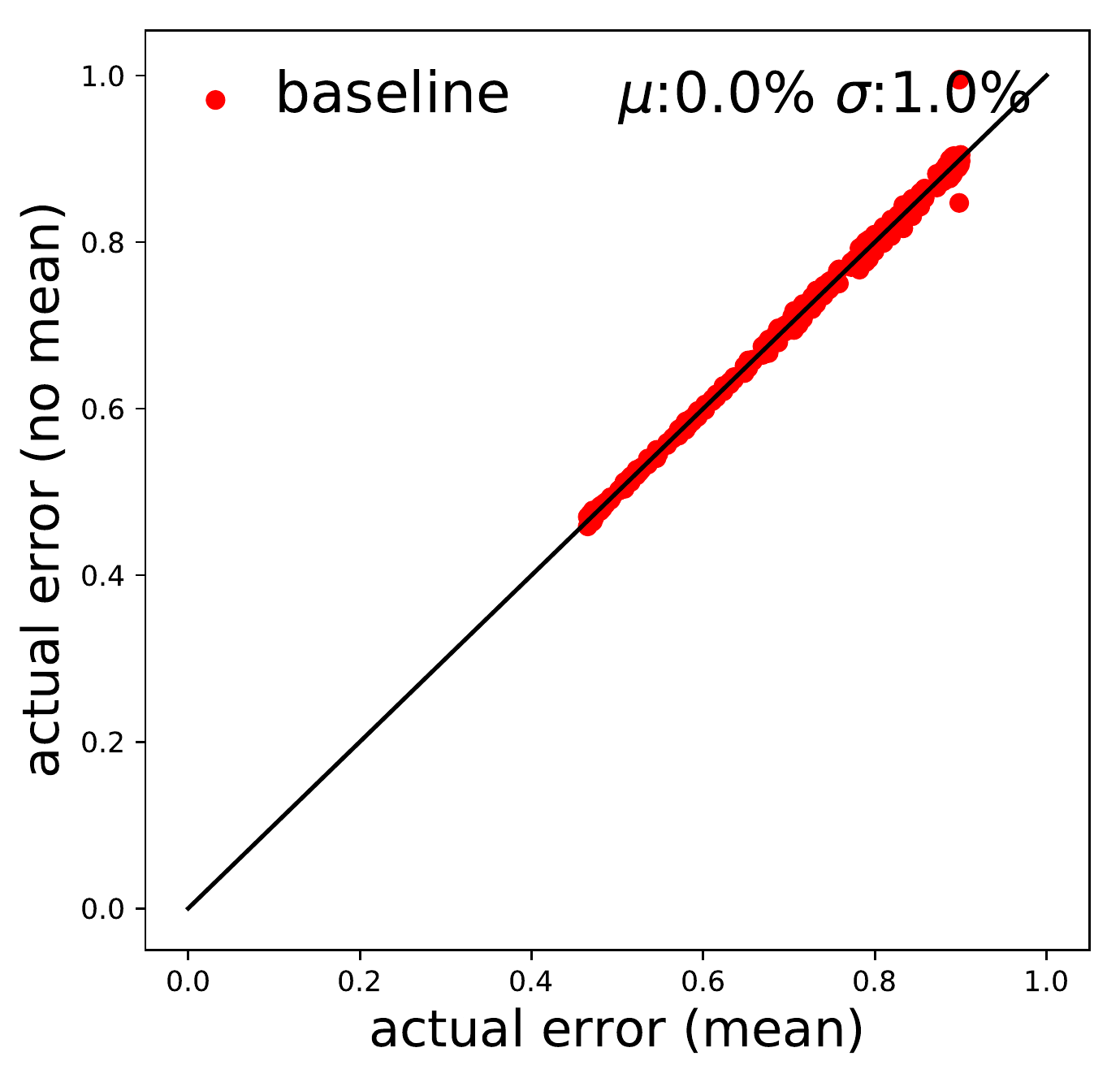}
\end{minipage}
\caption{Fit for ResNet-18 on TinyImageNet with IMP pruning.}
\vspace{0mm}
\label{fig:resnet-imagenet}
\end{figure}

We add these additional comparisons in the following Figures:

\begin{itemize}
    \item Figure \ref{fig:resnet-svhn}: ResNet-20 on SVHN with IMP as width varies. $|\mu|<2\%$, $\sigma<8\%$. 
    Notably, the measurement error in this case is large ($\sigma \sim 9.5\%$), dominating (over the approximation error) the total fit error. The fit averages out some of this error, resulting in a fit error which is lower than the measurement error. In general, experiments on SVHN are quite noisy, leading to significant measurement error.
    \item Figure \ref{fig:vgg-imp}: VGG-16 on CIFAR-10 with IMP as width varies. $|\mu|<3\%$, $\sigma<7\%$ (compared to measurement error 12\%).
    \item Figure \ref{fig:vgg-svhn}: VGG-16 on SVHN with IMP as width varies. $|\mu|<4\%$, $\sigma<15\%$ (compared to measurement error 20\%---measurement error is large at very low densities).
    \item Figure \ref{fig:densenet-cifar}: DenseNet-121 on CIFAR-10 with IMP as width varies. $|\mu|<1\%$, $\sigma<8\%$. 
    Bias is evident in the transition as we discuss in Section \ref{sec:design}.
    \item Figure \ref{fig:densenet-svhn}: DenseNet-121 on SVHN with IMP as width varies. $|\mu|<4\%$, $\sigma<19\%$ (compared to measurement error 16\%---measurement error is large at very low densities).
    \item Figure \ref{fig:resnet-imagenet}: ResNet-18 on TinyImageNet with IMP as width varies. $|\mu|<1\%$, $\sigma<1.3\%$ (compared to measurement error 1\%).

\end{itemize}


\begin{figure}[h!]
\centering
\begin{minipage}{0.3\textwidth}
    \includegraphics[width=\linewidth,trim={0.2 0 0 0.65cm},clip]{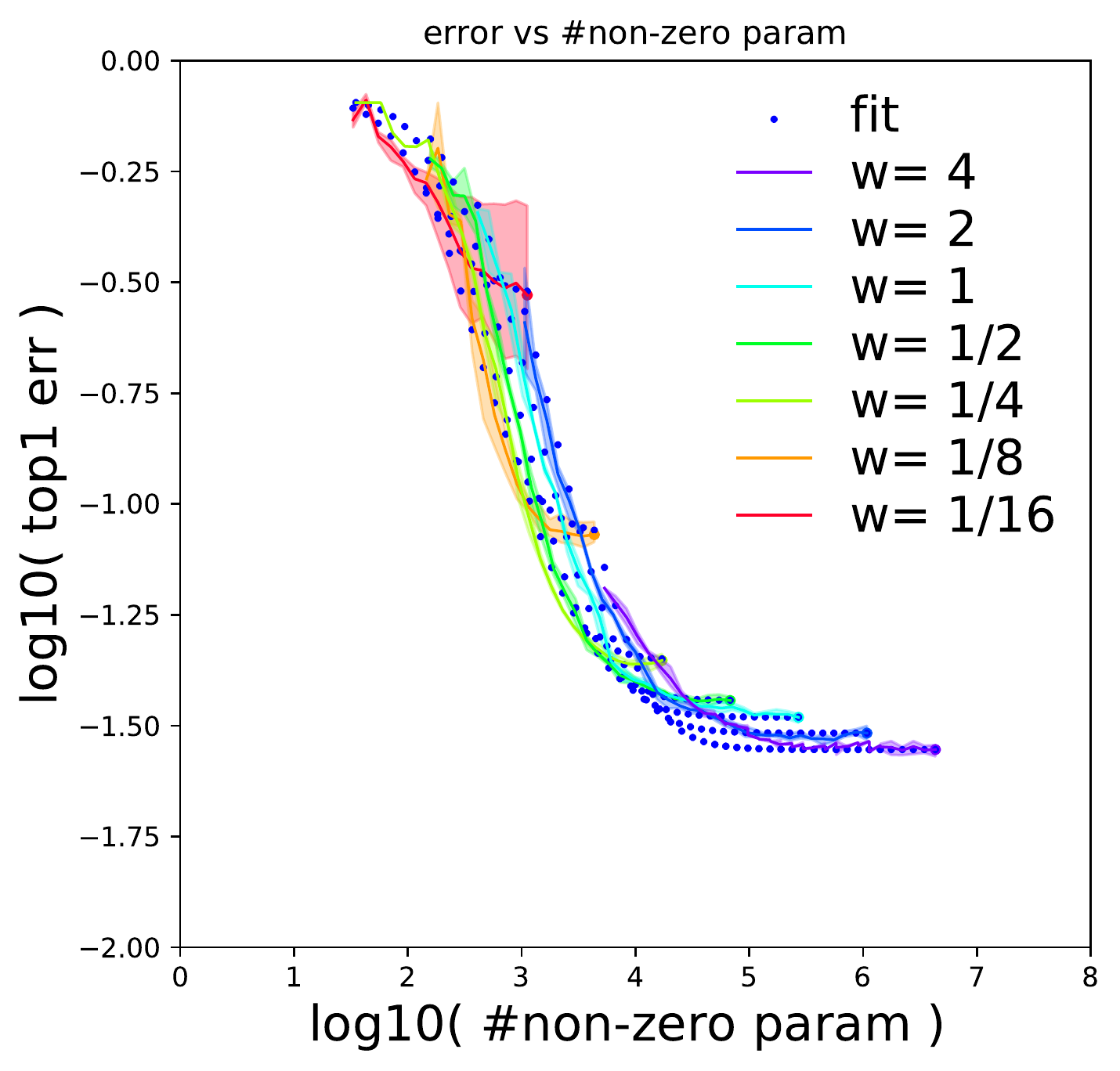}
\end{minipage}
\begin{minipage}{0.3\textwidth}
    \includegraphics[width=\linewidth]{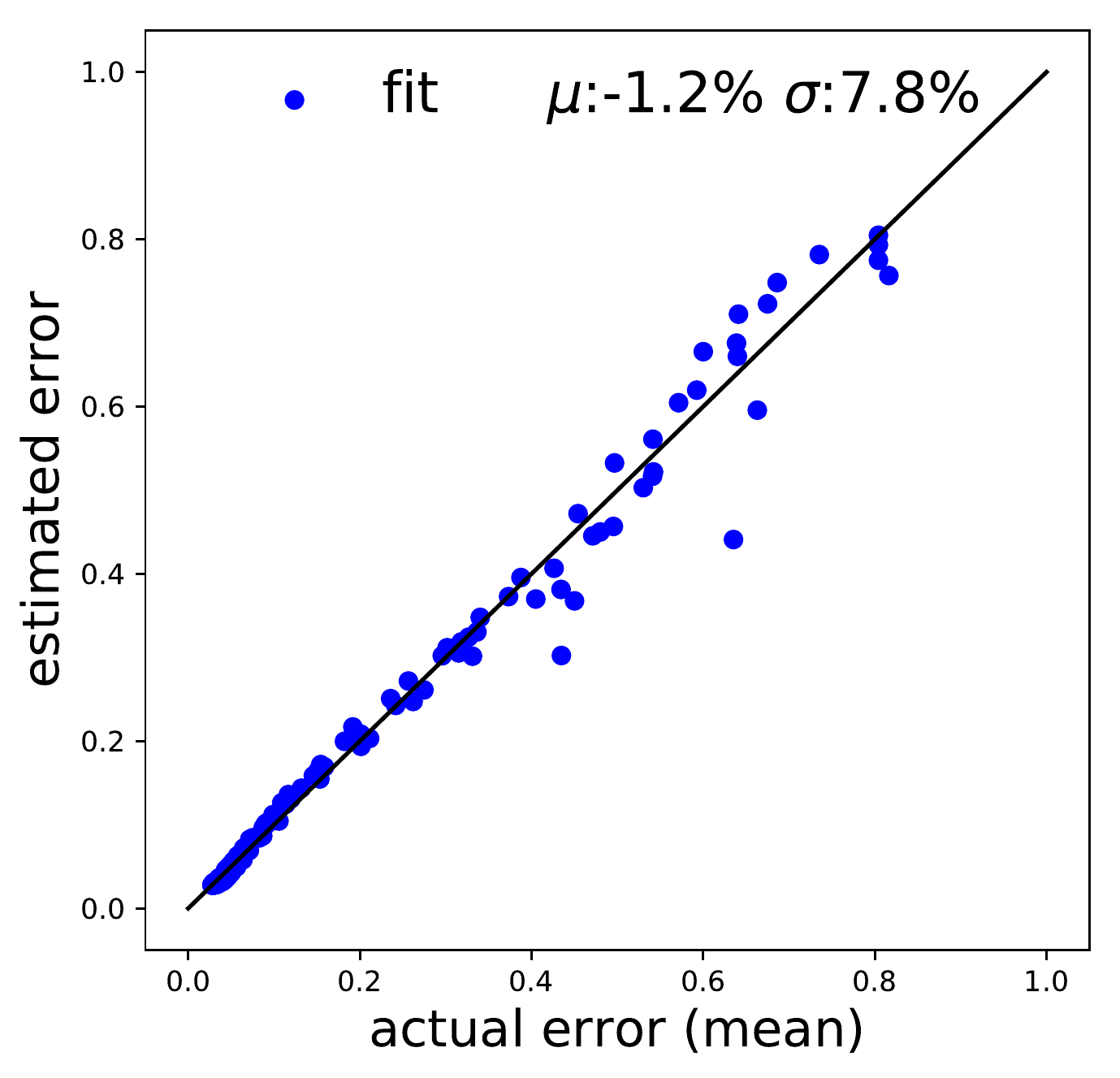}
\end{minipage}
\begin{minipage}{0.3\textwidth}
    \includegraphics[width=\linewidth]{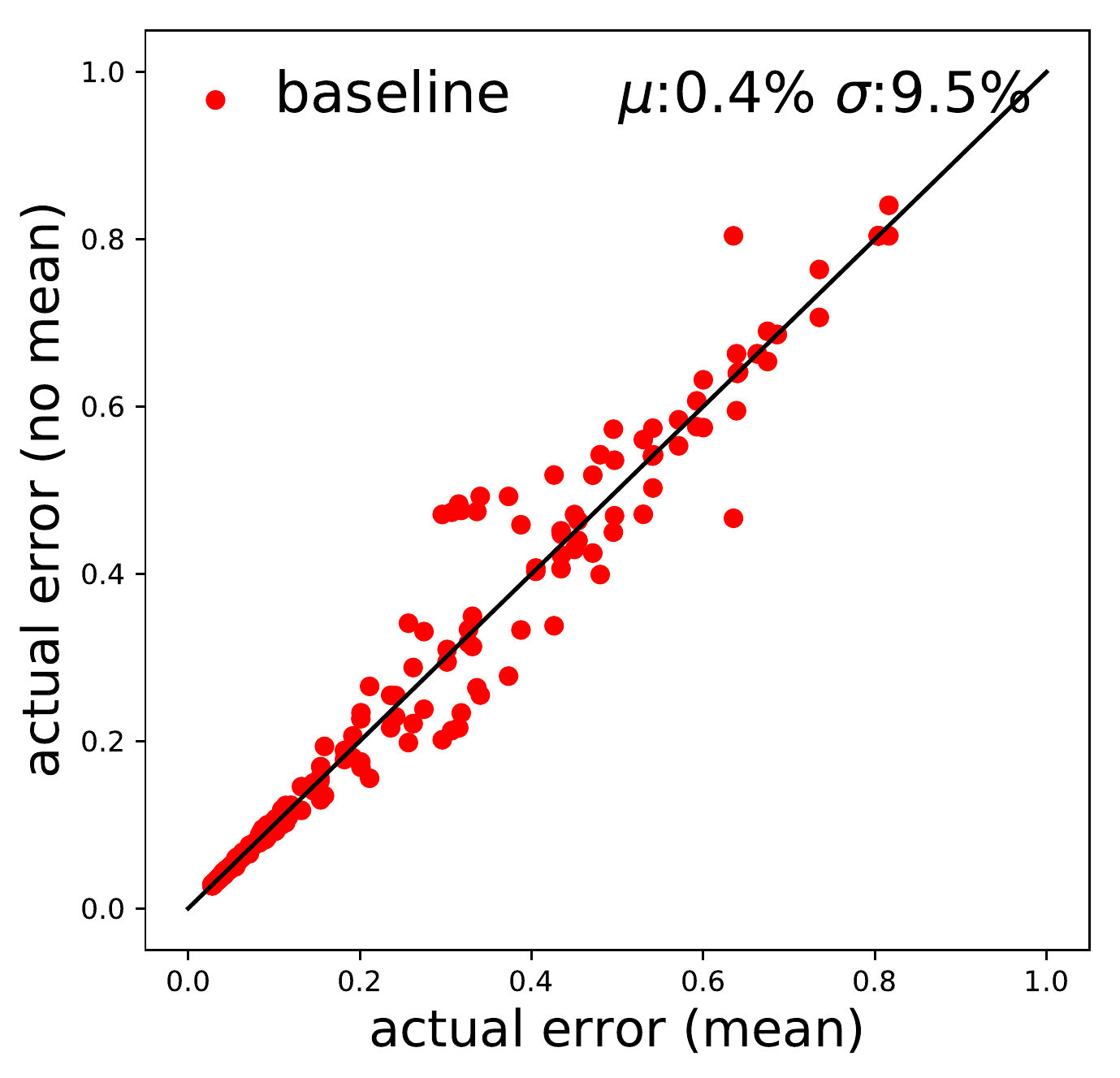}
\end{minipage}
\caption{Fit for ResNet-20 on SVHN with IMP pruning.}
\vspace{0mm}
\label{fig:resnet-svhn}
\end{figure}

\clearpage
\newpage

\section{Towards Extrapolation}
\label{app:more_extrapolations}

\textbf{Background.}
In the main body, we showed that our scaling law accurately fits the error of pruned neural networks.  As, such it has predictive power, allowing us to reason in a principled manner about pruning trade-offs.
Similarly, it allows to make predictions about what would happen at larger model and data scales than explored here.
Importantly, only a few experiments need be performed to find the coefficients for the scaling law (see Appendix \ref{app:interpolation}).

However, we could ask, how accurately can we estimate the scaling law parameters from even smaller scales?
That is, is it possible to fit our scaling law to data from networks with deliberately smaller depths, widths, and dataset sizes and accurately predict the error of larger-scale models? 
If so, we could make informed decisions about pruning large-scale models through small-scale experiments alone, saving the costs associated with large scale training and pruning.


Outside the context of pruning, the scaling laws of \cite{rosenfeld2020a} (for both language models and image classification) and \cite{kaplan2020scaling} (for predicting the expected performance of GPT-3 \citep{brown2020language} at very large scale) have been shown to extrapolate successfully in this manner.

\begin{figure}[h!]
    \centering
         \begin{minipage}{0.34\textwidth}
            \includegraphics[width=\linewidth,trim={0.2 0 0 0.65cm},clip]{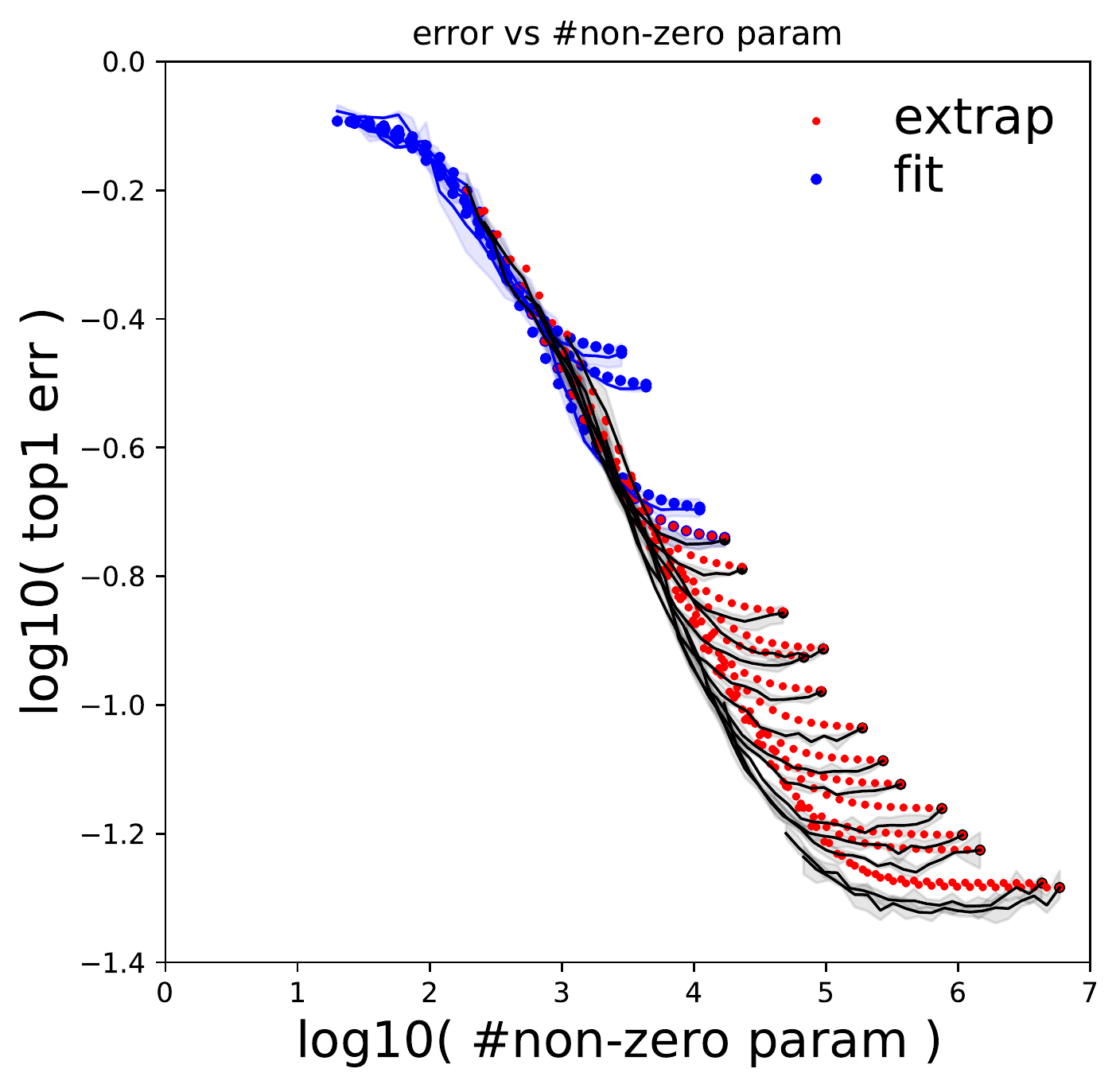}
            \label{fig:cifar_extrap}
        \end{minipage}%
        \begin{minipage}{0.335\textwidth}
            \centering
            \includegraphics[width=\linewidth,trim={0.2cm 0 0 0.3cm},clip]{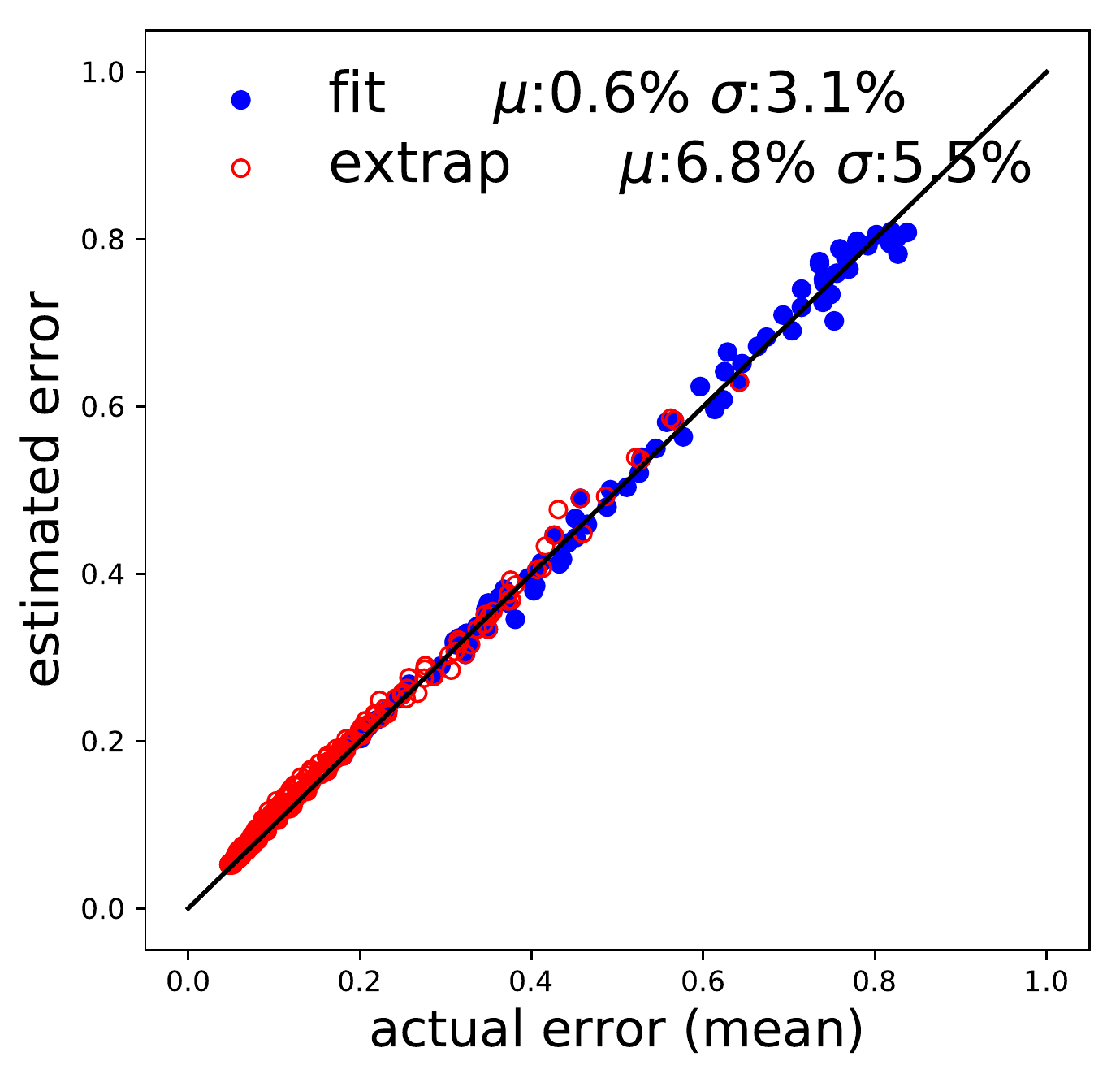}
            \label{fig:cifar_extrap_corr}
        \end{minipage}%

\caption{ Extrapolation results from four pruned networks on CIFAR10 $w=\frac{1}{8},\frac{1}{4}; l=14,20$ to all larger networks $(n=1)$. Fit results are in blue, extrapolation in red, actual in black.  Error versus number of non-zero parameters (left).  Estimated versus actual errors (right). 
}
\label{fig:extrap}
\end{figure}

\textbf{Results on CIFAR-10.}
In Figure \ref{fig:extrap}, we show the result of extrapolating from small-scale networks on CIFAR-10 ($w = \frac{1}{8}, \frac{1}{4}$; $l = 14, 20$) to all widths and depths on CIFAR-10.
Extrapolation prediction is still accurate: $\mu<7\%$, $\sigma<6\%$ (vs. $\mu<1\%$, $\sigma<6\%$ in the main body).

\textbf{Future work.}
However, extrapolation is particularly sensitive to systemic errors.
Specifically, the transitions and the error dips can lead to large deviations when extrapolating.
For ImageNet, the error dips (especially on small dataset sizes) are especially pronounced, preventing stable extrapolation.
In order to improve extrapolation performance, future work should explore the challenges we discuss in Section \ref{sec:design}: approaches to either model or mitigate these dips and to improve the fit of the transitions.



\setlength{\tabcolsep}{0pt}
\renewcommand{\arraystretch}{0.5}

\newpage

\section{Comparison of Pruning and Non-pruning Scaling Laws}
\label{app:comparison-to-rosenfeld}

In this appendix, we contrast the behavior of the error when pruning with the behavior of the error in the non-pruning setting. \citet{hestness2017deep} show the the error follows a saturating power-law form when scaling data (with both low and high-error plateaus) but does not model them.
Recall from Chaper \ref{sec:Dense} the error dependency on data and model size:

\begin{equation} \label{eq:norm_err} 
\vspace{-2pt}
    \tilde{\epsilon}(m,n) = an^{-\alpha} + bm^{-\beta} + c_\infty  
\end{equation}

\begin{equation} \label{eq:envelope}
    \hat{\epsilon}(m,n) = \epsilon_0 \left\Vert \frac{ \tilde{\epsilon}(m,n)}{\tilde{\epsilon}(m,n)-j\eta} \right\Vert 
\end{equation}

where $m$ is the total number of parameters and $n$ is the dataset size. $a,b,\alpha,\beta,c_\infty,$ and $\eta$ are constants, and $\epsilon_0$ plays a similar role of $\epsilon^\uparrow$ for the pruning case.

In Chapter \ref{sec:Dense} we modelled the upper transition---from power-law region to the high-error plateau---by a rational form in a fashion similar to the approach taken here.
The key difference is that we consider a power of the polynomials in the numerator and denominator of the rational form, where in Eq. \ref{eq:norm_err} the power is hidden in the term $\tilde\epsilon$.

The biggest difference arises when considering the lower transition (between the low-error plateau and the power-law region).
This transition is captured by Eq. \ref{eq:norm_err}.
Considering either the width or depth degrees of freedom $x \in \{w,l\}$, Eq. \ref{eq:norm_err} can be re-written as:

\begin{equation} 
\vspace{-2pt} \label{eq:adapted}
    \tilde{\epsilon}(x) = b_xx^{-\beta_x} + c_x
\end{equation}

Where $b_x$ and $\beta_x$ are constants and $c_x$ is a constant as a function of $x$ (it is only a function of the data size $n$).

Figure \ref{fig:mismatch_app} (right) shows the error versus depth for different dataset sizes.
In grey is the actual error, while in red is the best fit when approximating the error by Eq. \ref{eq:adapted}.
Qualitatively, one sees that the fit using Eq. \ref{eq:adapted} does indeed closely match the error in practice.

Recall that we are interested in comparing the errors as a function of the density. 
A requirement from any functional form used to model the dependency on the density is to degenerate to the error of the non pruned model $\epsilon_{np}$ at $d=1$.
We adapt Eq. \ref{eq:adapted} by solving the relation between $b_x$ and $c_x$ meeting this constraint, to arrive at:

\begin{equation} 
\vspace{-2pt} \label{eq:adapted_density}
    \tilde{\epsilon}(x) = b_xx^{-\beta_x} + \epsilon_{np}-b_x
\end{equation}

Contrast Eq. \ref{eq:adapted} with the functional form we propose in Eq. \ref{eq:prune_density}, re-written here for convenience:
{
\small
\begin{equation}
\label{eq:prune_density_2}
    \hat{\epsilon}(d,\epsilon_{np}~|~l, w, n) = \epsilon_{np} \left\Vert \frac{d-jp\left(\frac{\epsilon^\uparrow}{\epsilon_{np}}\right)^{\frac{1}{\gamma}}}{d-j p} \right\Vert^\gamma
    \mbox{where } j = \sqrt{-1}
\end{equation}
}

This can be simplified to capture only the lower transition---far enough from the upper transition ($d\gg p$)---to:

{
\small
\begin{equation}
\label{eq:prune_density_3}
    \hat{\epsilon}(d,\epsilon_{np}~|~l, w, n) = \epsilon_{np} \left\Vert \frac{d-jp\left(\frac{\epsilon^\uparrow}{\epsilon_{np}}\right)^{\frac{1}{\gamma}}}{d} \right\Vert^\gamma
\end{equation}
}

Figure \ref{fig:mismatch_app}  (left) shows error versus density for different widths.
In blue is the fit with Eq. \ref{eq:prune_density_3} which follows closely the actual error (black) while in red is the fit with Eq.  \ref{eq:adapted_density} which deviates noticeably in comparison.

\label{app:difference_in_powerlaws}
{

\begin{figure}[H]
    \centering
    \begin{minipage}{0.325\textwidth}
        \includegraphics[width=\linewidth,trim={0 0 0 0.7cm},clip]{figures_3/pruning_curves_comparison.pdf}
    \end{minipage}%
    \begin{minipage}{0.325\textwidth}
        \includegraphics[width=\linewidth,trim={0 0 0 0.7cm},clip]{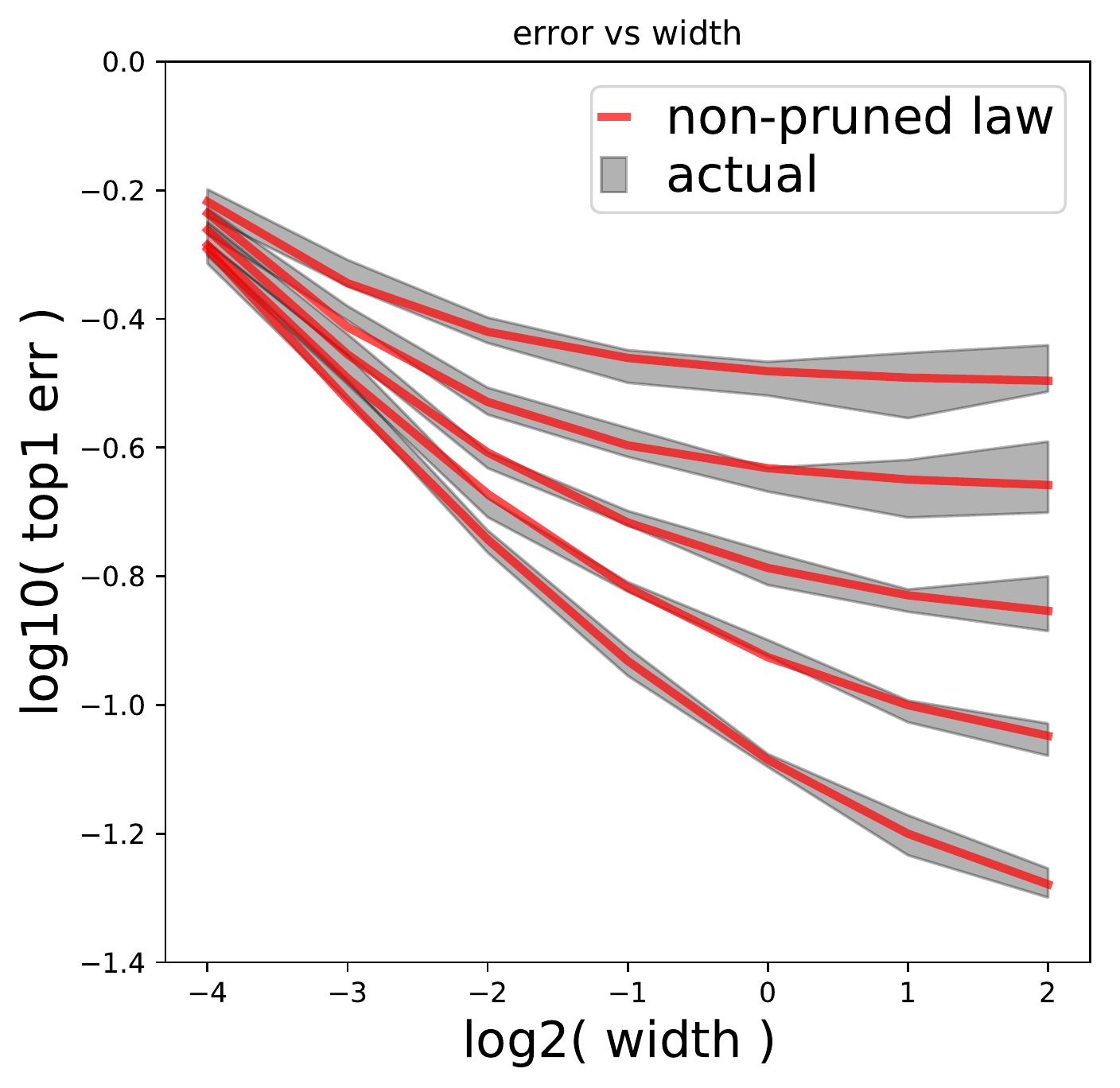}
    \end{minipage}%

    \caption{(Left) Error versus density for different widths. In blue is the fit \eqref{eq:intermediate_state} follows closely the actual error (black) while in red is the fit for the adapted from Chapter \ref{sec:Dense} which deviates noticeably in comparison. (Right) error of non-pruned networks versus width for different data, fit shown (solid red) for the non-pruning scaling from  Chapter \ref{sec:Dense}.
    }
    \label{fig:mismatch_app}

\end{figure} 

We have seen that in practice that the form of Eq. \ref{eq:adapted_density} does not match well the pruning case, where the mismatch originates from lower transition shape.
We have thus reached a phenomenological observation distinguishing the pruning and non-pruning forms; we leave the study of the origins of this phenomenon for future work. 

}

\newpage

\section{The effect of error dips on estimation bias}
\label{app:magic-one-percent}
In this appendix, we consider the effect of the error dips on our estimator as discussed in Section \ref{sec:joint}.
As we mention in that section, when pruning a network, error often dips below $\epsilon_{np}$ during the low-error plateau.

Recall that we find the parameters in our estimator (Equation \ref{eq:intermediate_state}) by minimizing the MSE of relative error $\delta$.
Our estimation has bias if $\mathbb{E}\left(\hat\epsilon -\epsilon\right) \neq 0 $ where the expectation is over all model and data configurations. Equivalently, the relative bias is $\mu \triangleq \mathbb{E}\delta = 0$ iff the estimator is unbiased. 
The Estimator captured by the joint form in Equation \ref{eq:intermediate_state} is a monotonically increasing function of the density.
It is also constrained such that at density $d=1$ it is equal to the non-pruned error $\epsilon_{np}$. 
It thus, can not reduce The MSE to zero, as it can not decrease to match the actual error dips. This results in the bias of the relative error $\mu$ which in practice is $\sim 1\%$.

\begin{singlespace}
\bibliography{main}
\bibliographystyle{plainnat}
\end{singlespace}

\end{document}